\documentclass[11pt]{article}
\usepackage[utf8]{inputenc}
\usepackage{amsthm,amsmath,amssymb,graphicx,gensymb} 
\usepackage{natbib}
\usepackage[normalem]{ulem}

\RequirePackage[colorlinks,citecolor=blue,urlcolor=blue]{hyperref}
\usepackage[dvipsnames]{xcolor}
\usepackage{color}
\usepackage{bm,bbm}
\usepackage[margin=2.1cm]{geometry}
\usepackage[skip=10pt plus1pt, indent=0pt]{parskip} 
\usepackage{caption}
\usepackage{subcaption}
\usepackage{pdflscape}
\usepackage{alphalph}

\newtheorem{definition}{Definition}
\newtheorem{proposition}{Proposition}

\title{\hrule Generative Machine Learning for Multivariate Angular Simulation} 
\author{Jakob Benjamin Wessel$^1$, Callum J.\ R.\ Murphy-Barltrop$^{2,3}$ and Emma S.\ Simpson$^4$\\ 
\normalsize{$^1$Department of Mathematics and Statistics, University of Exeter, UK}\\
\normalsize{$^2$Technische Universität Dresden, Institut Für Mathematische Stochastik, Dresden, Germany}\\
\normalsize{$^3$Center for Scalable Data Analytics and Artificial Intelligence (ScaDS.AI), Dresden/Leipzig, Germany}\\
\normalsize{$^4$Department of Statistical Science, University College London, UK\vspace{2.5pt}}\\
\normalsize{\textbf{Correspondence:} Jakob Wessel (j.wessel@exeter.ac.uk)}}
\date{\today\vspace{10pt}\hrule}

\begin{document}

\maketitle

\begin{abstract}
With the recent development of new geometric and angular-radial frameworks for multivariate extremes, reliably simulating from angular variables in moderate-to-high dimensions is of increasing importance. Empirical approaches have the benefit of simplicity, and work reasonably well in low dimensions, but as the number of variables increases, they can lack the required flexibility and scalability. Classical parametric models for angular variables, such as the von Mises--Fisher distribution (vMF), provide an alternative. Exploiting finite mixtures of vMF distributions increases their flexibility, but there are cases where, without letting the number of mixture components grow considerably, a mixture model with a fixed number of components is not sufficient to capture the intricate features that can arise in data. Owing to their flexibility, generative deep learning methods are able to capture complex data structures; they therefore have the potential to be useful in the simulation of multivariate angular variables. In this paper, we introduce a range of deep learning approaches for this task, including generative adversarial networks, normalizing flows and flow matching. We assess their performance via a range of metrics, and make comparisons to the more classical approach of using a finite mixture of vMF distributions. The methods are also applied to a metocean data set, with diagnostics indicating strong performance, demonstrating the applicability of such techniques to real-world, complex data structures. 
\end{abstract}


\vspace{0.25cm}\noindent{\bf Acknowledgements}\\ 
We would like to thank Ed Mackay for providing the data analysed in Section~\ref{sec:waves}, and for helpful discussions. Jakob Wessel was supported during this work by the Engineering and Physical Sciences Research Council (EPSRC) under grant award number 2696930. This work used JASMIN, the UK’s collaborative data analysis environment (\url{https://www.jasmin.ac.uk)}

\newpage
\section{Introduction}\label{sec:intro}

Generative AI methods have recently attracted considerable attention, offering flexible tools for modelling complex data structures. These approaches are proving popular in a range of application areas, from image generation \citep{rombach_high-resolution_2022} to protein structure prediction \citep{abramson_accurate_2024} and weather forecasting \citep{price_probabilistic_2025}. The multivariate extremes literature has also seen several recent contributions in this vein, and much more research combining extremes and AI is likely to emerge in the coming years. So far, there has been some focus on developing deep learning approaches that are able to generate multivariate data with heavy tails. In this setting, \cite{Bhatia2021} and \cite{Allouche2022} exploit generative adversarial networks (GANs), while \cite{Lafon2023} adopt a variational autoencoder approach and \cite{Hickling2024} use normalizing flows. Other recent contributions include the modelling of multivariate threshold exceedances, using GANs \citep{Allouche2025} or normalizing flows \citep{Hu2025}. While all of these approaches focus on extremes of $d$-dimensional random vectors in $\mathbb{R}^d$, a useful alternative in the study of multivariate extremes is to consider angular-radial representations of random variables. Very recent machine learning contributions in the angular-radial extremes context include the Wasserstein-Aitchison GAN approach of \cite{Lhaut2025} for generating angular data, and the approaches of \cite{Murphy-Barltrop2024} and \cite{DeMonte2025}, which both exploit deep learning methods in the framework of geometric extremes. In this paper, we focus on developing machine learning approaches specifically for generating multivariate angular data, with a view to these being used in the context of multivariate extremes modelling; further motivation for this is provided below.

As mentioned above, when studying a $d$-dimensional random vector $\boldsymbol{X}=(X_1,\dots,X_d)\in\mathbb{R}^d$, it can be convenient to consider a decomposition into a radial component $R$ and a vector of (pseudo-)angles $\boldsymbol{W}$, defined as 
\begin{equation}
R=\|\boldsymbol{X}\|_A, \qquad \boldsymbol{W} = \boldsymbol{X}/\|\boldsymbol{X}\|_B,
\label{eqn:RWdecomposition}
\end{equation}
for some choice of norms $\|\cdot\|_A$ and $\|\cdot\|_B$. By definition, $\boldsymbol{W}$ takes values on the  $(d-1)$-sphere, or hypersphere, denoted by $\mathbb{S}^{d-1}=\left\{\boldsymbol{x} \in \mathbb{R}^d:\|\boldsymbol{x}\|_B=1\right\}$. An example of where  decomposition~\eqref{eqn:RWdecomposition} often arises is under the classical framework of multivariate regular variation (MRV). Here, one considers the case where components of $\boldsymbol{X}$ have some common heavy-tailed and strictly positive marginal distribution, and the assumption of regular variation implies that $\boldsymbol{W}$ and $R$ become independent given $R>r$ as $r\rightarrow\infty$; see, e.g., \cite{Resnick2007} for details. 

More recently, there has been growing interest in so-called \textit{geometric} approaches for multivariate extremes, where the random vector $\boldsymbol{X}$ is assumed to have light-tailed margins. From a statistical perspective, \cite{WadsworthCampbell2024}  were the first to exploit this geometric framework for modelling the joint tails of $\boldsymbol{X}$. Their approach allows for theoretically-justified extrapolation to regions beyond those observed, i.e., estimation of small probabilities associated with the joint tail. Working on exponential margins and applying the $L^1$ norm to calculate $(R,\boldsymbol{W})$ in Eq.~\eqref{eqn:RWdecomposition}, they propose to model the upper tail of $R$ conditional on the value of $\boldsymbol{W}$; we note that this has echoes of the MRV framework, but the role of conditioning is now assigned to a different component of $(R,\boldsymbol{W})$. In this geometric approach, extrapolation is achieved via a two-step simulation method: first taking draws from a $(d-1)$-dimensional angular distribution, and subsequently from a univariate conditional radial distribution; these are then combined to return simulations on the original exponential scale, representing $\boldsymbol{X}$. Further approaches to inference in the context of geometric extremes have been proposed by \cite{Papastathopoulos2025}, who take a hierarchical Bayesian approach; \cite{Murphy-Barltrop2024}, who use neural networks to allow for estimation in higher dimensions; \cite{Campbell2024}, whose approach is based on piecewise-linear models and kernel density estimation; and \cite{DeMonte2025}, who exploit normalizing flows. This remains an active area of research.

The majority of existing geometric approaches tend to work with random vectors on the standard exponential scale. While it is mathematically convenient, working in exponential margins offers a less complete picture of joint tail behaviour, especially in the case of negative dependence structures; see \citet{Murphy-Barltrop2024b} for recent discussion. With this in mind, a natural extension is to instead consider $\boldsymbol{X}$ on a standard Laplace scale, increasing flexibility in the types of dependence features that can be studied. This is an approach taken, for instance, by \cite{Simpson2024b} in the estimation of \textit{environmental contours} and \cite{Papastathopoulos2025} for estimating \textit{return level sets}, and is considered in the supplementary material of \cite{Campbell2024} as an alternative to their piecewise-linear approach on exponential scale. We will bear this in mind later in the current article when considering suitable choices of marginal distribution for our comparative studies.

Adjacent to the introduction of the geometric approaches discussed above has been the development of semi-parametric angular-radial (SPAR) models for multivariate extremes. This modelling framework, first introduced by \citet{Mackay2023}, provides a robust and flexible way to model multivariate extreme events without assuming fixed marginal distributions. First, the joint density of $(R,\boldsymbol{W})$ is decomposed into the conditional form $f_{R,\boldsymbol{W}}(r,\boldsymbol{w}) = f_{\boldsymbol{W}}(\boldsymbol{w}) \,f_{R|\boldsymbol{W}}(r\mid\boldsymbol{w})$.
In this way, the problem of modelling multivariate extremes is transformed to that of modelling an angular density $f_{\boldsymbol{W}}$ and the tail of the conditional radial density $f_{R|\boldsymbol{W}}$. For a given angle $\boldsymbol{w}$, the density $f_{R|\boldsymbol{W}}(r\mid\boldsymbol{w})$ is univariate, and under the SPAR framework, the tail of $R\mid(\boldsymbol{W} = \boldsymbol{w})$ is modelled using a generalised Pareto distribution (GPD). Defining a threshold function $u(\boldsymbol{w})>0$ to be the quantile of $R\mid(\boldsymbol{W} = \boldsymbol{w})$ with exceedance probability $\tau\in(0,1)$, with $\tau$ close to 0, i.e., the solution of $\tau = \mathbb{P} \{R>u(\boldsymbol{w}) \mid \boldsymbol{W} = \boldsymbol{w}\}$, the SPAR model is formulated as
\begin{equation*} 
    f_{R,\boldsymbol{W}} (r,\boldsymbol{w}) = \tau f_{\boldsymbol{W}} (\boldsymbol{w}) f_{\text{GPD}} \{r - u(\boldsymbol{w}); \xi(\boldsymbol{w}), \sigma (\boldsymbol{w})\}, \quad r>u(\boldsymbol{w}),
\end{equation*}
where $f_{\text{GPD}}$ is the density function associated with the GPD, and $\xi(\boldsymbol{w}) \in \mathbb{R}$ and {$\sigma (\boldsymbol{w})>0$} are angle-dependent shape and scale parameters, respectively. Combined with the angular density $f_{\boldsymbol{W}} (\boldsymbol{w})$, the SPAR framework can be used to perform inference on the joint tail of $\boldsymbol{X}$ in regions containing both the lower and upper tails of each variable. Several implementations of the SPAR approach exist \citep{Murphy-Barltrop2024b,mackay_deep_2024,mackay2025spar}, and \citet{Mackay2023} demonstrate theoretically its links with several existing multivariate modelling frameworks, including those of \citet{Ledford1996} and \citet{Wadsworth2017}, as well as the geometric framework. As with the latter approach, extrapolation under the SPAR model can be achieved by drawing from the angular distribution, then subsequently simulating from the conditional GPD. 

Multivariate extreme value modelling approaches involving angular-radial decompositions have been applied to a wide variety of environmental datasets; these include structural design engineering \citep{Coles1994,deHaan1998}, air pollution monitoring \citep{Simpson2024b, Majumder2023}, flood risk management \citep{WadsworthCampbell2024} and metocean modelling \citep{Murphy-Barltrop2024,Papastathopoulos2025}. This demonstrates the practical utility of such modelling techniques, and hence the importance of accurately modelling the (pseudo-)angular vector. While our idea to study deep learning methods for angular simulation is motivated by geometric and angular-radial approaches in multivariate extremes, we believe the findings will also be useful more generally in other areas where circular or angular variables arise. Additional applications include, but are not limited to, forecasts of wind direction 
\citep[e.g.,][]{Lang.etal.2020}, modelling genomes \citep[e.g.,][]{Shieh.etal.2011}, and studies in marine biology \citep[e.g.,][]{Lund1999}. 

As outlined above, an essential component of inference in the SPAR, geometric extremes and MRV modelling frameworks is reliable estimation of the distribution of the (pseudo-)angular vector $\boldsymbol{W}$. Simulating from the distribution of $\boldsymbol{W}$ can be a useful way to achieve this without explicitly specifying the full density function, and facilitates the estimation of joint tail probabilities outside the range of observed values. In low dimensions, it may be reasonable to take an empirical approach to this simulation task (i.e., sampling only from the observed angular values), but this quickly becomes restrictive when moving into higher dimensions. Constructing parametric models that are able to capture the wide range of possible angular distributions is also a challenge. There is therefore room to investigate non-parametric approaches to flexibly simulate from an angular distribution based on observations, particularly in moderate-to-high dimensions.

The use of generative AI techniques for modelling angular data has been scarcely considered in the literature, and while deep learning techniques for data generation are well-established, their application in the context of angular modelling for (multivariate) extremes remains largely unexplored. In this work, we aim to address these shortcomings by introducing and rigorously assessing a range of generative AI approaches for angular data. We further aim to provide guidance and insights into best practices for applications requiring angular simulation.

Throughout this paper, we will focus on the case where both $\|\cdot\|_A$ and $\|\cdot\|_B$ in definition~\eqref{eqn:RWdecomposition} represent the $L^2$ (or Euclidean) norm. This leads to a generalisation of polar coordinates to the $d$-dimensional setting, usually referred to as \textit{spherical coordinates}. We still denote our radial component by $R\geq 0$ but, following notational convention elsewhere, the spherical angles are denoted by $\boldsymbol{\Theta}=\left(\Theta_1,\dots,\Theta_{d-1}\right)$; a $d$th component of $\boldsymbol{\Theta}$ is not required as $\boldsymbol{X}$ can be fully described by the radial component $R$ and $(d-1)$ angles. Equations linking the original random vector $\boldsymbol{X}$ to the 
$d$-dimensional spherical coordinates $(R, \boldsymbol{\Theta})$ are provided in Appendix~\ref{app:spherical}. Throughout this article, we use upper and lower case symbols to denote random variables and observations, respectively. For example, taking spherical angles, we use $\boldsymbol{\Theta}$ for the random vector and $\boldsymbol{\theta}$ for corresponding observations.

The outline of the paper is as follows. In Section~\ref{sec:methods}, we provide an overview of the machine learning approaches we consider, and introduce the novel extensions required for our angular setting. Section~\ref{sec:evaluation} details the evaluation metrics we use to test these methods for the task of multivariate angular simulation. We present a simulation study in Section~\ref{sec:simulation}, covering a range of dependence structures and marginal tail behaviours for $\boldsymbol{X}$, as well as considering results across different dimensions. The methods are demonstrated on a metocean application in Section~\ref{sec:waves}, and Section~\ref{sec:conclusion} concludes.

\section{Generative methods}\label{sec:methods}

We explore the use of a number of generative deep learning methods for the simulation of angular variables. In this section, we will briefly outline the methods used. For brevity, we do not provide complete introductions to these methods, but  additional resources are suggested for the interested reader. Additional details and a brief introduction to neural networks can be found in Appendix~\ref{app:NNdetails}. We have also made Jupyter notebooks available as supplementary material, demonstrating the implementation of each method. In addition to the methods outlined in the following, we explored the use of maximum mean discrepancy networks/energy score networks \citep{dziugaite_training_2015, chen_generative_2024, pacchiardi_probabilistic_2024, shen_engression_2024}, but found them less competitive for angular simulation, possibly due to challenges in designing kernels or scores that respect spherical geometry. Consequently, results for these models are omitted.

We start by introducing generative adversarial networks in Section~\ref{subsec:gan}, followed by two different normalizing flows in Section~\ref{subsec:normflows}: neural spline flows and masked autoregressive flows. In Section~\ref{subsec:flow_matching}, we introduce flow matching, while in Section~\ref{subsec:vMFmix} we propose a `baseline' approach that involves fitting finite mixtures of the parametric von Mises--Fisher distribution. Finally, in Section \ref{subsec:hyperparam} we discuss hyperparameter tuning and architecture choices, alongside further implementation details.

We note that, unlike standard applications of generative deep learning methods, which typically model variables in the Euclidean space $\mathbb{R}^d$, the methods below must accommodate the fact $\boldsymbol{W}$ lies on $\mathbb{S}^{d-1}$. This comes with unique challenges. For the GAN and normalizing flows,  we transform values on the hypersphere into the $(d-1)$ dimensional spherical coordinates $\boldsymbol{\Theta}$  mentioned in Section~\ref{sec:intro}, where the first $(d-2)$ coordinates are bounded on $[0, \pi]$ and the remaining coordinate is cyclic on $(-\pi, \pi]$. Consequently, the neural network architectures must account for this transformation. We remark that the transformation from $\mathbb{S}^{d-1}$ to spherical coordinates is not one-to-one, owing to non-uniqueness around the `poles' \citep{blumenson1960derivation}. However, for continuous data, this represents a set of finite points on the hypersphere with zero probability mass, and thus is unlikely to cause problems in practice. In contrast to the GAN and normalizing flows, the flow matching models will be defined directly on the hypersphere by adjusting the probability flows to follow the geometry of the problem. 

\subsection{Generative adversarial networks}\label{subsec:gan}

So-called \textit{generative adversarial networks} \citep[GANs, ][]{goodfellow_generative_2014} are one of the most widely-used and popular approaches to generative modelling, especially in the context of image modelling. In order to approximate a $(d-1)$-dimensional target distribution $p_\text{target}$, the GAN framework casts this as an adversarial learning problem \citep[e.g.,][]{lowd2005adversarial}. A GAN consists of a generator function $G(\boldsymbol{z})$ and discriminator function $D(\boldsymbol{x})$, both implemented as neural networks and linked in an adversarial training routine. The generator maps samples from a base distribution $\boldsymbol{z} \sim p_\text{base}$ to the target distribution $\boldsymbol{x} \sim p_\text{target}$ using a neural network. The discriminator, on the other hand, takes as its input a sample and outputs a value in $[0,1]$ which can be interpreted as the estimated probability that a given value comes from the data rather than the synthetic distribution defined by the generator. The training objective of the discriminator is thus to correctly distinguish between samples from the generator and samples from the data, whereas the generator's objective is to produce the most realistic looking samples by `fooling' the discriminator, thus minimising $\log [1-D\{G(\boldsymbol{z})\}]$. This leads a min-max problem, with loss function 
\begin{equation}
    \underset{G}{\min} \ \underset{D}{\max}\ V(D,G) = \mathbb{E}_{\boldsymbol{x} \sim p_\text{target}} \left[\log D(\boldsymbol{x})\right] + \mathbb{E}_{\boldsymbol{z} \sim p_\text{base}} \left[\log \left[1-D\{G(\boldsymbol{z})\}\right]\right].
    \label{eqn:GANtarget}
\end{equation}

In general, GAN training is known to be unstable, hence alternatives such as the Wasserstein GAN \citep{arjovsky_wasserstein_2017} have been proposed. This casts the learning target in Eq.~\eqref{eqn:GANtarget} as an adversarial game between a generator and a critic, with the critic returning values in $\mathbb{R}$ instead of $[0,1]$. However, in our examples, we found GAN traning to work well in all cases, and we proceed with the default formulation above. 

In our setting, GANs must approximate the distribution of the $(d-1)$ angular variables resulting from the spherical coordinate representation of $\mathbb{S}^{d-1}$. We recall that this differs from typical Euclidean data since the first $d-2$ angles are bounded on $[0,\pi]$ and the final angle is periodic on $(-\pi,\pi]$. We use feedforward neural networks \citep[][Chapter 6]{Goodfellow-et-al-2016} for both the generator and discriminator with rectified linear unit (ReLU) activation for the generator and leaky ReLU activation for the discriminator, following \citet{radford_unsupervised_2016}. To ensure the output of the generator network satisfies the relevant bounds for each angle, after the last generator layer, we use a sigmoid activation for the first $d-2$ angles and a hyperbolic tangent activation for the last angle, i.e.,
\begin{align*}
    \theta_i \leftarrow \pi \cdot \frac{1}{1 + e^{-x_i}} \text{ for } i = 1, \dots, d-2,\qquad \text{and}\qquad
    \theta_{d-1} \leftarrow \pi \cdot \tanh x_{d-1}.
\end{align*}
A circular wrapping using a modulo function would typically be used to impose periodicity on $\Theta_{d-1}$ in deep generative models (see the normalizing flows in Section~\ref{subsec:normflows}). However, for the GAN framework, we found that the circular wrapping resulted in instabilities during the training procedure, reducing the reliability of this approach. To overcome this, we opted to use the hyperbolic tangent to impose the variable bounds, and found this to work well in practice.

We note that, in principle, GANs can operate directly on the $(d-1)$-sphere, but this requires enforcing a norm constraint during training and complicates optimisation. We therefore work in the angular space, where constraints can be imposed through suitable output activations, as outlined above. We defer the exploration of GANs directly on $\mathbb{S}^{d-1}$ to future work.

\subsection{Normalizing flows}\label{subsec:normflows}

Normalizing flows provide a tractable but flexible way to build generative models, and have been successfully used in a variety of domains. Intuitively, normalizing flows aim to approximate a $(d-1)$-dimensional target density (spherical angles here) using a $(d-1)$-dimensional base distribution $p_\text{base}$ and a transformation $T$, that is 
\begin{equation}\label{eq:normflow}
    \boldsymbol{U} = T(\boldsymbol{X}) \text{ with } \boldsymbol{X} \sim p_\text{base}.
\end{equation}
If the transformation $T$ is a diffeomorphism, meaning it is invertible with both $T$ and $T^{-1}$ differentiable, then the standard change of variables formula gives the density of $\boldsymbol{U}$ as
\begin{equation}\label{eq:change_of_variables}
    p(\boldsymbol{u}) =  p_\text{base}\left\{T^{-1}(\boldsymbol{u})\right\} \left| \det\left(\frac{\partial T}{\partial \boldsymbol{u}}\right) \right|.
\end{equation}
As diffeomorphisms are composable, the transformation $T$ can be taken as a series of $M$ simpler diffeomorphisms $T = T_1 \circ \dots \circ T_M$, which sequentially modify the base density. Normalizing flows typically specify a series of transformations $T_i$, for $i=1,\dots,M$, using neural networks, resulting in a highly flexible density model.

Normalizing flows have the advantage that the density of the model is available, often in closed form. This makes the models tractable and allows for maximum likelihood based training. Furthermore, sampling from the model target density is possible by drawing from the base density and applying Eq.~\eqref{eq:normflow}. To make the neural network based transformations invertible, but also for efficient computation of the determinant of the Jacobian, different normalizing flows incorporate different constraints on the transformations $T_i$ $(i=1,\dots,M)$. This means that these models are sometimes considered to be less expressive than alternative generative models \citep{stimper_resampling_2022}. Nonetheless, the tractability provides the benefit of stable model training. See \cite{papamakarios_normalizing_2021} for a detailed overview.

We use \textit{neural spline flows} \citep{durkan_neural_2019} and \textit{masked autoregressive flows} \citep{papamakarios_masked_2017} for the simulation of angular variables, and abbreviate these approaches to NFNSF and NFMAF, respectively. NFNSFs give the transformation $T$ as a composition of so-called \textit{coupling transforms} and spline-based transforms using monotonic rational quadratic splines.  NFMAFs, on the other hand, compose the transformation $T$ as a sequence of invertible autoregressive transformations. Additional details are given in Appendix \ref{app:normflows}.

We again model the $(d-1)$-angular vector $\boldmath{\Theta}$, as this simplifies the definition of a diffeomorphism compared to defining normalizing flows directly on $\mathbb{S}^{d-1}$. We use a Gaussian distribution function transformation after the final NFNSF and NFMAF transformation to account for the bounded nature of the first $d-2$ angular variables. For the final angle $\theta_{d-1}$, we use a modulo transformation  to incorporate the fact that this variable is cyclic on $(-\pi, \pi]$, i.e., 
\begin{align*}
    \theta_i \leftarrow \pi \cdot \Phi(x_i) \text{ for } i = 1, \dots, d-2, \qquad\text{and}\qquad
    \theta_{d-1} \leftarrow (x_{{d-1}} \text{ mod } 2\pi) - \pi.
\end{align*}
A sigmoid activation, as used for the GANs (see Section~\ref{subsec:gan}) would also be possible for the first $d-2$ angular variables; however, we found that compared to the Gaussian distribution function, the sigmoid's slower convergence to zero and one at the lower and upper limits can result in poor approximations of the angular variable near the endpoints of the interval $[0,\pi]$. Our choice to circularly wrap the last angular coordinate follows \citet{rezende_normalizing_2020}. Analytic gradient expressions are available for the Gaussian distribution function, and the gradient of the modulo transformation is $1$ on $(-\pi, \pi)$. While the modulo function is not differentiable at $\pi$, we note that the left derivative still equals $1$, and thus we avoid issues when evaluating gradients in practice. This in turn allows us to calculate the log-likelihood using Eq.~\eqref{eq:change_of_variables}.

\subsection{Flow matching}\label{subsec:flow_matching}

Flow matching \citep[FM,][]{lipman_flow_2022} is a framework for generative modelling that has been successfully used in a variety of applications in recent years. It is a generalisation of the very popular class of diffusion models \citep{sohl-dickstein_deep_2015, ho_denoising_2020} which underlie many large-scale applications of generative AI methods, especially in image vision. For a detailed introduction to FM, we refer to \cite{lipman_flow_2024}.

In general terms, FM approximates a $(d-1)$-dimensional target distribution $p_\text{target}$ using a time-dependent velocity field $u(t, \cdot): [0,1] \times \mathbb{R}^{d-1} \to \mathbb{R}^{d-1}$. This velocity field is modelled using a neural network, specifically, a feedforward neural network taking input values in $[0,1] \times \mathbb{R}^{d-1}$ and outputting values in $\mathbb{R}^{d-1}$. The velocity field $u(t, \cdot)$ determines a flow $\Psi(t, \cdot): [0,1] \times \mathbb{R}^{d-1} \to \mathbb{R}^{d-1}$ via the ordinary differential equation
\begin{equation}
    \label{eq:fm_ode}
    \frac{\text{d}}{\text{d}t} \Psi(t, \boldsymbol{x}) = u\{t, \Psi(t, \boldsymbol{x})\}.
\end{equation}
The flow then gives a sequence of smooth transformations, known as \textit{probability paths}, between a base distribution $p_\text{base}$ and the target distribution $p_\text{target}$ as
\begin{equation*}
    \boldsymbol{X}_t := \Psi(t, \boldsymbol{X}_\text{base}), \qquad \boldsymbol{X}_\text{base} \sim p_\text{base},
\end{equation*}
with the target that $\boldsymbol{X}_1 = \Psi(1, \boldsymbol{X}_\text{base}) \sim p_\text{target}$ and under the constraint $\Psi(0, \boldsymbol{X}_\text{base}) \sim p_\text{base}$. Intuitively, FM defines a smooth mapping between the base and target distributions. This is achieved through use of a neural network to model the time-dependent velocity field, defining how much the base field changes at each time $t$. If one has trained such an FM model, samples that are similar to the target distribution can be obtained by solving the ODE in Eq.~\eqref{eq:fm_ode} until $t = 1$. 

The neural network-based velocity field $u(t, \cdot)$ is learned by approximating probability paths designed to interpolate between the base density and the target density. In practice, one common choice is the so-called \textit{conditional optimal transport}, with $\boldsymbol{X}_1 \sim p_\text{target}$, $\boldsymbol{X}_\text{base} \sim p_\text{base}$ and
\begin{equation*}
    \boldsymbol{X}_t = t\boldsymbol{X}_1 + (1-t) \boldsymbol{X}_{\text{base}}.
\end{equation*}
This defines a mapping between the base and target densities. The modelled velocity field $u(t, \cdot)$ is then regressed against the velocity field coming from the above probability path $u^\text{target}(t, \cdot)$, leading to the loss
\begin{equation}\label{eq:fm_loss}
    \mathcal{L} = \mathbb{E}_{t, \boldsymbol{X}_t} \left\Vert u(t, \boldsymbol{X}_t) - u^\text{target}(t, \boldsymbol{X}_t) \right\Vert^2 \text{ with } t\sim \text{Unif}[0,1].
\end{equation}
In practice, this loss is rarely tractable as $u^\text{target}(t, \cdot)$ is too complicated to compute. However, it simplifies strongly by conditioning on a single target leading to the so-called \textit{conditional flow-matching loss}. By minimizing this loss, parameter estimates can be obtained for the neural network representing the velocity field $u(t, \cdot)$.

Many different probability paths can be designed, leading to different FM models. In this work, we focus on the commonly-used conditional optimal transport above. FM has also been generalised to manifolds \citep{lou_neural_2020, mathieu_riemannian_2020, chen_flow_2023} by building flows that move along geodesic curves and using a Bregman divergence on the tangent plane of the manifold as flow matching loss. Such approaches can be adapted to the pseudo-angular vector under consideration, which lies on the hypersphere $\mathbb{S}^{d-1}$. Thus, unlike the other generative deep learning architectures, we do not map the hypersphere data into the space of angles, but rather define a flow directly on $\mathbb{S}^{d-1}$ that moves along great circles. For further mathematical details, we refer again to \cite{lipman_flow_2024}. 

\subsection{Mixture of von Mises--Fisher distributions}\label{subsec:vMFmix}

As a comparison to our proposed deep learning approaches, we consider a straightforward yet established method for flexible angular modelling. Specifically, for the reasons outlined below, we fit a mixture of von Mises--Fisher (vMF) distributions. The vMF distribution, which can be viewed as an extension of a Gaussian distribution in the circular setting, is the most widely-used parametric approach for modelling directional data in practice \citep{García–Portugués2013}. Furthermore, mixture distributions offer increased flexibility by allowing data to be modelled as a combination of multiple probability distributions, enabling one to capture complex structures that simple parametric models cannot represent; see \citet{Boldi2007} and \citet{MacDonald2011} for examples of applications in the extreme value context. Moreover, mixtures of vMF distributions have been successfully applied in a wide range of angular modelling settings, including clustering \citep{Banerjee2005}, image segmentation \citep{mcgraw2006segmentation} and radiation therapy \citep{bangert2010using}. \citet{mackay_deep_2024} also consider mixtures of vMF distributions in the context of the SPAR model.

Given \textit{concentration} $\kappa \geq 0$ and \textit{mean direction} $\boldsymbol{\mu} \in \mathbb{S}^{d-1}$ parameters, the density of the $d$-dimensional vMF distribution is given by 
\begin{equation} \label{eqn:vmf_dist}
    f_{vMF}(\boldsymbol{w} \mid \boldsymbol{\mu}, \kappa)=c_d(\kappa) e^{\kappa \boldsymbol{\mu}^T \boldsymbol{w}}, \; \; \boldsymbol{w} \in \mathbb{S}^{d-1},    
\end{equation}
where the normalising constant $c_d(\kappa)$ is given by 
$c_d(\kappa)=\kappa^{d / 2-1}/[(2 \pi)^{d / 2} I_{d / 2-1}(\kappa)],$
and $I_r(\cdot)$ is the modified Bessel function of the first kind and order $r$. The concentration parameter $\kappa$ indicates how much unit vectors drawn from the vMF distribution are concentrated about the mean direction $\boldsymbol{\mu}$. Furthermore, $\boldsymbol{\mu}^T \boldsymbol{w} \in [-1,1]$ in Eq.~\eqref{eqn:vmf_dist} is the cosine similarity between $\boldsymbol{\mu}$ and $\boldsymbol{w}$; see Section~\ref{subsec:cCRPS}. 

Let $\boldsymbol{\pi} \in \mathbb{R}^K$ be a vector of $K$ mixture probabilities, i.e., $\pi_k \in [0,1]$ for all $k = 1,\hdots,K$ and $\sum_{k=1}^K \pi_k = 1$. Then a mixture model of vMF distributions is given by 
\begin{equation} \label{eqn:vmf_mix_dens}
h(\boldsymbol{w} \mid \boldsymbol{\Lambda})=\sum_{k=1}^K \pi_k f_{vMF}\left(\boldsymbol{w} \mid \boldsymbol{\mu}_k,\kappa_k\right),
\end{equation}
where $h(\cdot \mid \cdot)$ denotes the mixture density, $\boldsymbol{\Lambda} = \{\boldsymbol{\mu}_1,\kappa_1,\hdots,\boldsymbol{\mu}_K,\kappa_K \}$, and $(\boldsymbol{\mu}_k,\kappa_k)$ are the parameters associated with mixture component $k$. In practice, given a fixed number of mixture components $K$, Eq.~\eqref{eqn:vmf_mix_dens} can be fitted using an expectation-maximisation (EM) algorithm for maximum likelihood estimation of the parameter vector $\boldsymbol{\Lambda}$ alongside the mixture probabilities $\boldsymbol{\pi}$. Moreover, $K$ can be selected as the value that optimises some metric of the resulting mixture model; a common choice is the Bayesian information criterion (BIC). We refer to \citet{hornik_movmf_2014} for a detailed overview of vMF mixture distributions and the EM algorithm for model fitting.

\subsection{Hyperparameter optimisation and implementation details}\label{subsec:hyperparam}

We conclude this section with a discussion of hyperparameter selection in the approaches we consider, and details on how they are implemented. Generative deep learning methods include many different hyperparameters, including the number of neural network layers (or number of flows for NFNSF and NFMAF), the number of neurons in each layer, training length, learning rates for optimisers and many more. These are usually set based on empirical performance rather than theoretical considerations. We optimised architectures for one of the simulation examples (the sparse Gaussian copula with double Pareto marginal distributions; see Section~\ref{sec:simulation}), as well as the wave data set studied in Section~\ref{sec:waves}, selecting neural network architectures with strong empirical performance across these instances. As hyperparameter optimisation can be very computationally expensive, optimising architectures for each individual case is infeasible. However, we found that differences within model performances were negligible given hyperparameter selections that were chosen to allow for some reasonable flexibility.

For the GAN generator and discriminator, we used neural networks containing 4 hidden layers with 128 units each. They are trained for a fixed number of $1\ 000$ epochs, which we find works well across examples. Similarly, the FM models also have 4 hidden layers with 128 hidden units and Swish activations \citep{ramachandran_swish_2017}. Unlike the GAN models, for the FM and normalizing flows, we split the training data into an 80\% training data set and 20\% validation data set. We use the validation data set to determine the training length, meaning that we evaluate the FM or normalizing flow loss on the validation data set and stop the training when no improvement can be seen for a certain number of epochs (the so-called \textit{patience}). This is known as \textit{early stopping} and is a common technique for regularisation in neural network training \citep[][Chapter 7.8]{Goodfellow-et-al-2016}. Early stopping is possible for the FM, NFNSF and NFMAF models, but not for the GAN models as the adversarial training nature makes the losses non-interpretable \citep{goodfellow_generative_2014}. We train the FM models to a maximum of $5\ 000$ epochs, with a fairly large patience of 500 epochs. Finally, for the normalizing flows, we find that 8 transforms work well for NFMAF, and 10 for NFNSF. We train both to up to 500 epochs, with a patience of 50. For all models, we use an Adam optimiser \citep{kingma_adam_2017} with a learning rate of $10^{-4}$. Moreover, we use minibatching with a batch size of 256 training samples. 

The base density $p_{\text{base}}$ used in all deep learning methods is often taken to be a simple distribution that is easy to evaluate and sample from, and this choice is often thought to be of minor importance. However, a number of recent works have pointed out that for standard generative deep learning applications, under weak conditions on the neural network mapping, the base distribution determines both tail heaviness and extremal dependence structures \citep{wiese_copula_2019, huster_pareto_2021, Lafon2023}. This has motivated further work exploring more flexible base distributions or tail adaptive neural network mappings \citep[see][]{Allouche2022, mcdonald_comet_2022}. In the present work, however, our focus is on modelling bounded, hyperspherical data, for which tail behaviour is largely irrelevant. This somewhat simplifies the modelling task, and we consequently find that the base distribution is of minor importance. We use a standard Gaussian for the GAN, NFMAF and FM models, and a $\text{Unif}[-\pi, \pi]$ distribution for NFNSF.

For implementing the deep learning methods, we use \texttt{pytorch} \citep{paszke_pytorch_2019}. We use the \texttt{zuko} package \citep{rozet2022zuko} to support implementation of the normalizing flows and the \texttt{flow\_matching} \citep{lipman_flow_2024} package for the FM models. The GAN models were implemented by the authors directly in \texttt{pytorch}. Once each of the deep learning approaches has been trained, it is straightforward to generate large samples of angular observations. Code with the implementation of all methods, alongside example notebooks, is available as supplementary material. For smaller data sizes ($\sim$ $1\ 000$ samples), the training of each neural network model takes less than a couple of minutes, while for medium ($\sim$ $10\ 000$ samples) and large ($\sim$ $100\ 000$ samples) sized  data sets, training times are in the order of 10 minutes and 1--2 hours, respectively, on a T4 GPU. Training is slightly slower, but not infeasibly so, on a CPU.

We use the R \texttt{movmf} package \citep{hornik_movmf_2014} for fitting the mixture of vMF distributions. We allowed the number of mixture components to vary for each of the copula examples introduced in Section~\ref{sec:simulation} and used BIC values to select the `optimum' number in each case. Comparing the selected $K$ values, we found that fixing $K=100$ appeared to offer a reasonable trade-off (i.e., enough flexibility without being over-parameterised) for capturing the wide range of dependence structures across two dimension sizes ($d = 5$ and $d=10$). As for the neural networks, the architecture could be optimised for each individual case (e.g., using BIC minimisation), but this would be computationally expensive in general and bespoke optimisation would give the vMF mixture model an unfair advantage over the deep learning approaches. Therefore, for the sake of comparison, we keep the tuning parameters fixed for each of the considered angular simulation approaches. Furthermore, for obtaining vMF parameter estimates, we run the EM algorithm for 10 iterations, as suggested by default in \citet{hornik_movmf_2014}; this appeared sufficient to achieve convergence for all parameters without too large a computational cost. 

\section{Evaluation}\label{sec:evaluation}

To evaluate the performance of each of the proposed modelling frameworks, we consider a variety of visual and numeric metrics. These metrics can be used to assess how well each model approximates a given set of observed angular variables. In Section~\ref{subsec:cCRPS}, we introduce a numerical metric that quantifies how well an estimated distribution represents the true, or observed, pseudo-angular process. In Section~\ref{subsec:visual_gof}, we introduce a range of visual goodness of fit metrics; namely, QQ and histogram plots for each of the marginal spherical angles, dependency scatterplots between each pair of spherical angles, and a probability comparison plots that summarises the dependence structure over $\mathbb{S}^{d-1}$.  

\subsection{Angular energy score} \label{subsec:cCRPS}
In many applications, scoring rules are used to assess how well an estimated distribution represents observed data. In particular, the continuous ranked probability score (CRPS), usually referred to as the `energy score' (ES) in multivariate settings, is a popular scoring rule, especially in meteorological settings; see, e.g., \citet{candille2005evaluation} or \citet{gneiting_calibrated_2005}. However, these scoring rules are defined for data observed on the real line, which offers little use for observations taking values on the $(d-1)$-sphere. To account for this limitation, in the $d = 2$ case, \citet{Grimit2006} introduced an angular analogue to the CRPS, termed the \textit{circular CRPS} (cCRPS). We now propose a natural multivariate extension of the cCRPS, which we term the \textit{angular energy score} (aES).\\

\begin{definition}\label{def:aES}
Given a (fitted) distribution $F'$ on $\mathbb{S}^{d-1}$ and a fixed angle $\boldsymbol{v} \in \mathbb{S}^{d-1}$, the angular energy score is given by
$$
\operatorname{aES}(F', \boldsymbol{v})=\mathbb{E}_{\boldsymbol{W} \sim F'}\{\alpha(\boldsymbol{W}, \boldsymbol{v})\}-\frac{1}{2} \mathbb{E}_{\boldsymbol{W}, \boldsymbol{W}^{'} \sim F'}\left\{\alpha\left(\boldsymbol{W}, \boldsymbol{W}^{'}\right)\right\},
$$
where $\alpha(\cdot,\cdot)$ is the angular distance between two points on $\mathbb{S}^{d-1}$, i.e., $\alpha(\boldsymbol{x},\boldsymbol{y}) = \cos^{-1}(\boldsymbol{x}^T\boldsymbol{y})$, and $\boldsymbol{W}$ and $\boldsymbol{W}'$ denote independent copies of the angular random variable with distribution $F'$. 
\end{definition}

\vspace{0.25cm}\begin{proposition}\label{prop:aES}
The angular energy score is a \textit{proper} scoring rule, relative to the set of probability measures $F$ with finite entropy, i.e., $\mathbb{H}_\alpha (F) := \tfrac{1}{2}\int_{\mathbb{S}^{d-1}}\int_{\mathbb{S}^{d-1}} \alpha(\boldsymbol{w},\boldsymbol{w}')\mathrm{d}F(\boldsymbol{w})\mathrm{d}F(\boldsymbol{w}') < \infty$.
\end{proposition}

We provide a proof of Proposition~\ref{prop:aES} in Appendix~\ref{app:aES}. Propriety of the aES implies that it will tend to favour honest \citep[i.e., there is no gain from hedging, see][]{garthwaite_statistical_2005} and sharp (informative) generative distributions. Furthermore, the aES provides a numerical means of quantifying how much a fitted model agrees with the observed distribution.

On its own, the aES can only be used to assess the quality of a model fit at some fixed angle $\boldsymbol{v} \in \mathbb{S}^{d-1}$; therefore, as a summary metric, we compute the expected aES over the `true' angular distribution $F$, i.e.,
\begin{equation*}
    \operatorname{aES}_{F, F'} = \mathbb{E}_{\boldsymbol{V} \sim F}\left\{\operatorname{aES}(F', \boldsymbol{V})\right\},
\end{equation*}
where $\boldsymbol{V}$ denotes a circular random vector with distribution $F$. This corresponds to the induced probability divergence between the fitted distribution $F'$ and the true distribution $F$. Here, smaller values of $\operatorname{aES}_{F, F'}$ indicate superior predictive performance for $F'$ given the observed distribution $F$.

In practice, closed-form expressions of $F$, and possibly $F'$, are often unavailable. However, given large samples $\boldsymbol{w}_1,\hdots,\boldsymbol{w}_M$ and $\boldsymbol{v}_1,\hdots,\boldsymbol{v}_N$ from $F'$ and $F$, respectively, we can estimate $\operatorname{aES}_{F, F'}$ using Monte Carlo techniques. First, we approximate $\operatorname{aES}(F', \boldsymbol{v})$ as   
$$
\widehat{\operatorname{aES}}\left(F', \boldsymbol{v}\right)=\frac{1}{M} \sum_{m=1}^M \alpha\left(\boldsymbol{w}_m, \boldsymbol{v}\right)-\frac{1}{2 M^2} \sum_{m=1}^M \sum_{r=1}^M \alpha\left(\boldsymbol{w}_m, \boldsymbol{w}_r\right).
$$
From this, we can obtain an estimator of $\operatorname{aES}_{F, F'}$ as 
\begin{equation*}
    \widehat{\operatorname{aES}}_{F, F'} = \frac{1}{N}\sum_{n=1}^N\widehat{\operatorname{aES}}\left(F', \boldsymbol{v}_n\right) 
\end{equation*}
In practice, we set $M=N= 100\ 000$; these values were found to result in negligible estimation uncertainty, giving us a robust metric for comparison. We use the estimates of $\widehat{\operatorname{aES}}_{F, F'}$ as a means of comparing simulations from each of the techniques introduced in Section~\ref{sec:methods}. Moreover, since part of our goal is to assess the feasibility of novel deep generative approaches against existing techniques for angular modelling, we compute skill ratios of the form
\begin{equation}\label{eqn:skill}
    \operatorname{Skill}(F_\text{new}) = \frac{\widehat{\operatorname{aES}}_{F_\text{new}, F}}{\widehat{\operatorname{aES}}_{F_{\text{base}}, F}},
\end{equation}
where $F_\text{new}$ and $F_\text{base}$ correspond to new and baseline fitted distributions. Skill values smaller than one indicate lower expected $\operatorname{aES}$ of the model $F_\text{new}$ compared to the baseline $F_\text{base}$ (and vice versa). 

\subsection{Visual goodness of fit metrics} \label{subsec:visual_gof}
Reducing performance evaluation to a single numerical metric, as in the case of $\operatorname{aES}_{F, F'}$, can sometimes be an over-simplification and hide important structure. We therefore support such numerical results with visual goodness of fit checks. We note that one can choose to apply diagnostics on either of the angular vectors, $\boldsymbol{\Theta}$ and $\boldsymbol{W}$. In many cases, the former is more convenient, owing to the spherical angles not being co-linear. However, there can also be advantages to performing goodness of fits tests directly on $\mathbb{S}^{d-1}$, particularly when evaluating the extremal dependence structure of $\boldsymbol{X}$, since one can compare probability masses across different orthants and assess whether the data exhibits certain forms of dependence. As such, we propose diagnostics for both $\boldsymbol{\Theta}$ and $\boldsymbol{W}$. 

Simulated angular data can be evaluated in two different ways; exploring how well the generated data recreate the marginal distributions of each angular component, and investigating if the same data recreate the dependence structure observed on $\mathbb{S}^{d-1}$. To explore the marginal distributions, we consider each of the spherical angular variables of $\boldsymbol{\Theta}$ in turn. In this setting, standard goodness of fit tests can be applied; in particular, we compute quantile-quantile (QQ) and histogram plots using the observed and generated data sets. Specifically, given the spherical variable $\Theta_i, i \in \{1,\hdots,d-1\}$ and letting $\boldsymbol{\theta}^o_i = (\theta^o_{1,i}, \hdots, \theta^o_{n,i})$ and $\boldsymbol{\theta}^g_i = (\theta^g_{1,i}, \hdots, \theta^g_{m,i})$ denote an observed sample of size $n$ and a generated sample of size $m$, respectively, we compute (empirical) quantiles and histograms over both samples. We would expect good agreement between the observed and generated data sets for an accurate fitted model. 

When presenting figures in Sections~\ref{sec:simulation} and \ref{sec:waves}, we overlay the marginal quantile estimates to produce one unified QQ plot. For this, we scale the final spherical angle to sit in the interval $[0,\pi]$, i.e., $ (\Theta_{d-1} + \pi)/2$. Overlaying the quantile estimates in this manner reduces the number of overall plots, thus providing a more concise and less tedious approach for assessing model fits. 

To assess how well the generated observations recreate the angular dependence structure on $\mathbb{S}^{d-1}$, we consider two further diagnostics. Firstly, we consider simple pairwise dependency plots between each pair of spherical angles $(\theta_i,\theta_j)$, $i,j \in \{1,\hdots,d-1\}$, $i\neq j$. Overlaying observed and generated samples, one can see how well the pairwise dependencies are captured by the fitted model. In this setting, one can also identify areas of sparsity in the angular distribution, i.e., regions of $\mathbb{S}^{d-1}$ with zero probability mass. One could also produce pairwise scatterplots of the marginal, pseudo-angular variables, but such plots are less straightforward to interpret given the co-linearity that exists between the marginal variables.

For high dimensions, visualising the full, joint distribution on \(\mathbb{S}^{d-1}\) becomes infeasible. However, in the Cartesian space, we can compute univariate quantities that provide a summary of dependence features and sparsity, allowing one to summarise the joint dependence structure in a univariate plot. First, observe that \(\mathbb{S}^{d-1}\) can be divided into \(2^d\) orthants, defined by the signs of each component of the angular variable $\boldsymbol{W}$ (equivalently, $\boldsymbol{X}$). Define \(\mathbbm{1}_k(\boldsymbol{W})\) as an indicator function for the \(k\)th orthant, where \(k \in \{1, \dots, 2^d\}\), i.e.,
$\mathbbm{1}_k(\boldsymbol{W}) =
1 \text{ if } \boldsymbol{W} \text{ is in the } k\text{th orthant}$ and $ 
\mathbbm{1}_k(\boldsymbol{W}) = 0  \text{ otherwise}$. Furthermore, let \(p_k = \mathbb{P}(\boldsymbol{W} \text{ is in the } k\text{th orthant})\) $= \mathbb{P}\{\mathbbm{1}_k(\boldsymbol{W}) = 1\}$, such that $\sum_{k=1}^{2^d} p_k = 1.$ Assuming that we have $n$ independent and identically distributed samples of the angular variable $\boldsymbol{W}$, it follows that
\[
(N_1, \dots, N_{2^d}) \sim \text{Multinomial}\left(n; p_1, \dots, p_{2^d}\right),
\]
where \(N_k\) is the number of samples in the \(k\)th orthant, and \(\sum_{k=1}^{2^d} N_k = n\). Note that even in the case of dependent data, there still exists some finite, discrete distribution parameterised by the orthant probabilities. For a large enough sample size $n$, the probabilities $ p_1, p_2, \dots, p_{2^d}$ can be estimated empirically. We therefore propose comparing these empirical estimates for observed and generated samples. In particular, letting $(\hat{p}^o_1, \dots, \hat{p}^o_{2^d})$ and $(\hat{p}^g_1, \dots, \hat{p}^g_{2^d})$ denote observed and generated probability estimates, respectively, for a given angular variable, we compare the tuples $\{(\log(\hat{p}^o_k + 1), \log(\hat{p}^g_k+1)) : k = 1,\hdots,2^d \}$, with good agreement indicating the dependence features on $\mathbb{S}^{d-1}$ are being captured appropriately. Note that the transformed log-scale is used for comparison, owing to the fact that as $d$ grows, the probabilities of observing data in certain orthants can be very close (or equal) to zero, corresponding to sparse regions in $\mathbb{S}^{d-1}$ (or $\mathbb{R}^d$). As such, the log-scale provides a better domain for assessing and comparing the range of probability estimates. This diagnostic provides information about the dependence features of both $\boldsymbol{W}$ and $\boldsymbol{X}$, illustrating the intricate link between these random vectors, as discussed in Section~\ref{sec:intro}. 

Example applications of each evaluation metric are given in Sections~\ref{sec:simulation} and \ref{sec:waves}. We note that considering such a wide range of metrics, incorporating both the marginal and dependence distributions of the angular variable $\boldsymbol{W}$, ensures that our proposed simulation techniques are rigorously tested and, if selected for inference, can accurately recreate a variety of relevant features in the angular data. 

\section{Simulation study}\label{sec:simulation}

\subsection{Overview and simulation setup}\label{subsec:sim_setup}
In this section, we present a simulation study to investigate the performance of each of our proposed generative deep learning methods. As explained in Section~\ref{subsec:vMFmix}, these approaches are compared to a simple baseline model, exploiting a mixture of vMF distributions. To allow for a thorough comparison, we study a range of dependence structures (defined through their copulas), two different distributions for the common margins of $\boldsymbol{X}$, a range of sample sizes and two different dimensions. We now detail each of these choices, while the remainder of the section is dedicated to the results of our simulation study. 

First, we justify and introduce our choice of five different copula models. In the study of multivariate extremes, we are traditionally interested in the classification of \textit{asymptotic dependence} vs.\ \textit{asymptotic independence} \citep{Coles1999}, with copula models for $\boldsymbol{X}$ falling into one of the two categories depending on whether or not the variables can take their very largest values simultaneously. For a thorough study motivated by extreme-value applications, we should therefore consider copulas that cover each of these cases. In addition, we are interested in investigating model performance in the presence of sparsity, i.e., where regions of $\mathbb{S}^{d-1}$ do not contain any angular mass; this is a particularly important consideration as we move into higher-dimensional settings \citep[see][]{Engelke2021}. An ideal method would be able to capture any true sparsity in a data set, and work reliably well across a range of different dependence features in $\boldsymbol{X}$ and $\boldsymbol{W}$. With these points in mind, we select the following five copula models, noting that the numbers assigned in this list will be used later in the section to distinguish between the different choices: 
\begin{enumerate}
    \item A Gaussian copula with positive semi-definite correlation matrix $\boldsymbol{\Sigma}$.
    \item A mixture distribution, consisting of a Gaussian copula and a student-$t$ copula. Here, the same correlation matrix $\boldsymbol{\Sigma}$ is used for both copulas, and for the latter, we also specify the degrees of freedom parameter, defined on $\mathbb{R}_+$, as $\nu = 0.3$. We use mixing probabilities $\boldsymbol{\pi} = (0.5,0.5)$.  
    \item A logistic, or Gumbel, copula with dependence parameter $\alpha \in (0,1]$ \citep[see][]{Gumbel1960,Tawn1990}. We set the dependence parameter to $\alpha = 0.5$. 
    \item A mixture distribution, consisting of a logistic copula and an independence copula. For the former, we set $\alpha = 0.5$. We use mixing probabilities $\boldsymbol{\pi} = (0.5,0.5)$, noting that this is a special case of the asymmetric logistic copula \citep[see][]{Tawn1990}.
    \item A `sparse' Gaussian copula with positive semi-definite correlation matrix $\boldsymbol{\Sigma}_s$, with $\boldsymbol{\Sigma}_s$ specified such that the variables are clustered into groups of dependent variables with independence between groups. 
\end{enumerate}
For copulas $1$ and $2$, the same correlation matrix $\boldsymbol{\Sigma}$ is used; this is simulated randomly using the methodology given in \citet{makalic2022efficient} and kept fixed throughout the study. Moreover, we impose that the correlation matrices are ordered with respect to $d$. That is, when we consider two copulas with dimensions $5$ and $10$, then the $5 \times 5$ correlation matrix is a submatrix of the $10 \times 10$ correlation matrix. For copula $5$, a `sparse' correlation matrix is constructed by forming a block matrix from smaller correlation matrices, then computing the nearest positive definite matrix to this block matrix \citep[see][]{higham2002computing}, giving us a valid correlation matrix $\boldsymbol{\Sigma}_s$. This induces a sparse, clustered structure within the resulting random vector, and we again impose that the correlation matrix is ordered with respect to $d$. Visualisations of both correlations matrices $\boldsymbol{\Sigma}$ and $\boldsymbol{\Sigma}_s$ for $d = 10$ are included in Appendix~\ref{app:add_sim_study}.  

In terms of marginal models, for simplicity, and because this is standard practice in many multivariate extremes approaches, we choose a common distribution across all $d$ variables in $\boldsymbol{X}$. Motivated by the discussion in Section~\ref{sec:intro}, i.e., to avoid restricting ourselves to only the positive dependence setting, we consider marginal distributions with support on the full real line. Additionally, we choose one marginal distribution with light-tails, and another with heavy-tails, reflecting that both cases are commonly required in the study of multivariate extremes. For the light-tailed case, we use the standard Laplace distribution, also referred to as a \textit{double exponential}, having distribution function 
\[
F_L(x)=\begin{cases}
\frac{1}{2}e^x, & ~x \leq 0, \\
1-\frac{1}{2}e^{-x}, & ~x > 0.
\end{cases}
\]
For the heavy-tailed case, we take inspiration from the Laplace distribution and define what we term the \textit{double Pareto distribution}, constructed so that it behaves like a standard Pareto distribution in both tails; see also \citet{yamamoto2024deformation} for related literature. We note that the support of the standard Pareto distribution is $[1,\infty)$, so we shift this down by 1, before reflecting the standard Pareto density about 0 to construct our double Pareto model. This ensures that the support of the double Pareto distribution is on the full range $(-\infty,\infty)$, i.e., without a gap at $[-1,1]$. Our double Pareto distribution function is therefore given by
\[
F_{DP}(x) = \begin{cases}
\frac{1}{2(1-x)}, & ~x \leq 0, \\
1-\frac{1}{2(1+x)}, & ~x > 0.
\end{cases}
\]
Illustrations of the double Pareto distribution are provided in Appendix~\ref{app:add_sim_study}. For each combination of the five copula and two marginal models given above, we consider samples of size $n \in \{1\ 000, 10\ 000, 100\ 000\}$ and dimensions $d\in\{5,10\}$. This leads to a total of $5\times 2\times 3\times 2=60$ combinations, i.e., an extensive and rigorous setup for our tests. We stress here that while the models are trained on different sample sizes, all evaluation metrics introduced in Section~\ref{sec:evaluation} are computed against a large generated sample (with $n = m = 100\ 000$) from each theoretical copula and fitted distribution. This allows us to test whether the full angular distribution can be recovered from only a small sample, a feature that is desirable when employing deep modelling approaches in practice.  

\subsection{Results} \label{subsec:results}
Following Section~\ref{subsec:cCRPS}, we assess the performance of each deep learning approach using the improvement relative to our baseline, by considering the skill ratio in Eq.~\eqref{eqn:skill}. Fixing the dimension to $d=5$, the results for each copula, marginal distribution and sample size are given in Table~\ref{tab:cCRPS_d_5}; the corresponding results for $d=10$ are provided in Table~\ref{tab:cCRPS_d_10} of Appendix~\ref{app:add_sim_study}. The lowest score in each row (i.e., for each simulation scenario) is shown in bold.

\begin{figure}[!h]
    \centering
     \begin{subfigure}[b]{0.2\textwidth}
        \centering
        \includegraphics[width=\textwidth]{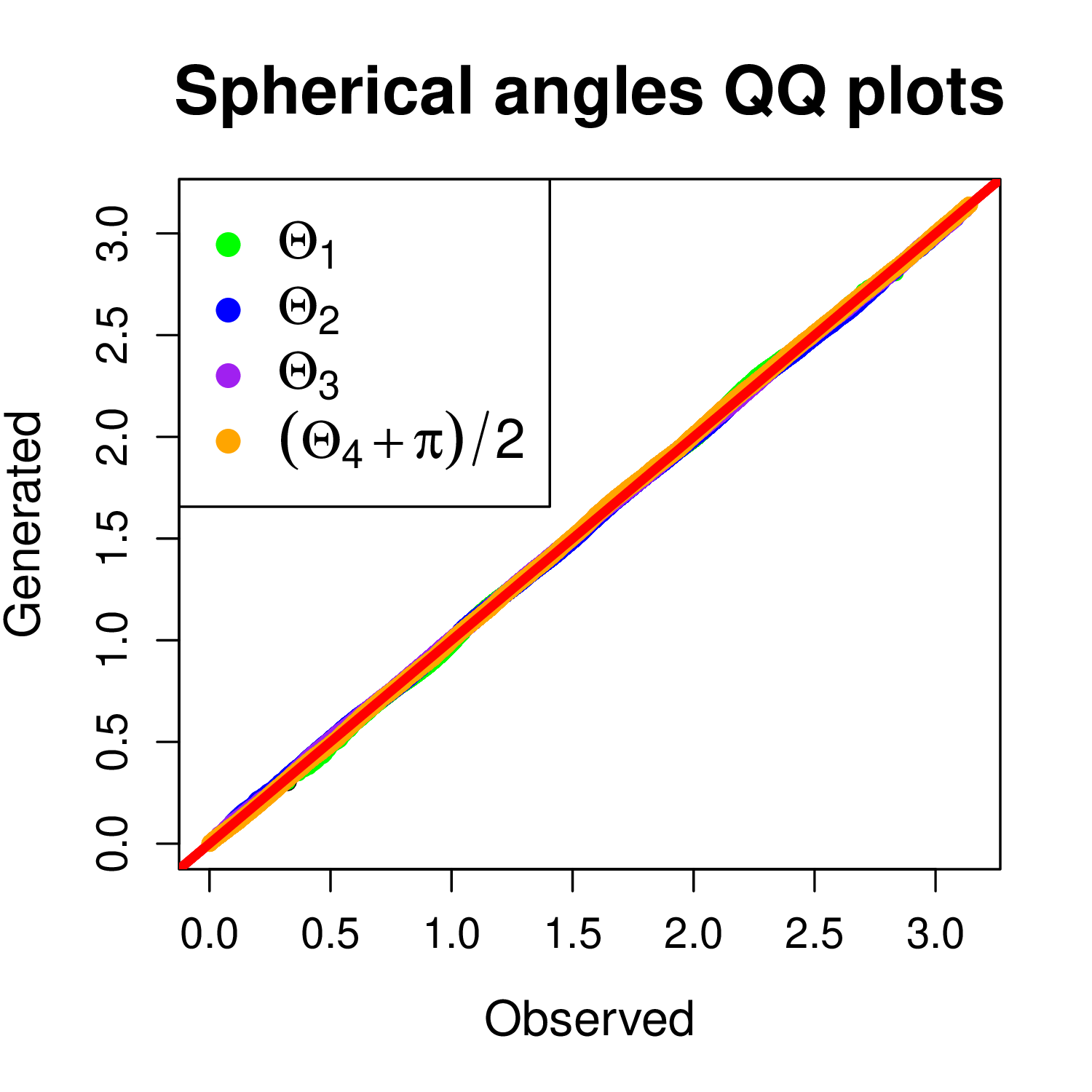}
    \end{subfigure}%
    \begin{subfigure}[b]{0.2\textwidth}
        \centering
        \includegraphics[width=\textwidth]{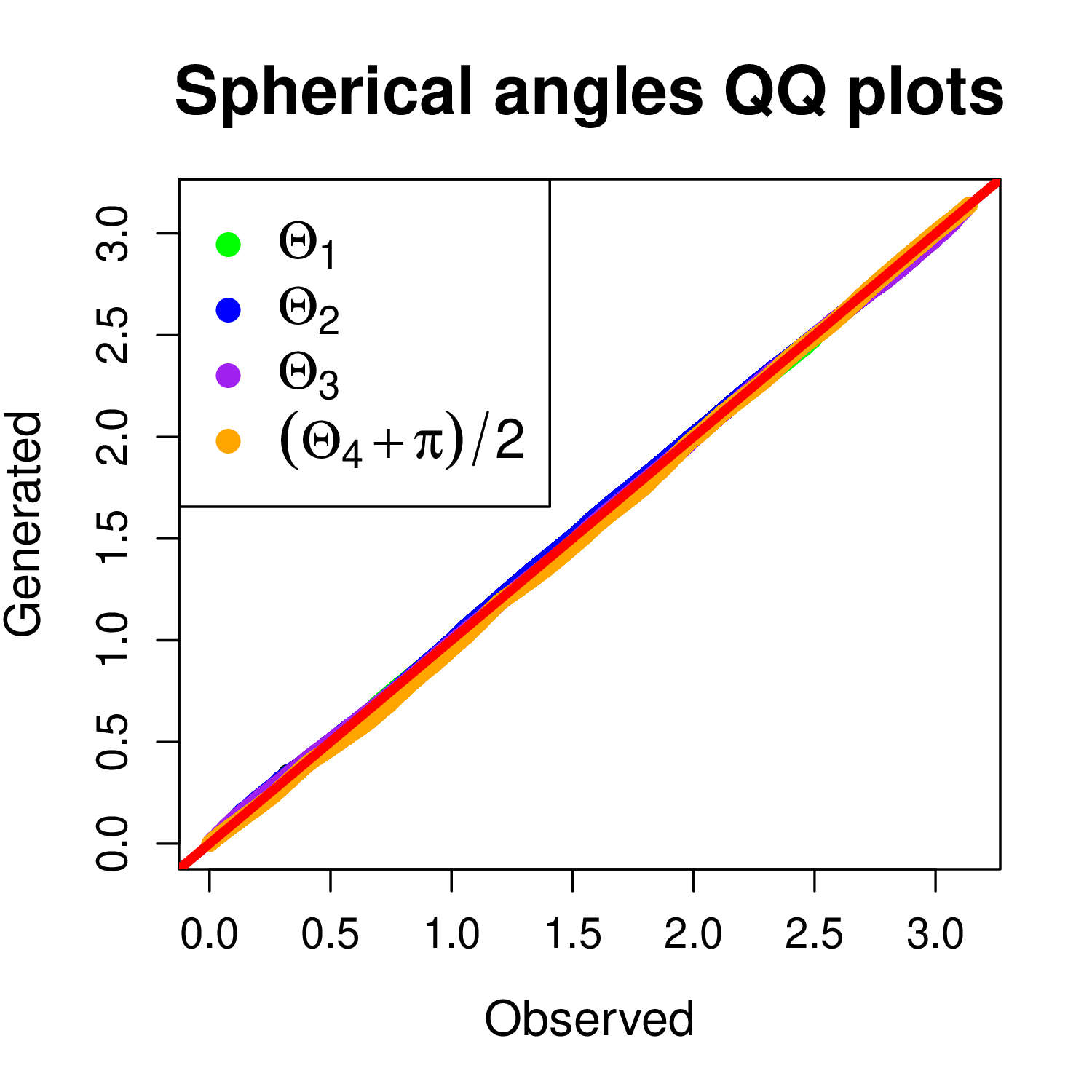}
    \end{subfigure}%
    \begin{subfigure}[b]{0.2\textwidth}
        \centering
        \includegraphics[width=\textwidth]{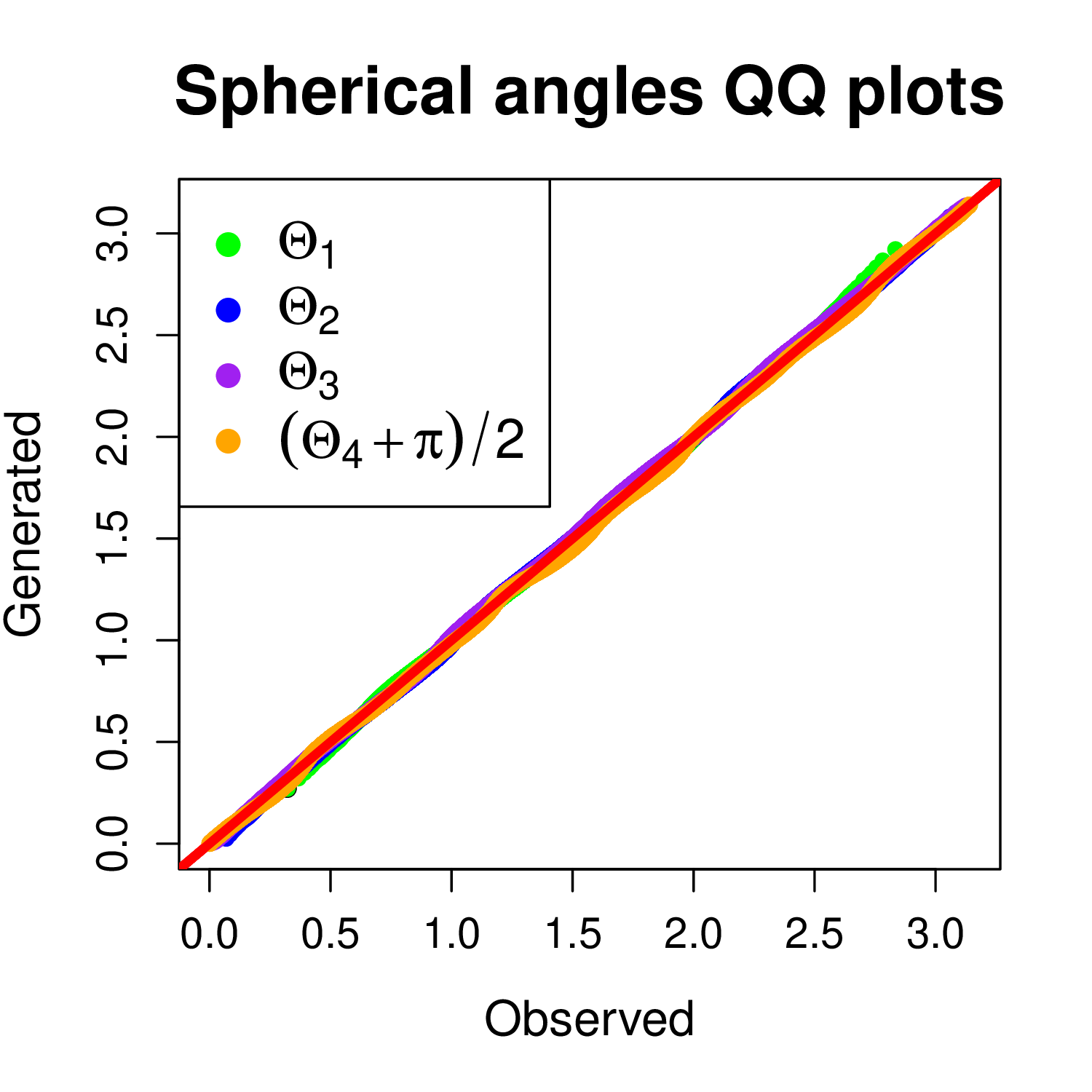}
    \end{subfigure}%
    \begin{subfigure}[b]{0.2\textwidth}
        \centering
        \includegraphics[width=\textwidth]{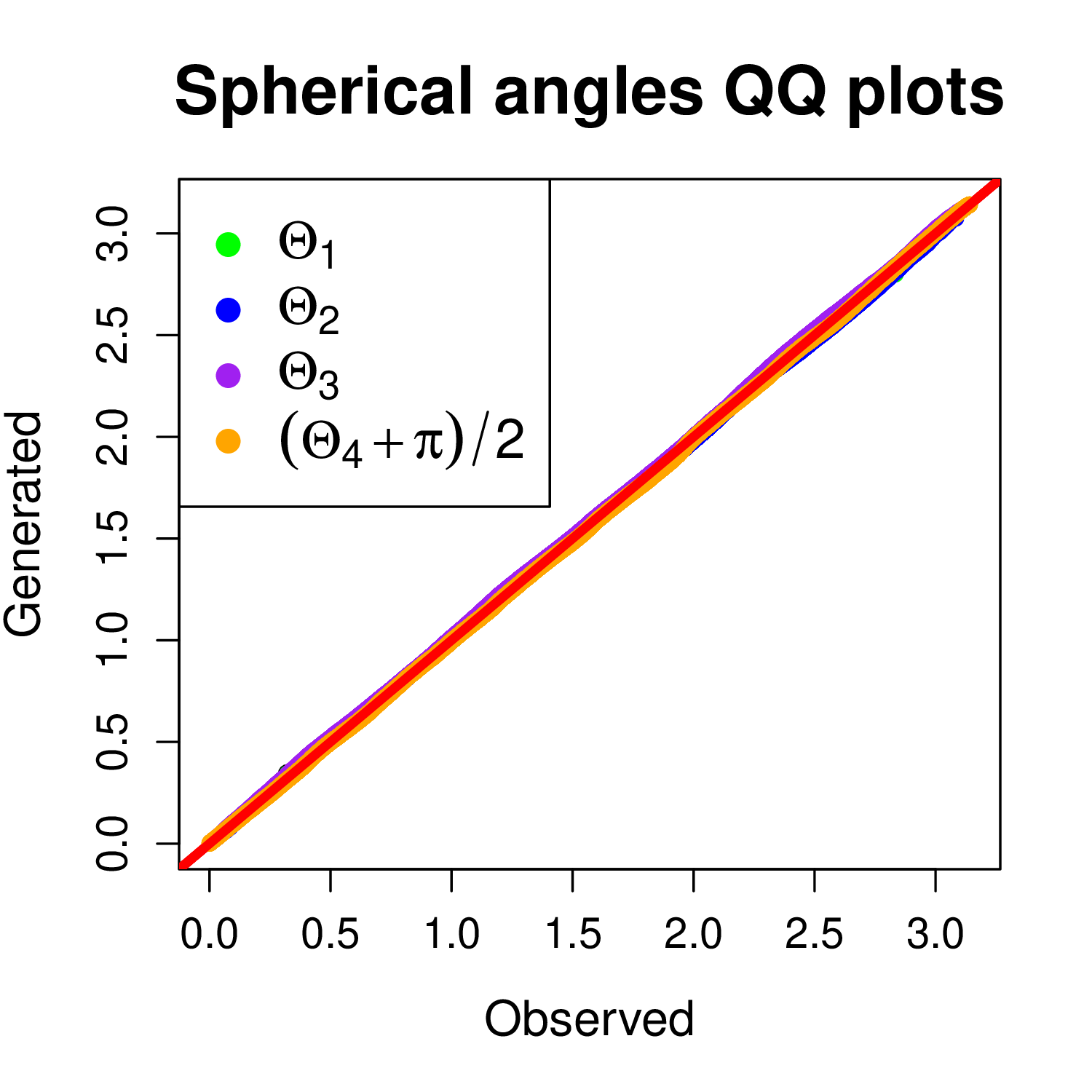}
    \end{subfigure}%
    \begin{subfigure}[b]{0.2\textwidth}
        \centering
        \includegraphics[width=\textwidth]{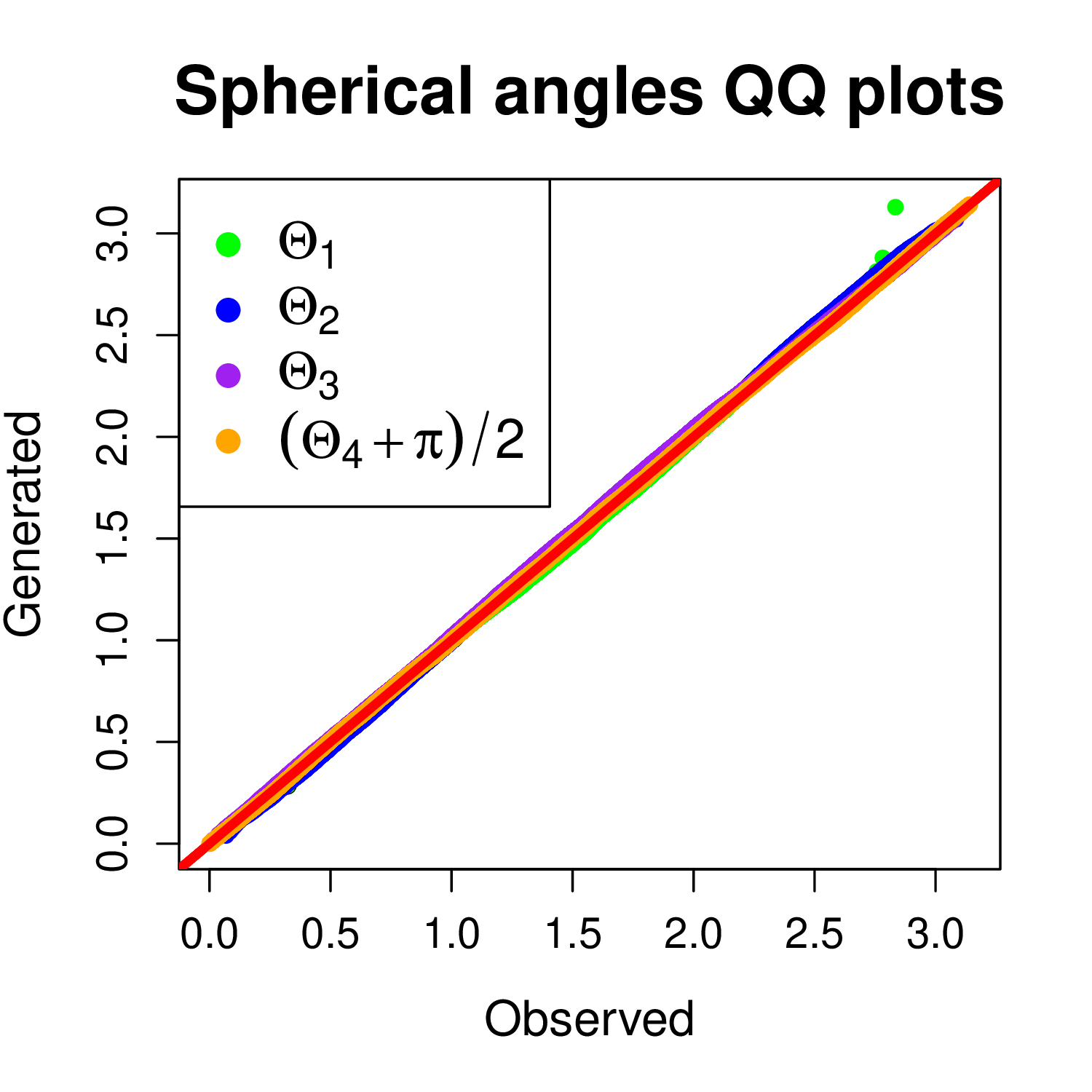}
    \end{subfigure}%

    \begin{subfigure}[b]{0.2\textwidth}
        \centering
        \includegraphics[width=\textwidth]{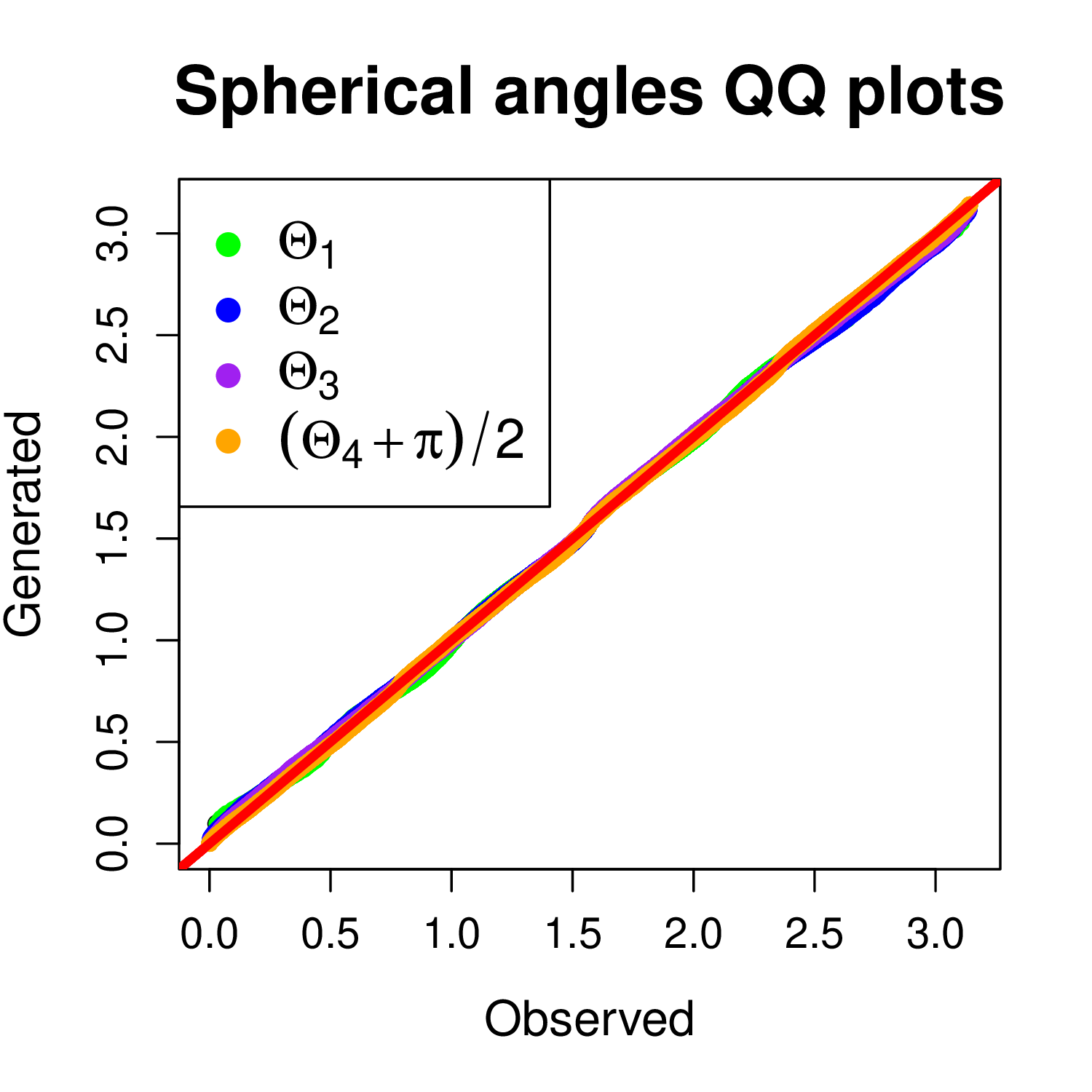}
    \end{subfigure}%
    \begin{subfigure}[b]{0.2\textwidth}
        \centering
        \includegraphics[width=\textwidth]{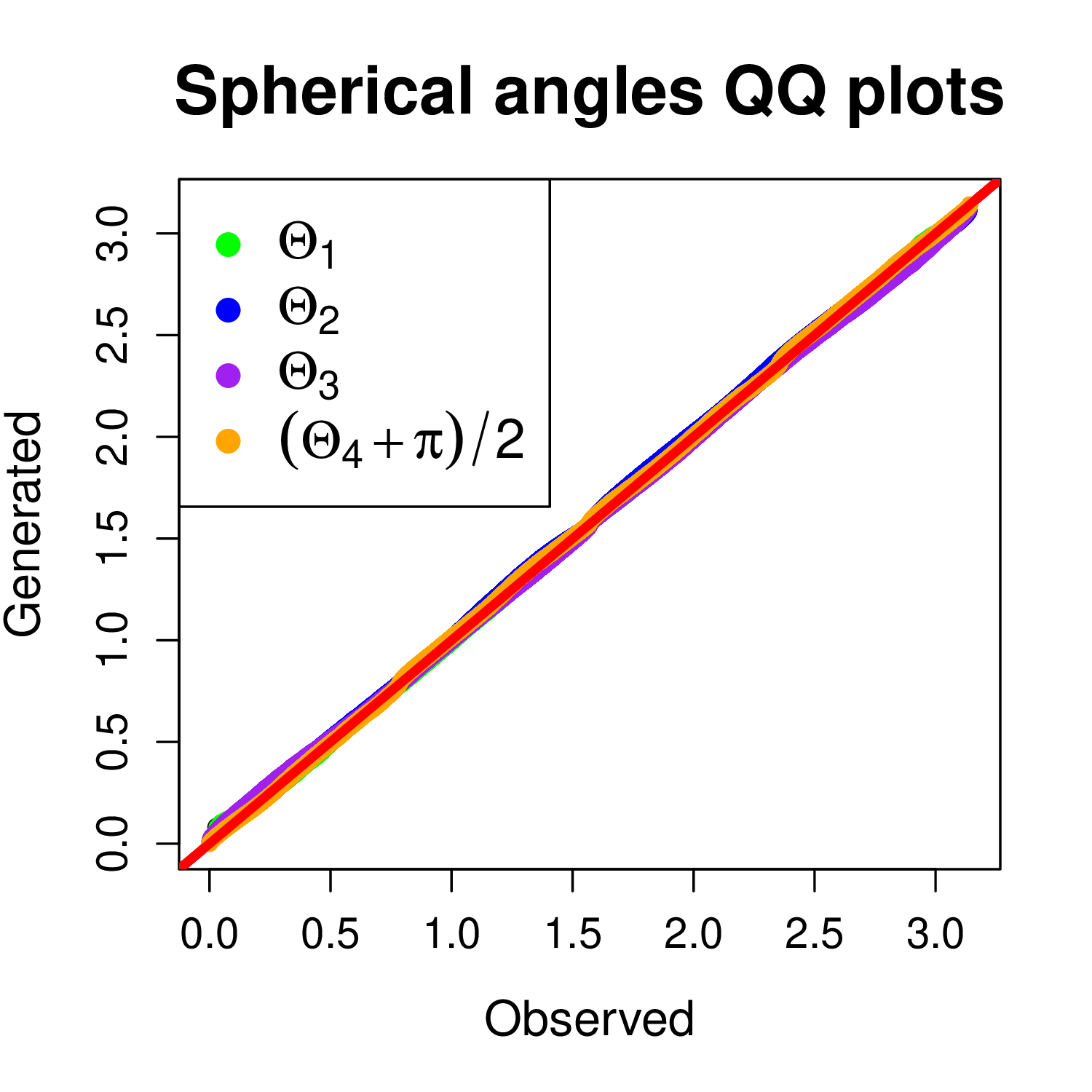}
    \end{subfigure}%
    \begin{subfigure}[b]{0.2\textwidth}
        \centering
        \includegraphics[width=\textwidth]{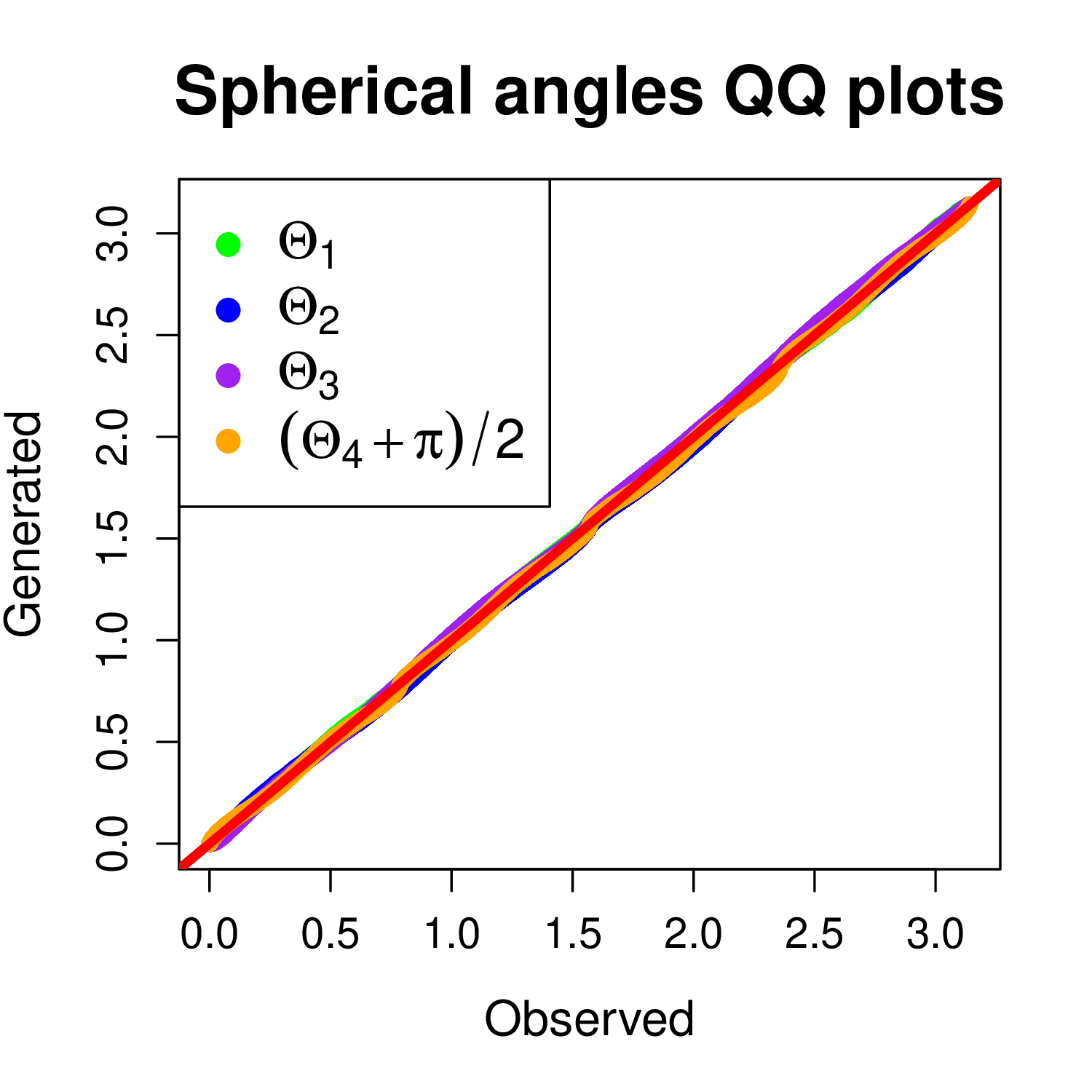}
    \end{subfigure}%
    \begin{subfigure}[b]{0.2\textwidth}
        \centering
        \includegraphics[width=\textwidth]{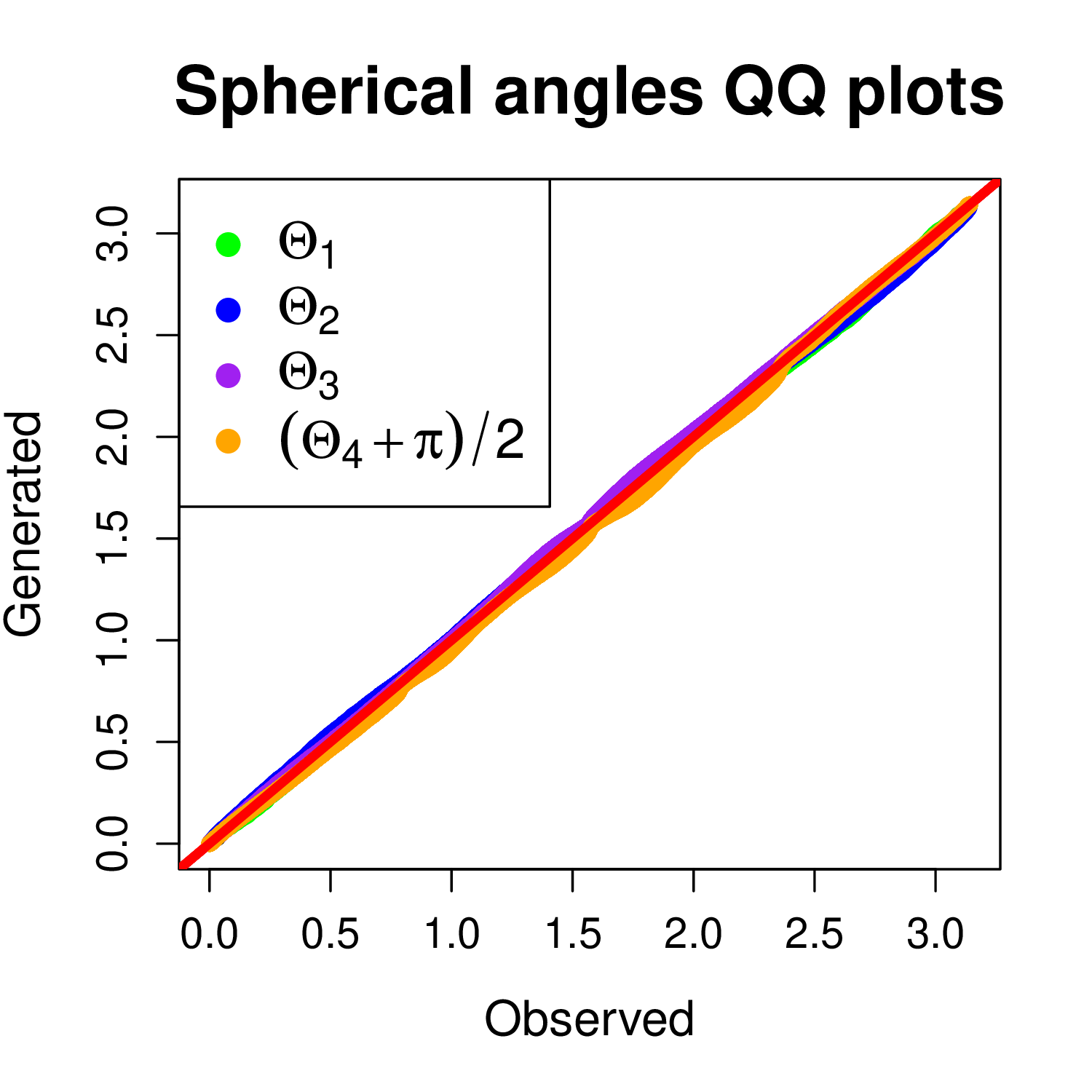}
    \end{subfigure}%
    \begin{subfigure}[b]{0.2\textwidth}
        \centering
        \includegraphics[width=\textwidth]{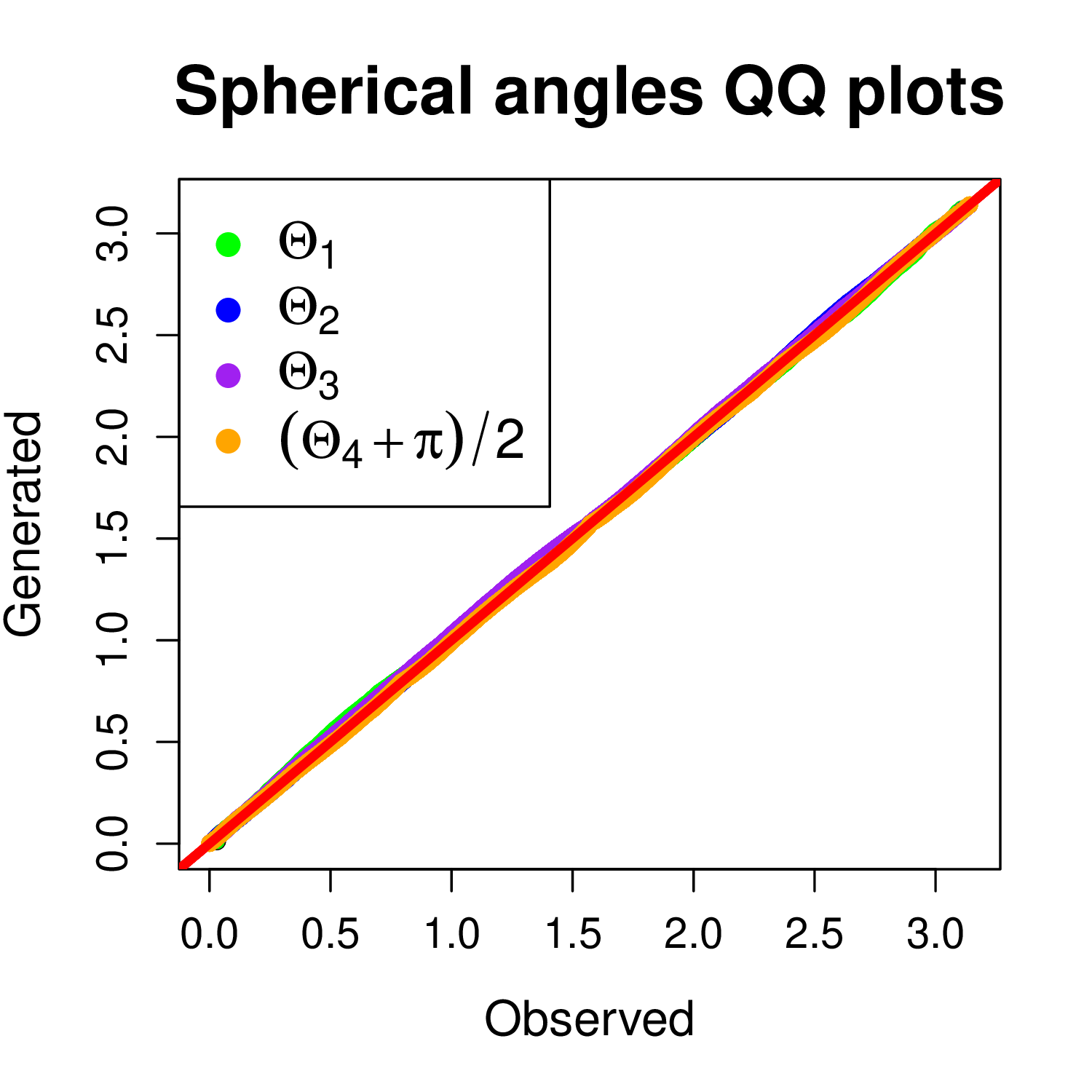}
    \end{subfigure}%
    \caption{Spherical angle QQ plots for copula 2 with $d=5$ and $n=100\ 000$. Top and bottom rows correspond to Laplace and double Pareto margins, respectively. The left, centre left, centre, centre right and right panels correspond to the vMF mixture, FM, NFMAF, GAN, and NFNSF approaches, respectively.}
    \label{fig:qq_plots_cop2_d5}

        \begin{subfigure}[b]{0.2\textwidth}
        \centering
        \includegraphics[width=\textwidth]{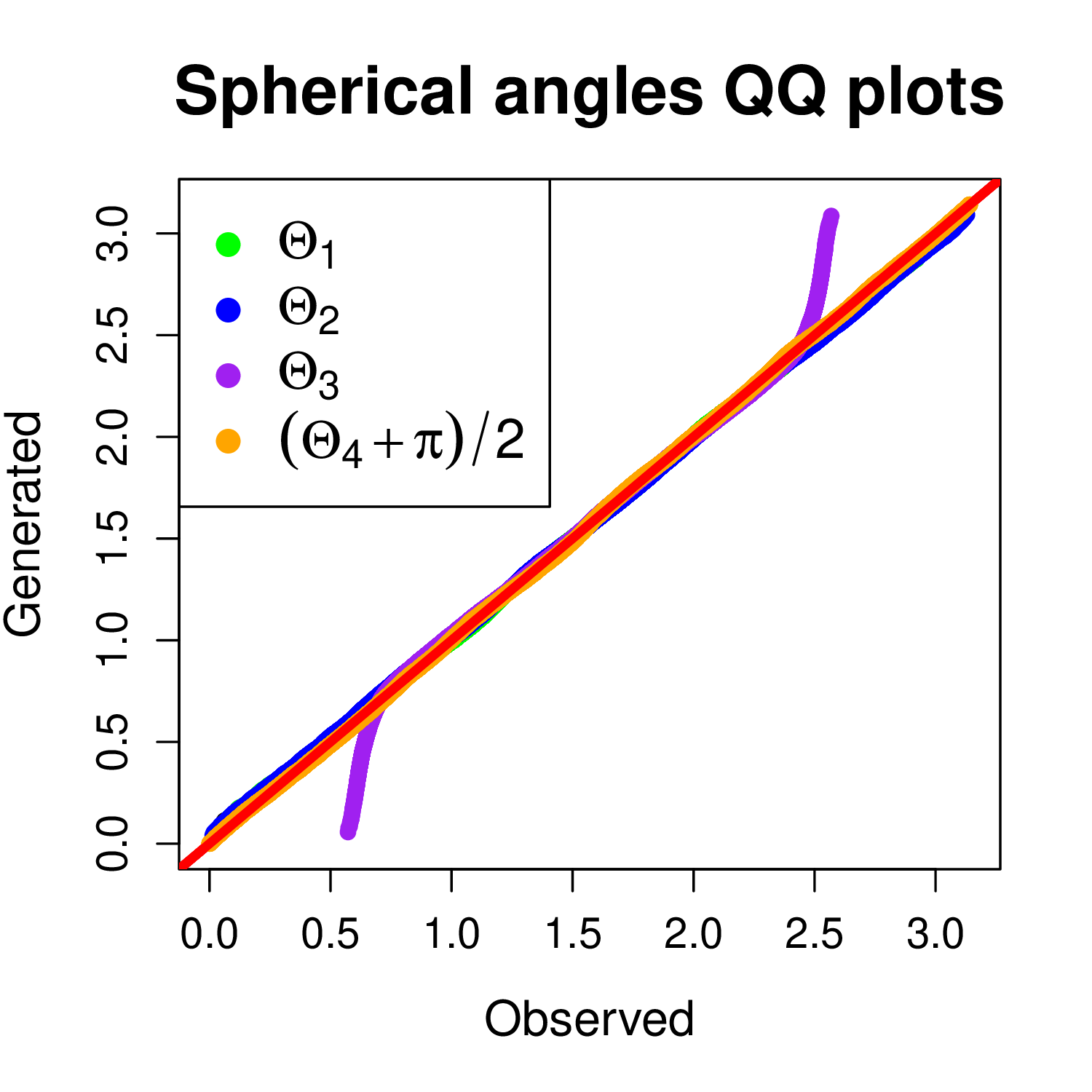}
    \end{subfigure}%
    \begin{subfigure}[b]{0.2\textwidth}
        \centering
        \includegraphics[width=\textwidth]{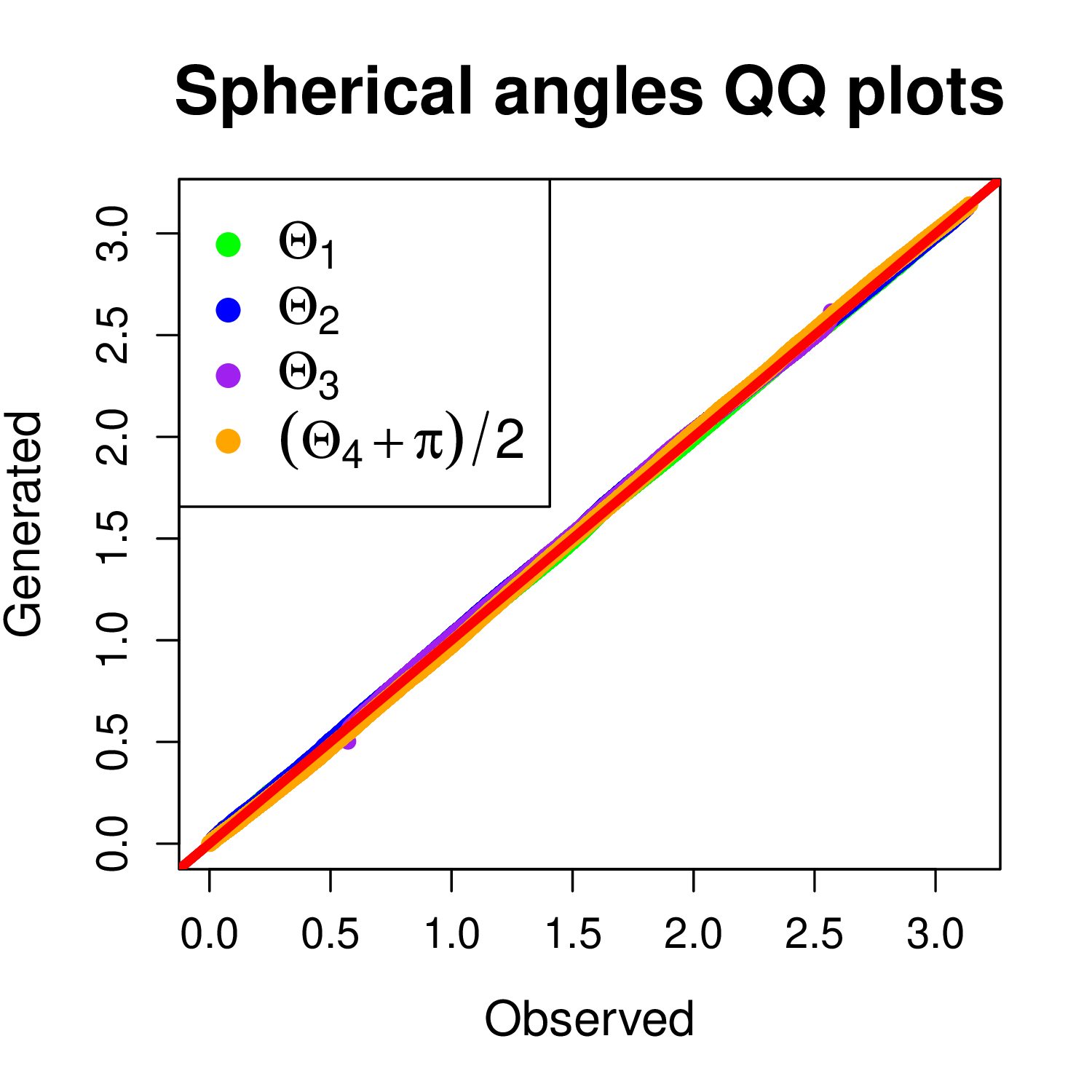}
    \end{subfigure}%
    \begin{subfigure}[b]{0.2\textwidth}
        \centering
        \includegraphics[width=\textwidth]{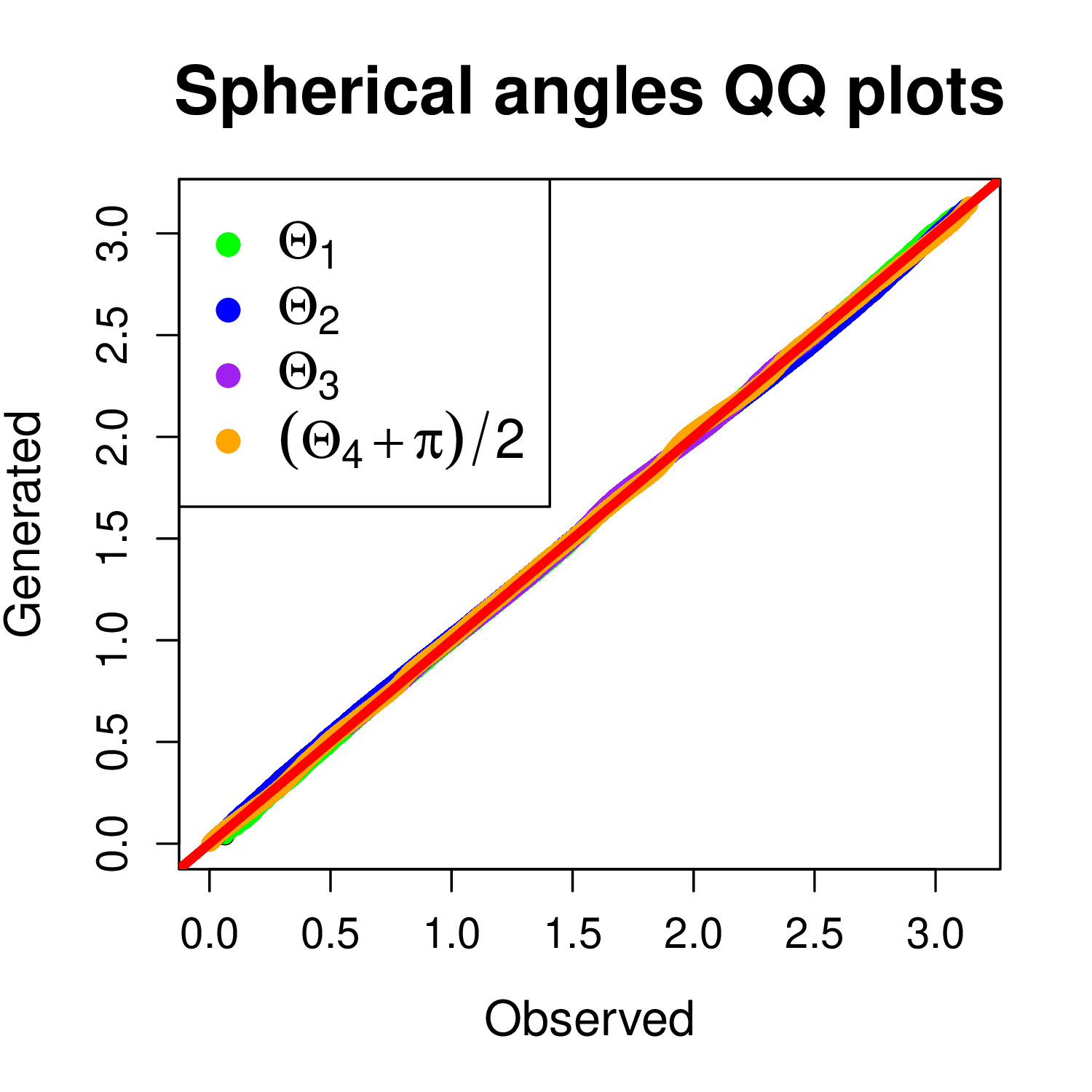}
    \end{subfigure}%
    \begin{subfigure}[b]{0.2\textwidth}
        \centering
        \includegraphics[width=\textwidth]{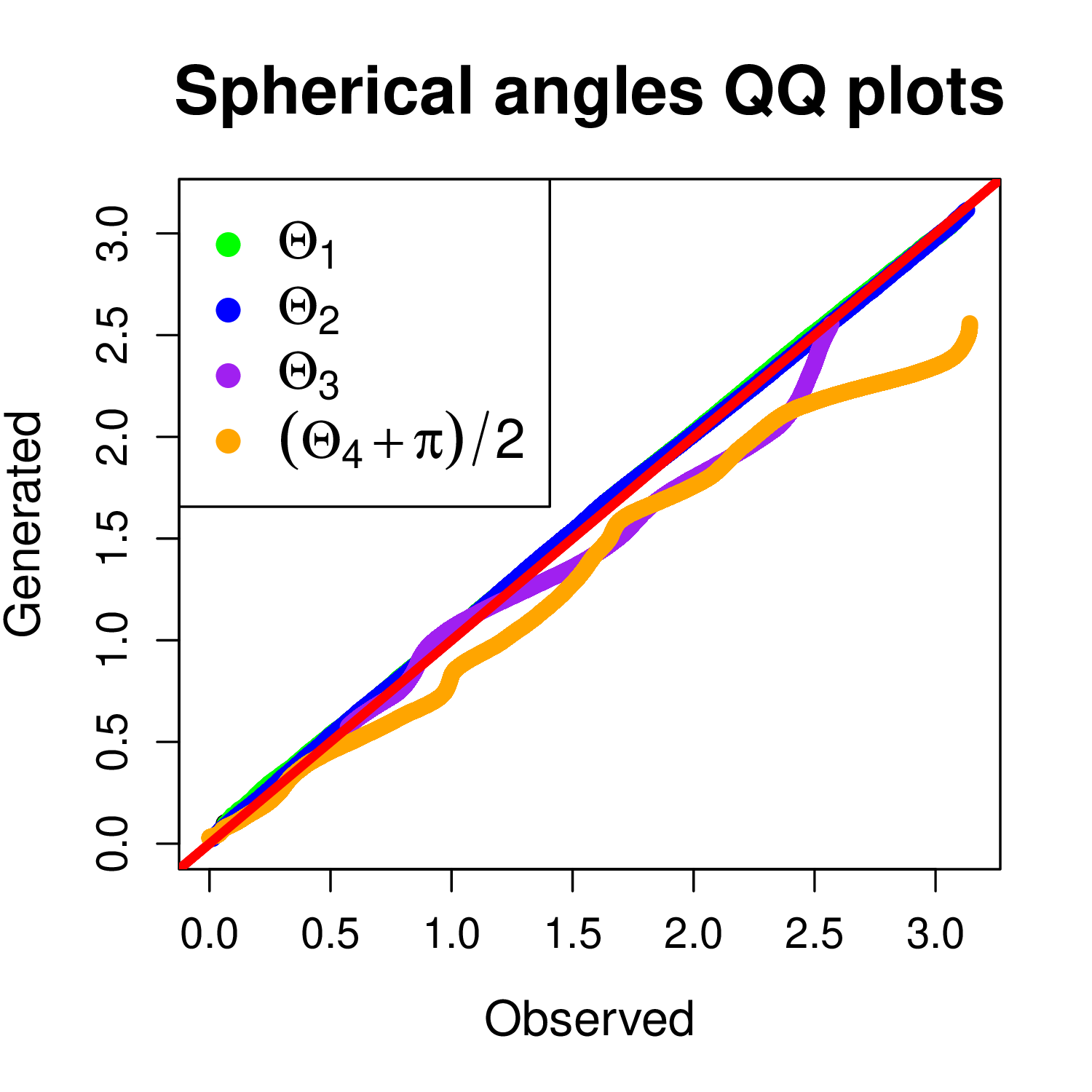}
    \end{subfigure}%
    \begin{subfigure}[b]{0.2\textwidth}
        \centering
        \includegraphics[width=\textwidth]{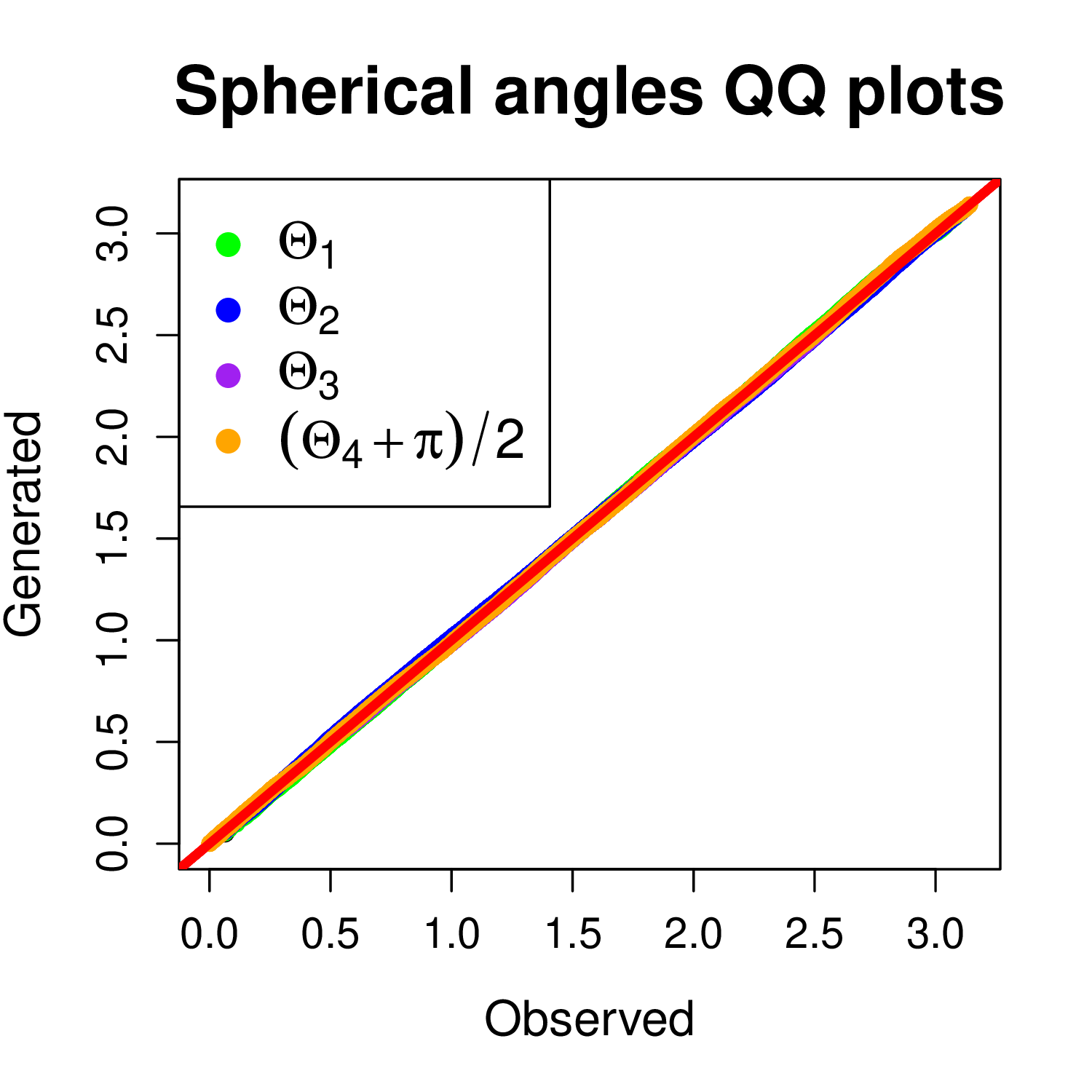}
    \end{subfigure}%

        \begin{subfigure}[b]{0.2\textwidth}
        \centering
        \includegraphics[width=\textwidth]{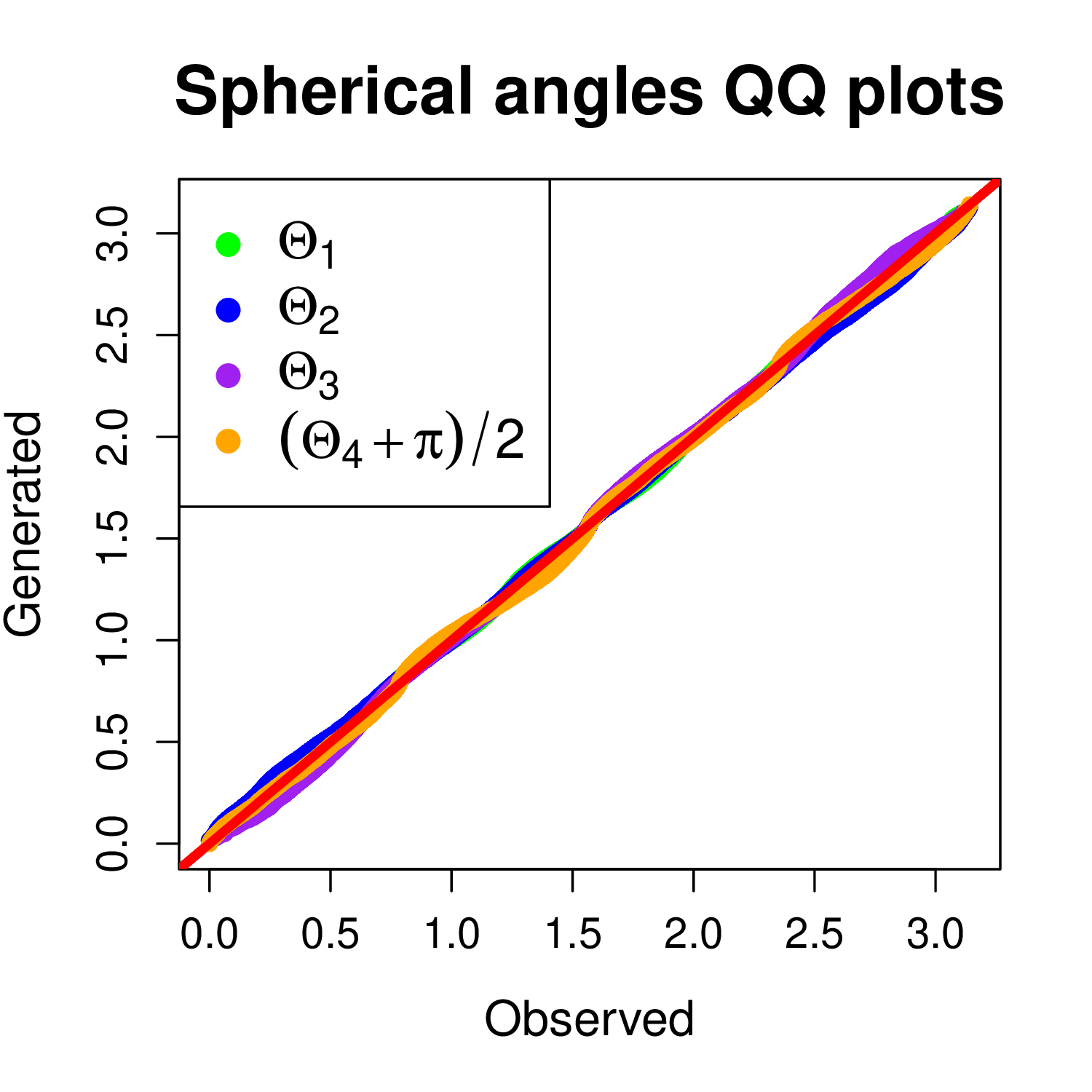}
    \end{subfigure}%
    \begin{subfigure}[b]{0.2\textwidth}
        \centering
        \includegraphics[width=\textwidth]{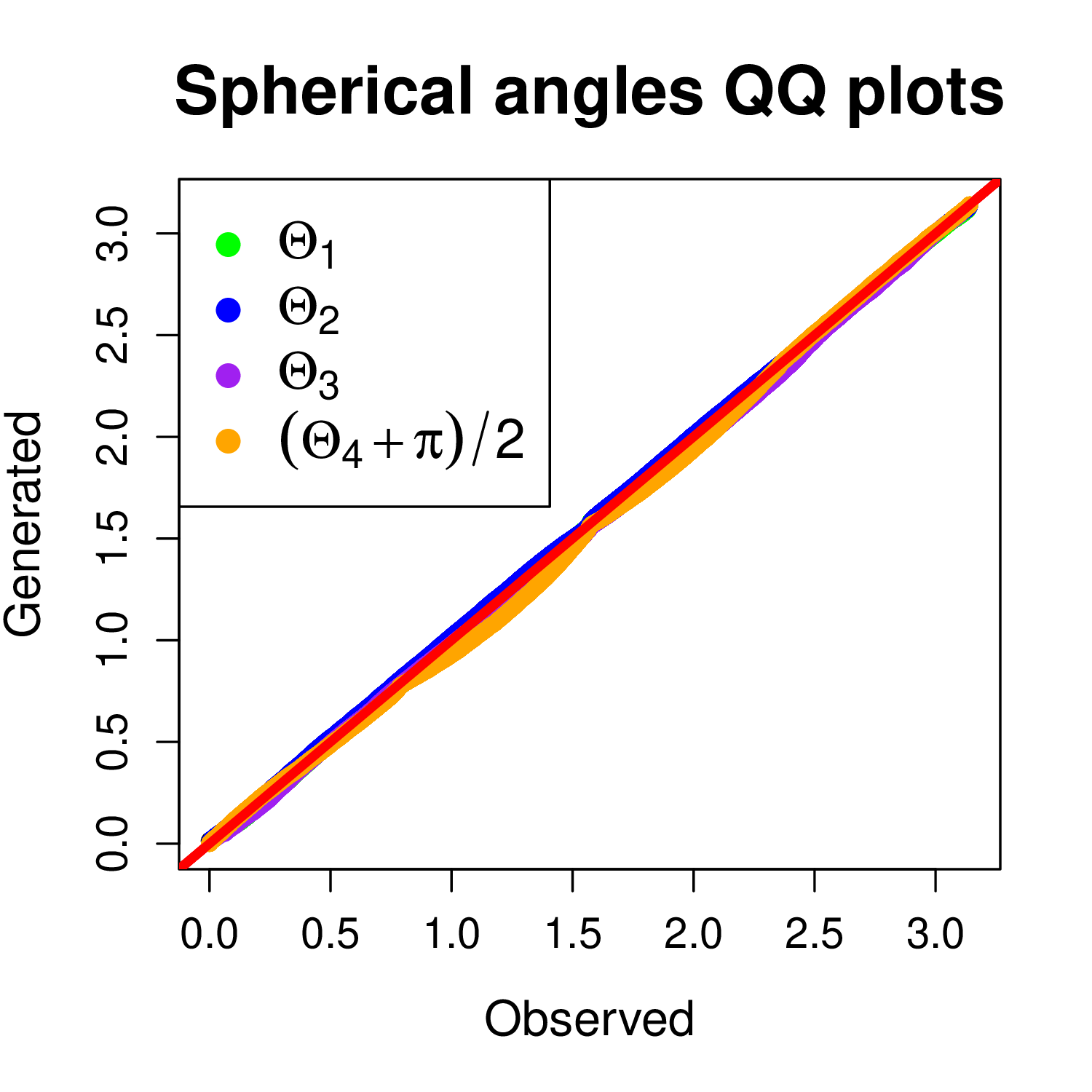}
    \end{subfigure}%
    \begin{subfigure}[b]{0.2\textwidth}
        \centering
        \includegraphics[width=\textwidth]{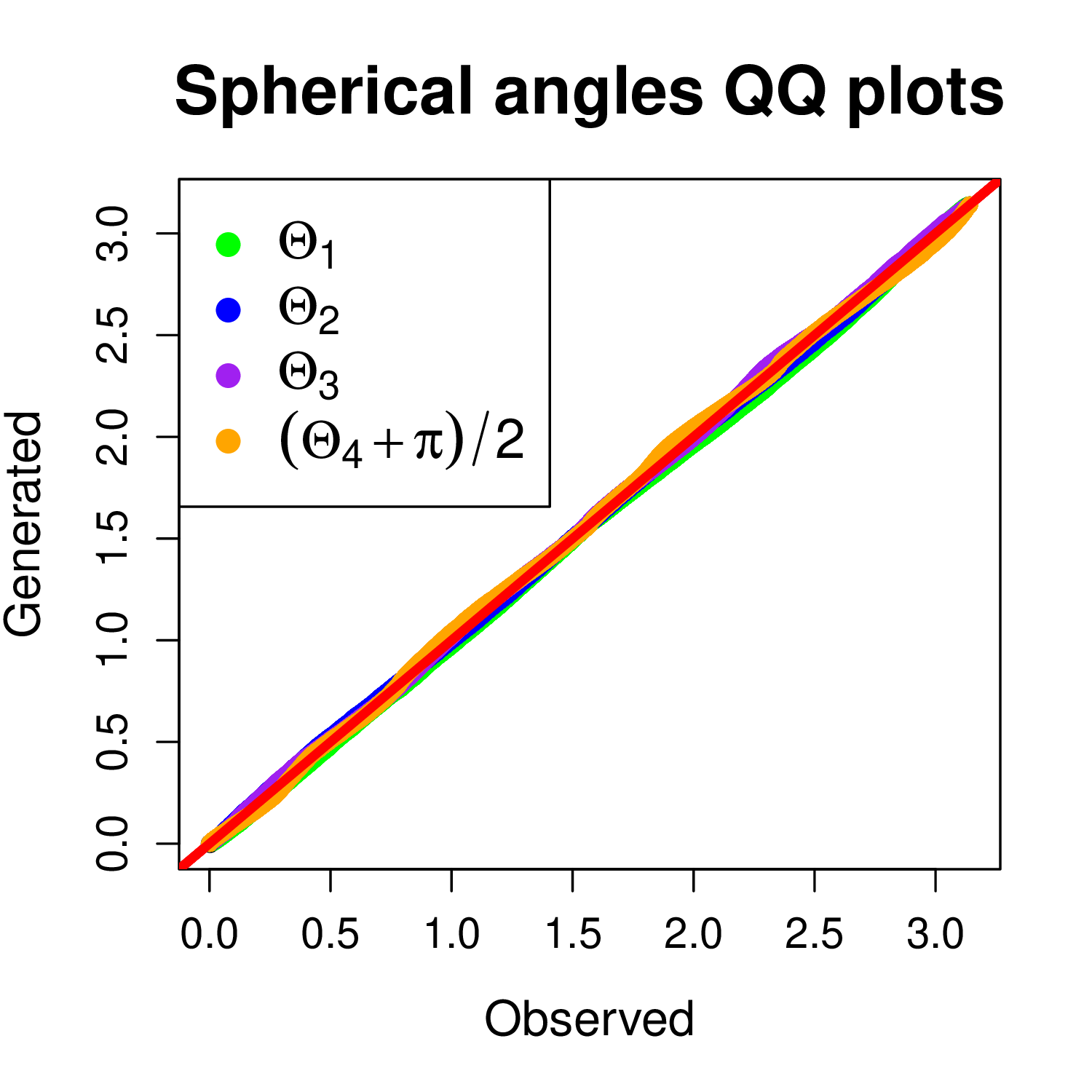}
    \end{subfigure}%
    \begin{subfigure}[b]{0.2\textwidth}
        \centering
        \includegraphics[width=\textwidth]{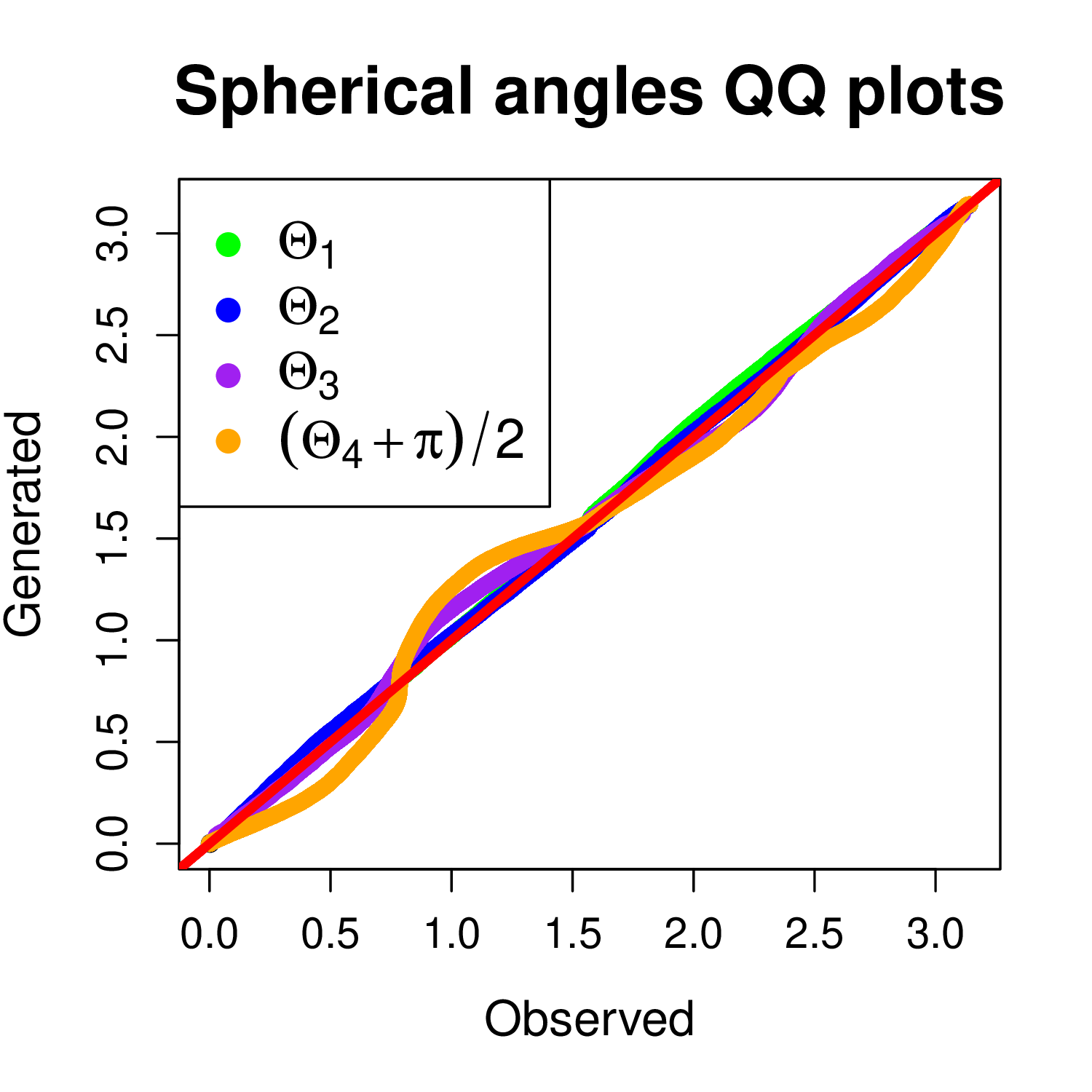}
    \end{subfigure}%
    \begin{subfigure}[b]{0.2\textwidth}
        \centering
        \includegraphics[width=\textwidth]{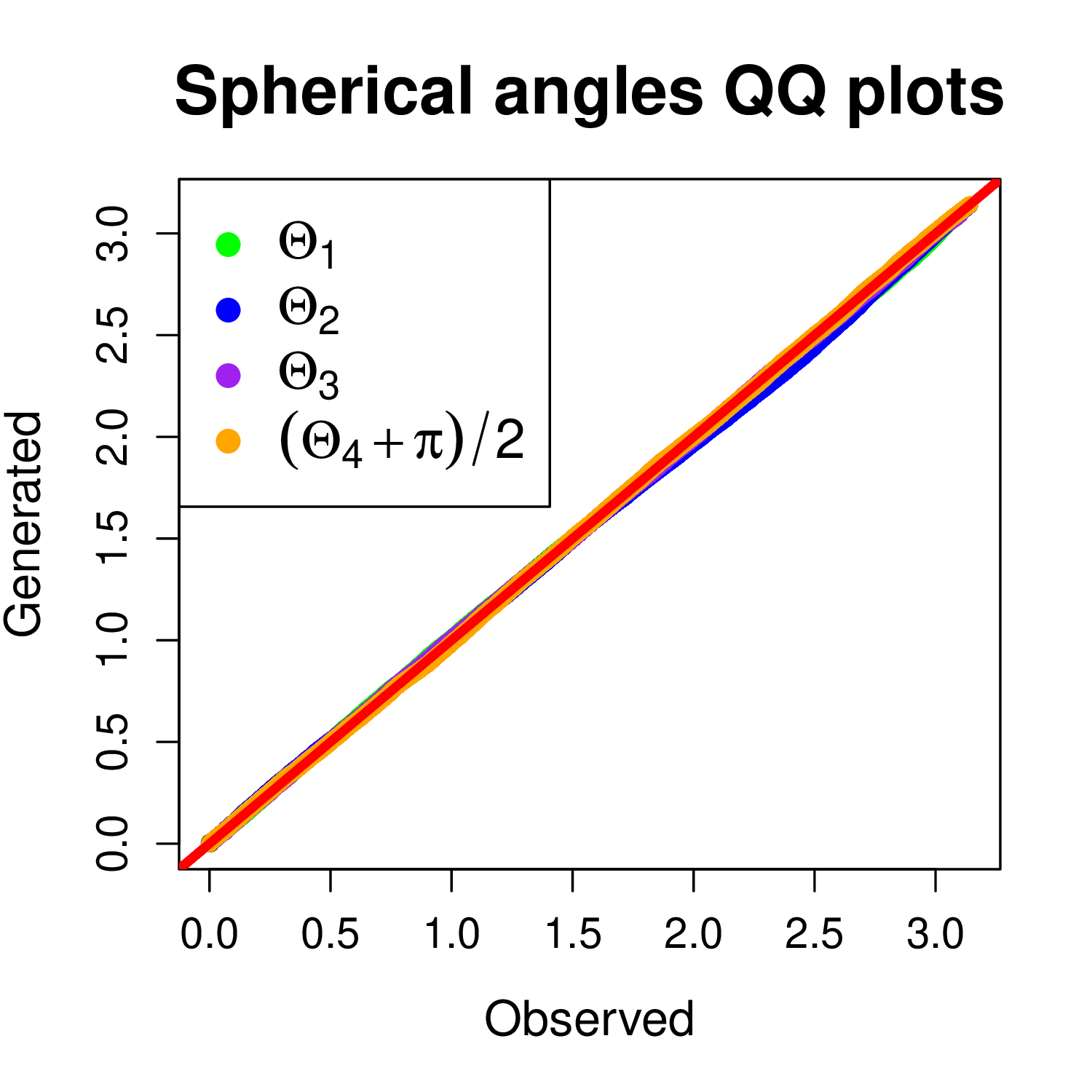}
    \end{subfigure}%
    \caption{Spherical angle QQ plots for copula 5 with $d=5$ and $n=100\ 000$. The ordering of methods and margins is as in Figure~\ref{fig:qq_plots_cop2_d5}. }
    \label{fig:qq_plots_cop5_d5}
\end{figure}

\begin{figure}[h!]
    \centering
      \begin{subfigure}[b]{0.2\textwidth}
        \centering
        \includegraphics[width=\textwidth]{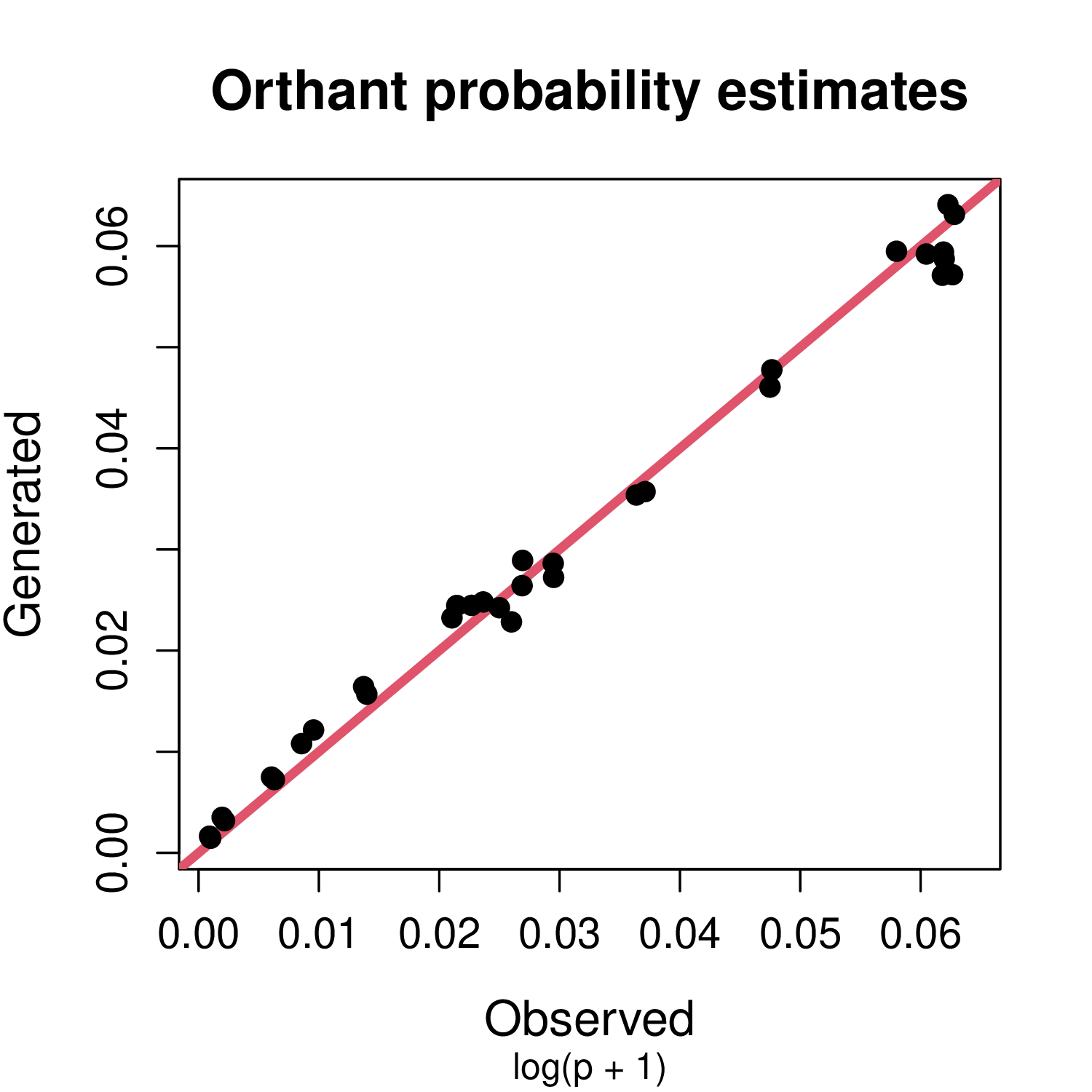}
    \end{subfigure}%
    \begin{subfigure}[b]{0.2\textwidth}
        \centering
        \includegraphics[width=\textwidth]{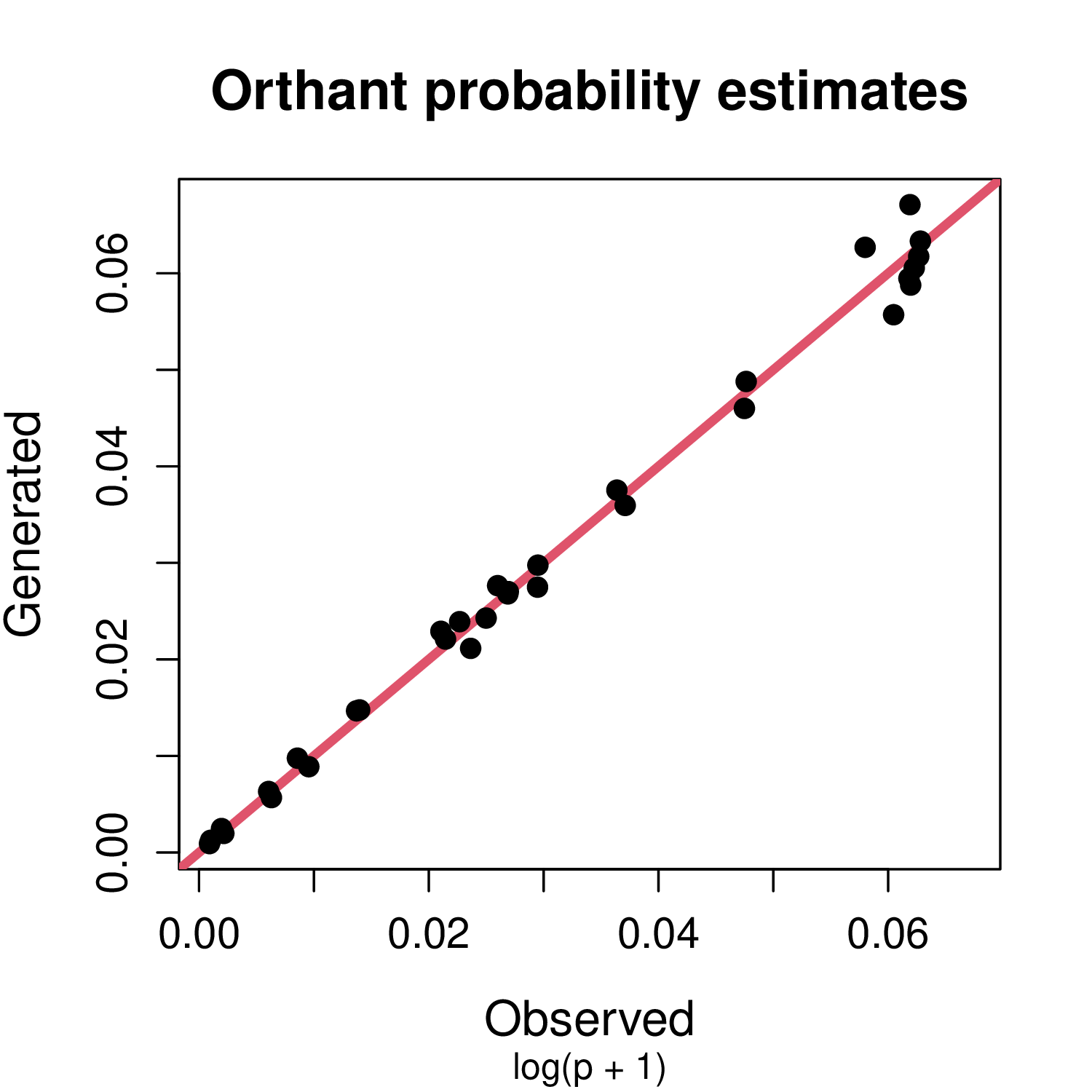}
    \end{subfigure}%
    \begin{subfigure}[b]{0.2\textwidth}
        \centering
        \includegraphics[width=\textwidth]{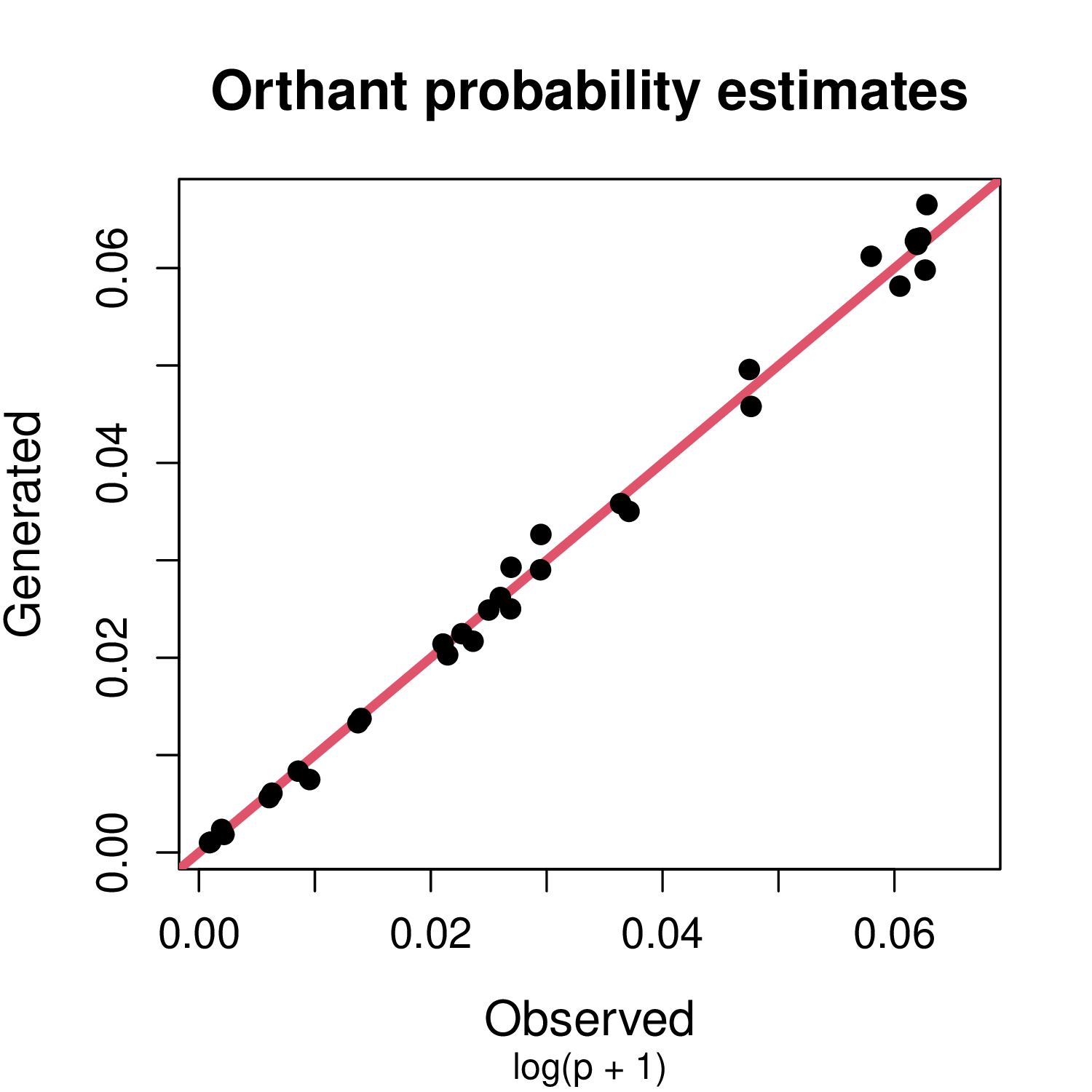}
    \end{subfigure}%
    \begin{subfigure}[b]{0.2\textwidth}
        \centering
        \includegraphics[width=\textwidth]{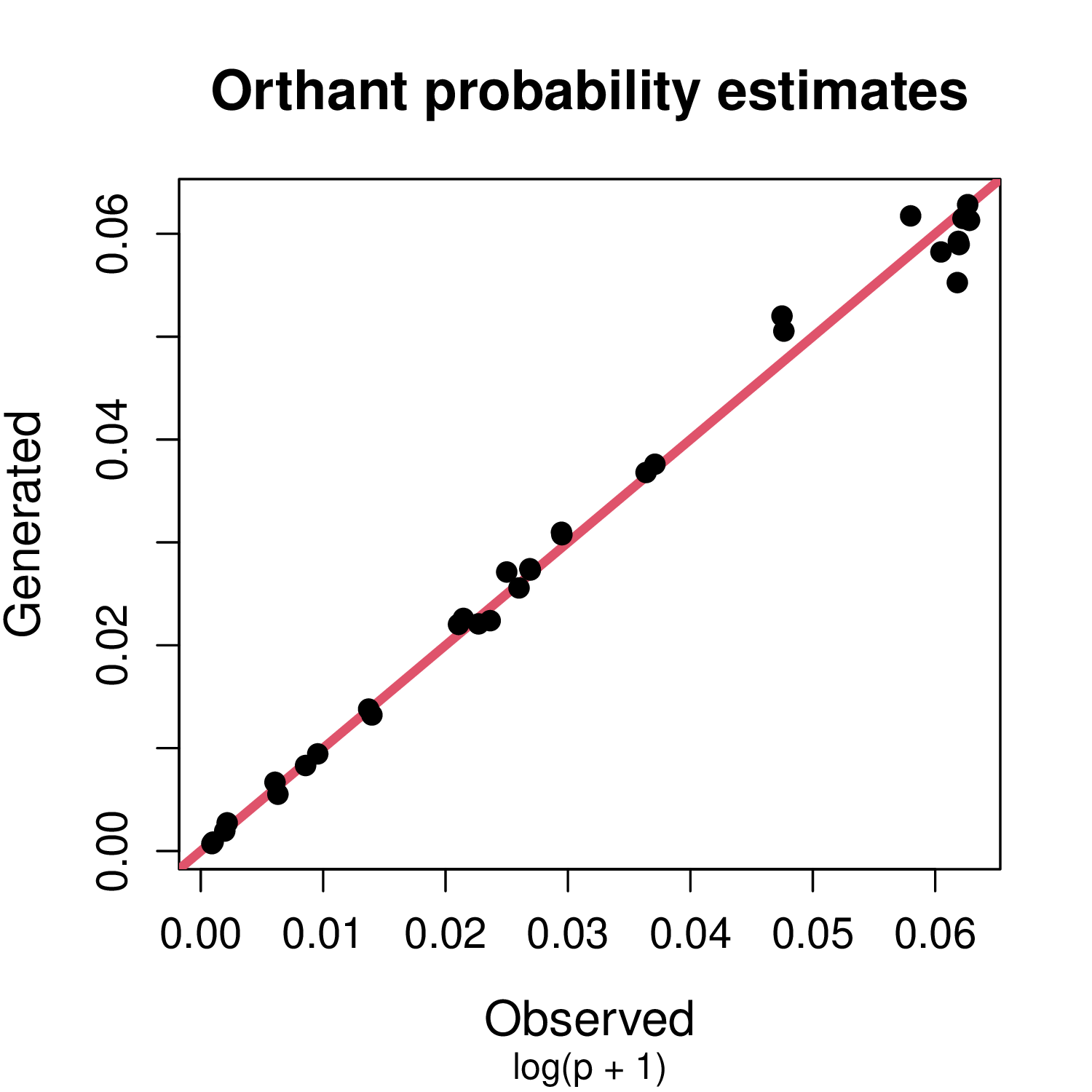}
    \end{subfigure}%
    \begin{subfigure}[b]{0.2\textwidth}
        \centering
        \includegraphics[width=\textwidth]{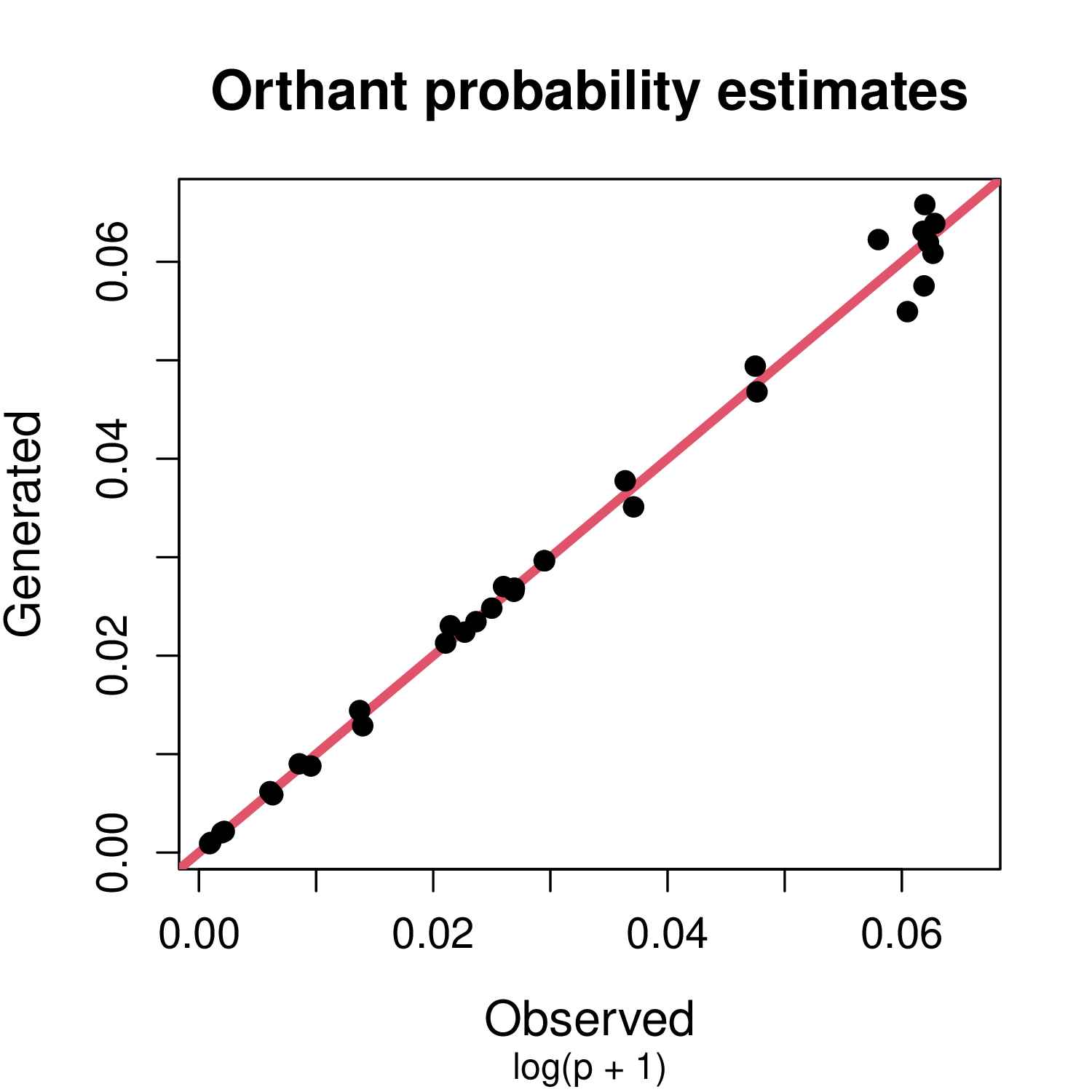}
    \end{subfigure}%

     \begin{subfigure}[b]{0.2\textwidth}
        \centering
        \includegraphics[width=\textwidth]{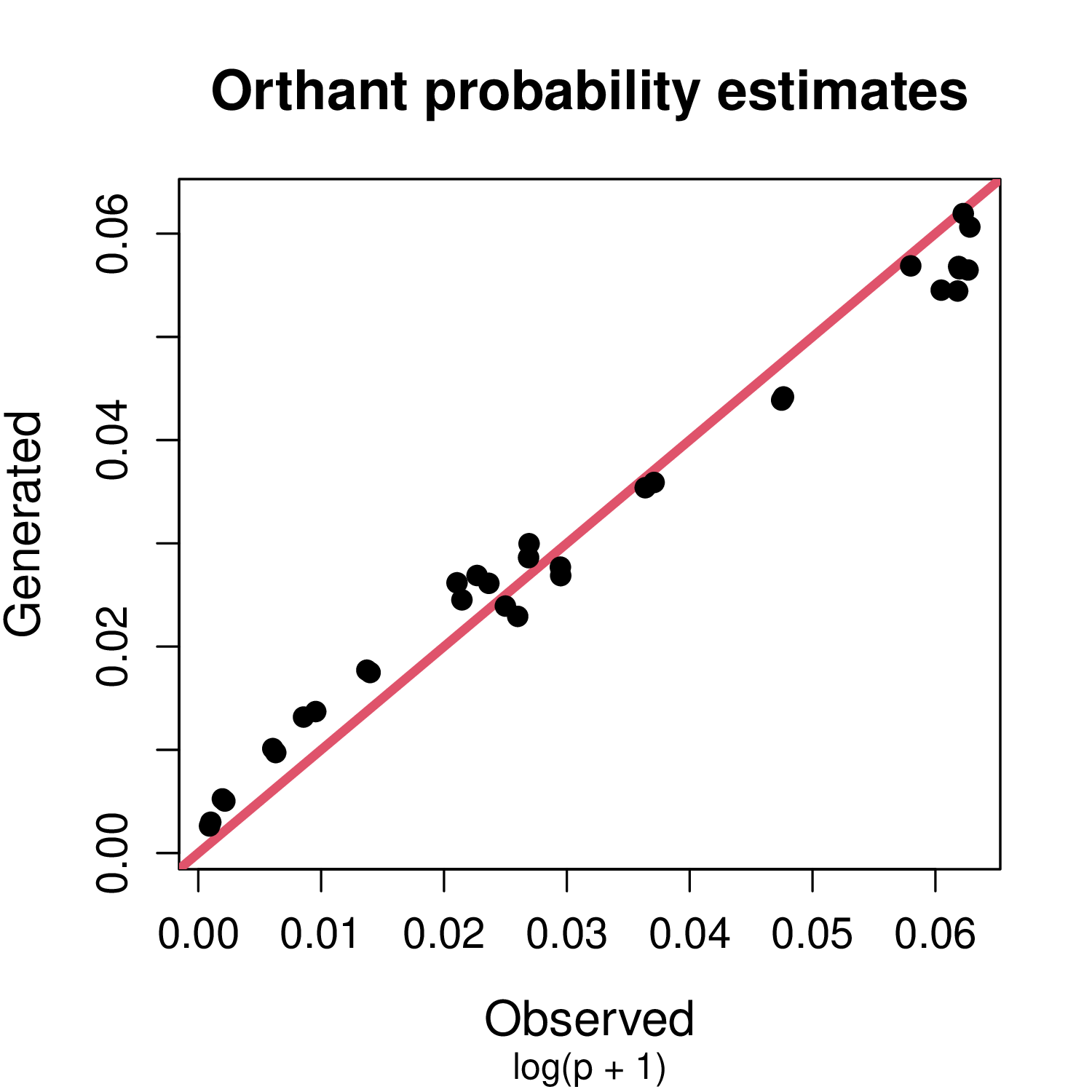}
    \end{subfigure}%
    \begin{subfigure}[b]{0.2\textwidth}
        \centering
        \includegraphics[width=\textwidth]{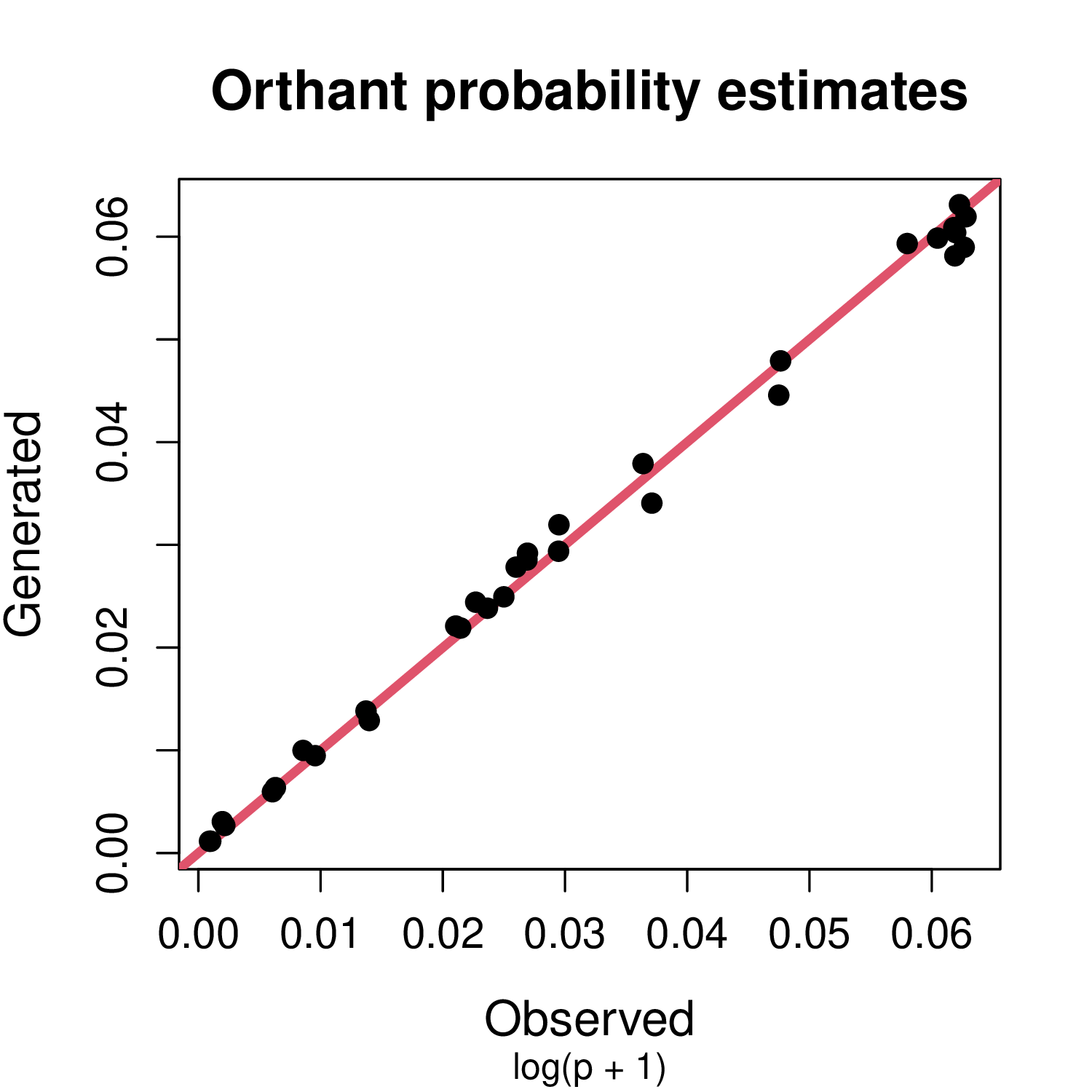}
    \end{subfigure}%
    \begin{subfigure}[b]{0.2\textwidth}
        \centering
        \includegraphics[width=\textwidth]{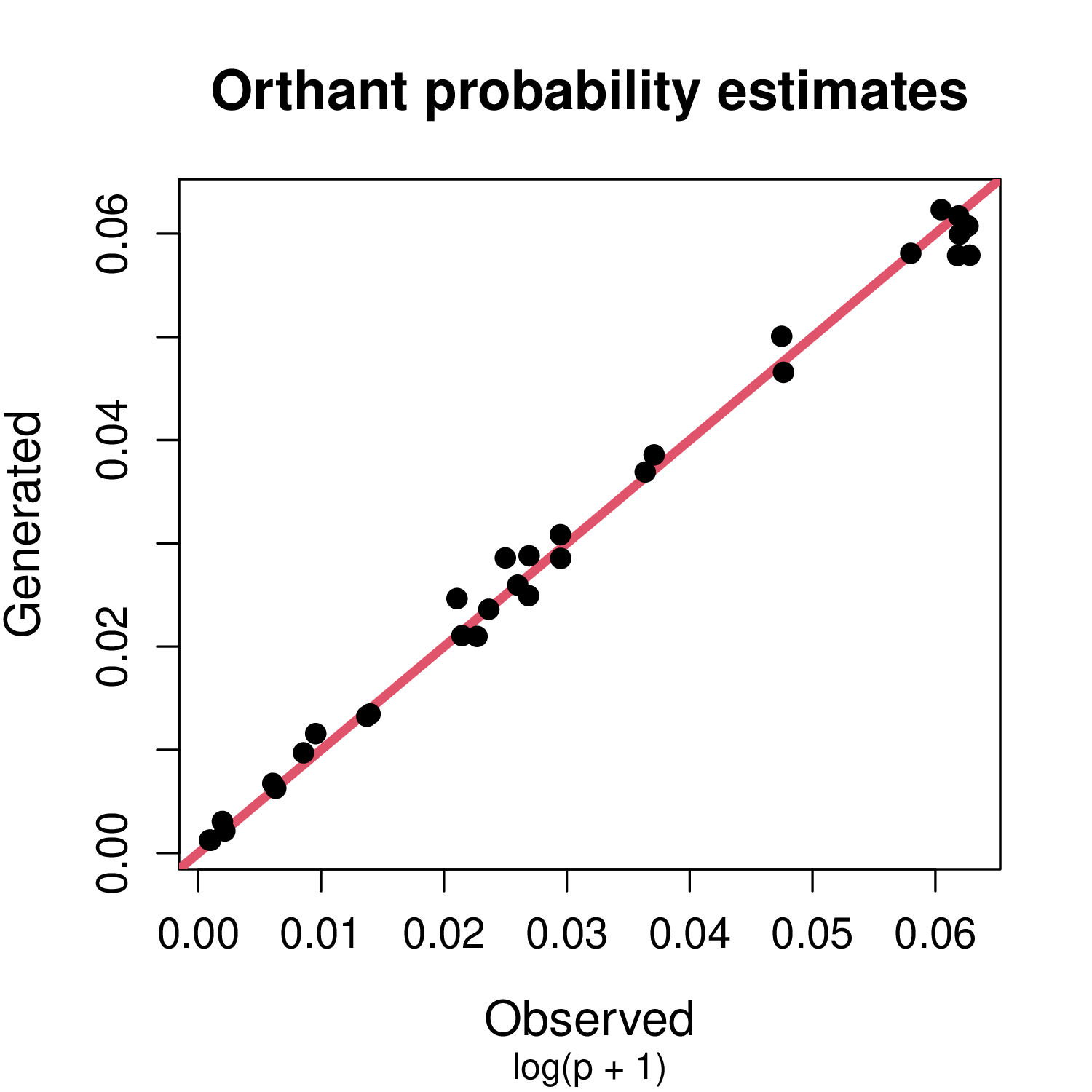}
    \end{subfigure}%
    \begin{subfigure}[b]{0.2\textwidth}
        \centering
        \includegraphics[width=\textwidth]{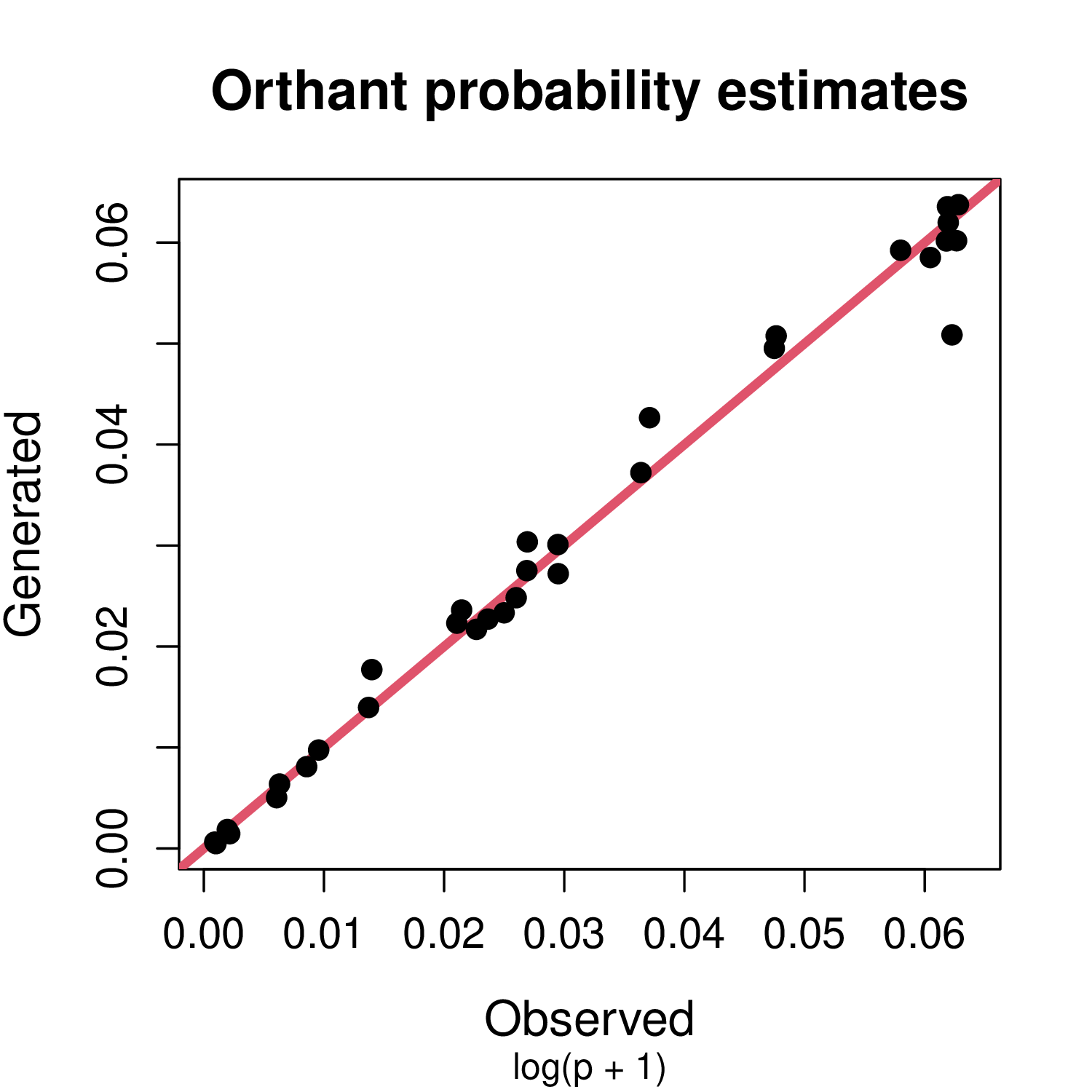}
    \end{subfigure}%
    \begin{subfigure}[b]{0.2\textwidth}
        \centering
        \includegraphics[width=\textwidth]{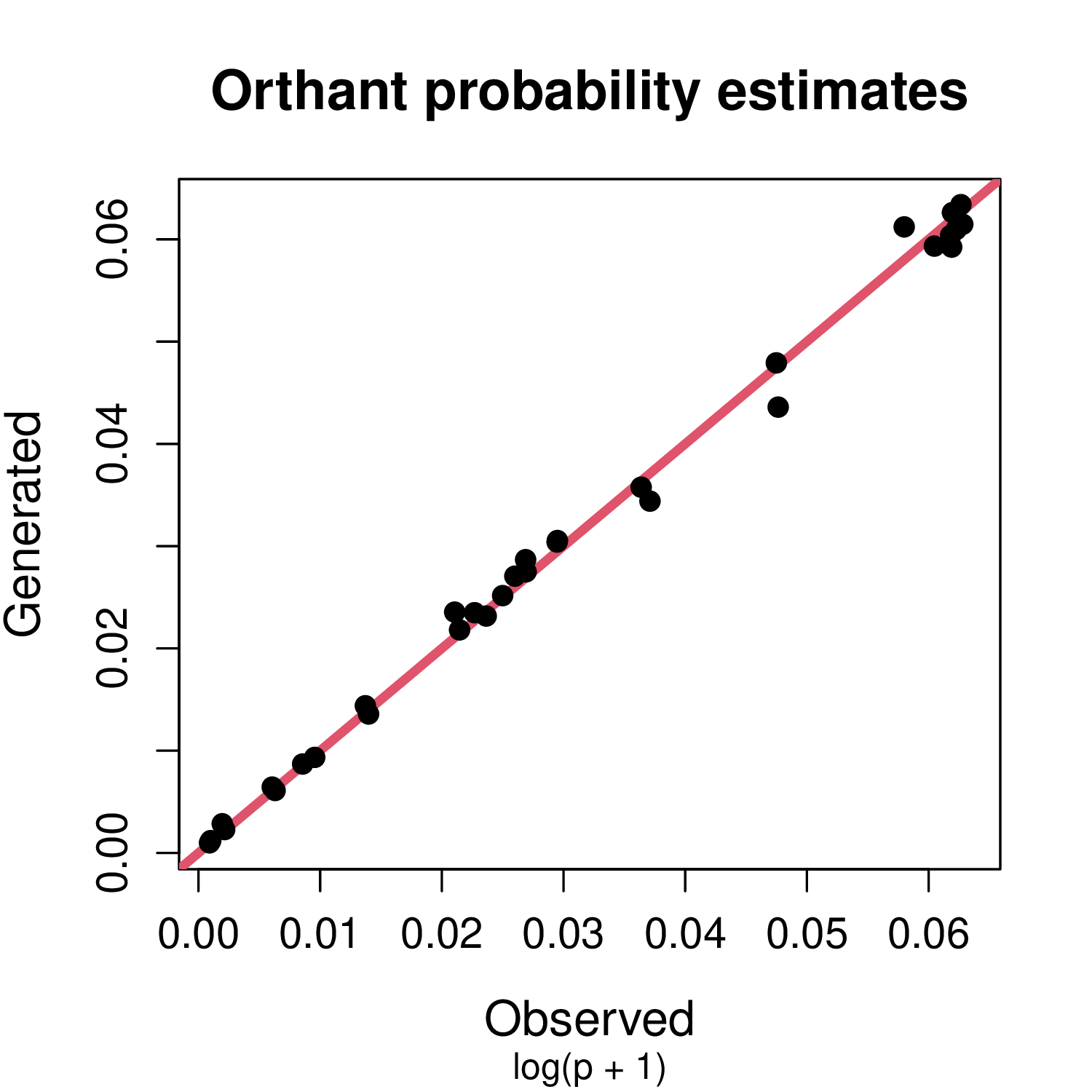}
    \end{subfigure}%
    \caption{Orthant probability plots for copula 2 with $d=5$ and $n=100\ 000$. The ordering of methods and margins is as in Figure~\ref{fig:qq_plots_cop2_d5}.}
    \label{fig:orthant_plots_cop2_d5}

\end{figure}

\begin{figure}[h!]
    \centering
        \begin{subfigure}[b]{0.2\textwidth}
        \centering
        \includegraphics[width=\textwidth]{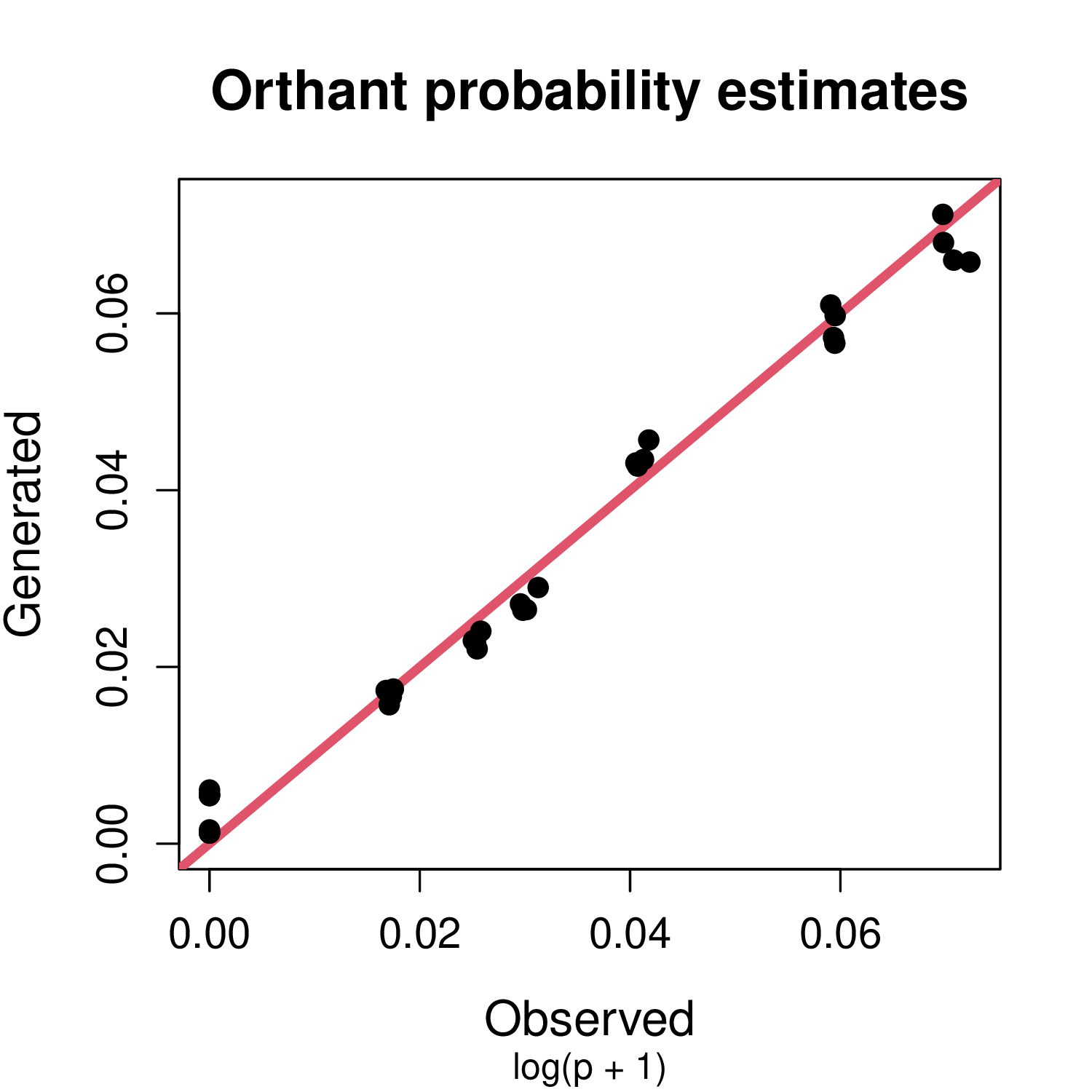}
    \end{subfigure}%
    \begin{subfigure}[b]{0.2\textwidth}
        \centering
        \includegraphics[width=\textwidth]{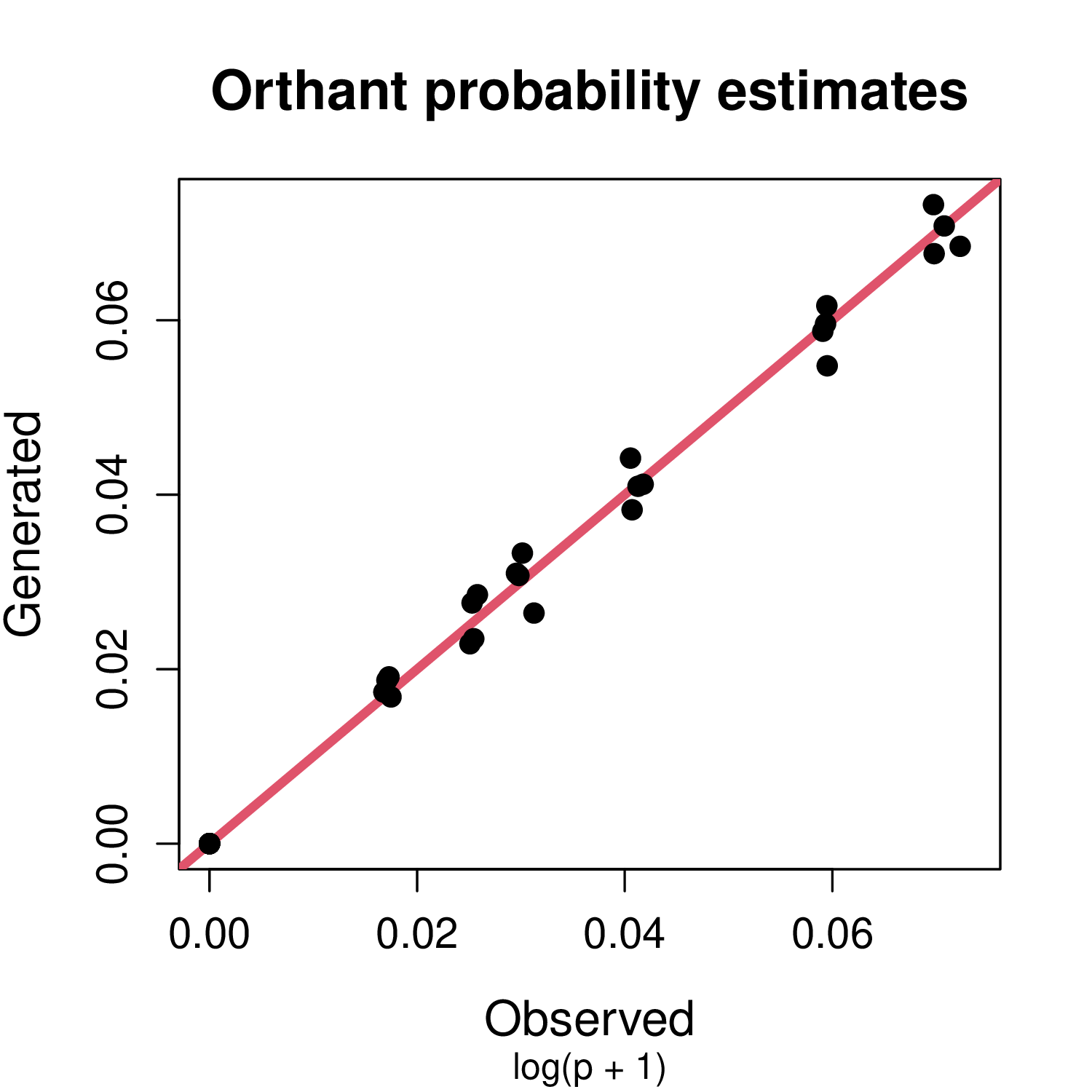}
    \end{subfigure}%
    \begin{subfigure}[b]{0.2\textwidth}
        \centering
        \includegraphics[width=\textwidth]{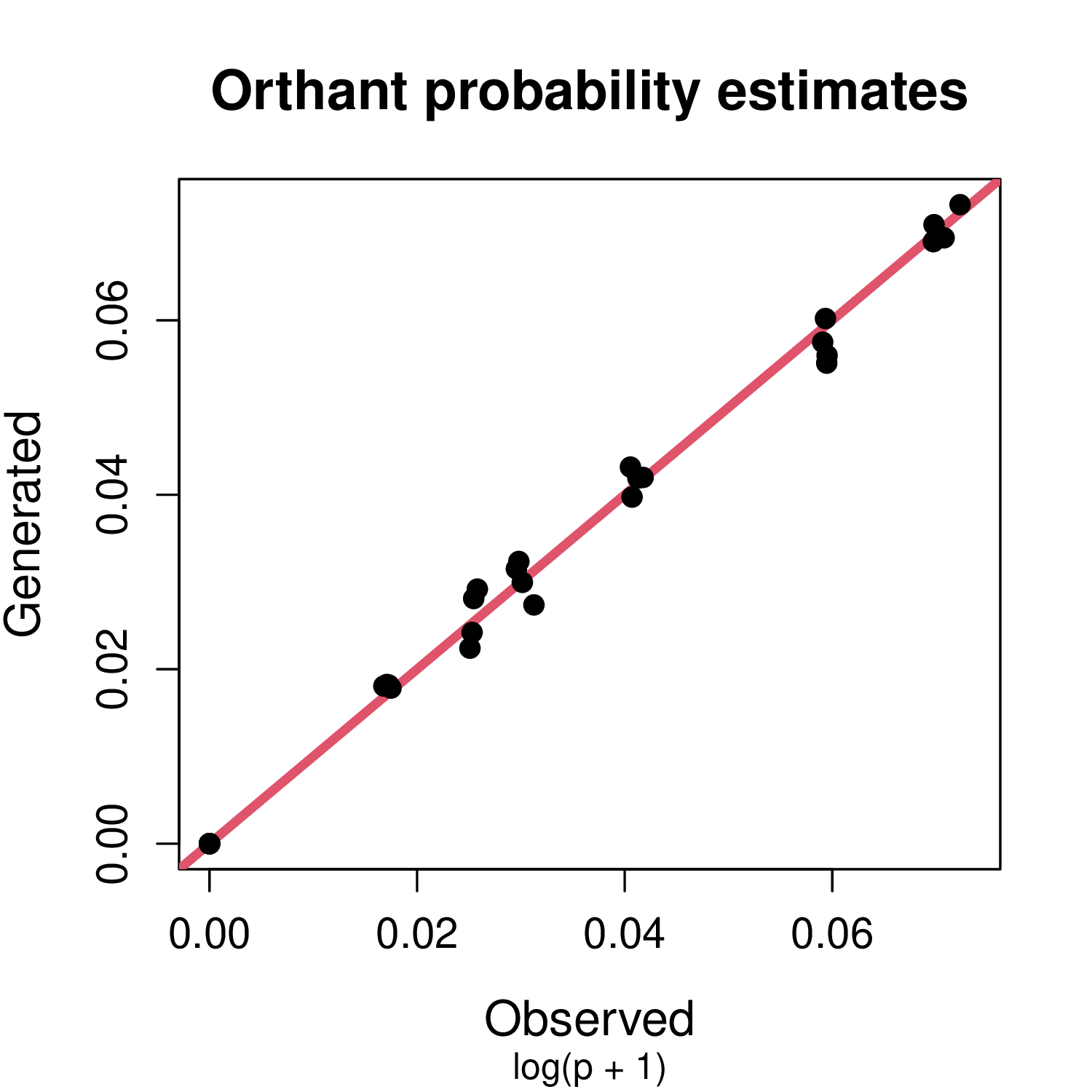}
    \end{subfigure}%
    \begin{subfigure}[b]{0.2\textwidth}
        \centering
        \includegraphics[width=\textwidth]{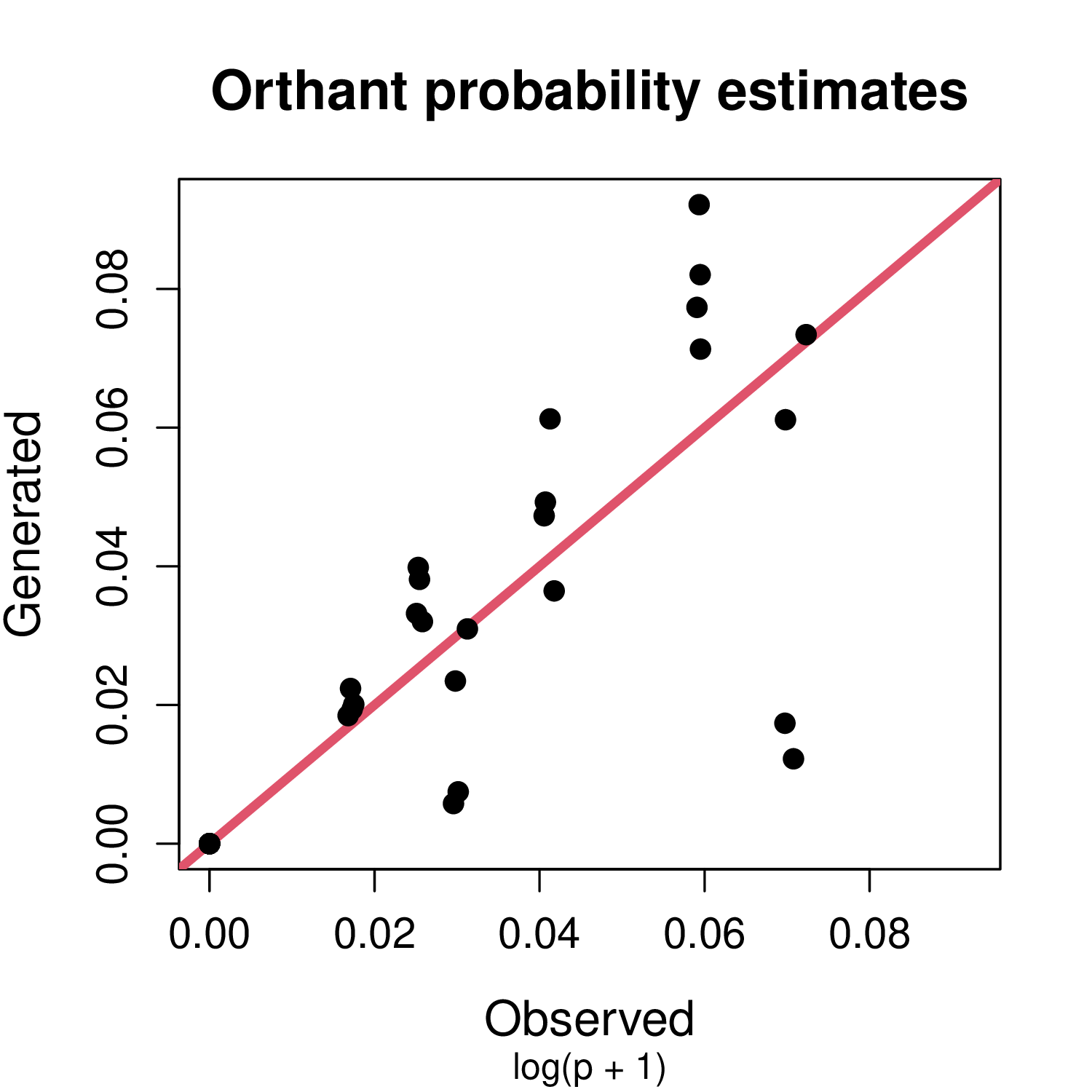}
    \end{subfigure}%
    \begin{subfigure}[b]{0.2\textwidth}
        \centering
        \includegraphics[width=\textwidth]{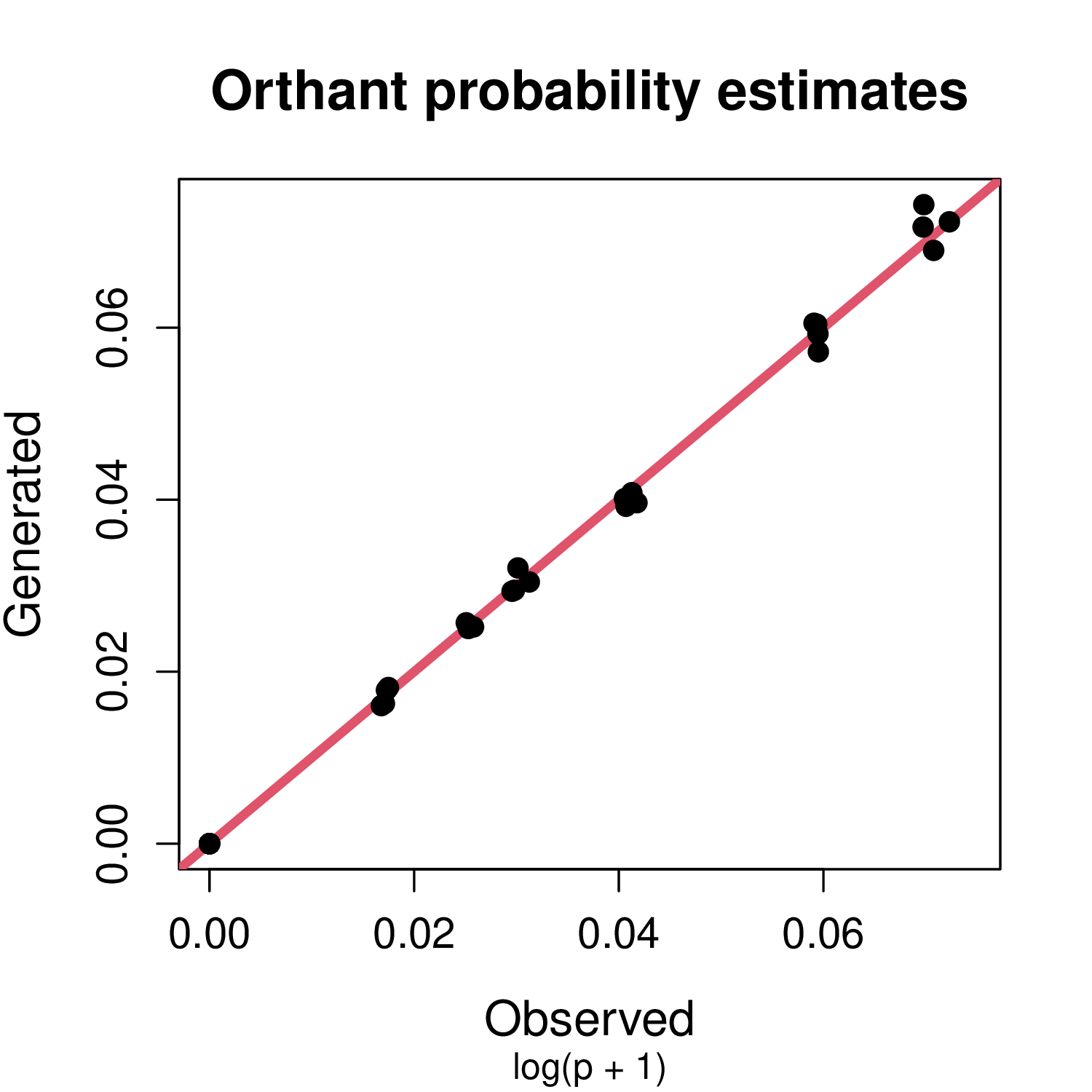}
    \end{subfigure}%

    \begin{subfigure}[b]{0.2\textwidth}
        \centering
        \includegraphics[width=\textwidth]{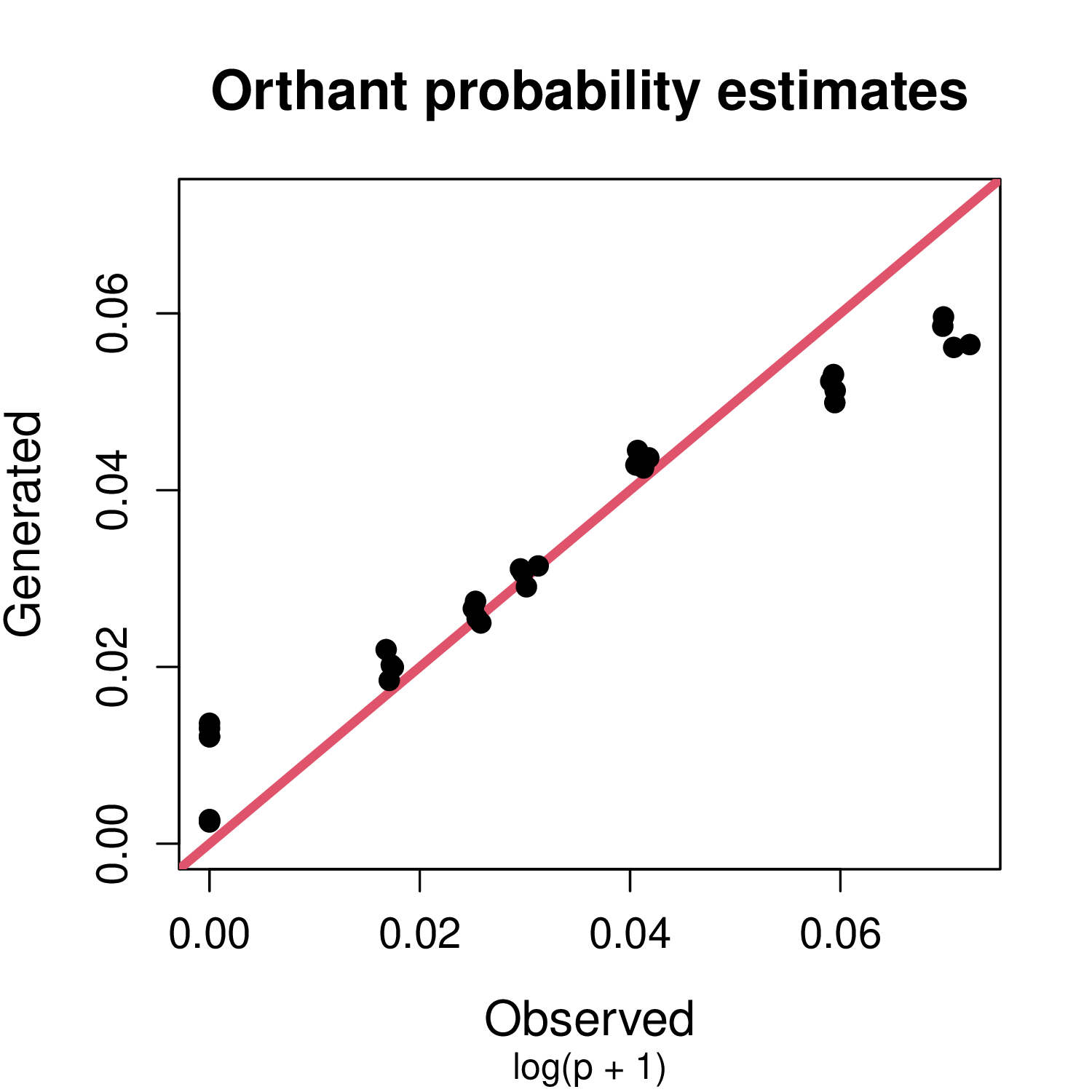}
    \end{subfigure}%
    \begin{subfigure}[b]{0.2\textwidth}
        \centering
        \includegraphics[width=\textwidth]{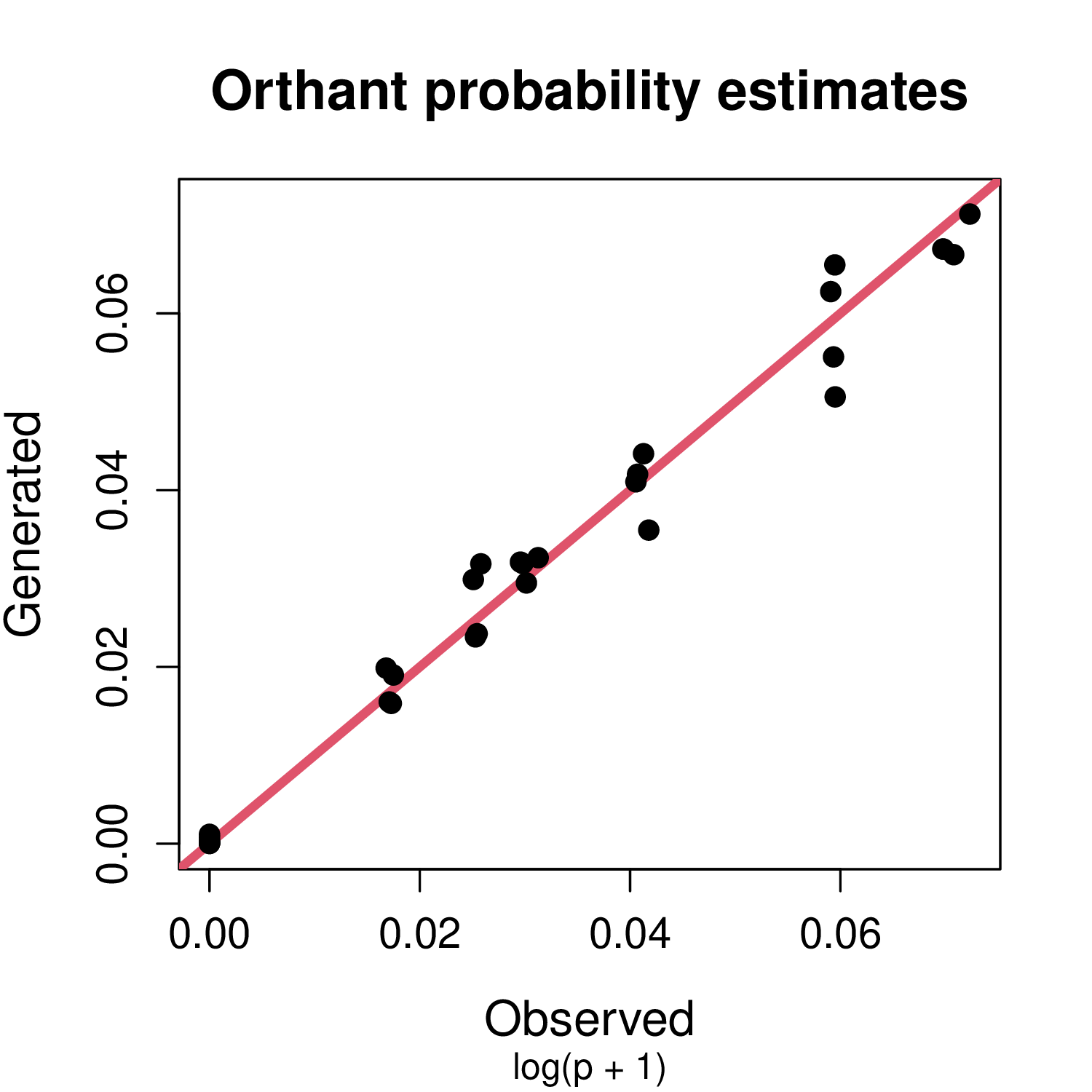}
    \end{subfigure}%
    \begin{subfigure}[b]{0.2\textwidth}
        \centering
        \includegraphics[width=\textwidth]{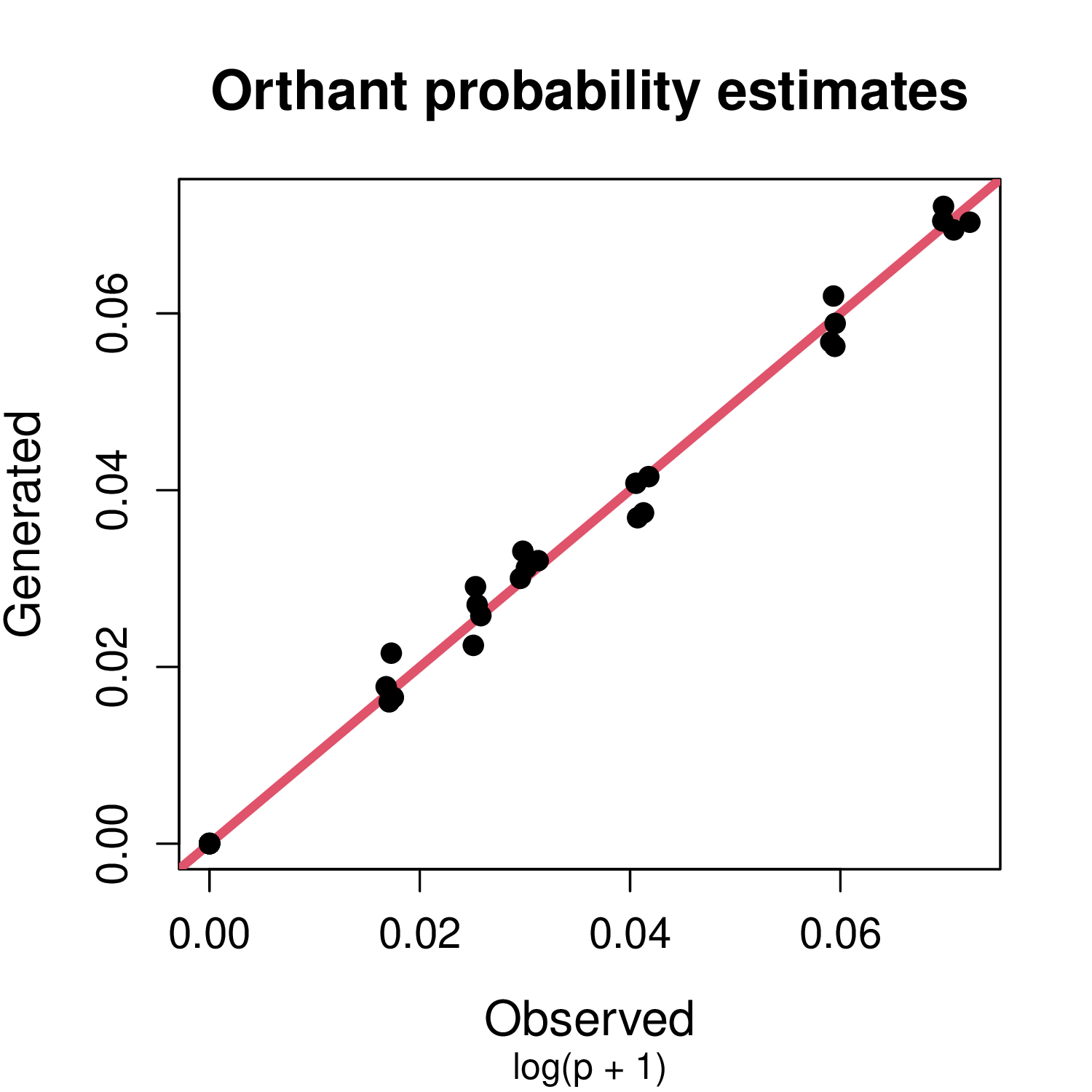}
    \end{subfigure}%
    \begin{subfigure}[b]{0.2\textwidth}
        \centering
        \includegraphics[width=\textwidth]{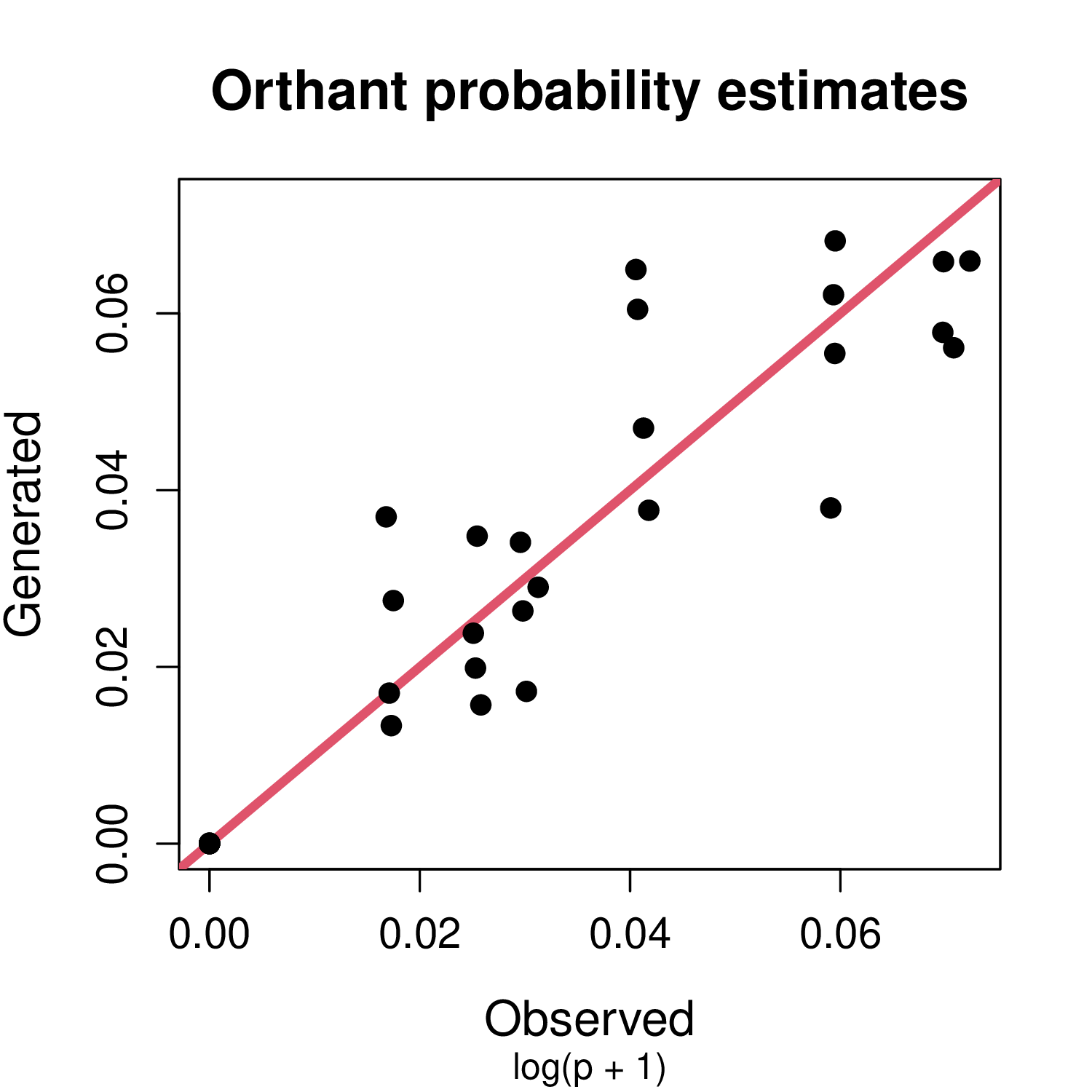}
    \end{subfigure}%
    \begin{subfigure}[b]{0.2\textwidth}
        \centering
        \includegraphics[width=\textwidth]{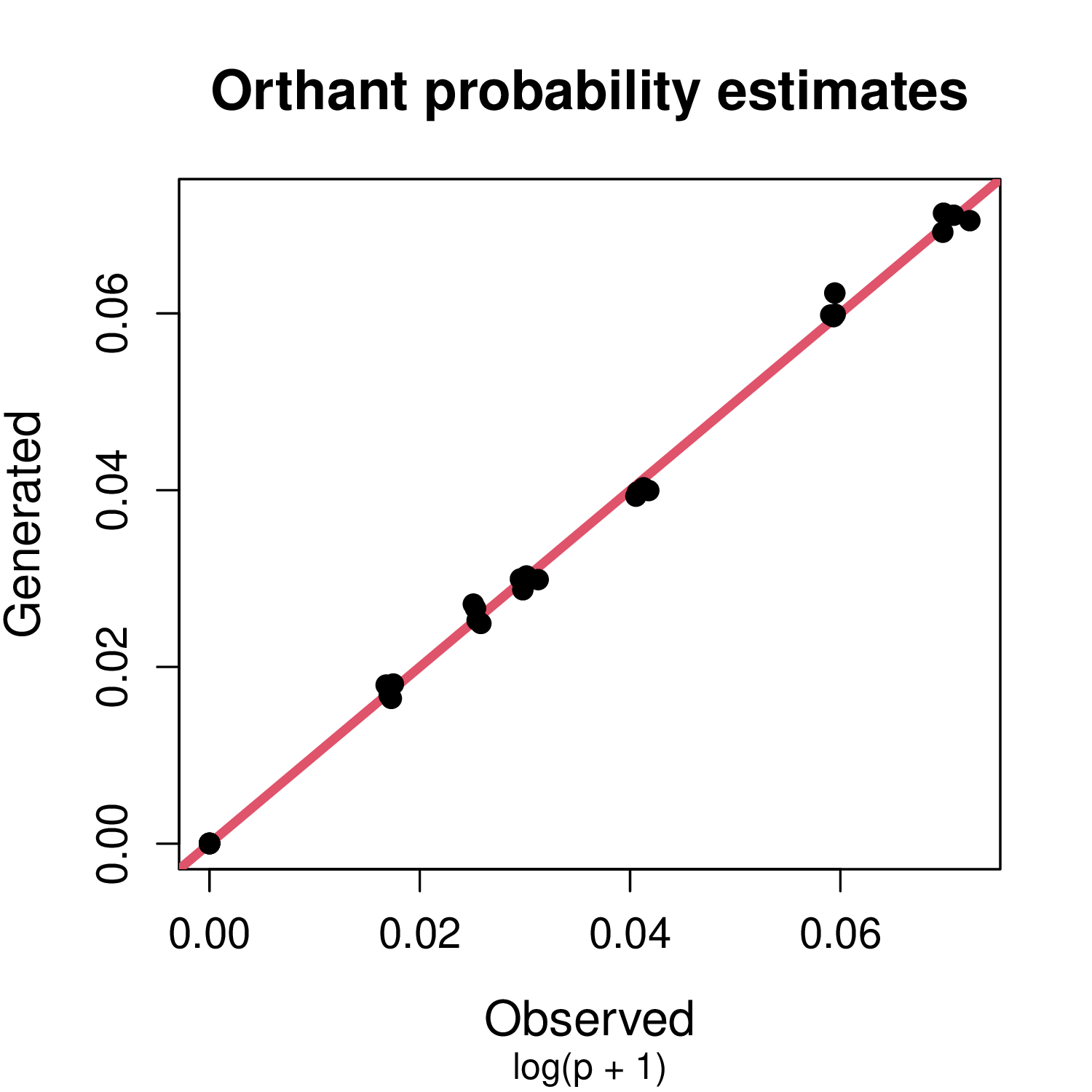}
    \end{subfigure}%
    \caption{Orthant probability plots for copula 5 with $d=5$ and $n=100\ 000$. The ordering of methods and margins is as in Figure~\ref{fig:qq_plots_cop2_d5}. }
    \label{fig:orthant_plots_cop5_d5}
\end{figure}

Our numerical skill metric is supported by a range of visual diagnostics. Illustrating each of these plots for every possible setting introduced in Section~\ref{subsec:sim_setup} would be both laborious and tedious. Therefore, we have curated a small subset of interesting results to present here; specifically, the mixture distribution defined by copula 2, and the sparse distribution defined by copula 5. We fix the sample size as the maximum $n = 100\ 000$ and consider both $d = 5$ and $d = 10$, as well as both proposed marginal distributions. For $d = 5$, Figures~\ref{fig:qq_plots_cop2_d5}--\ref{fig:qq_plots_cop5_d5} give the combined QQ plots for each deep generative approach, while Figures~\ref{fig:orthant_plots_cop2_d5}--\ref{fig:orthant_plots_cop5_d5} give the corresponding orthant probability plots; the equivalent figures for $d = 10$ are given in Appendix~\ref{app:add_sim_study}. Visual diagnostics for the remaining simulation scenarios are available as supplementary material.

One can make several observations from these results. Firstly, in terms of the expected aES, the baseline approach gives the lowest value in the vast majority of cases. However, the results for the deep generative approaches are all broadly very close to the baseline method, indicating a similar model performance according to this metric. The relative ordering of each of the deep generative approaches appears to agree with intuition when one considers the associated visual diagnostics, at least in the case of the QQ plots, i.e., a lower skill score usually corresponds to a better marginal goodness of fit. This indicates that while there may appear to be very little difference between the estimated skill scores, the induced ordering is still somewhat informative. We stress here that the skill score is computed over a very large sample size, resulting in negligible uncertainty within the estimates for a given data set. We also note that all scores are very close to the minimal achievable score from the true distribution.

\begin{table}[!h]
\centering\centering
\caption{$\operatorname{Skill}(F_{*})$ scores (to 6 significant figures) of each deep generative approach ($* \in \{\text{FM},\text{NFMAF},\text{GAN},\text{NFNSF} \} $) across all combinations for $d = 5$.\\}
\label{tab:cCRPS_d_5}
\centering
\begin{tabular}[t]{c|c|c|c|c|c|c|c}
\hline
 &  &  &  & \multicolumn{4}{c}{\textbf{Model}}\\
\hline
\textbf{Copula} & \textbf{Margins} & $n$ & $d$ & FM & NFMAF & GAN & NFNSF\\
\hline
1 & Laplace & $10^{3}$ & 5 & \textbf{1.00069} & 1.00156 & 1.01045 & 1.00236\\
\hline
1 & Laplace & $10^{4}$ & 5 & 1.00057 & \textbf{1.00038} & 1.00096 & 1.00045\\
\hline
1 & Laplace & $10^{5}$ & 5 & \textbf{1.00036} & 1.00046 & 1.0006 & 1.00046\\
\hline
1 & Double Pareto & $10^{3}$ & 5 & 1.00797 & 1.00262 & \textbf{1.00147} & 1.00229\\
\hline
1 & Double Pareto & $10^{4}$ & 5 & 1.0006 & \textbf{1.00037} & 1.00479 & 1.00135\\
\hline
1 & Double Pareto & $10^{5}$ & 5 & 1.00057 & \textbf{1.00006} & 1.00124 & 1.0001\\
\hline
2 & Laplace & $10^{3}$ & 5 & \textbf{0.998882} & 1.00023 & 1.0089 & 0.99968\\
\hline
2 & Laplace & $10^{4}$ & 5 & 1.00075 & 1.00046 & 1.00063 & \textbf{1.00031}\\
\hline
2 & Laplace & $10^{5}$ & 5 & 1.00021 & \textbf{1.00003} & 1.00004 & 1.00022\\
\hline
2 & Double Pareto & $10^{3}$ & 5 & 0.999417 & 1.00144 & 1.01425 & \textbf{0.999399}\\
\hline
2 & Double Pareto & $10^{4}$ & 5 & 1.00031 & \textbf{1.00013} & 1.00304 & 1.0002\\
\hline
2 & Double Pareto & $10^{5}$ & 5 & 1.00027 & 1.00032 & 1.00025 & \textbf{1.00003}\\
\hline
3 & Laplace & $10^{3}$ & 5 & \textbf{1.00248} & 1.00305 & 1.01212 & 1.00357\\
\hline
3 & Laplace & $10^{4}$ & 5 & \textbf{1.00012} & 1.00032 & 1.00338 & 1.00019\\
\hline
3 & Laplace & $10^{5}$ & 5 & 1.00016 & \textbf{1.00004} & 1.00085 & 1.00019\\
\hline
3 & Double Pareto & $10^{3}$ & 5 & 1.00417 & \textbf{1.00151} & 1.0066 & 1.00245\\
\hline
3 & Double Pareto & $10^{4}$ & 5 & 1.00013 & 1.00013 & 1.01177 & \textbf{1.00003}\\
\hline
3 & Double Pareto & $10^{5}$ & 5 & 1.00036 & 1.00012 & 1.00077 & \textbf{1.0001}\\
\hline
4 & Laplace & $10^{3}$ & 5 & 1.00149 & 1.00139 & 1.00421 & \textbf{1.00088}\\
\hline
4 & Laplace & $10^{4}$ & 5 & 1.00077 & 1.00073 & 1.00858 & \textbf{1.00023}\\
\hline
4 & Laplace & $10^{5}$ & 5 & 1.00023 & 1.00034 & 1.00485 & \textbf{1.00021}\\
\hline
4 & Double Pareto & $10^{3}$ & 5 & 1.00236 & 1.00297 & 1.00286 & \textbf{1.00146}\\
\hline
4 & Double Pareto & $10^{4}$ & 5 & 1.00028 & \textbf{1.00021} & 1.01189 & 1.00029\\
\hline
4 & Double Pareto & $10^{5}$ & 5 & 1.0006 & 1.00019 & 1.00034 & \textbf{1.00016}\\
\hline
5 & Laplace & $10^{3}$ & 5 & \textbf{1.00067} & 1.00177 & 1.00614 & 1.00233\\
\hline
5 & Laplace & $10^{4}$ & 5 & 1.00026 & \textbf{1.00016} & 1.01154 & 1.00021\\
\hline
5 & Laplace & $10^{5}$ & 5 & 1.00037 & \textbf{1.00001} & 1.00866 & 1.00004\\
\hline
5 & Double Pareto & $10^{3}$ & 5 & \textbf{1.00064} & 1.00069 & 1.00331 & 1.00084\\
\hline
5 & Double Pareto & $10^{4}$ & 5 & \textbf{1} & 1.00019 & 1.00054 & 1.00033\\
\hline
5 & Double Pareto & $10^{5}$ & 5 & 1.00056 & 1.00044 & 1.00186 & \textbf{1.00004}\\
\hline 
\end{tabular}
\end{table}

Considering the visual diagnostics as a whole, we observe that, with the exception of the GAN approach, each of the tested deep generative approaches appears able to recreate the marginal \textit{and} dependence structures from the angular variables in almost every case. It is also worth remarking that even though the baseline approach had a lower expected aES, the visual diagnostics do not indicate that this approach performed best \textit{overall}. For many cases, we found the baseline approach performed poorly, which was not the case for all of the deep generative approaches; see, for instance, Figures~\ref{fig:qq_plots_cop5_d5} and \ref{fig:orthant_plots_cop5_d5}, alongside Figures~\ref{fig:qq_plots_cop5_d10} (QQ plots) and \ref{fig:orthant_plots_cop5_d10} (orthant probability plots) of Appendix~\ref{app:add_sim_study}. These diagnostics clearly indicate poor performance of the baseline approach relative to what we observe from the deep generative techniques. Consequently, while the expected aES provides a simple numerical means of comparing between fitted models, the resulting scores do not necessarily tell the full story. It is worth noting that the energy score has been shown to be fairly insensitive towards misspecification of the dependence structure, with its behaviour tending to be dominated by the marginal distributions \citep{pinson_discrimination_2013, bjerregard_introduction_2021}; our angular extension may well have similar shortcomings. Alternative proper scores have been proposed \citep{scheuerer_variogram-based_2015}, but to the authors' knowledge have not yet been generalised to multivariate angular variables. This might help to explain why the vMF mixture distribution, which often correctly reproduces the marginal behaviour, tends to perform best according to the aES. However, as illustrated by e.g. Figure~\ref{fig:qq_plots_cop5_d5}, this is not always the case, and therefore it is unclear why the aES metric consistently tends to favour the vMF approach. We therefore stress caution when interpreting the numerical results given in this study, since they only provide a simplistic summary of the overall predictive power of each fitted model.

We note that the choice of marginal distribution can affect the quality of model fits. In particular, for some model setups that perform well in the case of standard Laplace margins, we obtain low quality model fits when data are simulated on double Pareto margins (see, e.g., the plots for the GAN approach in Figures~\ref{fig:qq_plots_cop2_d5}--\ref{fig:orthant_plots_cop5_d5}). This illustrates the importance of considering different choices of marginal distribution when comparing generative approaches for angular variables.

Ideally, we would compute skill scores over a large number of simulated samples and take the average as a summary of the overall model performance. However, due to the high computational and memory demands required under the proposed deep learning frameworks, this was infeasible for our study and thus we have just considered single samples from each setup. In Appendix~\ref{app:add_sim_study}, we briefly consider the effect of repeated sampling for some specific setups; in particular, copula 5 with $d=10$, and varying sample size and marginal distributions. As one may expect, the relative ranking of each method changes slightly when using average scores, but generally there is little difference between the scores for all of the deep learning techniques (besides the GAN model) and confidence intervals across the averaged scores intersect in many cases. This once again demonstrates the limitations of the aES metric for comparing model fits. Moreover, the resulting visual diagnostics over different simulated samples show the deep learning approaches almost always outperform the benchmarking technique for these particular cases.

Our key finding is that all of the deep learning approaches introduced in Section~\ref{sec:methods} (besides the GAN) appear to generally work well for simulating a range of angular distributions, offering robustness over marginal choices and dimensions. This indicates that such approaches are generally suitable for simulating from angular variables when the `true' angular distribution is unknown.

On a final note, we remark that the skill score does not always tend to decrease as the training sample size increases, as would ordinarily be expected. There are several reasons why this may occur, but we suspect it may be due to lower sample sizes resulting in more `sharpness' (i.e., joint distributions exhibiting strong multi-modalities) within the generated angular simulations, which is rewarded by CRPS-type metrics \citep{gneiting_strictly_2007, buchweitz_asymmetric_2025}. When less information is available, we tend to find the simulated distributions have lower variability and concentrate around regions with high probability mass. In practice, this would not be desirable, and consequently, the increasing scores over sample size do not necessarily indicate that the model performance is worse when more information is available. This statement is supported by the visual diagnostics, which generally show clear improvements across all methods for larger training sample sizes.

\subsection{Higher dimensional setting} \label{subsec:d_50}

To further evaluate the proposed methodology, we consider some higher dimensional examples. In particular, we fix $d = 50$ and simulate data from a sparse Gaussian copula, constructed as described in Section~\ref{subsec:sim_setup}, using both choices of marginal distribution. We also consider data generated directly from a GAN model with randomly initialised weights and biases. Combined, these samples provide complex scenarios for which one might not expect simpler parametric models (such as the vMF mixture distribution) to perform adequately. Furthermore, this allows us to test the scalability of the proposed deep learning approaches, which are generally considered to perform well in higher dimensional settings \citep{Goodfellow-et-al-2016}. For parsimony, we use the architectures selected in Section~\ref{subsec:hyperparam}. 

The resulting aES metrics and visual diagnostics tell a similar story to the results presented in Section~\ref{subsec:cCRPS}; the vMF mixture model still gives the lowest overall scores, while the QQ plots reiterate that the deep learning methods tend to outperform this model, particularly for sparse structures. All QQ plot diagnostics and scores for the $50$-dimensional examples are provided in Appendix~\ref{app:add_sim_study}. It was not possible to provide the remaining visual diagnostics for such a dimension, since the orthant probability diagnostic would require the computation of $2^{50}$ probabilities (requiring approximately $8\; 000$TB of data), and the histogram diagnostic would produce 49 separate plots for each simulated sample and model. These findings once again highlight the lack of interpretability and usefulness of the aES metric, but in general, provide further evidence that deep learning approaches outperform the benchmark technique overall. 

\section{Case study}\label{sec:waves}

In this section, we apply the modelling approaches introduced in Section~\ref{sec:methods} to a hindcast data set consisting of 31-years of wind and wave variables from 01/01/1990 to 31/12/2020 for a site off the south-west coast of the UK. In this context, such variables are often referred to as \textit{metocean} variables \citep[e.g.,][]{Jonathan2013}. This data set, obtained as part of an EU INTERREG TIGER funded project \citep{mackay2022joint}, was recently considered by \citet{mackay_deep_2024}, who applied the SPAR model to successfully capture joint extremes of metocean variables. To simulate from the angular component, the authors used an empirical approach, but this has limitations, as discussed in Section~\ref{sec:intro}. We remark that \citet{mackay_deep_2024} also considered fitting mixtures of vMF distributions to model the angular component, but found the resulting simulations to be inadequate for SPAR model inference.  

As noted by \citet{mackay_deep_2024}, the observation site has been identified as a possible location for the development of floating wind farm projects. In such developments, robust structural analyses are required to ensure floating wind turbines are able to withstand the most extreme ocean events. Viewed through the angular-radial paradigm, this requires understanding of which directions in $\mathbb{S}^{d-1}$ one could reasonably expect to observe (extreme) ocean events. Therefore, accurate modelling of the angular variable is vital for this application. 

The data consist of hourly values of significant wave height $(H_s)$, mean wave period $(T_m)$, mean wave direction $(\theta_{wave})$, hourly mean wind speed at 10 m above sea level $(U_{10})$, and wind direction $(\theta_{wind})$, with $n=271\ 704$ total observations. Such variables directly influence floating structures such as wind turbines, and understanding their joint extremes is crucial for design analysis. Following \citet{mackay_deep_2024}, we work with the $x$- and $y$-components of wave height and wind speed, defined as $H_x := H_s \cos(\theta_{wave})$, $H_y := H_s \sin(\theta_{wave})$, $U_x := U_{10} \cos(\theta_{wind})$, and $U_y := U_{10} \sin(\theta_{wind})$, accounting for the fact the directional variables are periodic. Furthermore, we set $L_T :=\log(T_m) \in(-\infty,\infty)$; this step ensures all variables are observed on $\mathbb{R}$, which was found to improve the reliability of the SPAR model fitting procedure. Moreover, all variables are normalised and an origin is specified using knowledge of the physics of the process; see \citet{mackay_deep_2024} for further discussion. We remark that since we are not fitting the SPAR framework, or any other angular-radial model, in this work, the exact choice of origin or scale of variables is less important for this application.  

\begin{figure}[!h]
    \centering
    \includegraphics[width=.6\linewidth]{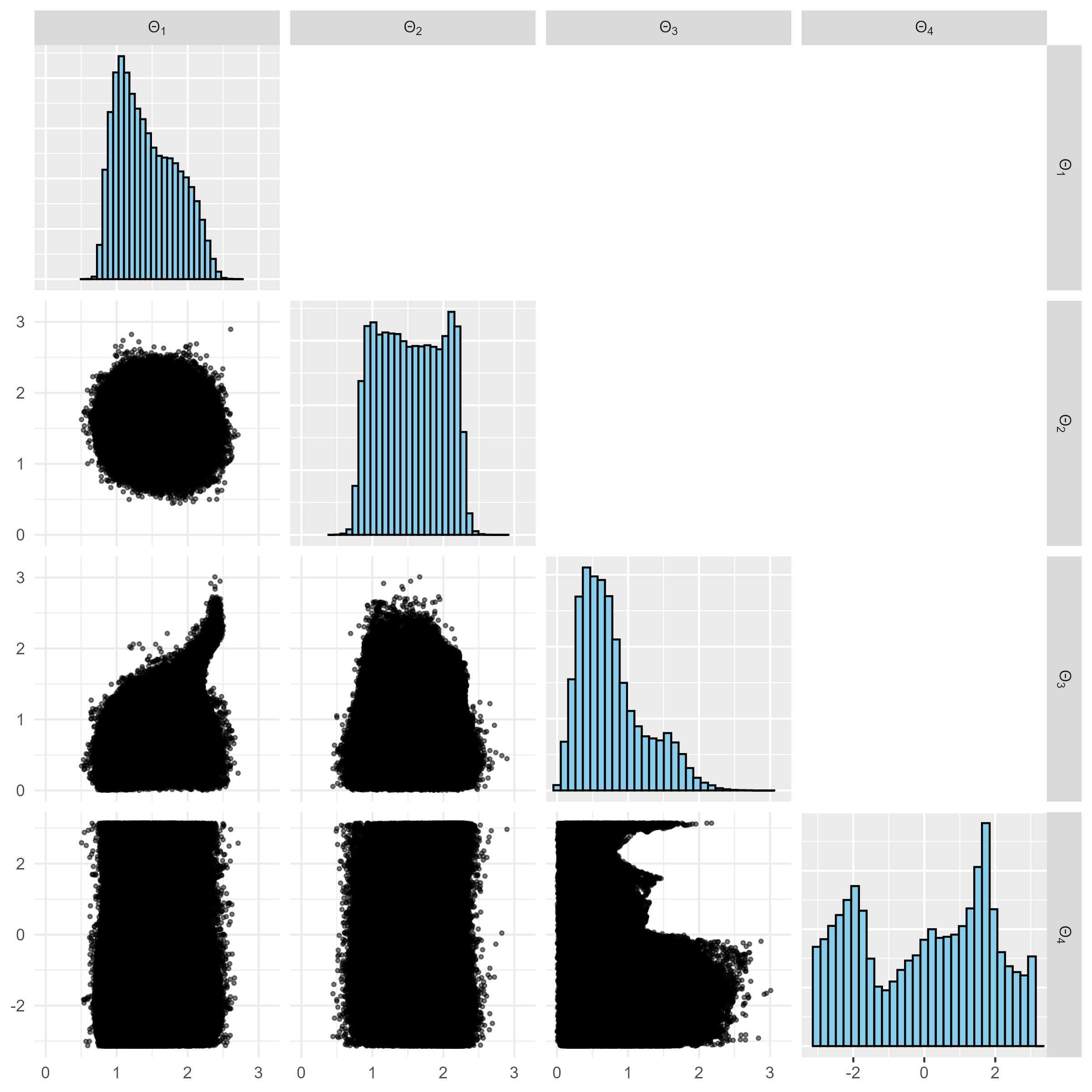}
    \caption{Pairwise scatterplots of spherical angles for metocean variables, alongside marginal histograms.}
    \label{fig:scatter}
\end{figure}

The pairwise scatterplots between the spherical angles of the normalised variables are shown in Figure~\ref{fig:scatter}, alongside the marginal histograms. The plots demonstrate the complex dependence structure exhibited by these data, particularly the fact there exist regions of the joint angular domain where we obtain no angular observations at all. These features make this data set a relevant example for testing the accuracy and flexibility of the angular modelling techniques introduced in Section~\ref{sec:methods}. We remark that while metocean processes are often assumed to be approximately stationary in time, the data exhibit non-negligible temporal dependence \citep{MACKAY2021110092}. Since de-clustering is non-trivial in the multivariate setting and best practices are yet to be established, we opt to treat the data as independent for the purpose of our illustrative analysis. Modelling approaches developed for independent data can still be successfully implemented for data exhibiting temporal dependence, so long as the additional uncertainty that arises from this feature is accounted for \citep{Kunsch1989,Politis1994}. See \citet{Murphy-Barltrop2024b} for further discussion, and \citet{Keef2013a} and \citet{Murphy-Barltrop2023} for examples within the extremes literature. We remark that one could alternatively apply techniques that explicitly account for temporal dependence, such as those which employ blocked assignment of observations to validation sets \citep{Richards2022,richards2023insights,Pasche2024a}.

Using the tuning parameters and architectures selected in Section~\ref{subsec:hyperparam}, we apply each of the methodologies from Section~\ref{sec:methods} and generate angular data. For this, a random subset consisting of 20\% of the full sample is computed. Each approach is applied to this subset, with the subsequent generated data compared to the remaining randomised 80\% of observations, which form our test set. With this, we aim to see whether features of the joint angular distribution can be captured using a relatively small subsample from the full data. This also accounts for the fact it is unlikely that we will have as many as $n=271\ 704$ observations in most practical applications.

Table~\ref{table:case_study_table} gives the skill scores for each of the deep generative approaches. As in Section~\ref{sec:simulation}, there is very little difference between the competing methods, yet the ordering of skill scores agrees with intuition when the visual diagnostics are considered. For example, the diagnostics from the GAN approach, as illustrated in Appendix~\ref{app:case_figs}, suggest a reasonable model fit, yet the upper and lower ends of some spherical variables are not well captured. Furthermore, the GAN simulations do not appear to completely capture the observed joint dependence structure. Moreover, it is encouraging to see that as in Section~\ref{sec:simulation}, the NFMAF and FM techniques appear to come out on top.

\begin{table}[!h]
\centering
\caption{$\operatorname{Skill}(F_{*})$ scores (to 6 significant figures) of each deep generative approach ($* \in \{\text{FM},\text{NFMAF},\text{GAN},\text{NFNSF} \} $) for the metocean data.}

\label{table:case_study_table}
\centering
\begin{tabular}[t]{c|c|c|c}
\hline
\multicolumn{4}{c}{\textbf{Model}}\\
\hline
FM & NFMAF & GAN & NFNSF\\
\hline
1.000299 & \textbf{1.000005} & 1.002580 & 1.000545 \\
\hline
\end{tabular}
\end{table}

\begin{figure}[h!]
    \centering
       \begin{subfigure}[b]{0.195\textwidth}
        \centering
        \includegraphics[width=\textwidth]{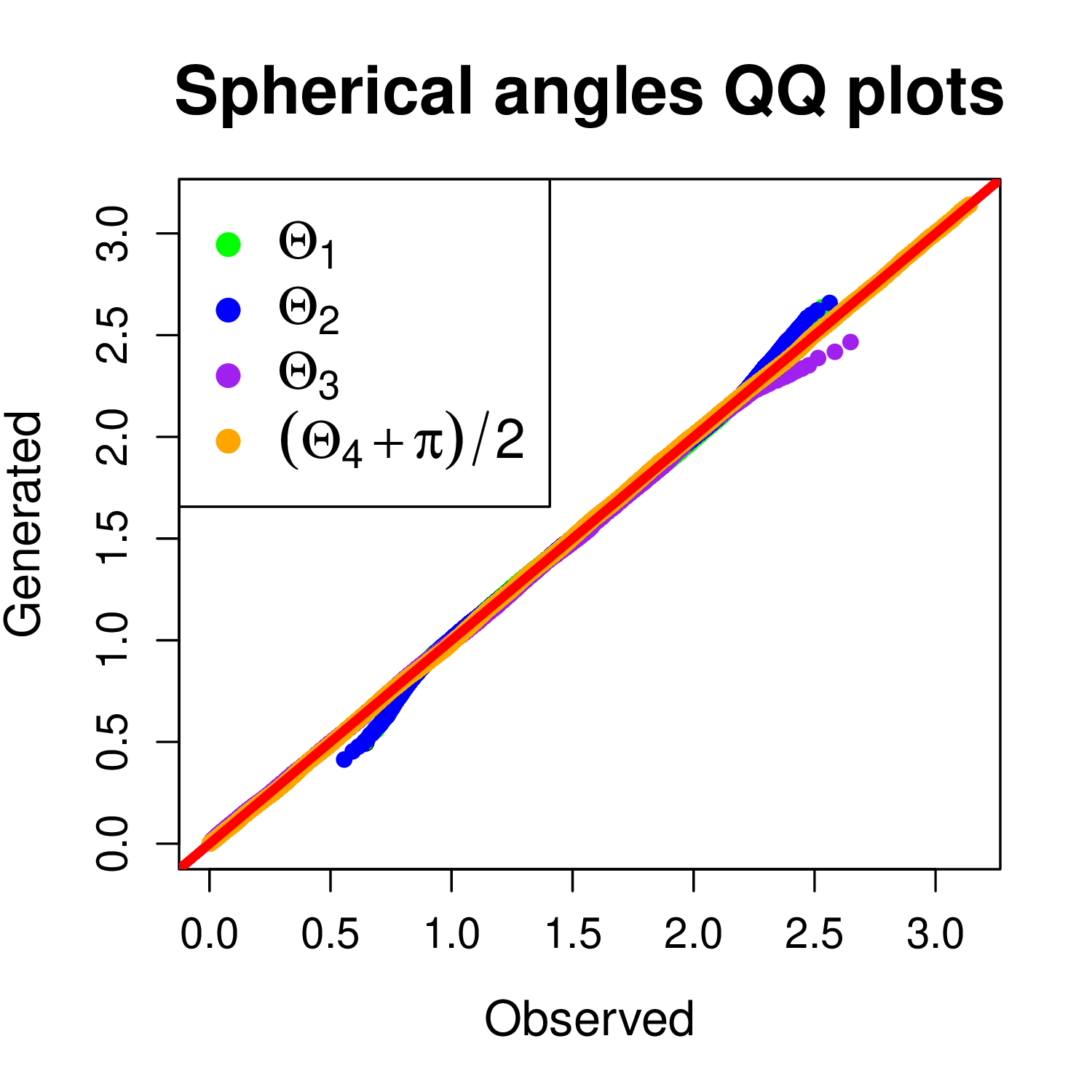}
    \end{subfigure}%
    \begin{subfigure}[b]{0.195\textwidth}
        \centering
        \includegraphics[width=\textwidth]{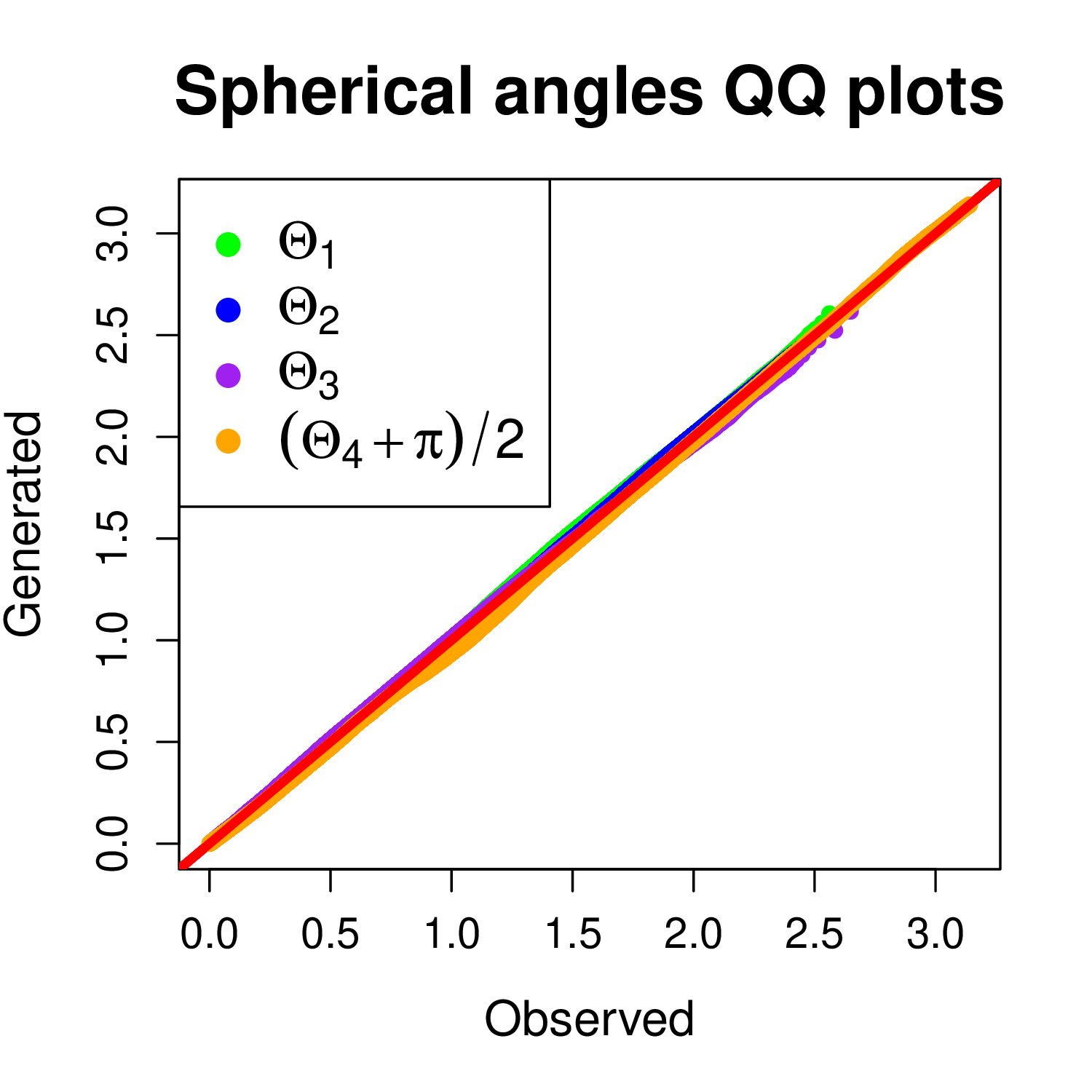}
    \end{subfigure}%
    \begin{subfigure}[b]{0.195\textwidth}
        \centering
        \includegraphics[width=\textwidth]{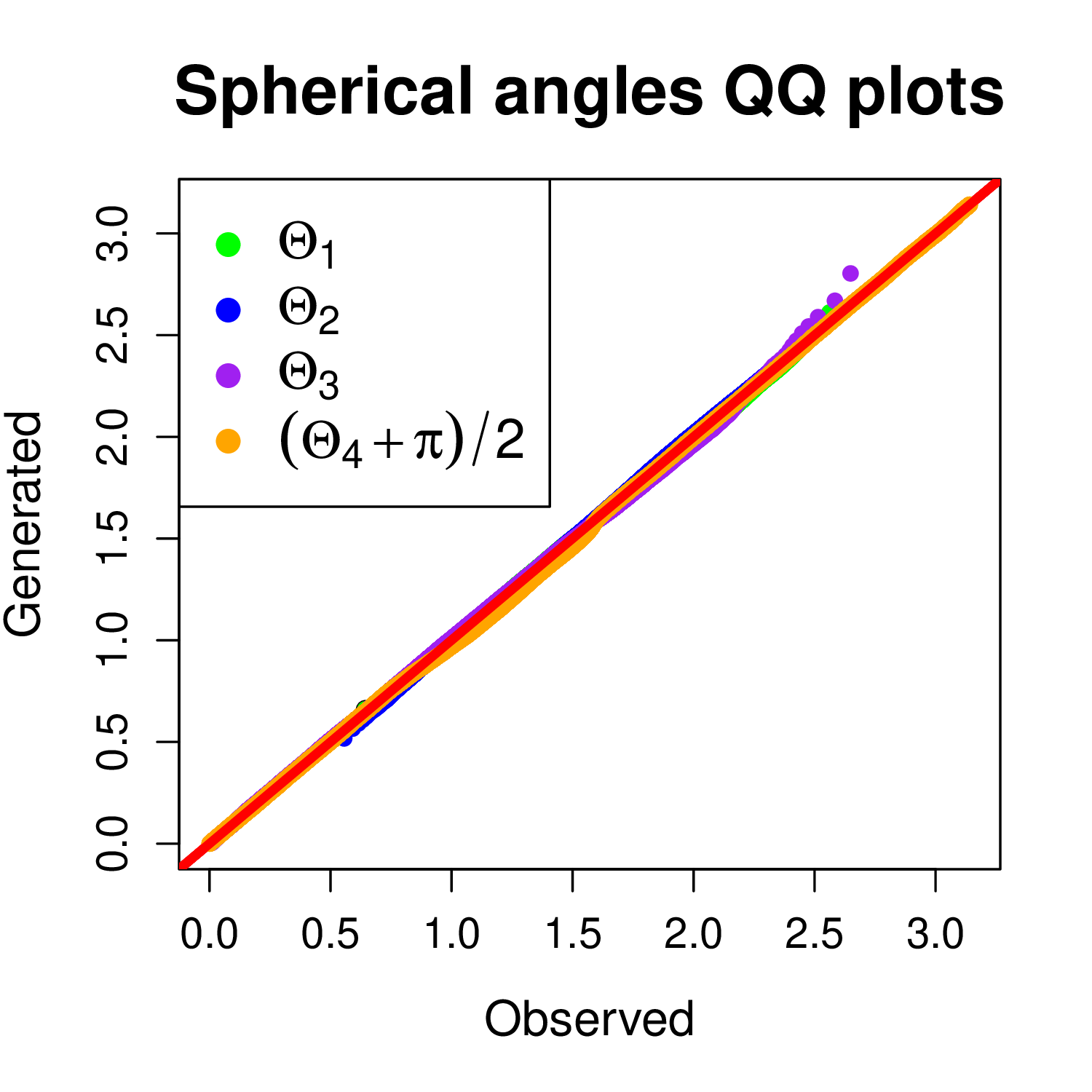}
    \end{subfigure}%
    \begin{subfigure}[b]{0.195\textwidth}
        \centering
        \includegraphics[width=\textwidth]{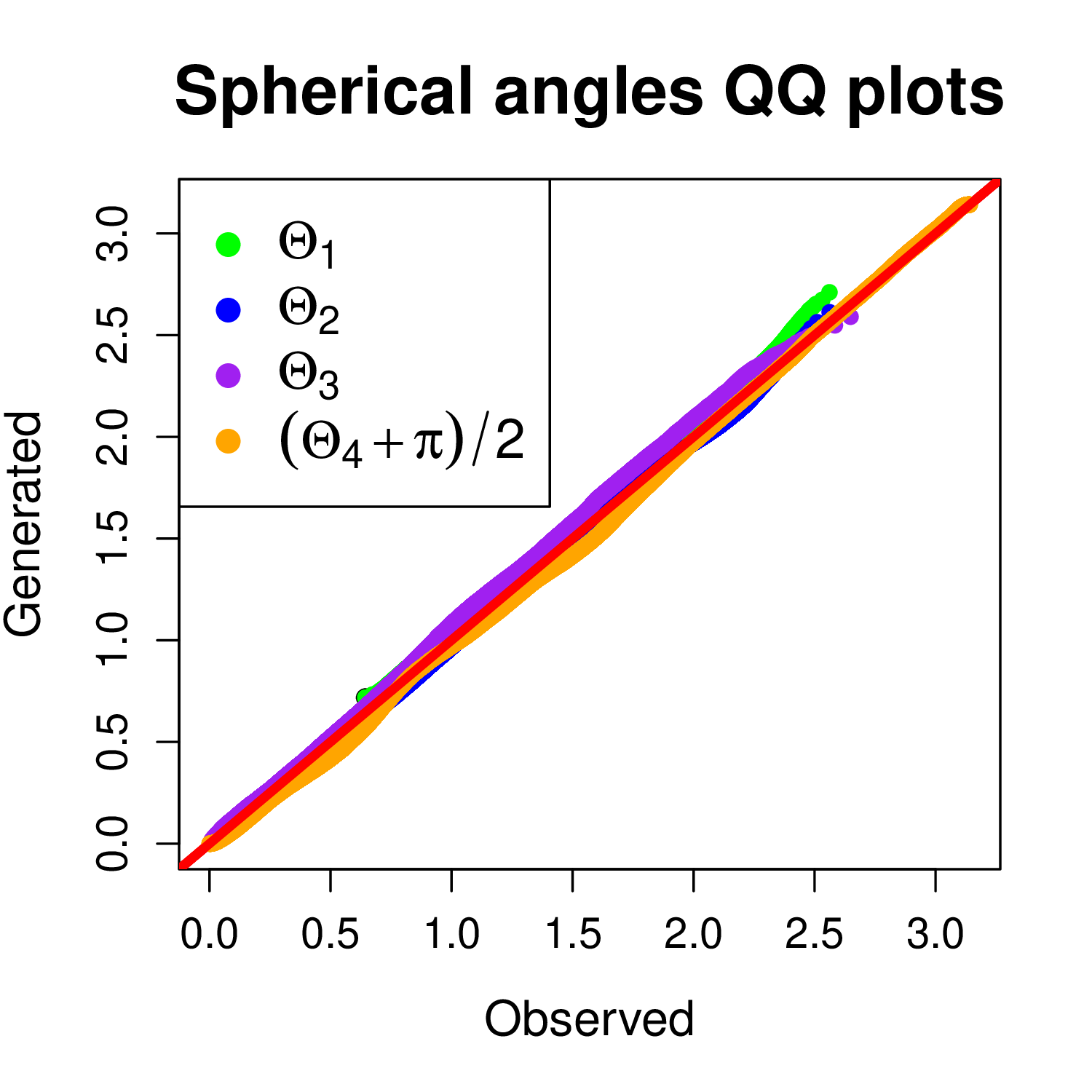}
    \end{subfigure}%
    \begin{subfigure}[b]{0.195\textwidth}
        \centering
        \includegraphics[width=\textwidth]{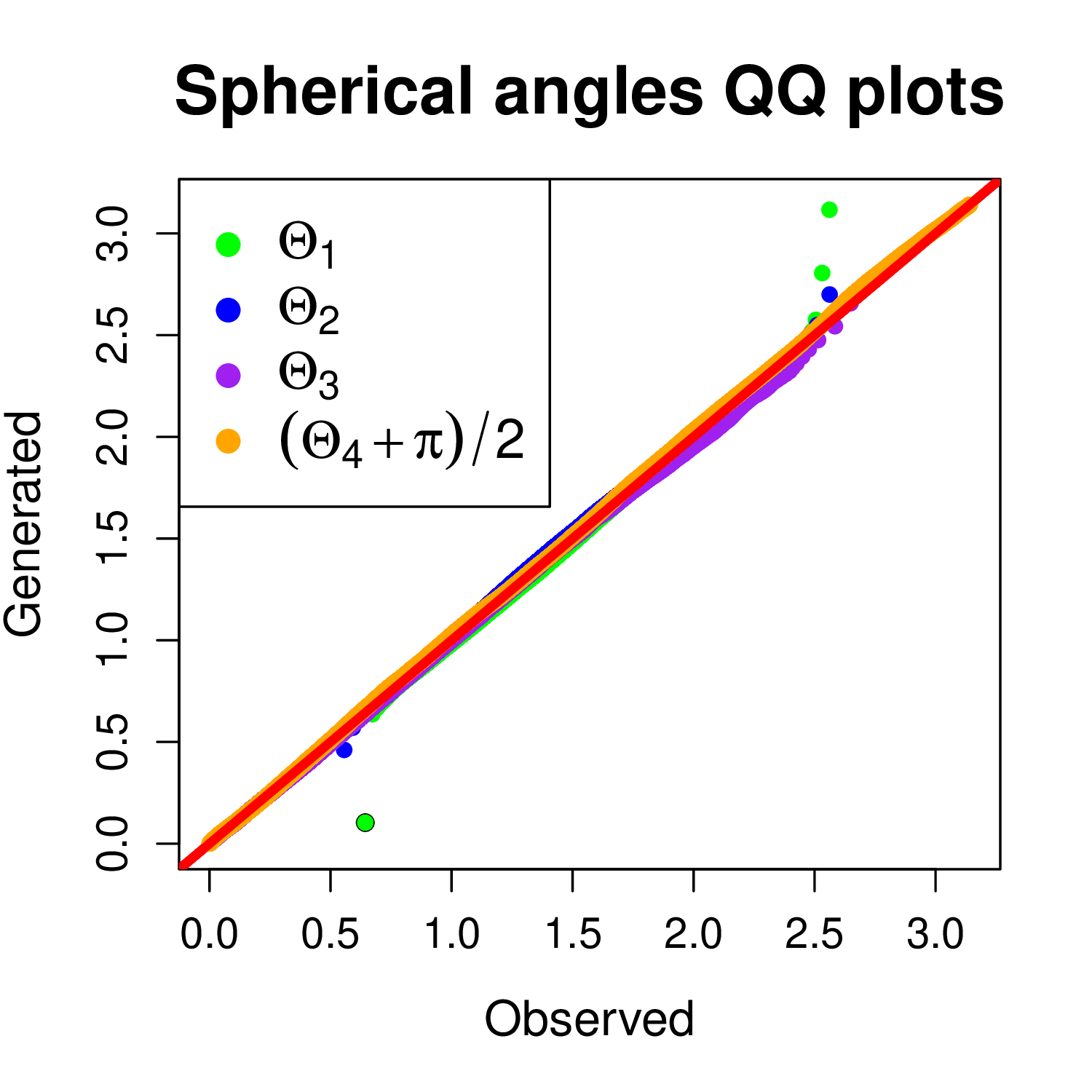}
    \end{subfigure}%

    \begin{subfigure}[b]{0.195\textwidth}
        \centering
        \includegraphics[width=\textwidth]{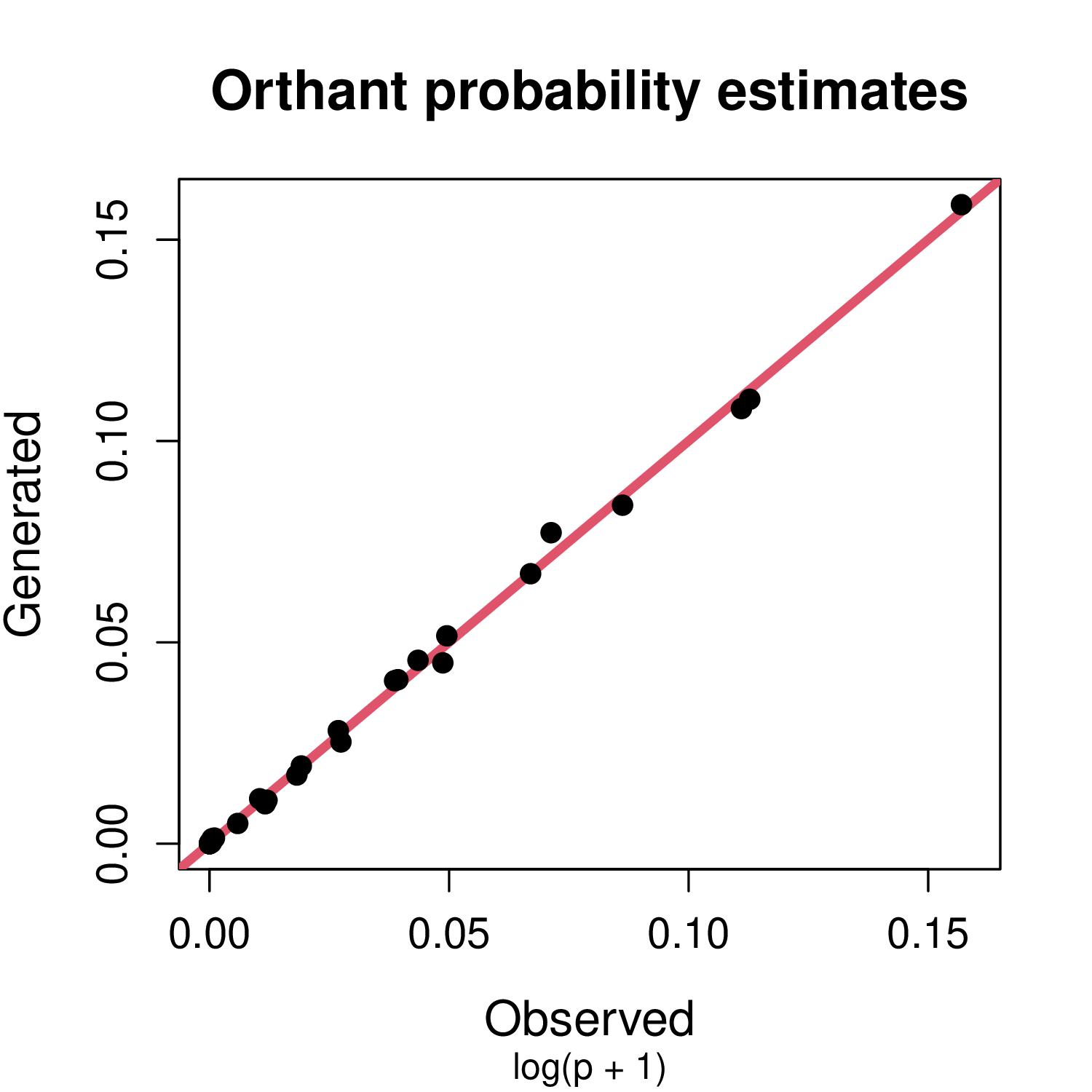}
        \end{subfigure}
    \begin{subfigure}[b]{0.195\textwidth}
        \centering
        \includegraphics[width=\textwidth]{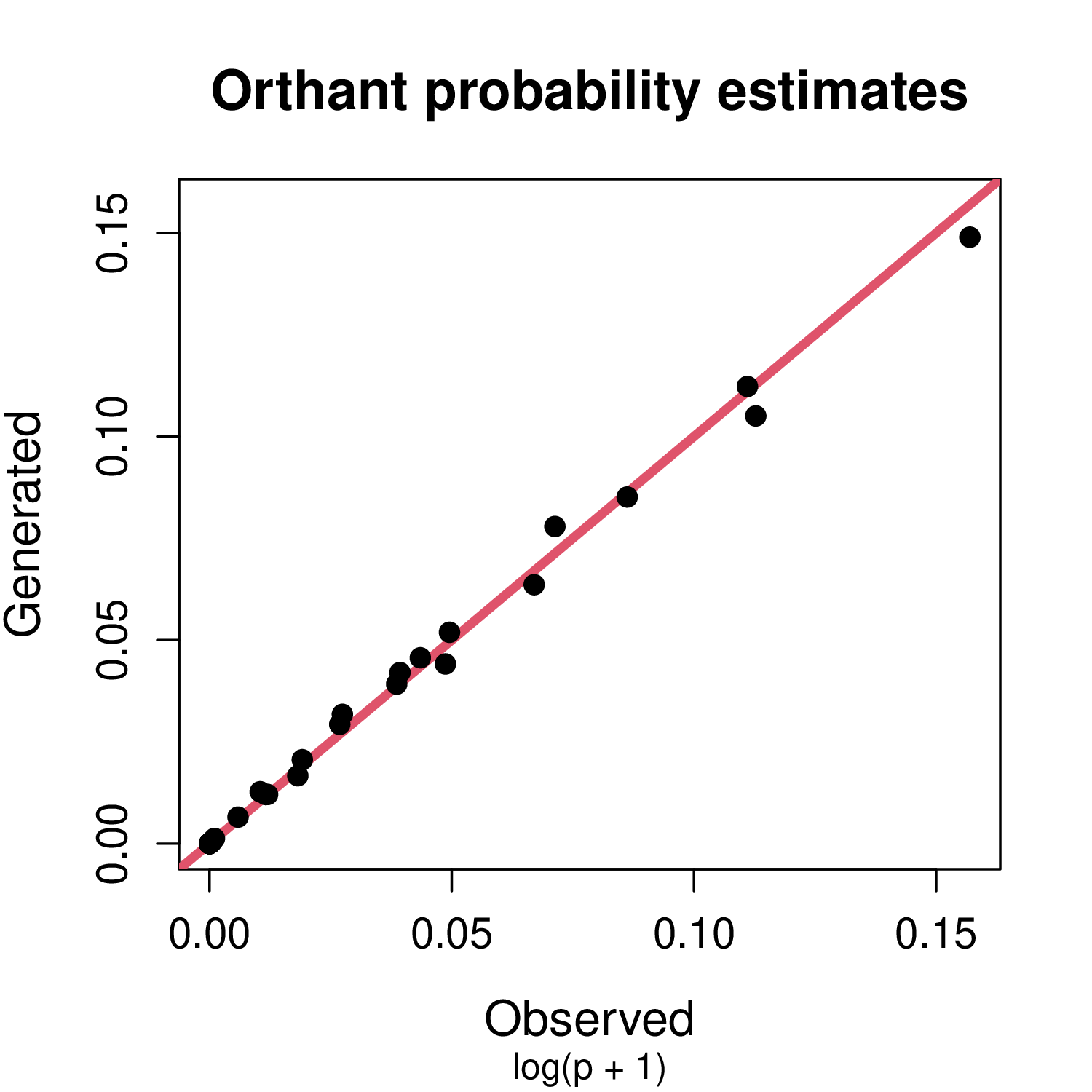}
    \end{subfigure}%
    \begin{subfigure}[b]{0.195\textwidth}
        \centering
        \includegraphics[width=\textwidth]{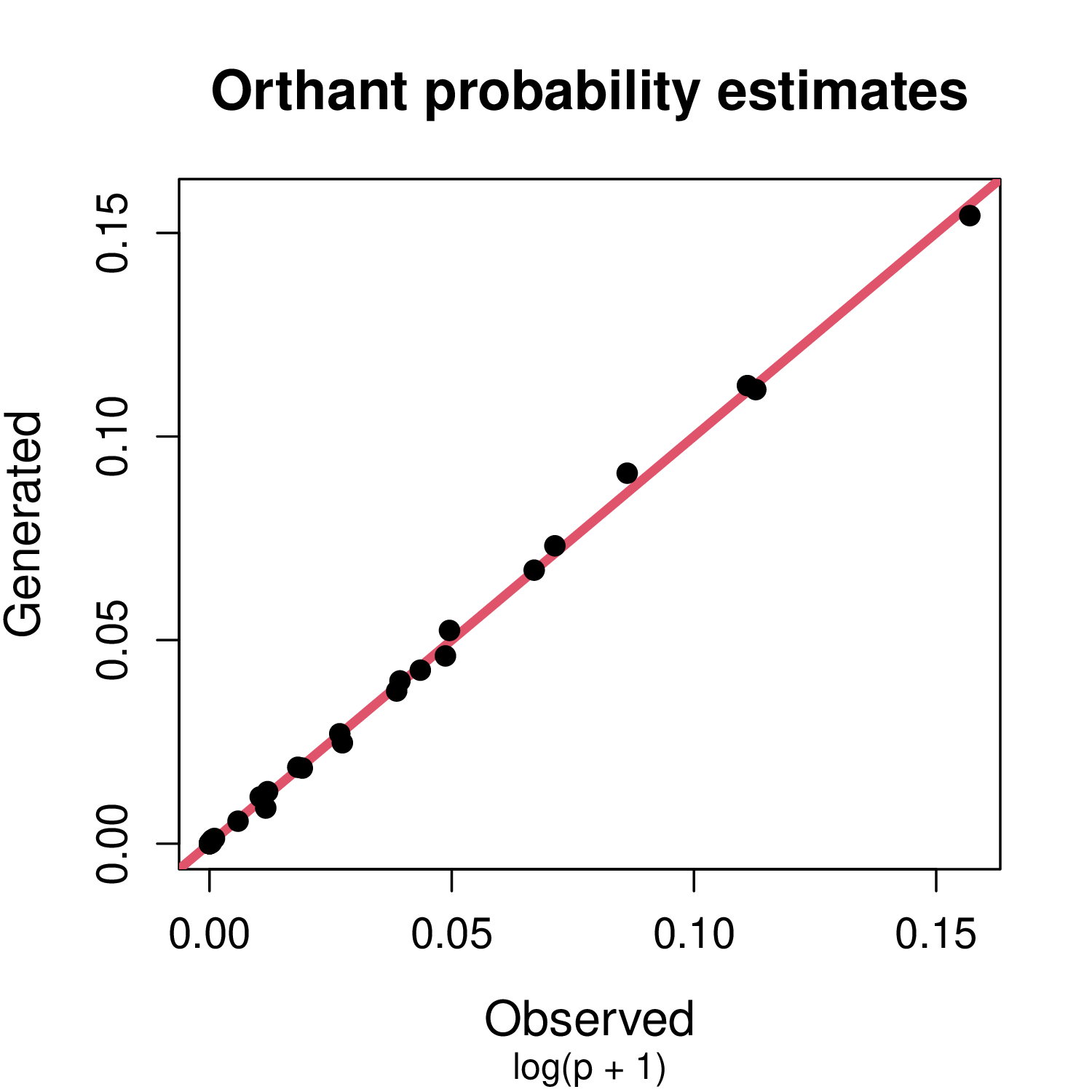}
    \end{subfigure}%
    \begin{subfigure}[b]{0.195\textwidth}
        \centering
        \includegraphics[width=\textwidth]{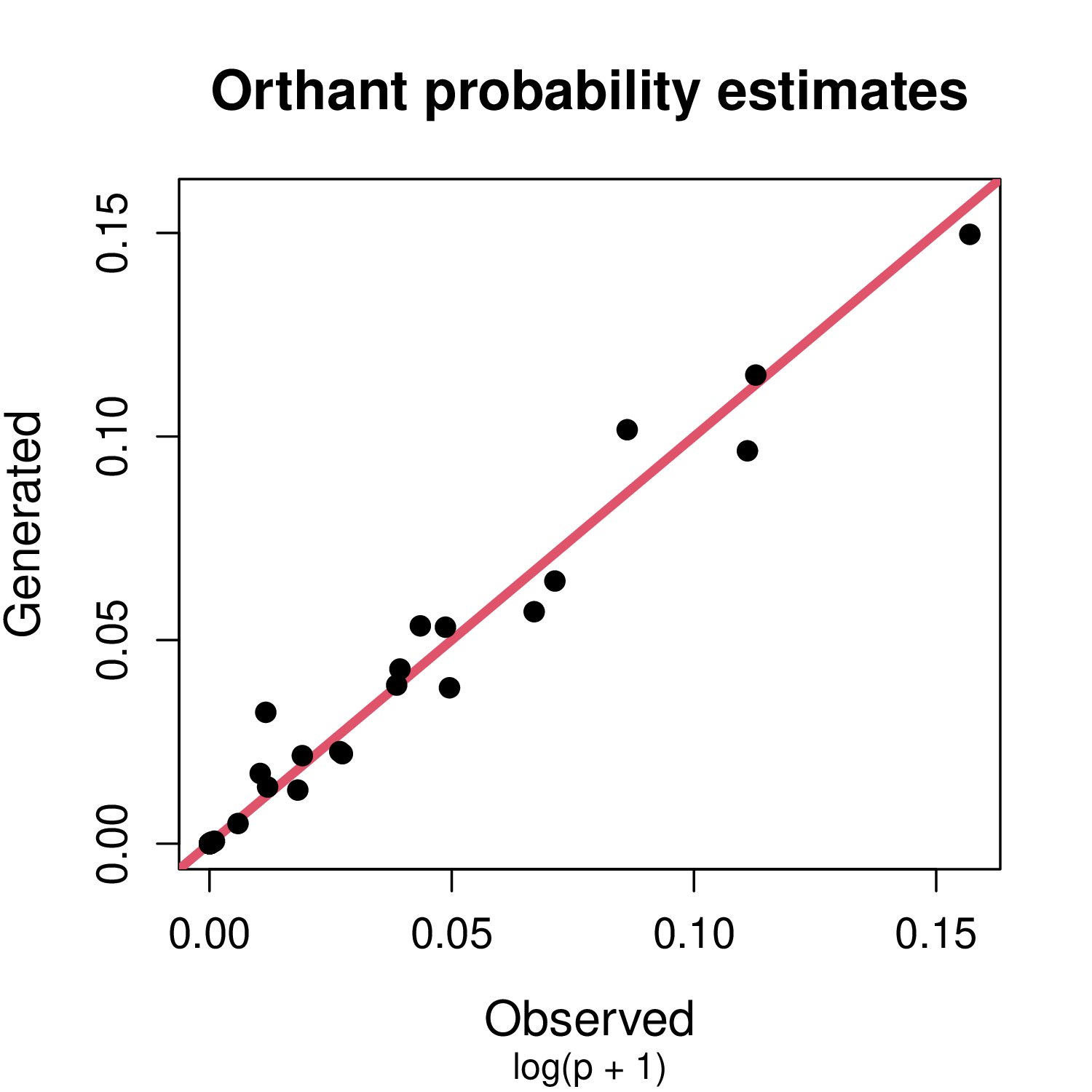}
    \end{subfigure}%
    \begin{subfigure}[b]{0.195\textwidth}
        \centering
        \includegraphics[width=\textwidth]{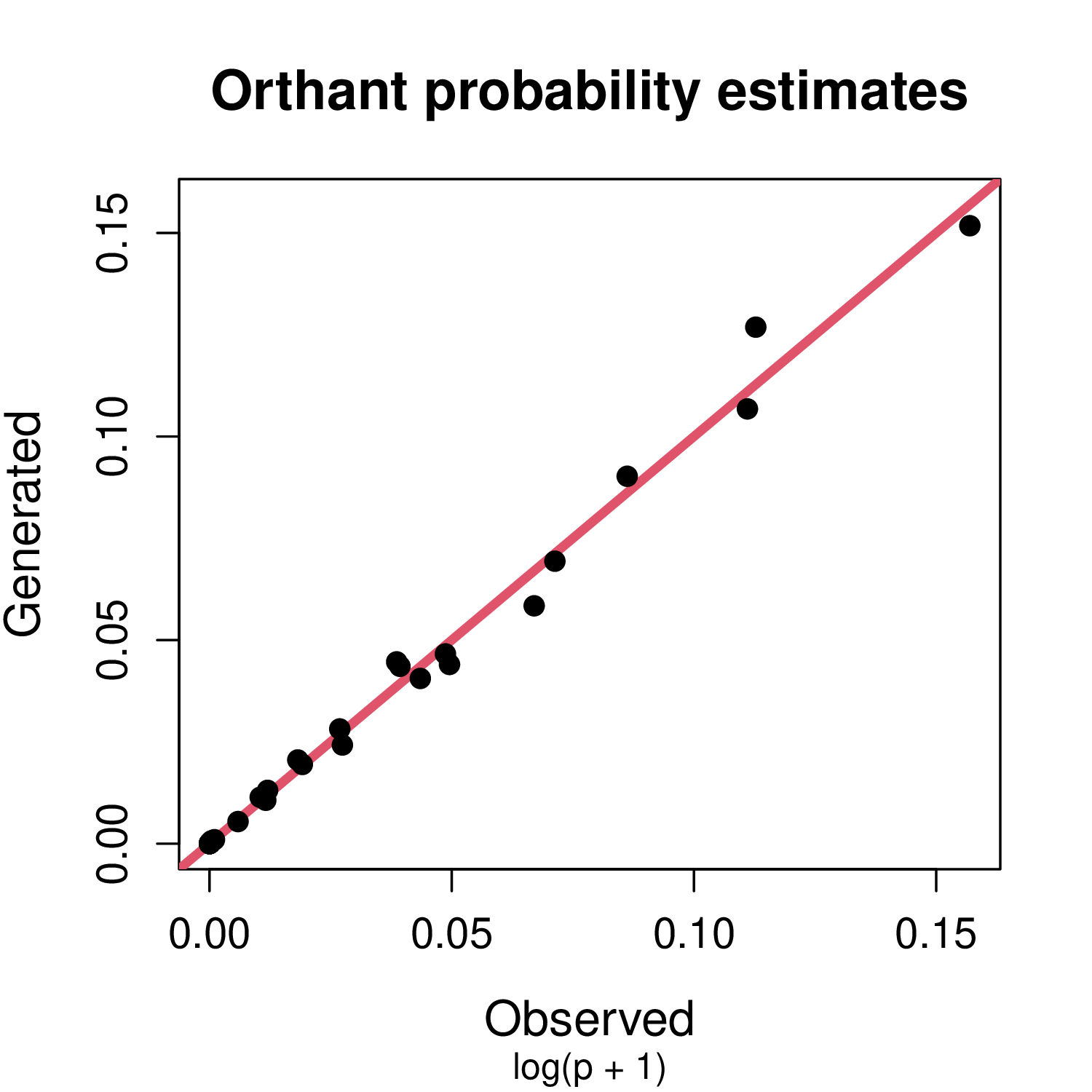}
    \end{subfigure}%
    \caption{Spherical angle QQ plots (top) and orthant probability plots (bottom) for the metocean data set. The ordering of methods is as in Figure~\ref{fig:qq_plots_cop2_d5}. }
    \label{fig:qq_orthant_plots_wave}
\end{figure}

Comparing the (relative) score values to visual diagnostics, we obtain consistency in our findings. Figure~\ref{fig:qq_orthant_plots_wave} gives the QQ and orthant probability plots for each deep learning approach. One can observe generally good performance across all approaches, yet the diagnostics from the FM and NFNSF techniques appear slightly superior to the other deep learning techniques.

Furthermore, when compared against the fitted vMF mixture model, the visual diagnostics indicate slightly superior performance for some of the deep generative methods. Take the NFMAF and FM approaches for instance; the scatterplot and histogram diagnostics for the former are given in Figures~\ref{fig:hist_wave_nf_naf}--\ref{fig:scatter_wave_nf_naf}, while those for the latter are presented in Appendix~\ref{app:case_figs}. For these techniques, the generated angular data appear almost indistinguishable from the validation set, indicating accurate model fits. The same cannot be said for the vMF mixture model, which produces non-negligible  discrepancies, especially when considering the angular marginal quantiles and pairwise scatterplots. The remaining diagnostics for the NFNSF and vMF approaches are also given in Appendix~\ref{app:case_figs}. 

\begin{figure}
    \centering
    \includegraphics[width=\linewidth]{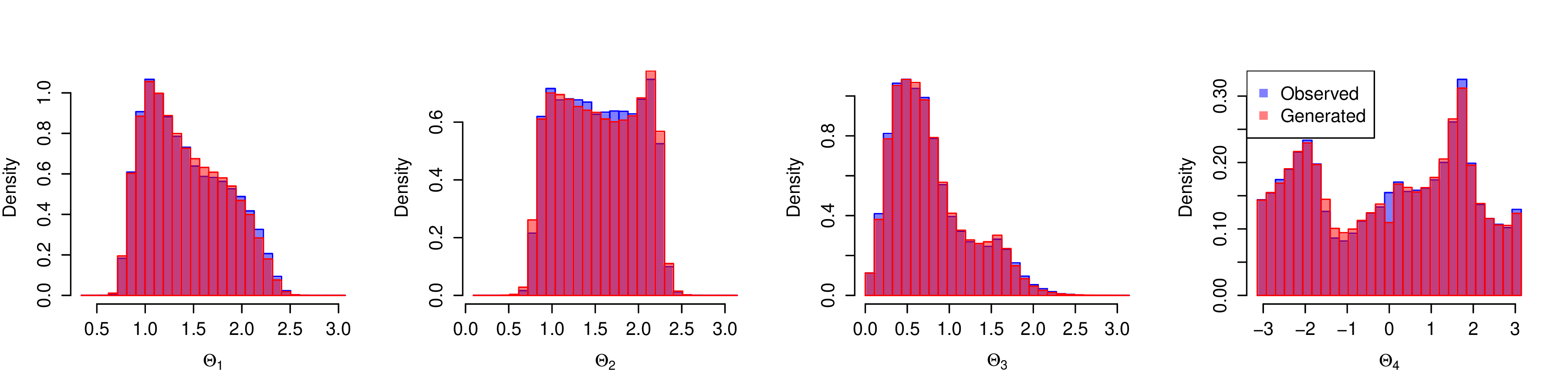}
    \caption{Histogram plots for spherical angles obtained from the NFMAF approach.}
    \label{fig:hist_wave_nf_naf}
\end{figure}

\begin{figure}
    \centering
    \includegraphics[width=.6\linewidth]{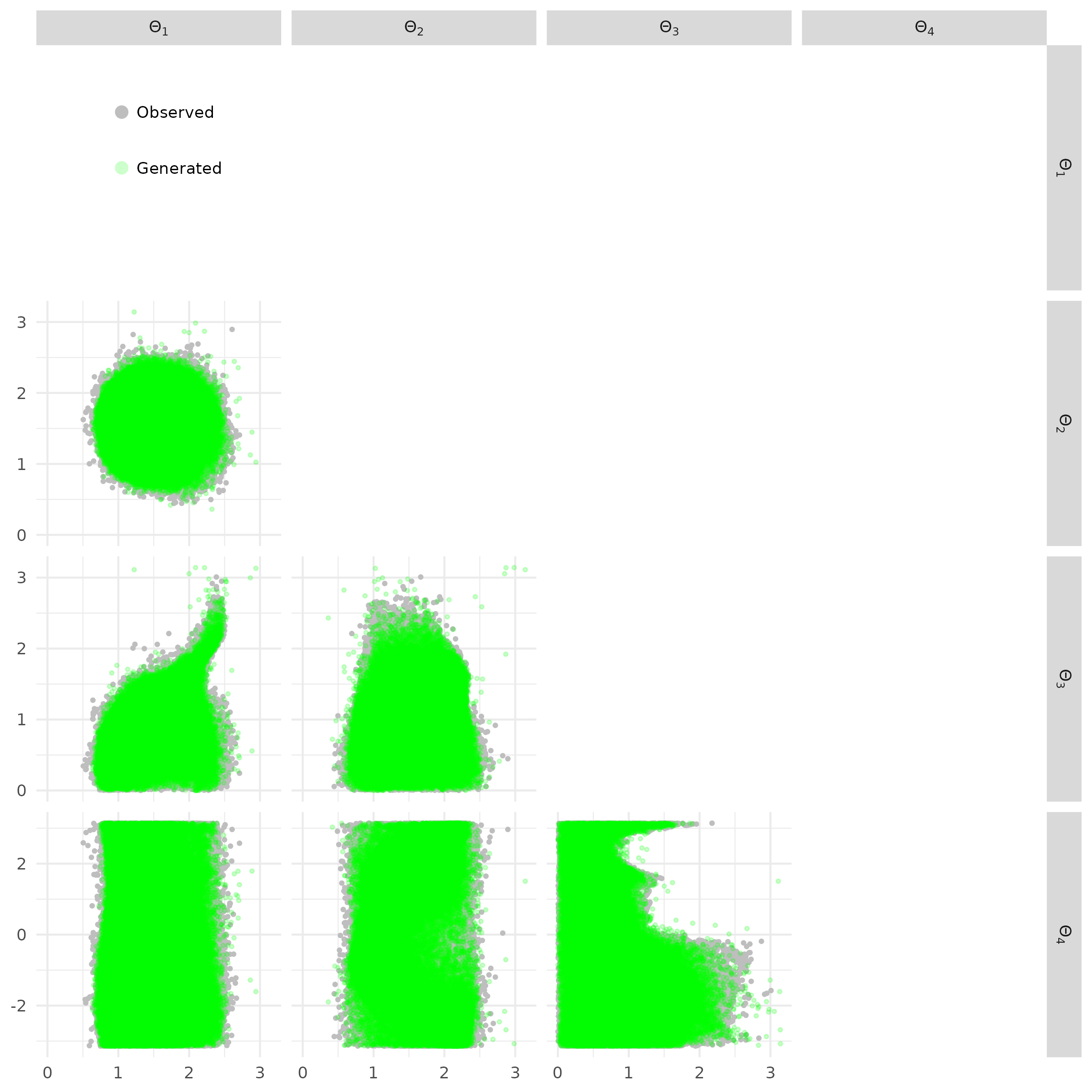}
    \caption{Pairwise scatterplots for spherical angles obtained from the NFMAF approach.}
    \label{fig:scatter_wave_nf_naf}
\end{figure}

Overall, we believe the results from this example data set clearly demonstrate the feasibility of deep generative approaches for non-parametrically modelling angular variables in applied settings, including those stemming from applications of extreme value theory. Generally, each tested approach appears to work well, but we have found certain deep learning techniques (namely FM and NFMAF) to give simulated angular distributions that are almost indistinguishable from the observed process. However, since there is no single approach that appears obviously superior to the others, we recommend in practice testing a range of deep generative techniques, comparing the resulting diagnostics and selecting the model that most closely matches the observations. 

\section{Conclusion and discussion}\label{sec:conclusion}

The aim of this paper was to introduce and investigate the use of several deep learning approaches in the task of modelling, and simulating from,  multivariate angular variables. The flexibility and scalability offered by these non-parametric approaches make them suitable candidates for this task. By studying results from a range of simulated data sets and a metocean application, and using a variety of validation metrics, we were able to thoroughly test the utility of these deep learning approaches in the context of angular simulation.

For many of the dependence structures we considered, the deep learning approaches were at least competitive with, if not superior to, the more classical approach of fitting a mixture of von Mises--Fisher distributions. This was particularly the case for the sparse data examples in the simulation study and the complex dependence structures observed in our metocean application. Therefore, while the baseline approach should in no way be discounted entirely, there appears to be room for deep learning approaches to be used in this domain. However, since there is no clear winner among the introduced deep learning techniques, further work is needed to establish best practices for applying such models.

One drawback of our simulation study is where we generate just a single data set for each combination of copula, marginal model, dimension and sample size. It is normally desirable within statistical inference procedures  to assess uncertainty in results, e.g., by implementing bootstrapping, with resampling and refitting carried out a large number of times. However, as discussed in Section~\ref{subsec:results}, the computational complexity of deep learning approaches, alongside the lack of interpretability for the aES metric, makes this somewhat infeasible for such a large scale comparative study. We also stress the need for reliable uncertainty quantification techniques in the deep learning setting. Despite these limitations, we believe that our findings still clearly demonstrate the feasibility and potential of deep learning approaches in the novel context of angular simulation.

As there was no clear individual `winner' in our study, i.e., no single method out-performed the others across all data sets, model selection is an important consideration. In particular, our results show that the type of metric used to assess the methods can influence their ranking. It is therefore wise to consider a range of validation metrics, and ensure that these are tailored to the application at hand. Our motivation for this study was primarily based on the use of angular-radial representations in multivariate extremes. In such settings, simulating angular data will generally be one step in a larger modelling procedure. In general, diagnostics will be tailored to the end modelling goal, but assessment of intermediate steps such as this can be useful for identifying potential areas for improvement. 

While the literature on validation metrics in general is very large, only a handful of these have been extended to angular variables; this offers a potential opportunity for further work. It would be useful to see a richer pool of methods for evaluation in this context, particularly the development of metrics that are able to fully assess the dependence features of angular distributions, i.e., without placing too much emphasis on marginal behaviour. Furthermore, it might be relevant to develop `local' metrics of performance, emphasizing important directional ranges during evaluation. The framework of weighted and kernelized scoring rules \citep{allen_weighted_2023} or of aggregation and transformation \citep{pic_proper_2025} could be used to extend the angular energy score to these ends.

We chose the generative approaches considered in this paper based on the most widely-used deep learning techniques in general settings. However, this is an area of very active research, and the introduction of new generative approaches, or the development of existing ones, has the potential to further improve upon the results shown here. For instance, advancements in computational efficiency could enhance the feasibility of applying bootstrapping techniques for uncertainty assessment. In addition, deep generative techniques are often criticised for their lack of theoretical guarantees. In our context, theoretical developments in terms of hyperparameter selection would be particularly welcome, as this is currently a time-consuming and delicate task.

To summarise, we believe that our study has shown the potential for deep learning approaches to be successfully used for angular variables, and hope that this will be a first step in the development of further generative approaches for angular modelling and simulation.

\vspace{0.25cm}\noindent{\bf Code}\\
Example code to implement the methods and goodness of fit metrics presented in this paper is available from the following GitHub repository: \url{https://github.com/callumbarltrop/DeGeMoH}.

\vspace{0.25cm}\noindent{\bf Supplementary Material}\\
{\bf Supplementary Figures:} A collection of .zip folders providing all of the generated visual diagnostics figures for the simulation study detailed in Section~\ref{sec:simulation}. These folders are freely available online via the following link: \url{https://datashare.tu-dresden.de/s/wNHmAfRJH25AwaX} (.zip files)

\bibliographystyle{apalike}
\bibliography{bibliography.bib}

\vspace{1cm}
\appendix

\renewcommand{\thefigure}{\AlphAlph{\value{figure}}}
\setcounter{figure}{0} 
\renewcommand{\thetable}{\Alph{table}}
\setcounter{table}{0} 

\newpage
{\bf\Large{Appendix}}
\section{Spherical coordinate transform}\label{app:spherical}

Here, we detail the relationship between the $d$-dimensional random vector $\boldsymbol{X}\in\mathbb{R}^d$ and the spherical coordinates $(R,\boldsymbol{\Theta})$, first introduced by \citet{blumenson1960derivation}. This is the decomposition of $\boldsymbol{X}$ used throughout our study for simulating angular variables. The Cartesian and spherical coordinates are related via the equations
\begin{align*}
  X_1 &= R \cos(\Theta_1), \\
  X_2 &= R \sin(\Theta_1) \cos(\Theta_2), \\
  X_3 &= R \sin(\Theta_1) \sin(\Theta_2) \cos(\Theta_3), \\
      &\qquad \vdots\\
  X_{d-1} &= R \sin(\Theta_1) \cdots \sin(\Theta_{d-2}) \cos(\Theta_{d-1}), \\
  X_d     &= R \sin(\Theta_1) \cdots \sin(\Theta_{d-2}) \sin(\Theta_{d-1}),
\end{align*}
with $R\geq 0$, $\Theta_1, \dots \Theta_{d-2}\in [0, \pi]$, and $\Theta_{d-1}\in(-\pi, \pi]$. To obtain $(R,\boldsymbol{\Theta})$ from $\boldsymbol{X}$, the equations above can simply be rearranged, yielding 
\begin{align*}
R             &=  \sqrt{{X_d}^2 + {X_{d-1}}^2 + \cdots + {X_2}^2 + {X_1}^2},     \\
\Theta_1     &= \text{atan2} \left({\textstyle \sqrt{{X_d}^2 + {X_{d-1}}^2 + \cdots + {X_2}^2}}, X_{1}\right), \\
\Theta_2     &= \text{atan2} \left({\textstyle \sqrt{{X_d}^2 + {X_{d-1}}^2 + \cdots + {X_3}^2}}, X_{2}\right), \\
       &\qquad \vdots\\
\Theta_{d-2} &= \text{atan2} \left({\textstyle \sqrt{{X_d}^2 + {X_{d-1}}^2}}, X_{d-2}\right), \\
\Theta_{d-1} &= \text{atan2} \left(X_d, X_{d-1}\right),
\end{align*}
where $\text{atan2}$ denotes the two argument arctangent function. We stress here that unlike the first $d-2$ angles, the last angle $\Theta_{d-1}$ is defined on the half-open set $(-\pi, \pi]$ and is cyclic in nature. We note this transformation is not one-to-one. 

\section{Neural network details}\label{app:NNdetails}
\subsection{Multi-layer perceptrons}

The neural network models used in this study are built from so-called \textit{multi-layer perceptrons} (MLPs), also known as \textit{feedforward neural networks} \citep[Ch.~6]{Goodfellow-et-al-2016}. MLPs are a loosely biologically-inspired machine learning architecture. They define a mapping from an input vector $\boldsymbol{x} \in \mathbb{R}^n$ to an output vector $f(\boldsymbol{x}) \in \mathbb{R}^m$ via a sequence of affine transformations, followed by non-linear transformation functions. Formally,
\begin{equation}\label{eq:mlp}
    f(\boldsymbol{x}) = \phi_L(\boldsymbol{A}_L\, \phi_{L-1}(\cdots \phi_2(\boldsymbol{A}_2\phi_1(\boldsymbol{A}_1 \boldsymbol{x} + \boldsymbol{b}_1) + \boldsymbol{b}_2)\cdots ) + \boldsymbol{b}_L),
\end{equation}
where, for a given $\ell \in \{1,\hdots,L \}$, $\boldsymbol{A}_\ell$ and $\boldsymbol{b}_\ell$ denote weight matrices and bias vectors, respectively, and $\phi_\ell$ is a non-linear activation function, typically applied elementwise. The number of layers corresponds to the number of transformations $L$, while the row dimension of $\boldsymbol{A}_\ell$ gives the number of neurons in that layer. The resulting parameter vector for a MLP is given by $\boldsymbol{\Gamma} := (\boldsymbol{A}_\ell, \boldsymbol{b}_\ell)_{\ell = 1}^L$. 

The activation functions in Eq.~\eqref{eq:mlp} ensure that the resulting mapping is non-linear and allow the composition to approximate highly flexible transformations. Commonly-used activation functions include the Rectified Linear Unit (ReLU)
\[
\text{ReLU}(x) = \max(0, x),
\]
and its extension in the LeakyReLU
\[
\text{LeakyReLU}(x):
=
\begin{cases}
x, & x \ge 0,\\[2mm]
\alpha x, & x < 0,
\end{cases}\qquad \alpha > 0,
\]
which we use for the generator and discriminator components of the GAN. Here $\alpha$ is often chosen very small, $\alpha = 0.2$ in our case. For the flow matching models we use a Swish activation \citep{ramachandran_swish_2017}, defined by
\[
\mathrm{Swish}(x) = \frac{x}{1 + e^{-\beta x}}, \qquad \beta > 0,
\]
which interpolates smoothly between a linear function and the ReLU activation;  we use $\beta = 1$. As outlined in Section~\ref{sec:methods}, the activation functions in the final layer can be used to constrain the output to certain domains. For example, we use a sigmoid function for the first $d-2$ coordinates of the GAN model to ensure the outputs are bounded in $[0,\pi]$.

MLP models are trained using \textit{stochastic gradient descent} (SGD) methods. For this article, we opt to use the popular Adam algorithm \citep{kingma_adam_2017}. Suppose $\mathcal{Y}$ denotes a training data set: SGD methods work by evaluating a loss function, $\mathcal{L}(\cdot)$, over batches (small subsets) of the training data, say $\boldsymbol{y} \subset \mathcal{Y}$. The parameters of the MLP are then updated in the direction of the loss function gradients, i.e., $\nabla_{\boldsymbol{\Gamma}} \mathcal{L}(\boldsymbol{y} | \boldsymbol{\Gamma})$. This is computed via \textit{backpropagation}, which applies the chain rule through the nested structure described in Eq.~\eqref{eq:mlp}. The appropriate loss functions vary across the different modelling frameworks introduced in this paper: the GAN model use the min-max formulation described in Eq.~\eqref{eqn:GANtarget}, the normalizing flows rely on the log-likelihood, and the flow matching models minimize the loss in Eq.~\eqref{eq:fm_loss}. For more details on the individual model architectures, including the choices of layers and neurons, we refer to Section~\ref{subsec:hyperparam}.

\subsection{Normalizing flows}\label{app:normflows}

As outlined in Section~\ref{subsec:normflows}, normalizing flows define a model via a sequence of diffeomorphic transformations (so-called \emph{flows}), mapping a simple base distribution to a more flexible one. MLPs are usually not invertible, meaning additional constraints are necessary. In this work, we consider two types of flow models: masked autoregressive flows (MAF) and neural spline flows (NSF), outlined in the following.

\subsubsection{Masked autoregressive flows}

In each flow, a MAF \citep{papamakarios_masked_2017} defines each output coordinate as a scaled and shifted version of the corresponding input coordinate, conditioned on the preceding coordinates, i.e.,
\[
x_i  = \mu_i(\boldsymbol{z}_{<i})
+
\exp\left[\sigma_i(\boldsymbol{z}_{<i})\right]\, z_i,
\qquad i=1,\ldots,d.
\]
Here $\mu_i \in \mathbb{R}$ and $\sigma_i\in \mathbb{R}$ are outputs of an MLP whose input is restricted to $\boldsymbol{z}_{<i}$ by binary masks. This transformation is immediately invertible even if the MLP model (the so-called \textit{conditioner}) is not. The restriction of the inputs $\boldsymbol{z}_{<i}$ to the MLPs ensures that the Jacobian of the inverse transformation (required for the log-likelihood) is diagonal, meaning that its determinant is easily computable during training and reduces to
\[
\log\left|
\det \nabla_{\boldsymbol{z}} T(\boldsymbol{z})
\right|
=
\sum_{i=1}^d \sigma_i(\boldsymbol{z}_{<i}).
\]
Thus, MAFs allow for exact and efficient likelihood evaluation, whilst being able to approximate highly complex transformations. In fact, they are universal function approximators. A full MAF is obtained by concatenating multiple affine transformations and switching the order of coordinates between blocks. We use three-layer MLPs with 64 hidden units and ReLU activations for the MAF flows.

\subsubsection{Neural spline flows}

The NSF \citep{durkan_neural_2019} transformation replaces the affine mapping in each autoregressive step by a monotonic rational--quadratic spline. For each $i = 1, \dots, d$, we set
\[
x_i = S_i\!\left(z_i;\,
\boldsymbol{\Omega}_i(\boldsymbol{z}_{<i})\right),
\]
where $S_i$ is an increasing rational--quadratic spline with $K$ knots, and the spline parameters $\boldsymbol{\Omega}_i$ (the derivatives at the knots) are generated by an MLP receiving $\boldsymbol{z}_{<i}$ as input. The rational-quadratic in the $k$th bin (between knots $\{x^{(k)}, y^{(k)}\}$ and $\{x^{(k+1)}, y^{(k+1)}\}$) is given as
\[
f(\xi) = y^k + \frac{(y^{(k+1)} - y^{(k)}) [s^{(k)} \xi^2 + \delta^{(k)}\xi(1-\xi)]}{s^{(k)} + [\delta^{(k+1)} + \delta^{(k)} - 2s^{(k)}]\xi(1-\xi)}, 
\]
where $\delta^{(k)}$ are the derivatives at each knot $k$ and with 
\[
s^{(k)} = \frac{y^{(k+1)}-y^{(k)}}{x^{(k+1)} - x^{(k)}}, \quad 
\xi(z) = \frac{z-x^{(k)}}{x^{(k+1)}-x^{(k)}}.
\]
Monotonicity of the spline ensures invertibility. Moreover, the structure of the conditioner ensures the Jacobian of the inverse mapping has diagonal structure, meaning the determinant is easily computable.  

NSFs are composed of richer transformations compared to the affine ones of MAF flows. They are stacked in a similar way by concatenating transformations and changing the order of coordinates in between. Again, we use a three-layer MLP with 64 units and ReLU activations, while we specify $K = 7$ equally spaced knots (leading to $8$ bins) on the respective domains of the angular coordinates for the spline formulation.

\section{Propriety of the angular energy score}\label{app:aES}
We defined the angular energy score (aES) in Definition \ref{def:aES}. This provides a multivariate generalisation of the circular CRPS (cCRPS), analogous to how the energy score generalises the classical CRPS. The aES reduces to the cCRPS for distributions and observations on the 1-sphere (i.e., unit circle).  We now show the propriety of this score. Let $\mathcal{D}_\alpha$ be the set of distributions on the $(d-1)$-sphere with finite entropy, i.e., $F \in \mathcal{D}_\alpha \Rightarrow \mathbb{H}_\alpha (F) := \tfrac{1}{2}\int_{\mathbb{S}^{d-1}}\int_{\mathbb{S}^{d-1}} \alpha(\boldsymbol{w},\boldsymbol{w}')\mathrm{d}F(\boldsymbol{w})\mathrm{d}F(\boldsymbol{w}') < \infty.$\\

\begin{proposition}[Proposition~\ref{prop:aES} restated]
The angular energy score is a \textit{proper} scoring rule with regard to all measures in $\mathcal{D}_\alpha$, meaning that for all $F, G \in \mathcal{D}_\alpha$ we have:
\begin{equation}\label{eq:propriety}
    \operatorname{aES}(G, G) \leq \operatorname{aES}(F, G),
\end{equation}
where $\operatorname{aES}(F, G) = \mathbb{E}_{g \sim G} \operatorname{aES}(F, g)$ corresponds to the induced divergence.
\end{proposition}

\begin{proof}
    The angular energy score in Definition \ref{def:aES} is a kernel score, with the angular distance given as the symmetric kernel $\alpha(\boldsymbol{x},\boldsymbol{y}) = \cos^{-1}(\boldsymbol{x}^T\boldsymbol{y})$. As we show below, this kernel is conditionally negative definite. Thus, according to Theorem 15 of \cite{waghmare_proper_2025}, the associated kernel score is proper relative to $\mathcal{D}_\alpha$.

    In order to show that the angular measure is a conditionally negative definite kernel we can consider the power series representation:
    $$\alpha(\boldsymbol{x},\boldsymbol{y}) = \frac\pi2 - \sum_{k\ge0}\binom{2k}{k}\frac{(\boldsymbol{x}^T \boldsymbol{y})^{2k+1}}{4^k(2k+1)}.$$
    The second part of this sum is a positive definite (p.d.) kernel, as $\boldsymbol{x}^T \boldsymbol{y}$ is a p.d.\ kernel (the so-called linear kernel), exponents and conical sums of p.d.\ kernels are p.d., and, if it exists, the limit of a sequence of p.d.\ kernels is p.d. A constant minus a p.d.\ kernel leads to a conditionally negative definite kernel. This proof of the conditional negative definiteness of the angular measure was first stated in \cite{mathoverflow}.
\end{proof}

The angular energy score is an intuitive construction of a kernel score on the hypersphere, using the natural angular measure as a kernel. By similarity with the classical energy score, one could expect this score to be strictly proper, meaning that equality in Eq.~\eqref{eq:propriety} holds only when $F=G$. However, that is not the case, as the following example shows\footnote{This counterexample was inspired by interactions with the generative AI model ChatGPT. The details were developed and checked by the authors. The authors have not found this counterexample in the published literature.}.

Let $\boldsymbol{u}=(1,0)$ and $\boldsymbol{v}=(0,1)$ and define
\[
F=\tfrac12(\delta_{\boldsymbol{u}}+\delta_{-\boldsymbol{u}}),\qquad G=\tfrac12(\delta_{\boldsymbol{v}}+\delta_{-\boldsymbol{v}}),
\]
where $\delta_{\boldsymbol{x}}$ is the Dirac density at $\boldsymbol{x}$. We will show $\operatorname{aES}(G,G) = \operatorname{aES}(F,G)$, or equivalently:
\begin{align*}
    \operatorname{aES}(F,G) - \operatorname{aES}(G,G) &= \mathbb{E}_{\boldsymbol{w} \sim F,\boldsymbol{w}' \sim G} \left[\alpha(\boldsymbol{w},\boldsymbol{w}')\right] - \frac{1}{2} \mathbb{E}_{\boldsymbol{w}, \boldsymbol{w}' \sim F}\left[\alpha(\boldsymbol{w},\boldsymbol{w}')\right] - \frac{1}{2} \mathbb{E}_{\boldsymbol{w}, \boldsymbol{w}' \sim G}\ \left[\alpha(\boldsymbol{w},\boldsymbol{w}') \right]  \\
    &= \frac{1}{2} \int_{\mathbb{S}^{d-1}}\int_{\mathbb{S}^{d-1}} \alpha(\boldsymbol{w}, \boldsymbol{w}') \mathrm{d}(G-F)(\boldsymbol{w}) \mathrm{d}(G-F)(\boldsymbol{w}') \\
    &= 0
\end{align*}
With the above $F$ and $G$ the last integral becomes:
\begin{align}\label{eq:quadratic_form}
    \frac{1}{2} \int_{\mathbb{S}^{d-1}}\int_{\mathbb{S}^{d-1}} \alpha(\boldsymbol{w}, \boldsymbol{w}') \mathrm{d}(G-F)(\boldsymbol{w}) \mathrm{d}(G-F)(\boldsymbol{w}') = \frac{1}{2} \sum_{i=1}^4 \sum_{j = 1}^4 p(\boldsymbol{w}_i) p(\boldsymbol{w}_j) \alpha(\boldsymbol{w}_i, \boldsymbol{w}_j)
\end{align}
where $(\boldsymbol{w}_i)_{i=1}^4 = (\boldsymbol{w}'_i)_{i=1}^4 = (\boldsymbol{v}, -\boldsymbol{v}, \boldsymbol{u}, -\boldsymbol{u})$ and $p(\boldsymbol{v}) = p(-\boldsymbol{v}) = \tfrac{1}{2}, p(\boldsymbol{u}) = p(-\boldsymbol{u}) = -\tfrac{1}{2}$. Using the definition of the angular measure, we have:
\begin{itemize}
    \item $\alpha(\boldsymbol{v},\boldsymbol{v}) = \alpha(\boldsymbol{u},\boldsymbol{u}) = 0$.
    \item $\alpha(\boldsymbol{v},-\boldsymbol{v}) = \alpha(\boldsymbol{u},-\boldsymbol{u}) = \pi$.
    \item $\alpha(\boldsymbol{v}, \boldsymbol{u}) = \alpha(\boldsymbol{v}, -\boldsymbol{u}) = \alpha(-\boldsymbol{v}, \boldsymbol{u}) = \alpha(-\boldsymbol{v}, -\boldsymbol{u}) = \tfrac{\pi}{2}$.
\end{itemize}
The rest of the cases follow from the symmetry of $\alpha(\boldsymbol{w},\boldsymbol{w}')$. Straightforward computations show that the quadratic form in Eq.~\ref{eq:quadratic_form} is therefore zero, illustrating that $\operatorname{aES}(G,G) = \operatorname{aES}(F,G)$ but $F\neq G$.

Some characterisations of strongly positive kernels on the hypersphere are given by \cite{gneiting_strictly_2013} and \cite{ steinwart_strictly_2021}. We defer the definition of a \emph{strictly} proper scoring rule on the hypersphere to future work.

\section{Additional simulation study figures and results}\label{app:add_sim_study}
\begin{figure}[!h]
    \centering
    \includegraphics[width=0.9\linewidth]{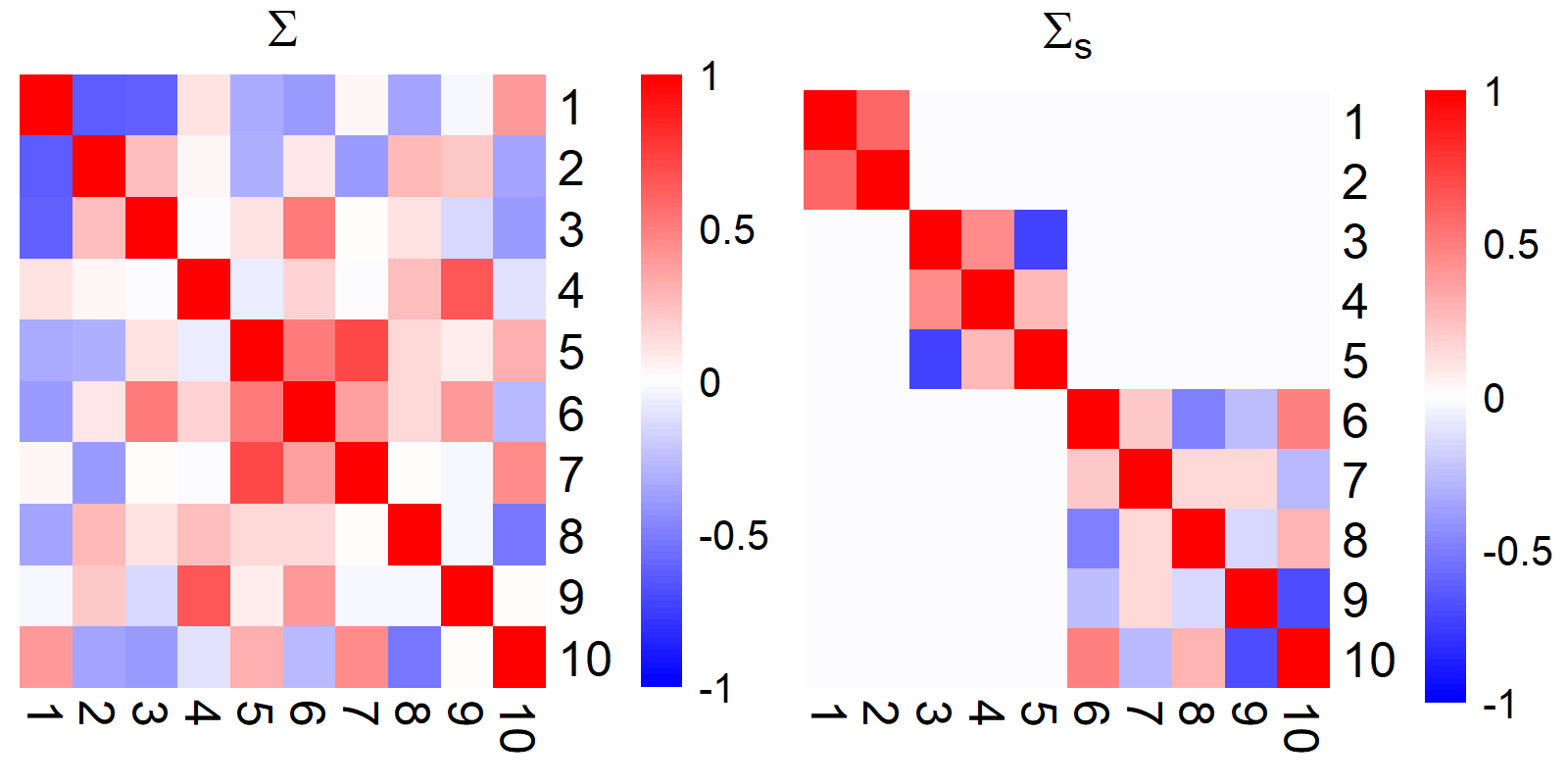}
    \caption{Heatmap illustrations of the correlation matrices $\boldsymbol{\Sigma}$ and $\boldsymbol{\Sigma}_s$.}
    \label{fig:cov_mats}
    \end{figure}

Figure~\ref{fig:cov_mats} illustrates the correlation matrices introduced in Section~\ref{sec:simulation} for simulating data from various copulae. Observe the range of dependence relationships considered, including both positive and negative correlations. 

\begin{figure}[!h]
    \centering
    \includegraphics[width=0.85\linewidth]{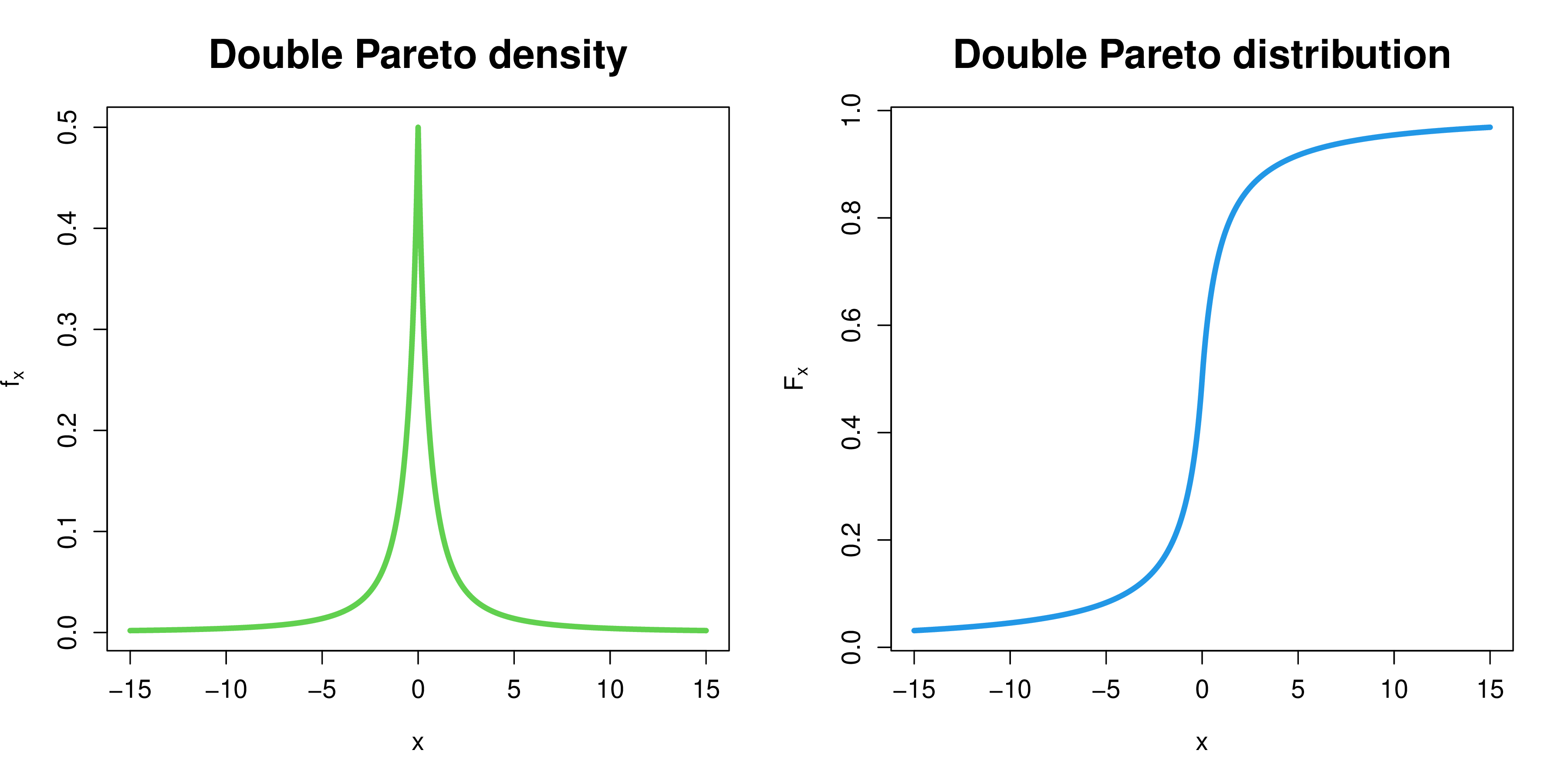}
    \caption{Probability density function (left) and cumulative distribution function (right) for the double Pareto distribution.}
    \label{fig:double_pareto_dist}
    \vspace{2cm}\includegraphics[width=0.85\linewidth]{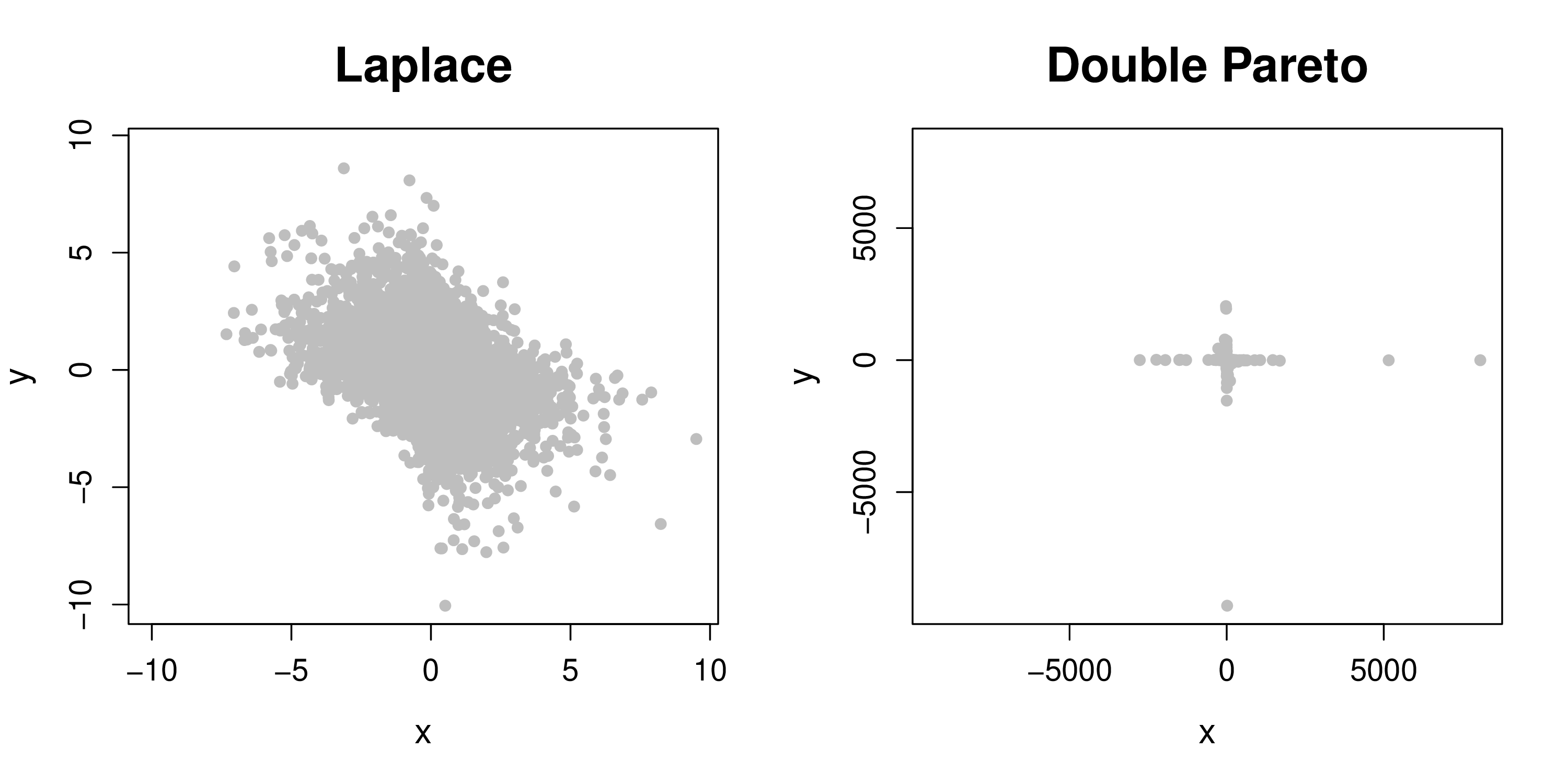}
    \caption{Simulated Gaussian data on Laplace (left) and double Pareto (right) margins.}
    \label{fig:sim_lap_par_data}
\end{figure}

Figure~\ref{fig:double_pareto_dist} illustrates the density and distribution functions of the double Pareto distribution introduced in Section~\ref{subsec:sim_setup}. Note the heaviness of the distribution in both the upper and lower tails.

Furthermore, Figure~\ref{fig:sim_lap_par_data} illustrates data simulated from a Gaussian copula on both Laplace and double Pareto margins. One can observe the stark difference in joint tail behaviour between the two marginal structures. 

Table~\ref{tab:cCRPS_d_10} gives the skill scores for each combination setup with the dimension fixed at $d = 10$.   

\begin{table}[!h]
\centering\centering
\caption{$\operatorname{Skill}(F_{*})$ scores (to 6 significant figures) of each deep generative approach ($* \in \{\text{FM},\text{NFMAF},\text{GAN},\text{NFNSF} \} $) across all combinations at $d = 10$.\\}
\label{tab:cCRPS_d_10}
\centering
\begin{tabular}[t]{c|c|c|c|c|c|c|c}
\hline
 &  &  &  & \multicolumn{4}{c}{\textbf{Model}}\\
\hline
\textbf{Copula} & \textbf{Margins} & $n$ & $d$ & FM & NFMAF & GAN & NFNSF\\
\hline
1 & Laplace & $10^{3}$ & 10 & 1.0011 & 1.00203 & 1.03251 & \textbf{1.0005}\\
\hline
1 & Laplace & $10^{4}$ & 10 & 1.00037 & \textbf{0.999984} & 1.01386 & 1.00056\\
\hline
1 & Laplace & $10^{5}$ & 10 & 1.00033 & \textbf{1.00015} & 1.00447 & 1.00041\\
\hline
1 & Double Pareto & $10^{3}$ & 10 & \textbf{1.00144} & 1.0023 & 1.00653 & 1.00312\\
\hline
1 & Double Pareto & $10^{4}$ & 10 & 1.00034 & 1.0004 & 1.00021 & \textbf{1.00009}\\
\hline
1 & Double Pareto & $10^{5}$ & 10 & 1.00054 & \textbf{1.00015} & 1.00053 & 1.0003\\
\hline
2 & Laplace & $10^{3}$ & 10 & 1.00268 & \textbf{1.00206} & 1.02785 & 1.00263\\
\hline
2 & Laplace & $10^{4}$ & 10 & \textbf{0.999934} & 1.00016 & 1.01829 & 1.00049\\
\hline
2 & Laplace & $10^{5}$ & 10 & 1.00037 & \textbf{1.0003} & 1.00154 & 1.00049\\
\hline
2 & Double Pareto & $10^{3}$ & 10 & 1.00107 & 1.00119 & 1.12247 & \textbf{1.00054}\\
\hline
2 & Double Pareto & $10^{4}$ & 10 & 1.00035 & \textbf{1.00005} & 1.00132 & 1.00043\\
\hline
2 & Double Pareto & $10^{5}$ & 10 & 1.00035 & 1.00038 & 1.00069 & \textbf{1.00007}\\
\hline
3 & Laplace & $10^{3}$ & 10 & \textbf{1.00011} & 1.00115 & 1.04368 & 1.00194\\
\hline
3 & Laplace & $10^{4}$ & 10 & \textbf{1.00009} & 1.00068 & 1.01169 & 1.00056\\
\hline
3 & Laplace & $10^{5}$ & 10 & \textbf{1.00005} & 1.00028 & 1.00129 & 1.00057\\
\hline
3 & Double Pareto & $10^{3}$ & 10 & \textbf{1.00047} & 1.00218 & 1.05332 & 1.00167\\
\hline
3 & Double Pareto & $10^{4}$ & 10 & \textbf{1.00012} & 1.00035 & 1.0007 & 1.00077\\
\hline
3 & Double Pareto & $10^{5}$ & 10 & \textbf{1.00034} & 1.00039 & 1.00035 & 1.00037\\
\hline
4 & Laplace & $10^{3}$ & 10 & 1.00095 & 1.00054 & 1.0658 & \textbf{1.00041}\\
\hline
4 & Laplace & $10^{4}$ & 10 & \textbf{1.00034} & 1.0005 & 1.01088 & 1.00056\\
\hline
4 & Laplace & $10^{5}$ & 10 & \textbf{1.00016} & 1.00028 & 1.00081 & 1.00055\\
\hline
4 & Double Pareto & $10^{3}$ & 10 & 1.00085 & 1.0012 & 1.09156 & \textbf{1.00074}\\
\hline
4 & Double Pareto & $10^{4}$ & 10 & 1.00038 & \textbf{1.00034} & 1.00083 & 1.00065\\
\hline
4 & Double Pareto & $10^{5}$ & 10 & 1.00074 & 1.00023 & 1.00073 & \textbf{1.00009}\\
\hline
5 & Laplace & $10^{3}$ & 10 & \textbf{1.00073} & 1.00164 & 1.05407 & 1.0024\\
\hline
5 & Laplace & $10^{4}$ & 10 & 1.0002 & \textbf{1.0001} & 1.00422 & 1.00012\\
\hline
5 & Laplace & $10^{5}$ & 10 & 1.00064 & 1.0004 & 1.01601 & \textbf{1.00039}\\
\hline
5 & Double Pareto & $10^{3}$ & 10 & \textbf{1.00024} & 1.00061 & 1.00493 & 1.00079\\
\hline
5 & Double Pareto & $10^{4}$ & 10 & \textbf{1.00007} & 1.00033 & 1.00056 & 1.00035\\
\hline
5 & Double Pareto & $10^{5}$ & 10 & 1.00038 & 1.00009 & 1.00101 & \textbf{1.00008}\\
\hline
\end{tabular}
\end{table}

\clearpage

\begin{figure}[h!]
    \centering
    \begin{subfigure}[b]{0.2\textwidth}
        \centering
        \includegraphics[width=\textwidth]{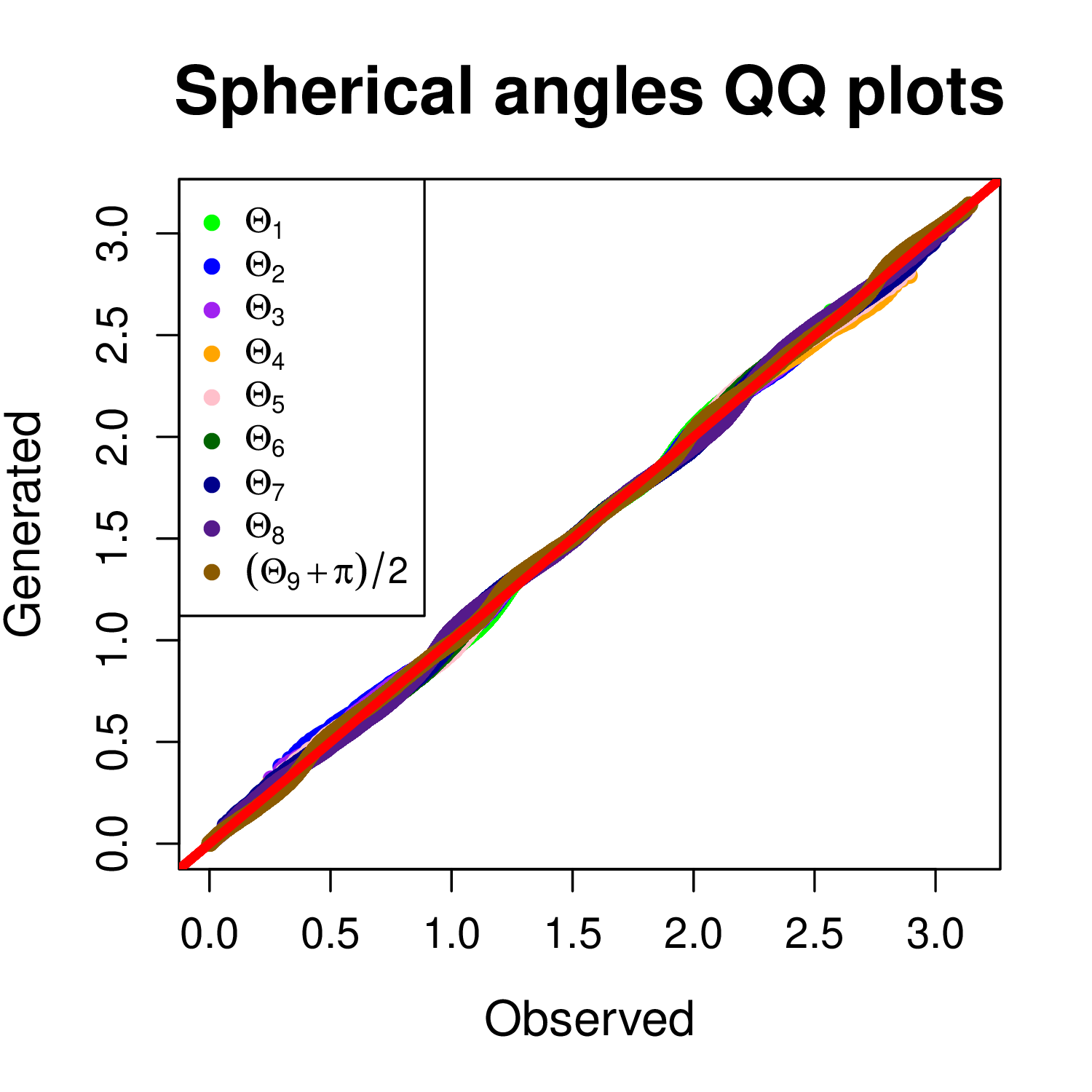}
    \end{subfigure}%
    \begin{subfigure}[b]{0.2\textwidth}
        \centering
        \includegraphics[width=\textwidth]{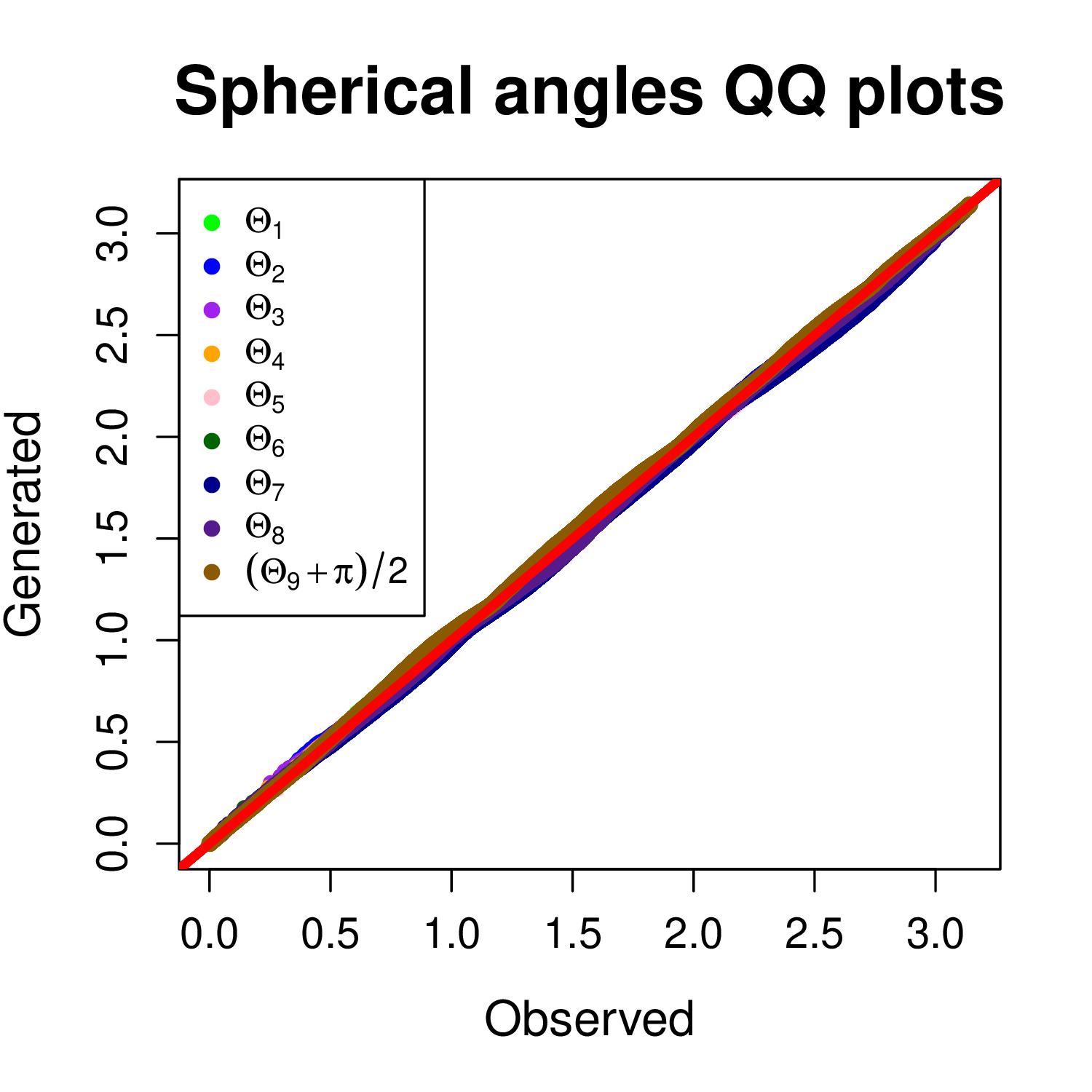}
    \end{subfigure}%
    \begin{subfigure}[b]{0.2\textwidth}
        \centering
        \includegraphics[width=\textwidth]{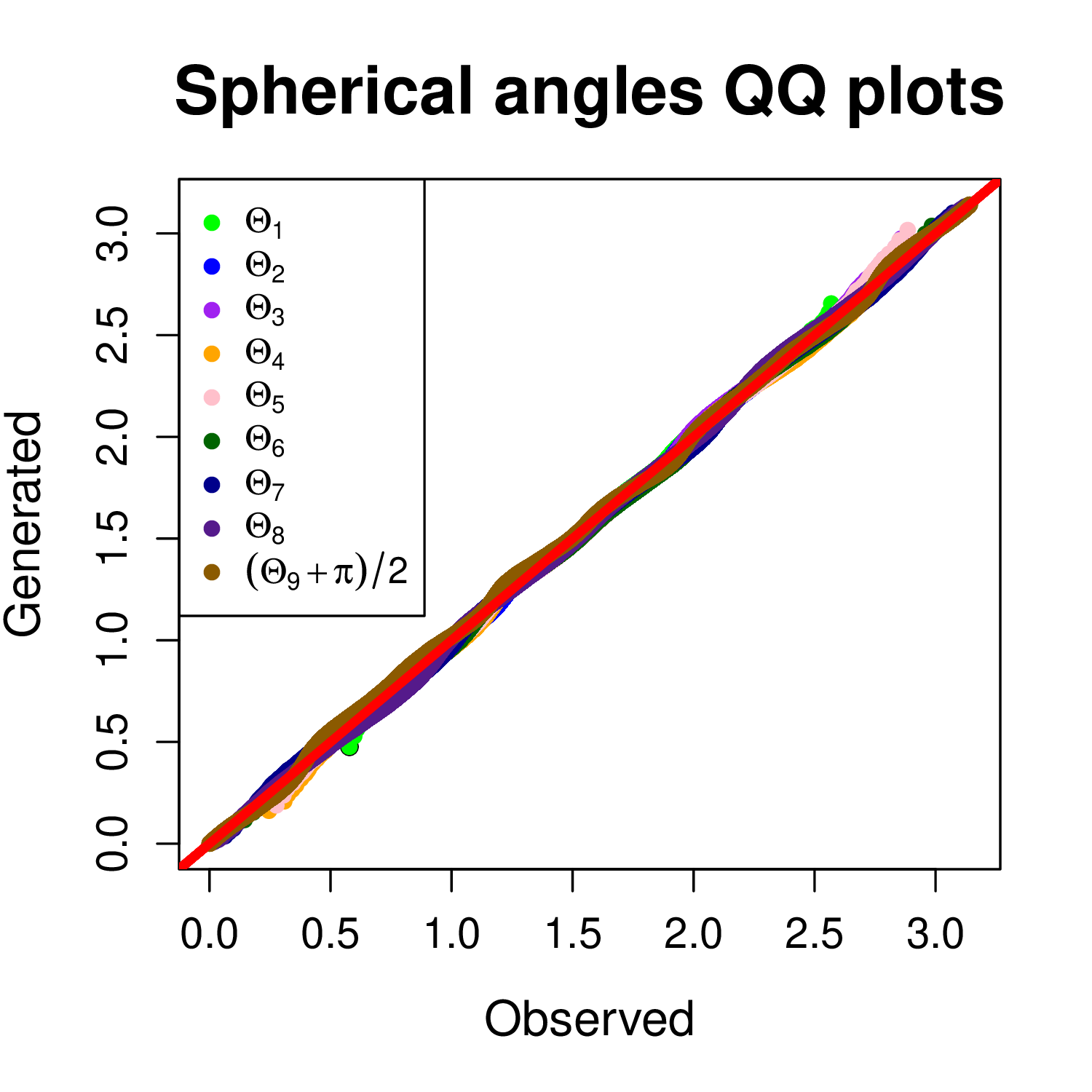}
    \end{subfigure}%
    \begin{subfigure}[b]{0.2\textwidth}
        \centering
        \includegraphics[width=\textwidth]{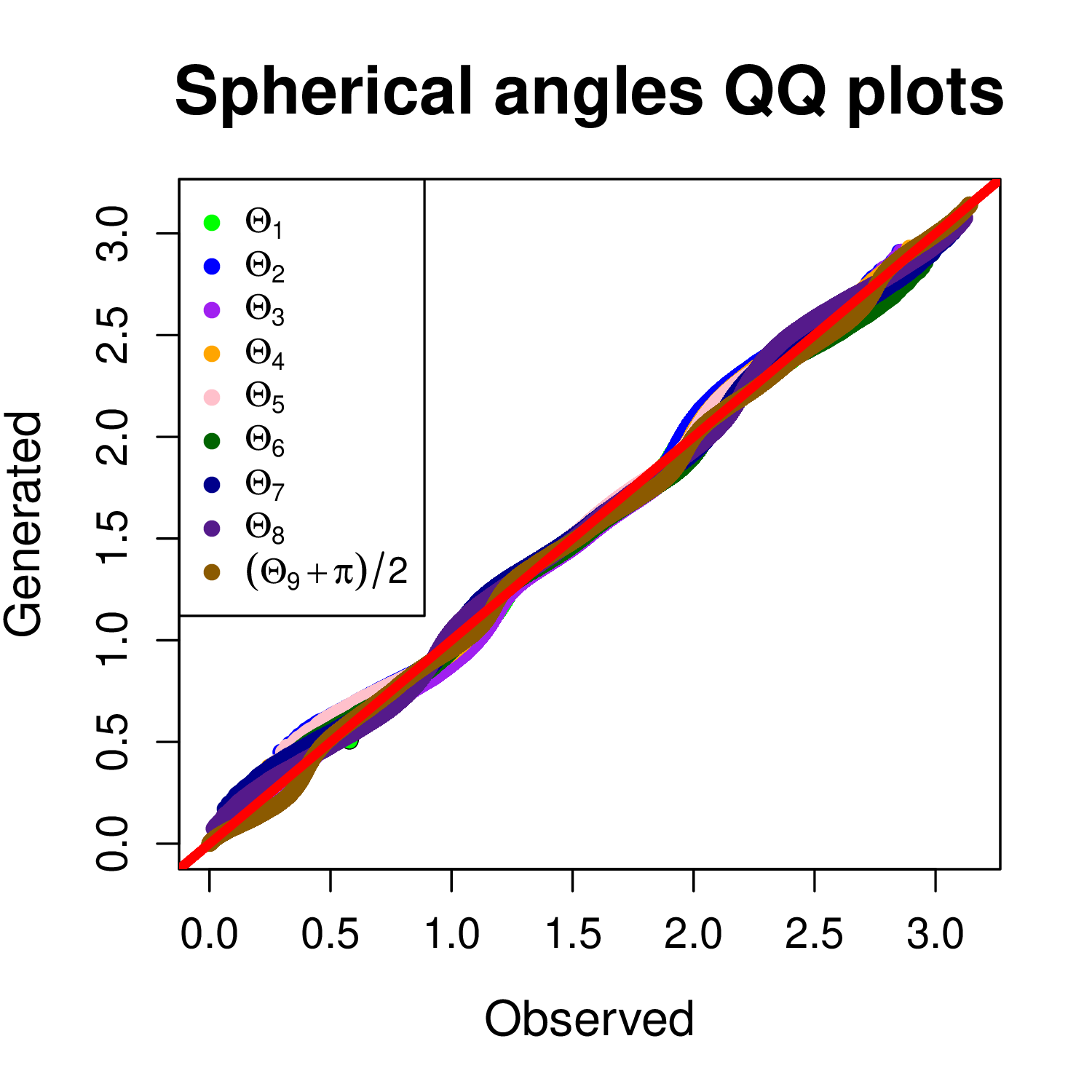}
    \end{subfigure}%
    \begin{subfigure}[b]{0.2\textwidth}
        \centering
        \includegraphics[width=\textwidth]{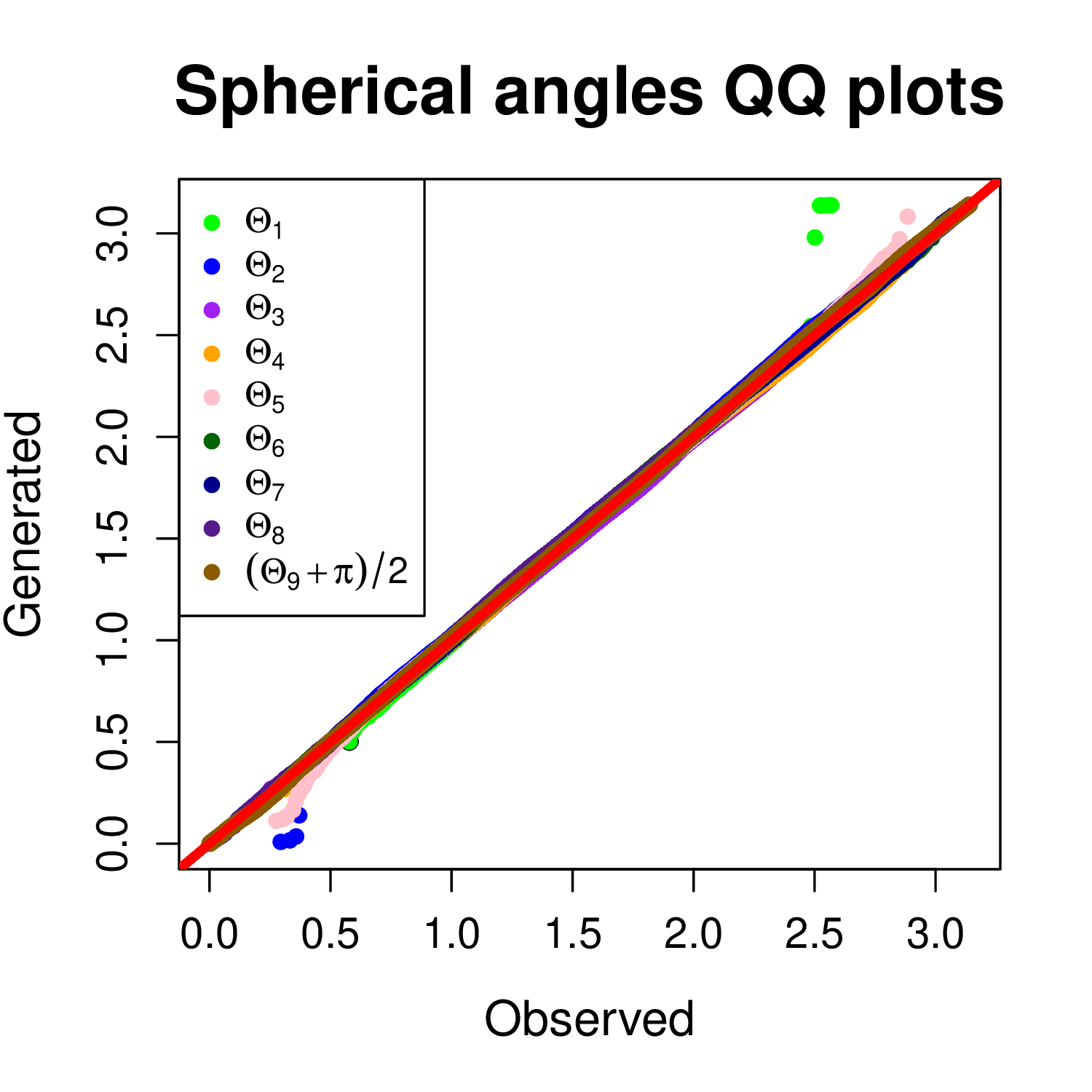}
    \end{subfigure}%

    \begin{subfigure}[b]{0.2\textwidth}
        \centering
        \includegraphics[width=\textwidth]{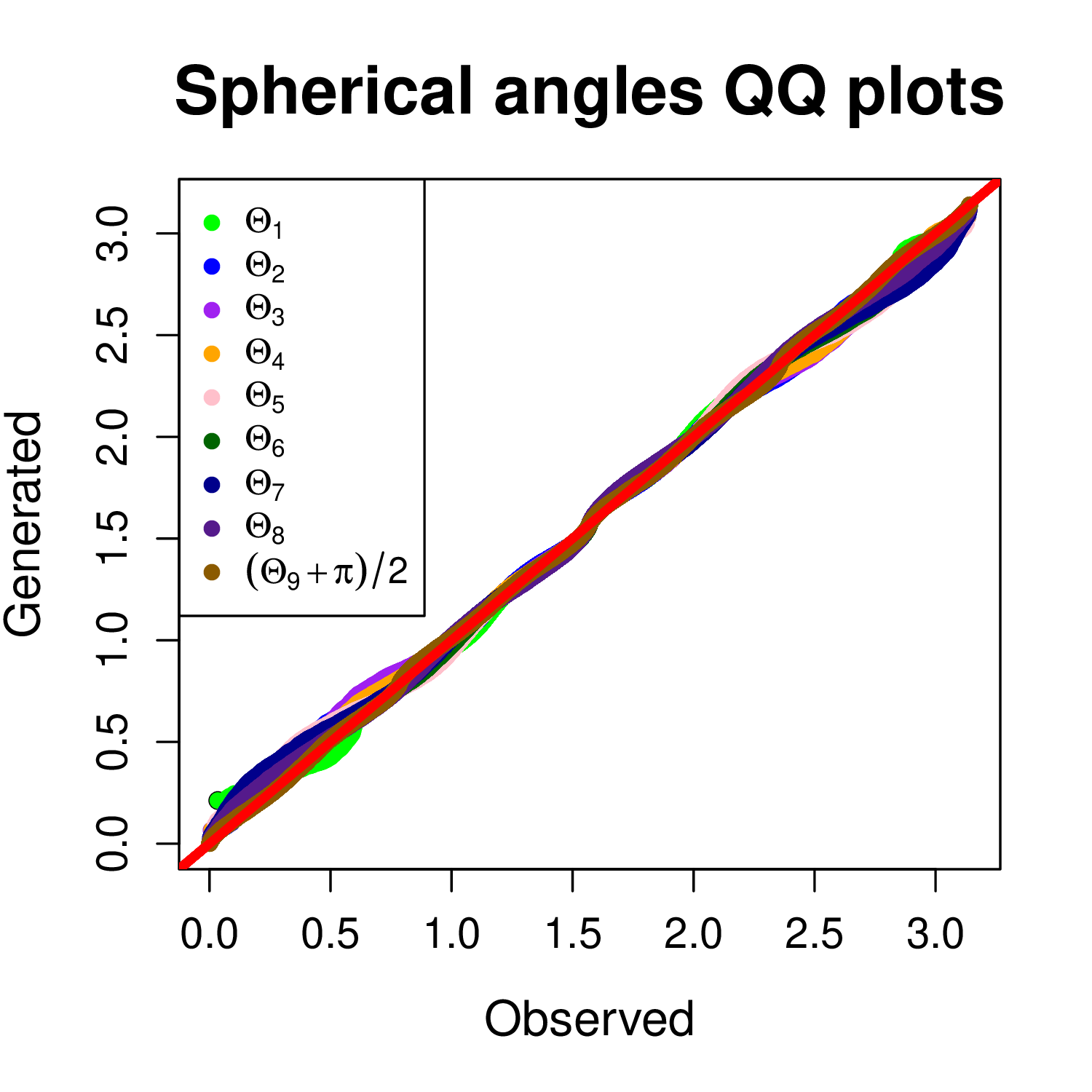}
    \end{subfigure}%
    \begin{subfigure}[b]{0.2\textwidth}
        \centering
        \includegraphics[width=\textwidth]{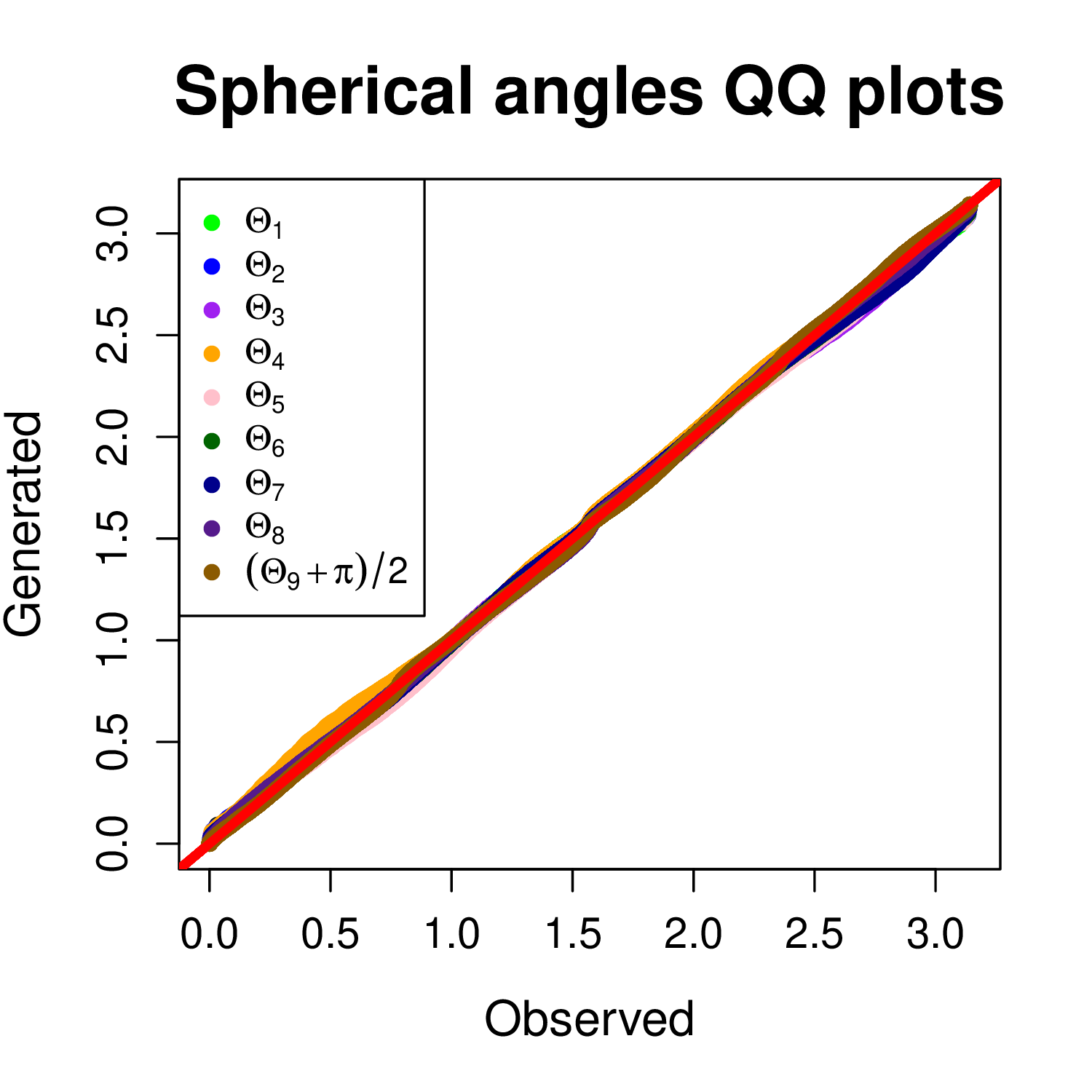}
    \end{subfigure}%
    \begin{subfigure}[b]{0.2\textwidth}
        \centering
        \includegraphics[width=\textwidth]{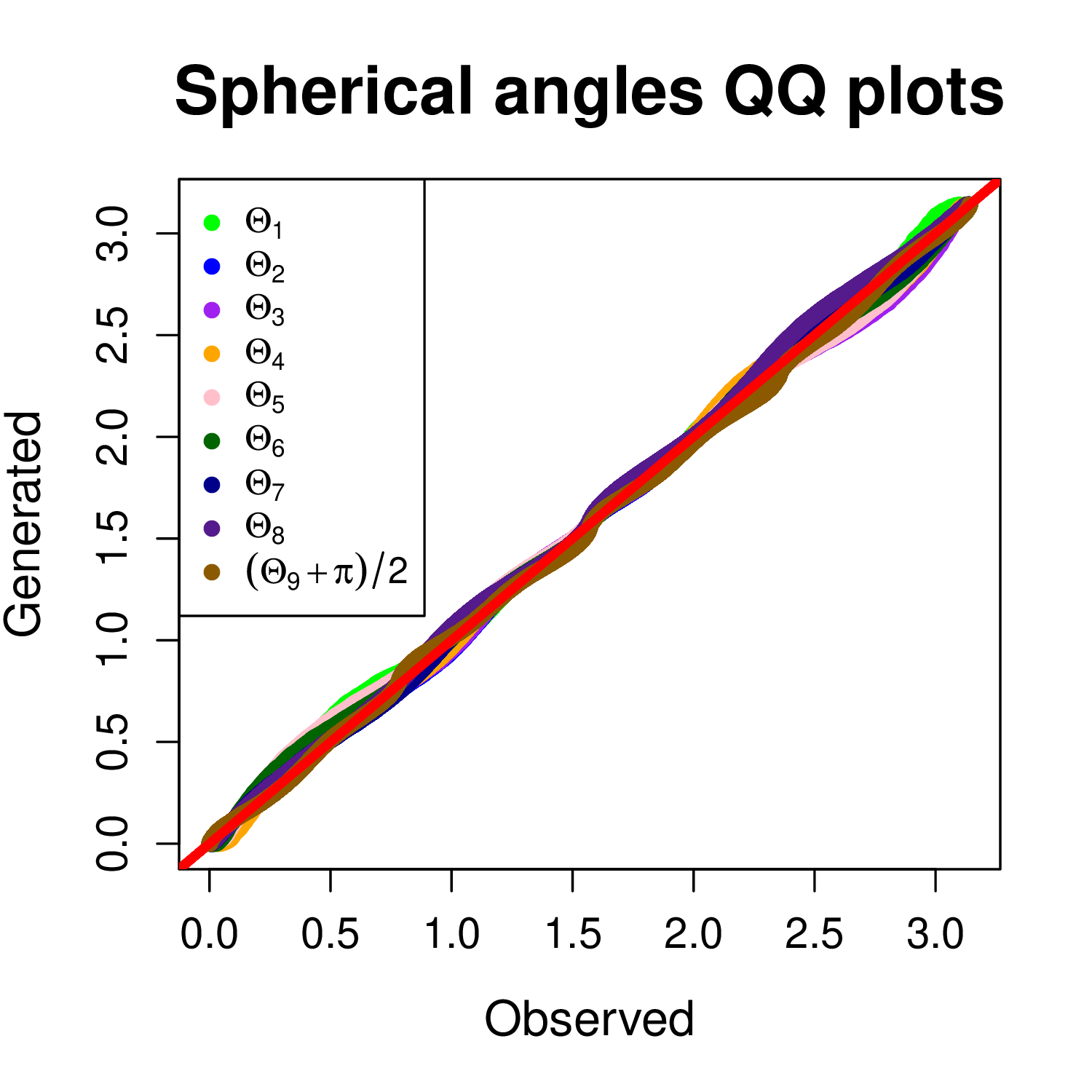}
    \end{subfigure}%
    \begin{subfigure}[b]{0.2\textwidth}
        \centering
        \includegraphics[width=\textwidth]{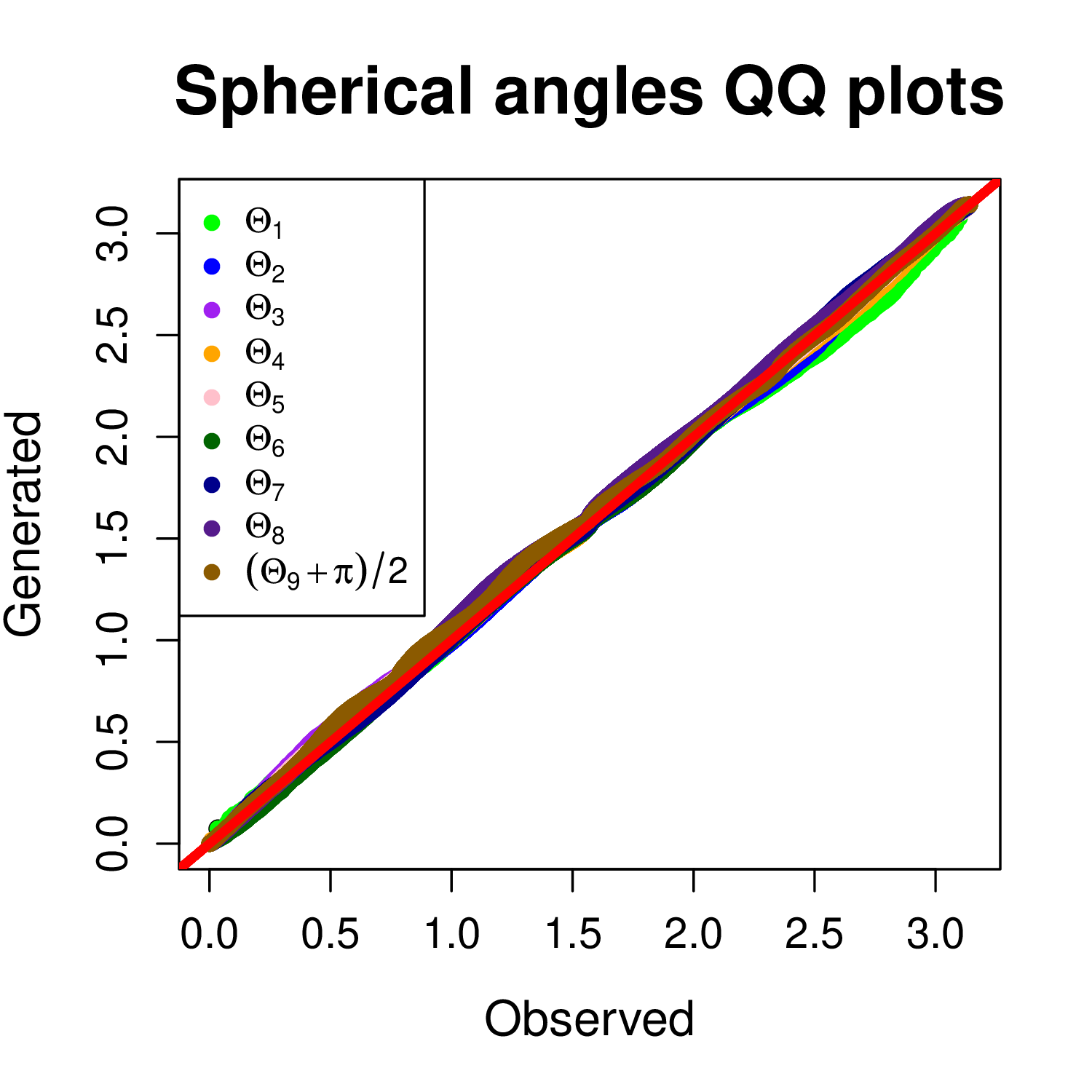}
    \end{subfigure}%
    \begin{subfigure}[b]{0.2\textwidth}
        \centering
        \includegraphics[width=\textwidth]{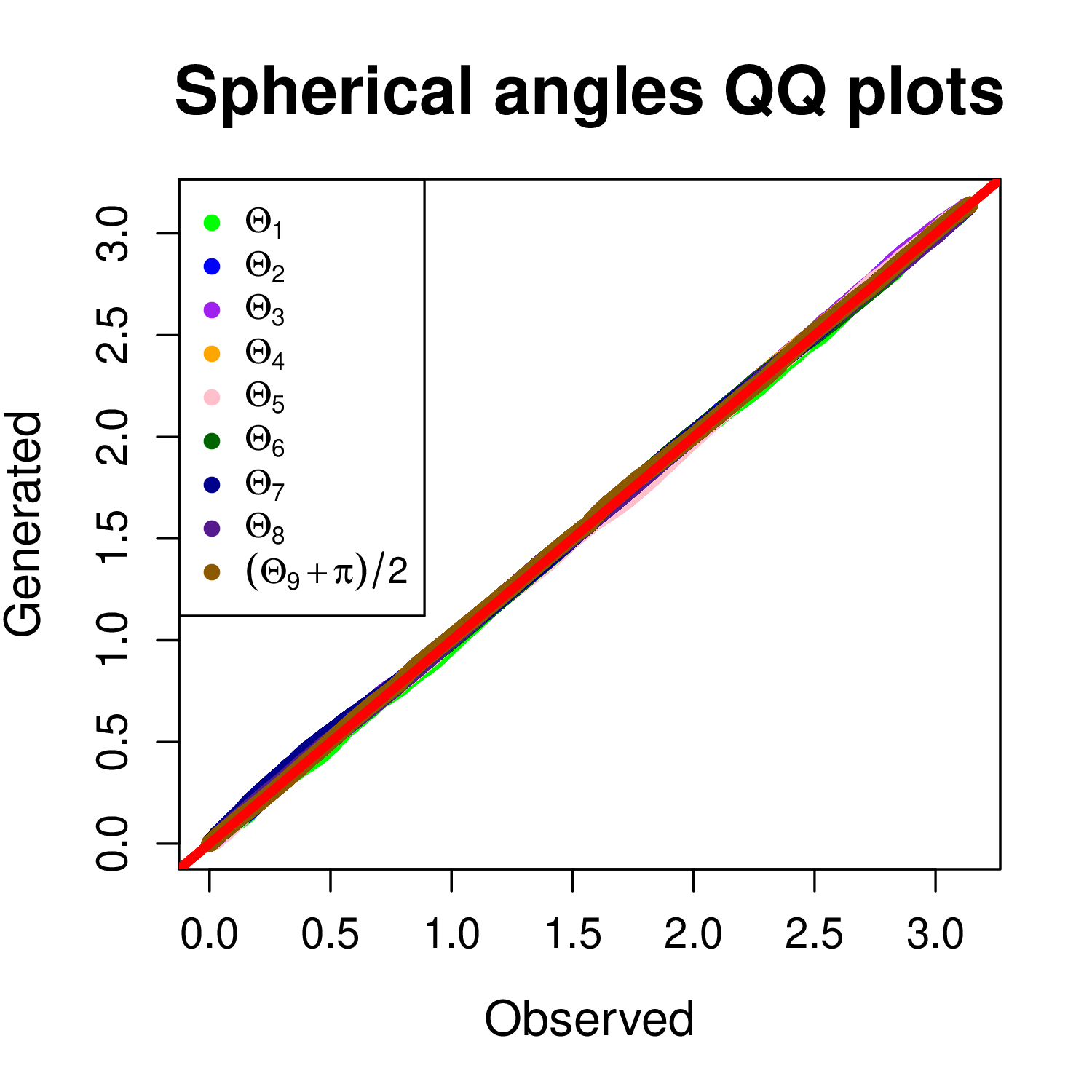}
    \end{subfigure}%
    \caption{Spherical angle QQ plots for copula 2 with $d=10$ and $n=100\ 000$.  }
    \label{fig:qq_plots_cop2_d10}
\end{figure}

\begin{figure}[h!]
    \centering
    \begin{subfigure}[b]{0.2\textwidth}
        \centering
        \includegraphics[width=\textwidth]{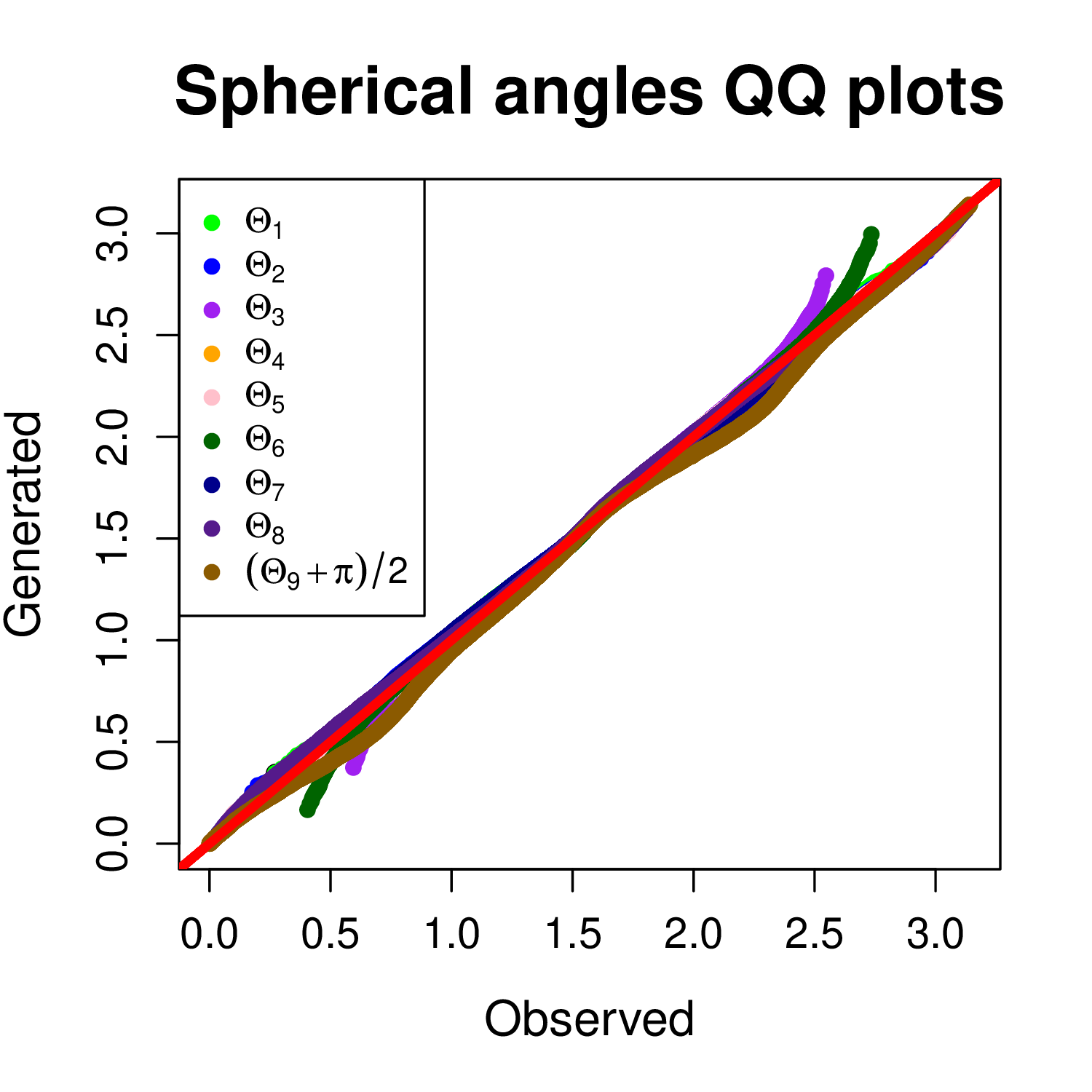}
    \end{subfigure}%
    \begin{subfigure}[b]{0.2\textwidth}
        \centering
        \includegraphics[width=\textwidth]{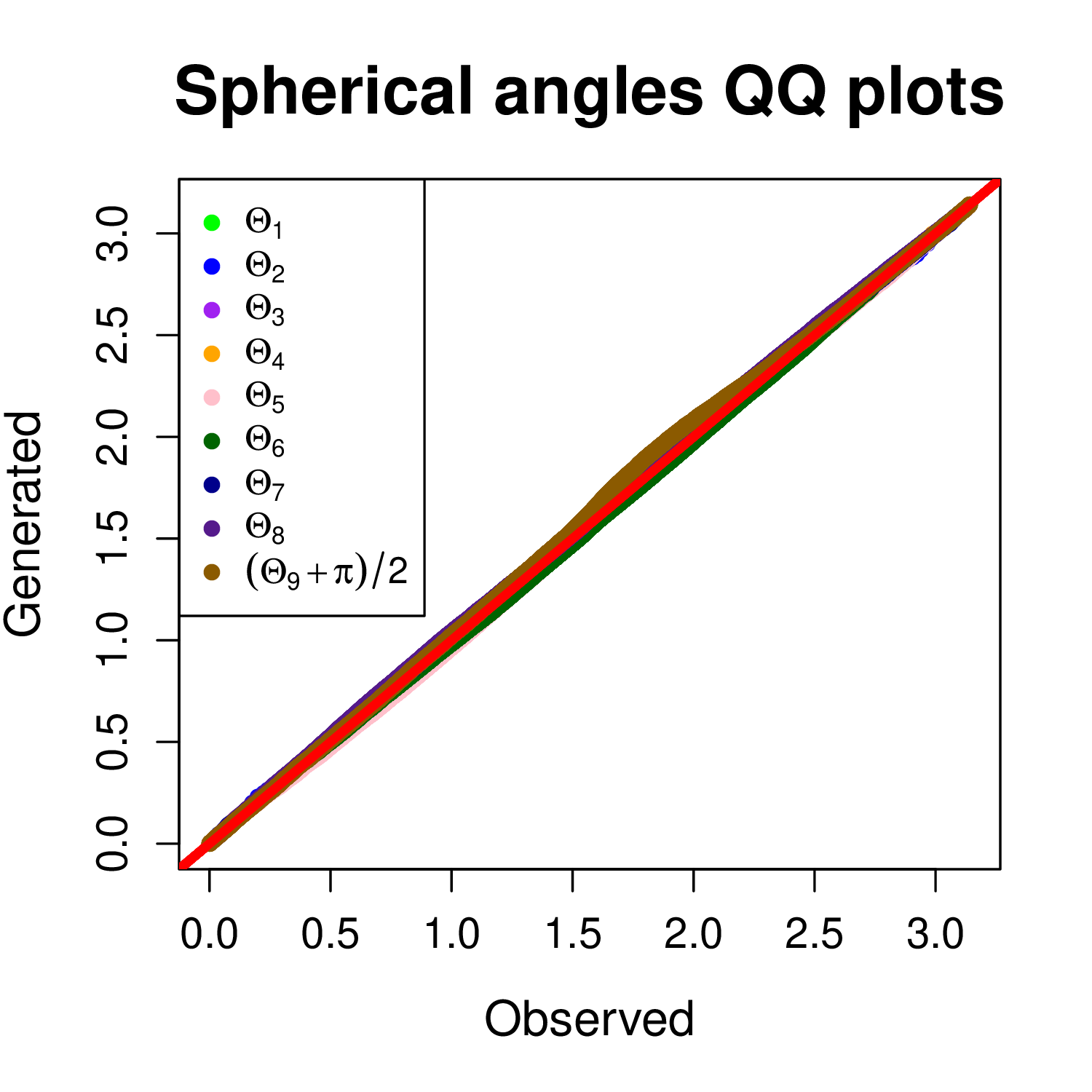}
    \end{subfigure}%
    \begin{subfigure}[b]{0.2\textwidth}
        \centering
        \includegraphics[width=\textwidth]{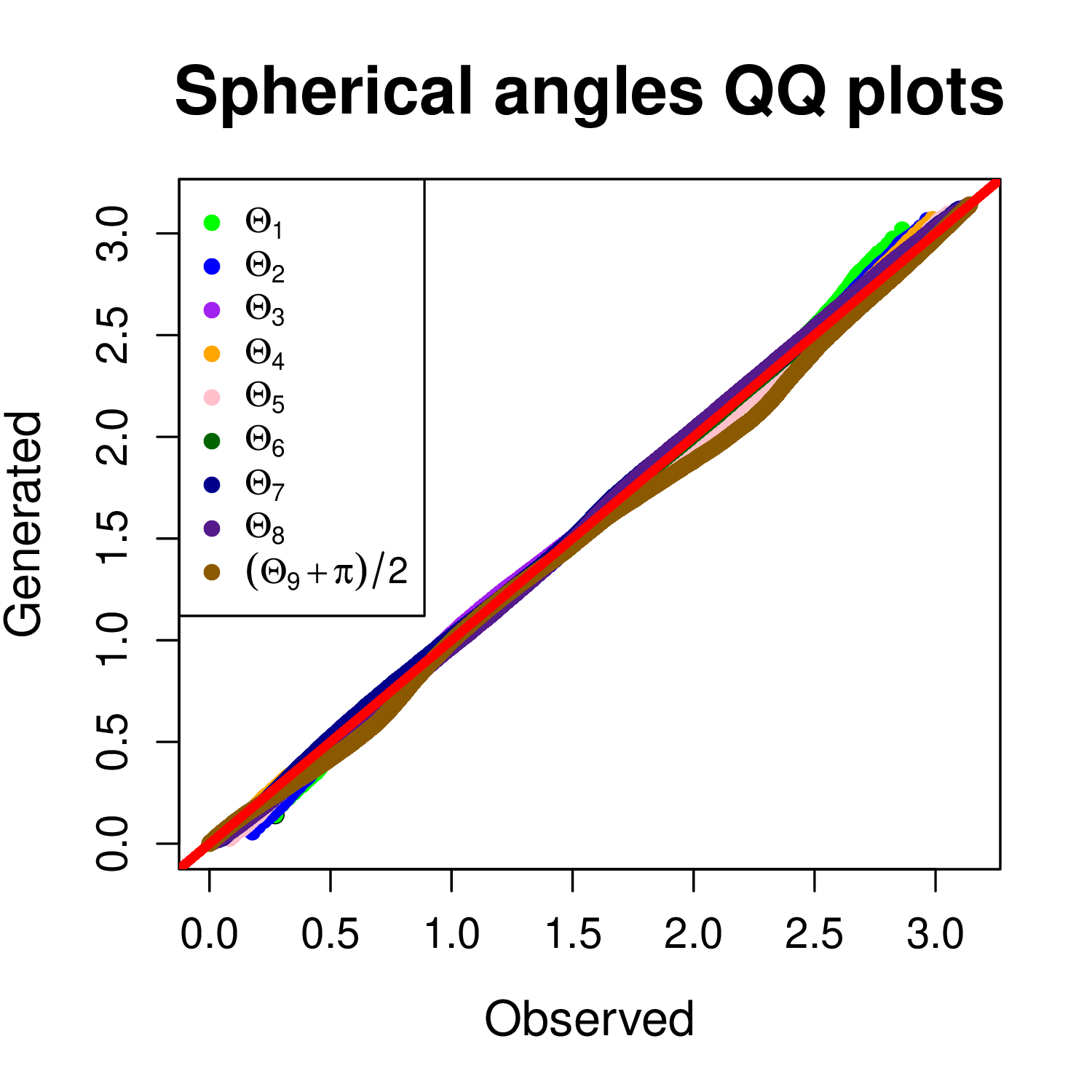}
    \end{subfigure}%
    \begin{subfigure}[b]{0.2\textwidth}
        \centering
        \includegraphics[width=\textwidth]{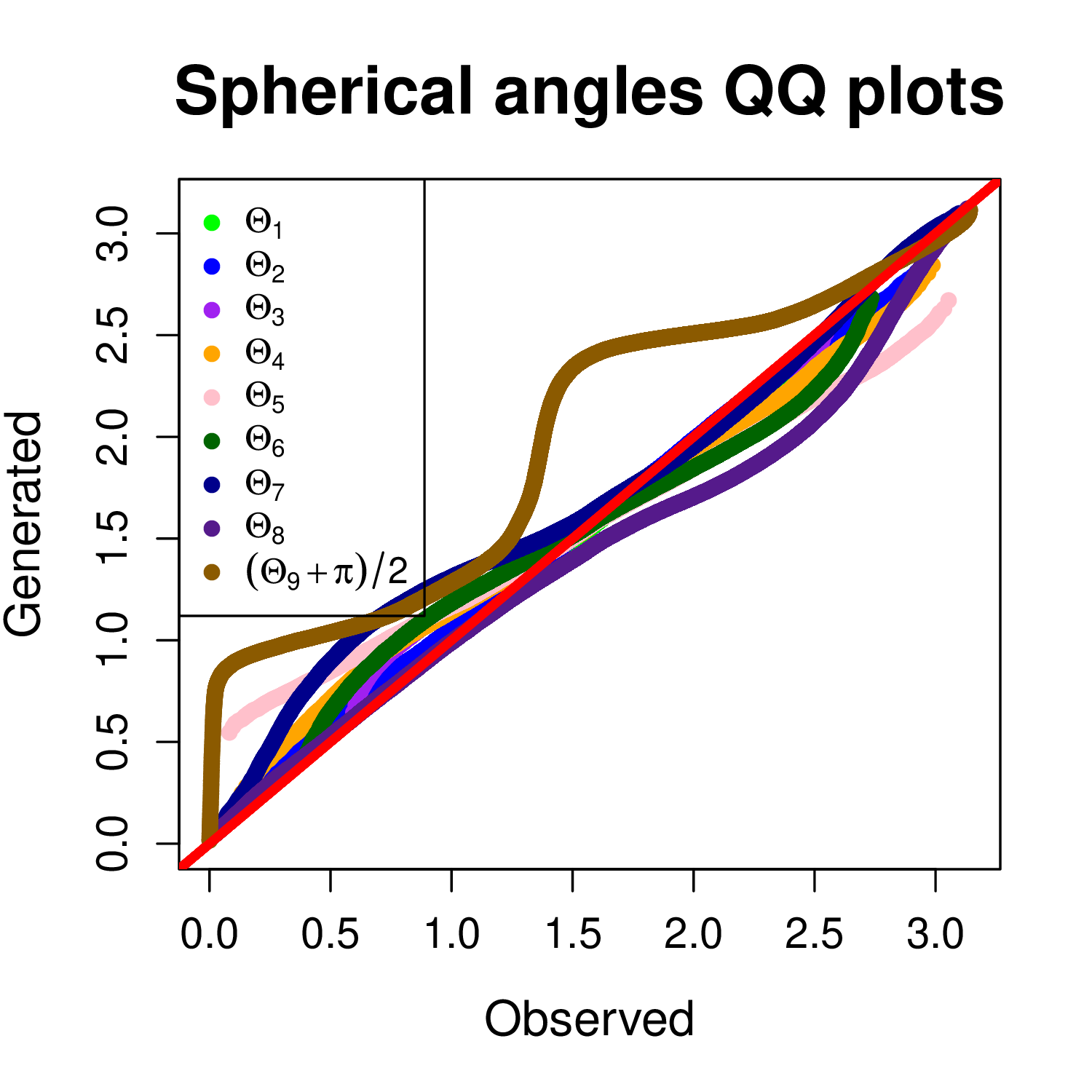}
    \end{subfigure}%
    \begin{subfigure}[b]{0.2\textwidth}
        \centering
        \includegraphics[width=\textwidth]{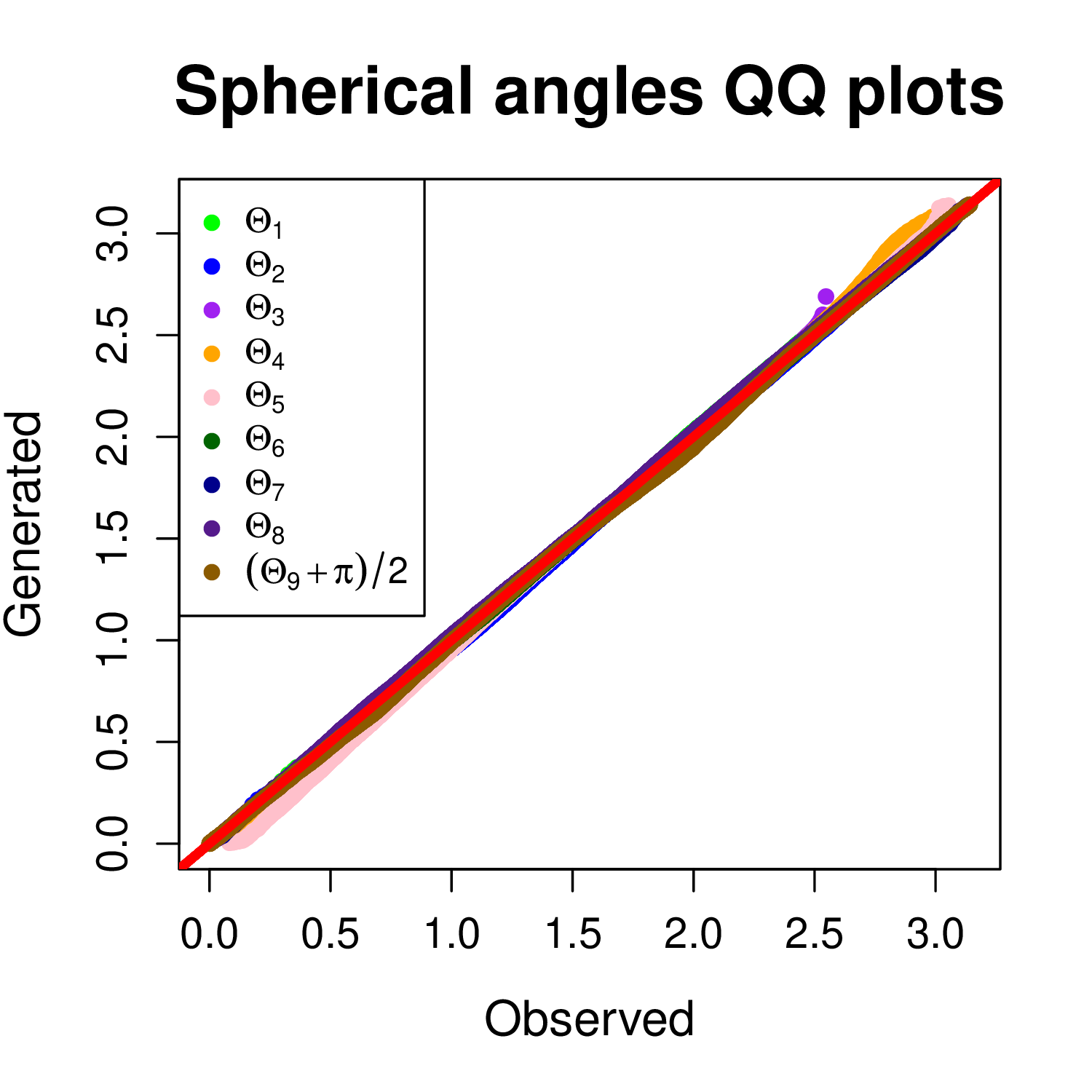}
    \end{subfigure}%

    \begin{subfigure}[b]{0.2\textwidth}
        \centering
        \includegraphics[width=\textwidth]{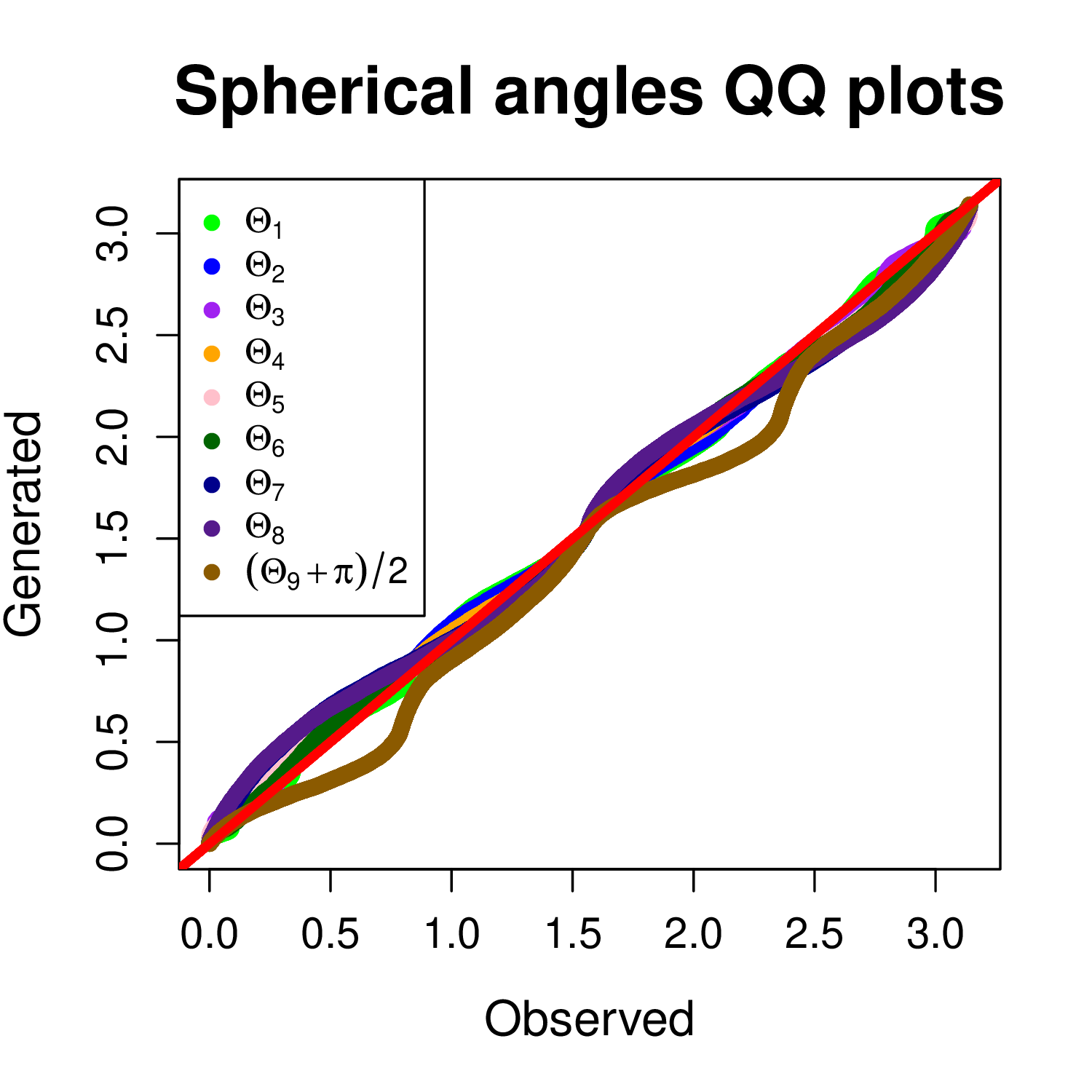}
    \end{subfigure}%
    \begin{subfigure}[b]{0.2\textwidth}
        \centering
        \includegraphics[width=\textwidth]{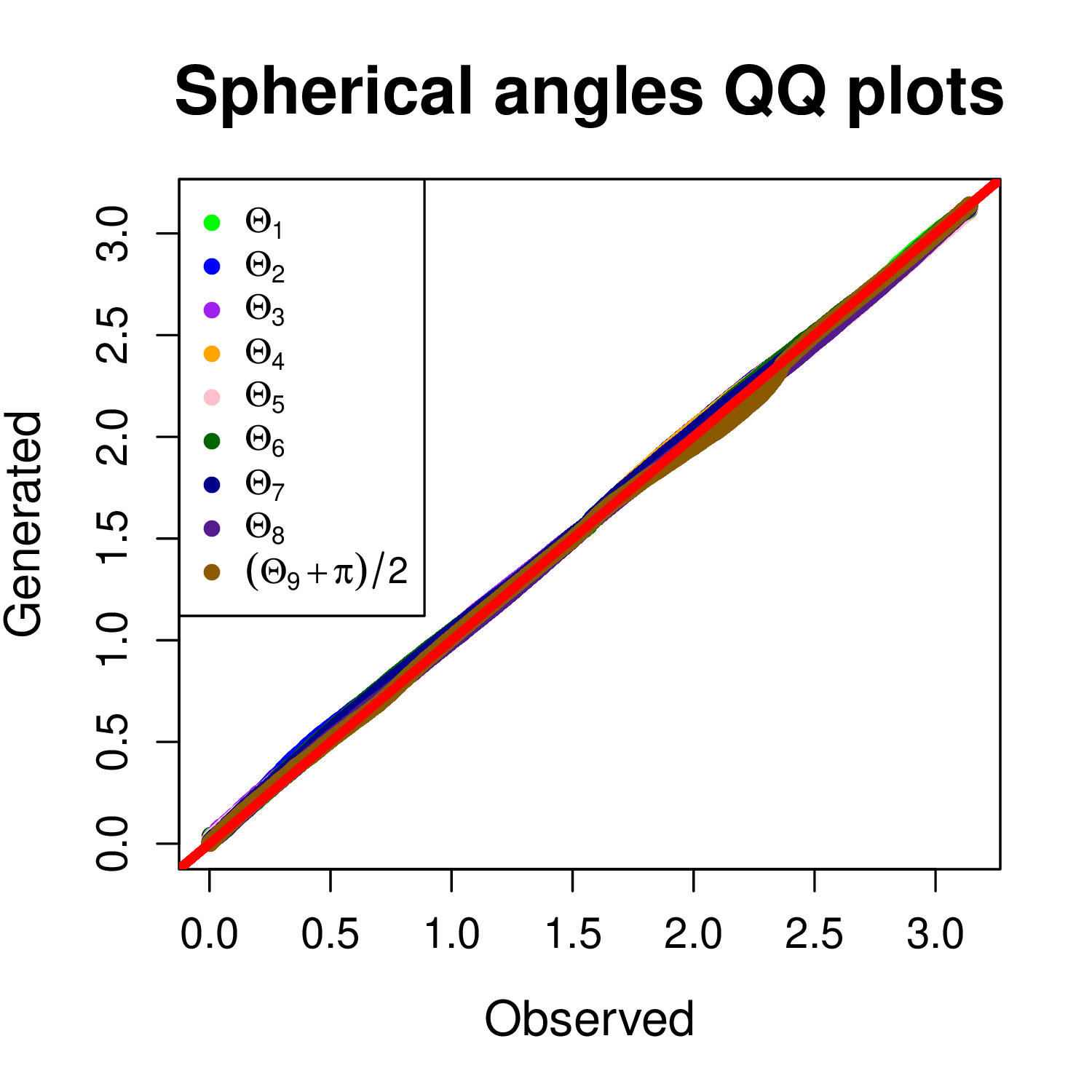}
    \end{subfigure}%
    \begin{subfigure}[b]{0.2\textwidth}
        \centering
        \includegraphics[width=\textwidth]{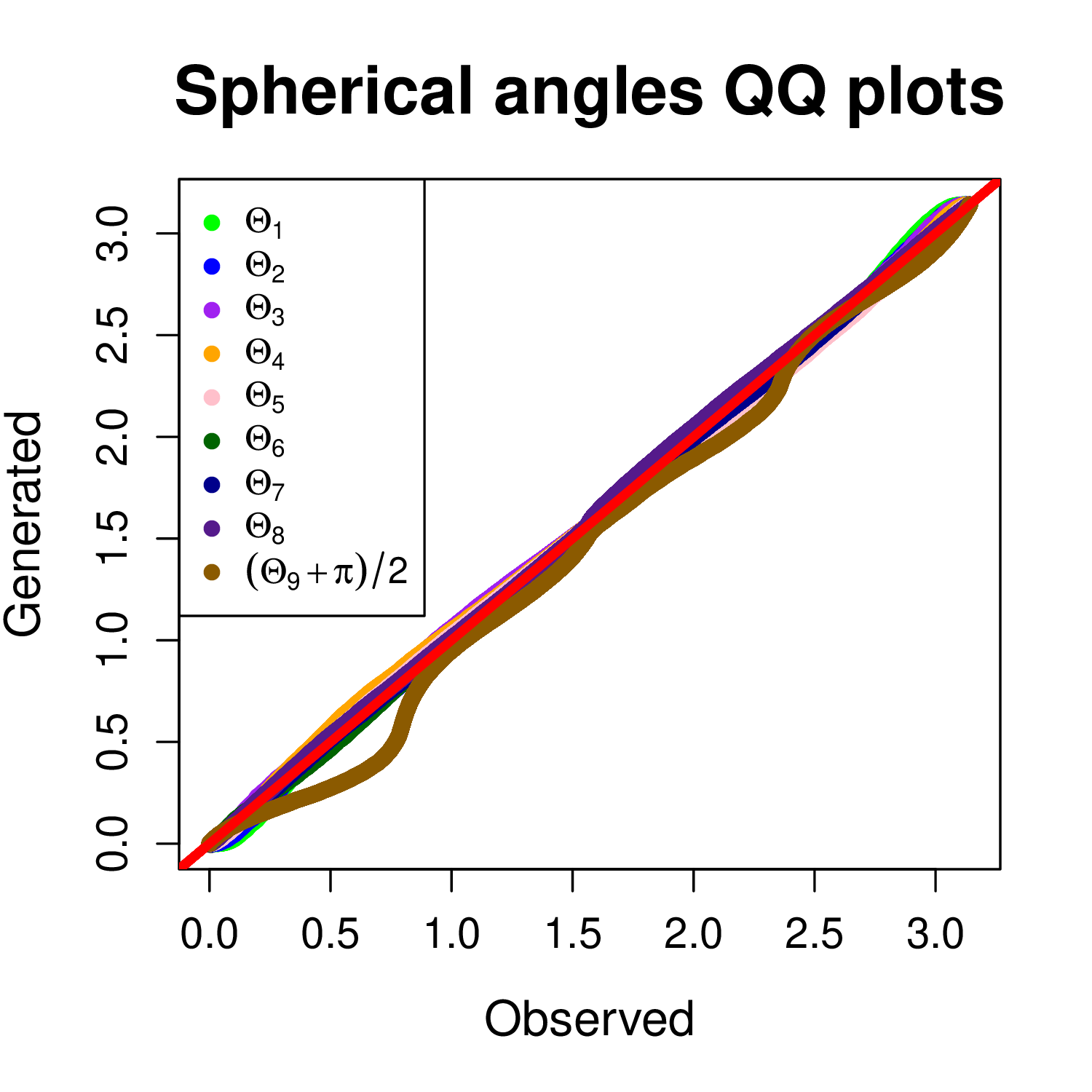}
    \end{subfigure}%
    \begin{subfigure}[b]{0.2\textwidth}
        \centering
        \includegraphics[width=\textwidth]{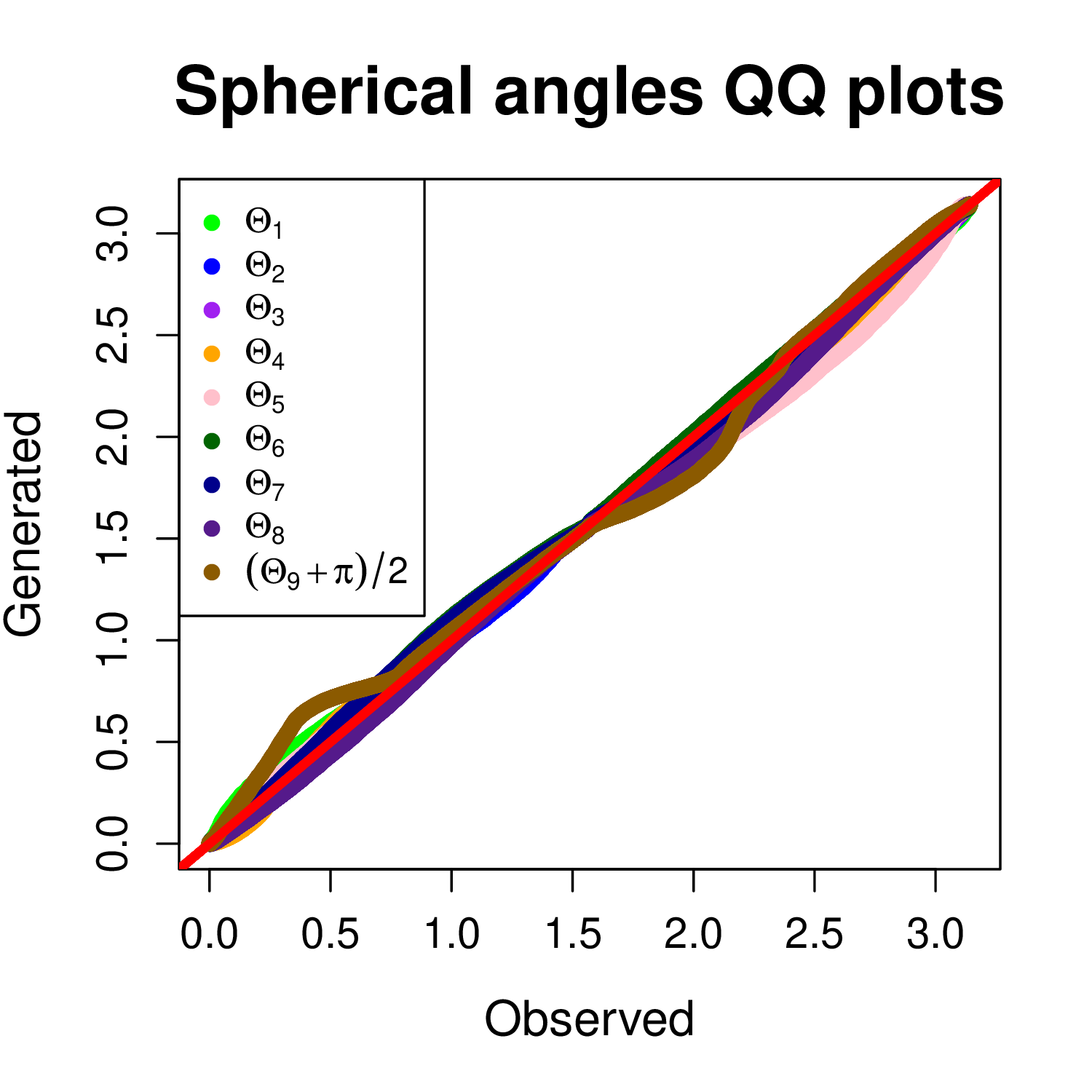}
    \end{subfigure}%
    \begin{subfigure}[b]{0.2\textwidth}
        \centering
        \includegraphics[width=\textwidth]{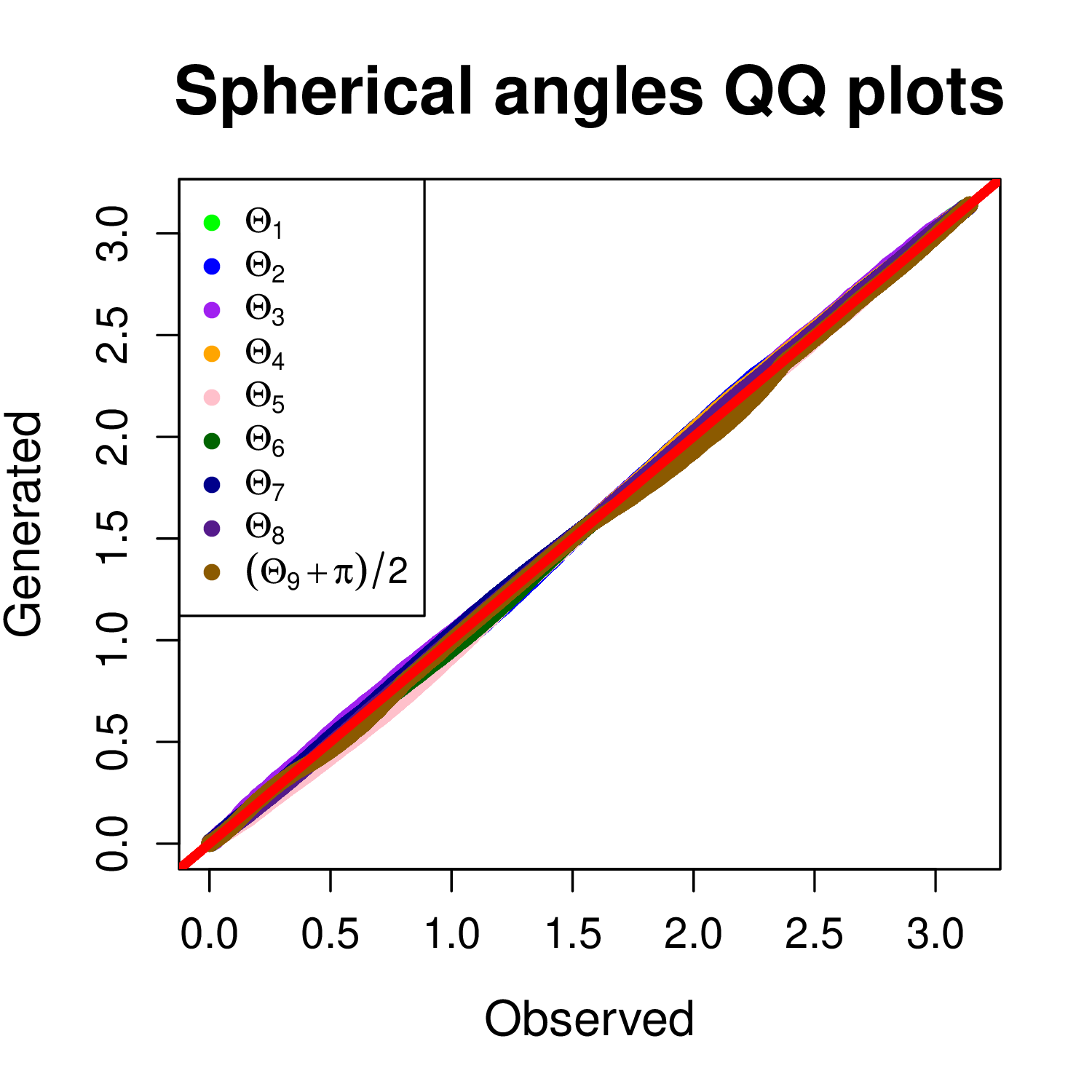}
    \end{subfigure}%
    \caption{Spherical angle QQ plots for copula 5 with $d=10$ and $n=100\ 000$.  }
    \label{fig:qq_plots_cop5_d10}
\end{figure}

\begin{figure}[h!]
    \centering
    \begin{subfigure}[b]{0.2\textwidth}
        \centering
        \includegraphics[width=\textwidth]{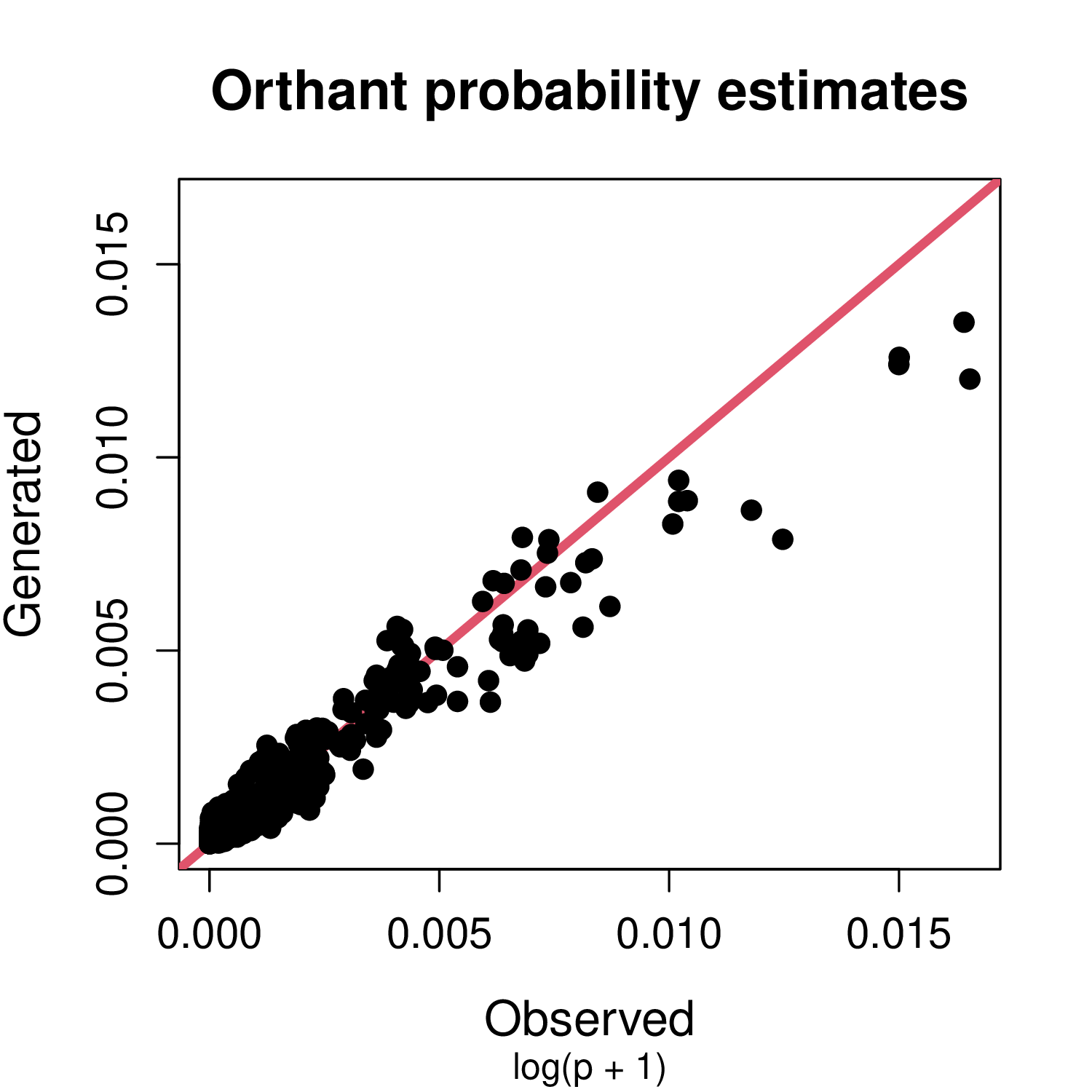}
    \end{subfigure}%
    \begin{subfigure}[b]{0.2\textwidth}
        \centering
        \includegraphics[width=\textwidth]{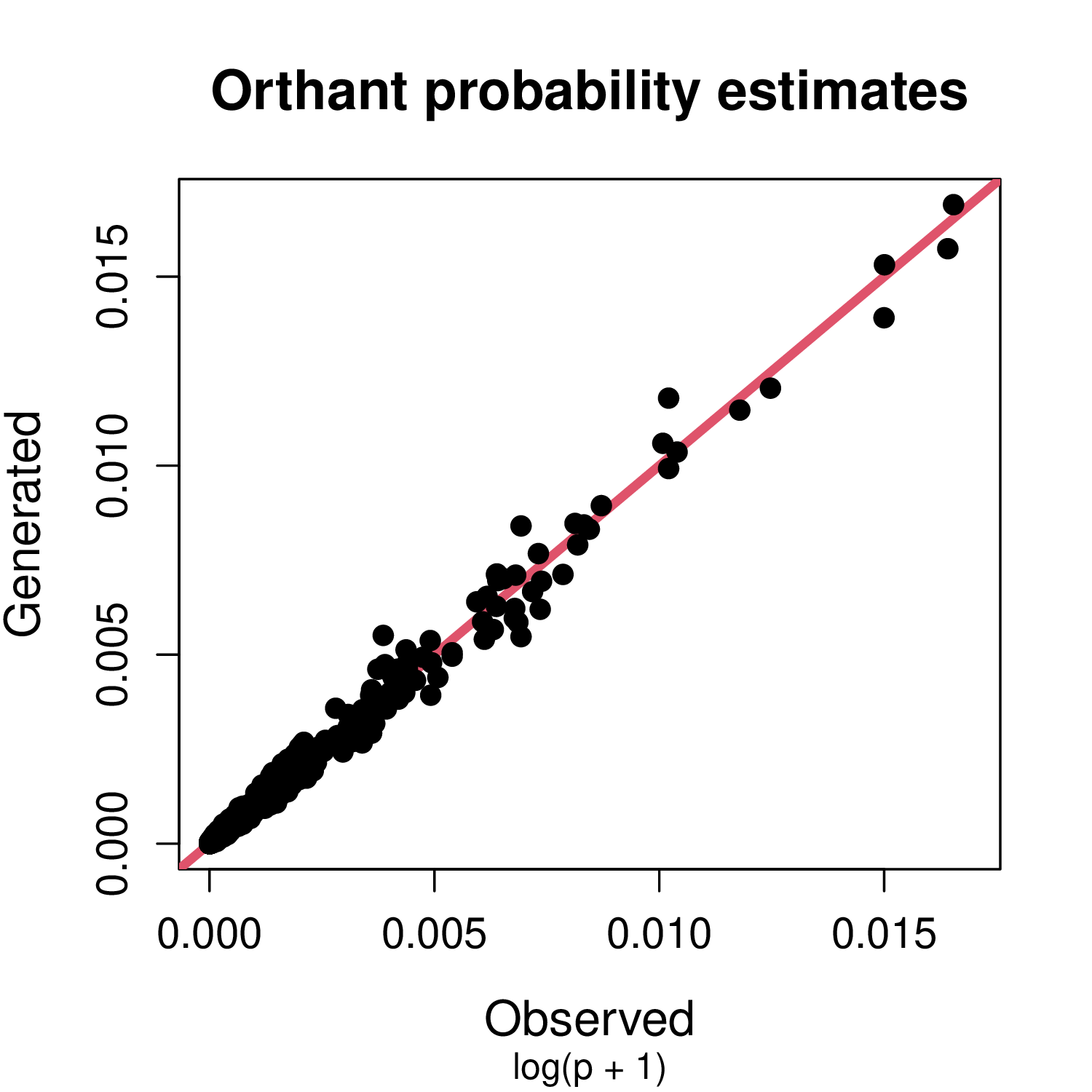}
    \end{subfigure}%
    \begin{subfigure}[b]{0.2\textwidth}
        \centering
        \includegraphics[width=\textwidth]{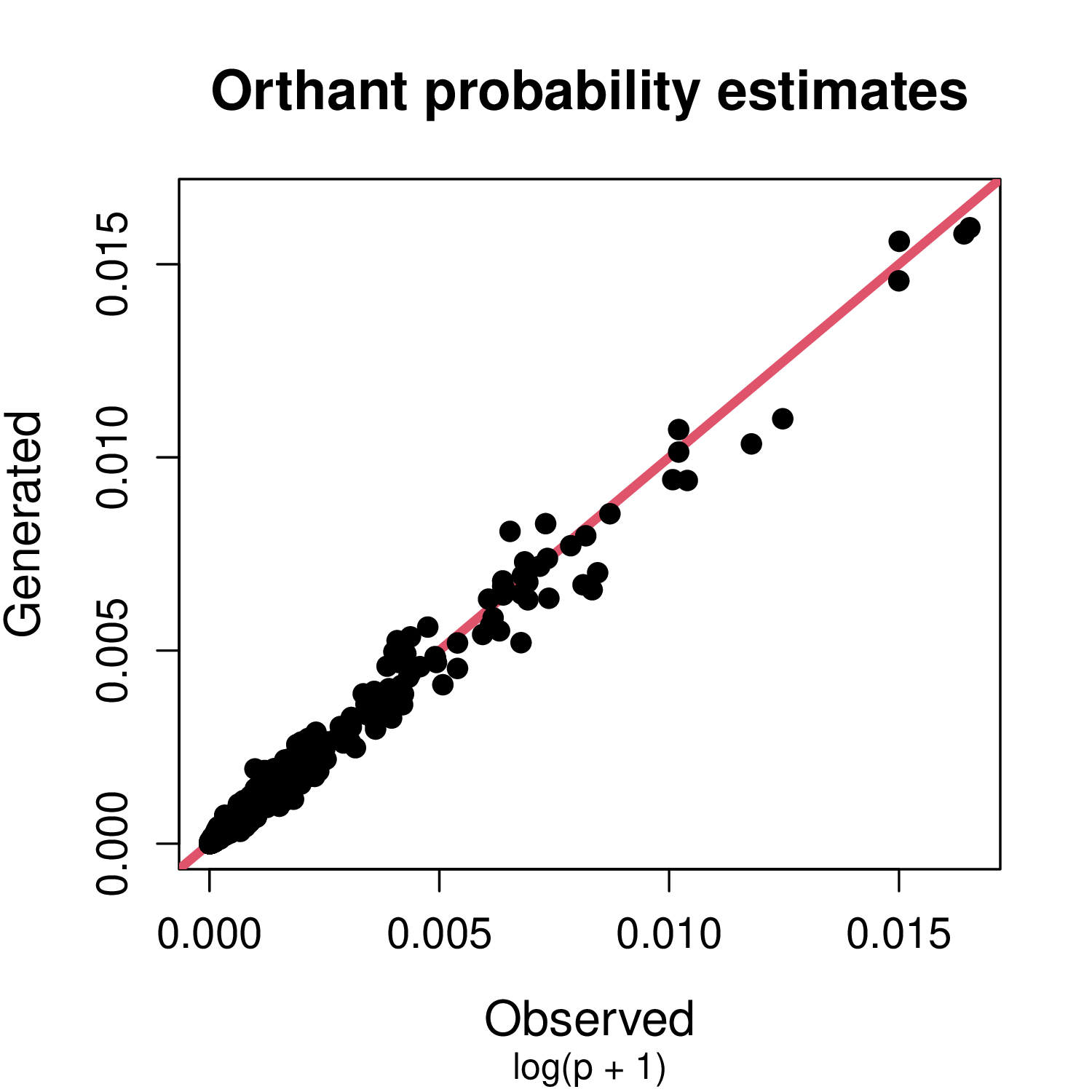}
    \end{subfigure}%
    \begin{subfigure}[b]{0.2\textwidth}
        \centering
        \includegraphics[width=\textwidth]{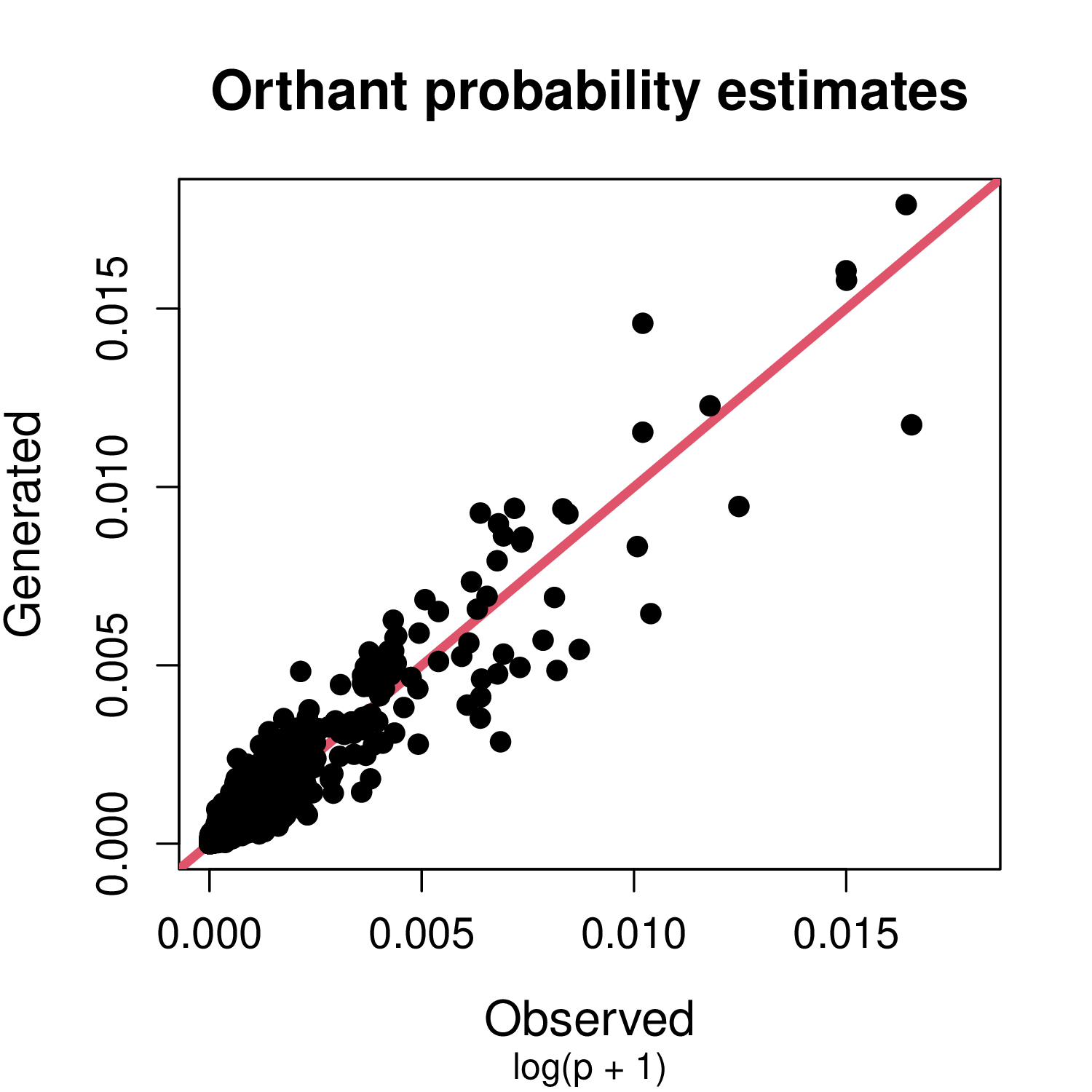}
    \end{subfigure}%
    \begin{subfigure}[b]{0.2\textwidth}
        \centering
        \includegraphics[width=\textwidth]{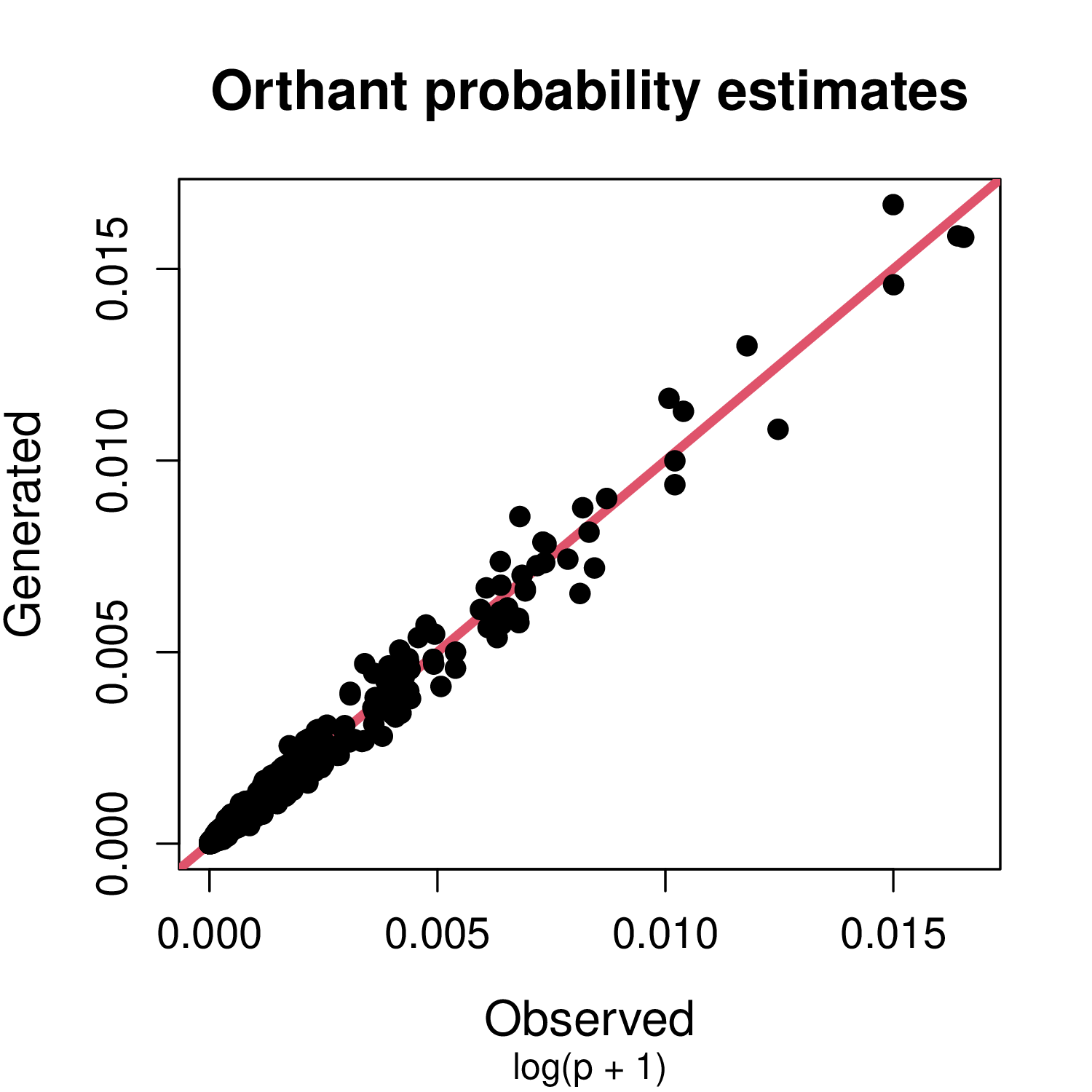}
    \end{subfigure}%

    \begin{subfigure}[b]{0.2\textwidth}
        \centering
        \includegraphics[width=\textwidth]{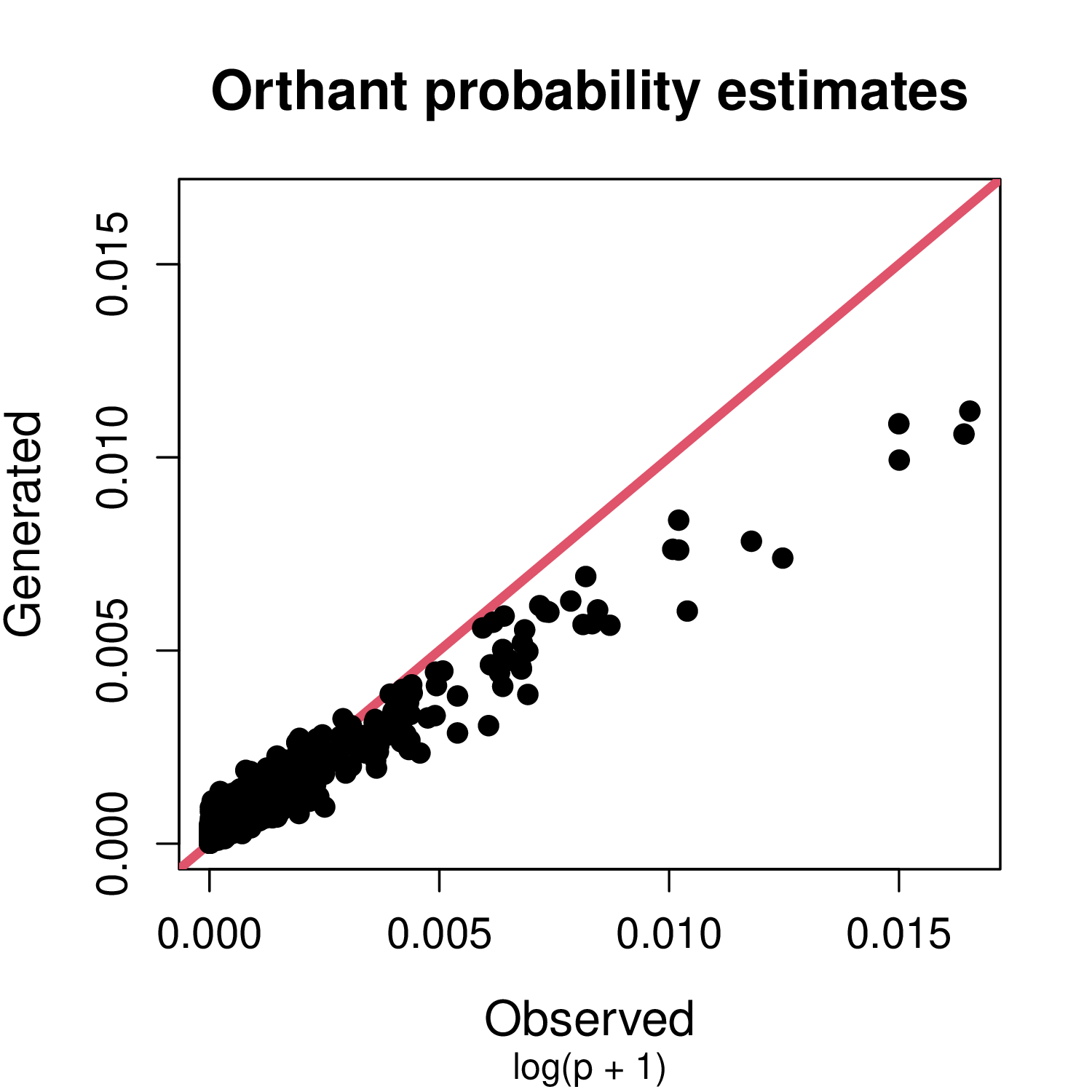}
    \end{subfigure}%
    \begin{subfigure}[b]{0.2\textwidth}
        \centering
        \includegraphics[width=\textwidth]{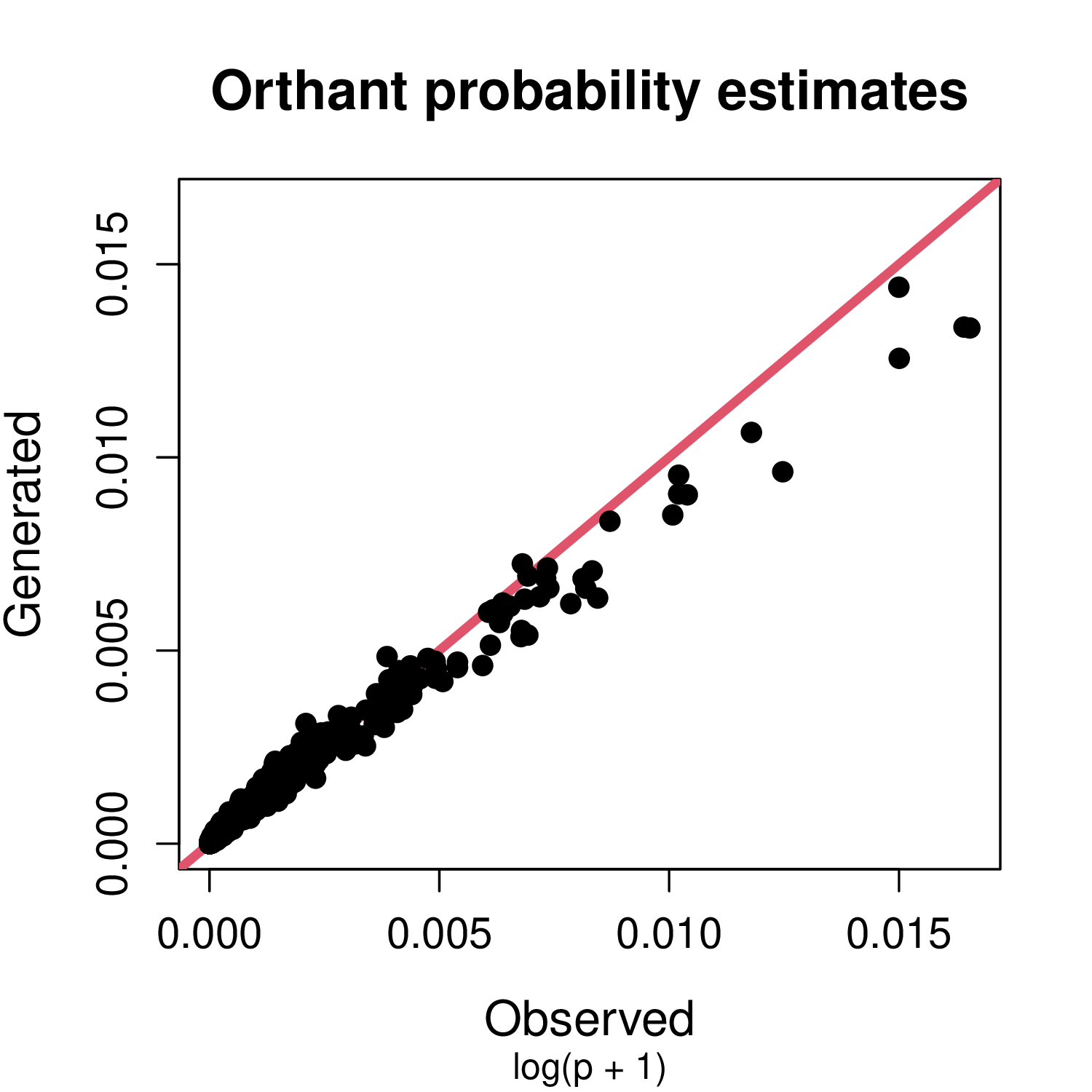}
    \end{subfigure}%
    \begin{subfigure}[b]{0.2\textwidth}
        \centering
        \includegraphics[width=\textwidth]{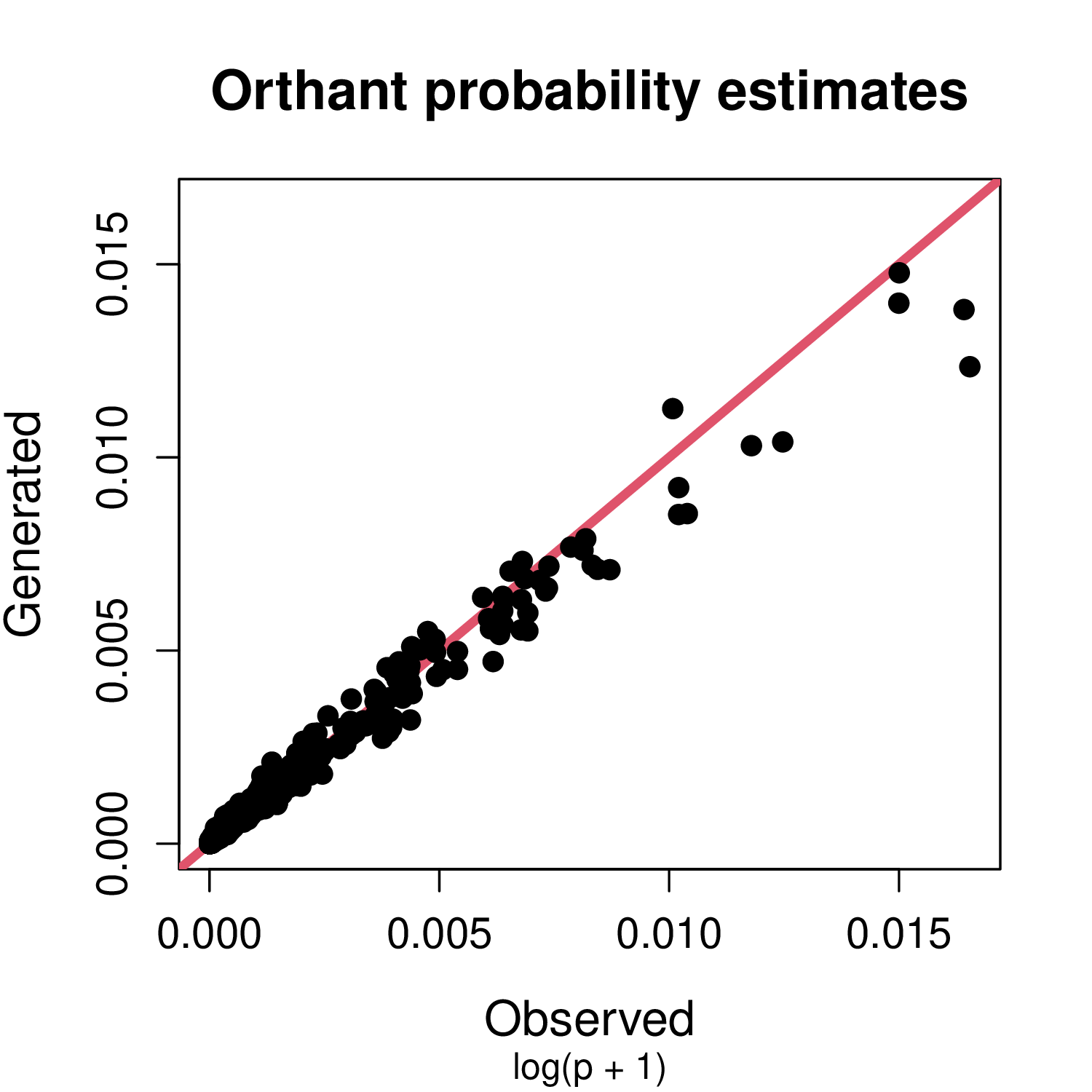}
    \end{subfigure}%
    \begin{subfigure}[b]{0.2\textwidth}
        \centering
        \includegraphics[width=\textwidth]{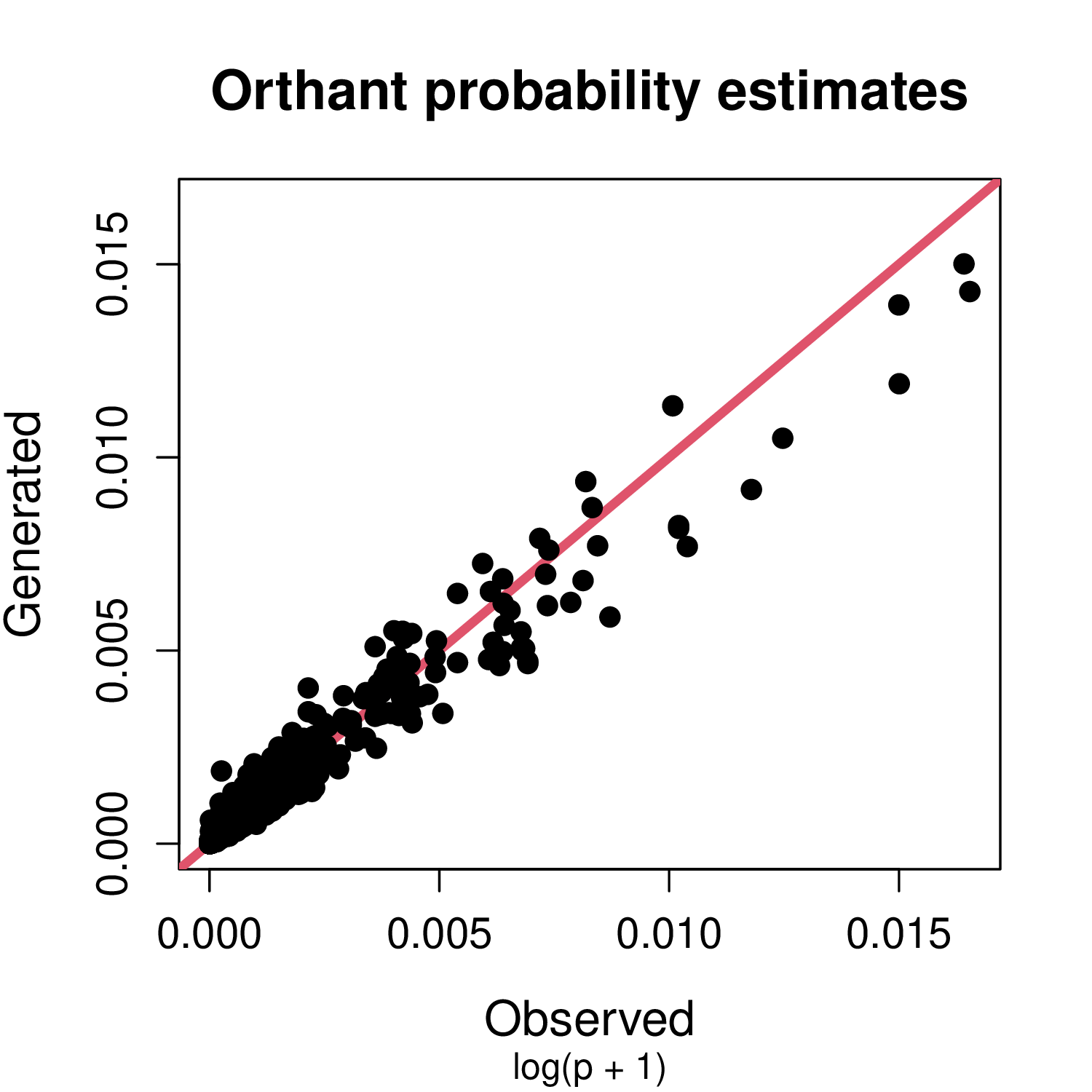}
    \end{subfigure}%
    \begin{subfigure}[b]{0.2\textwidth}
        \centering
        \includegraphics[width=\textwidth]{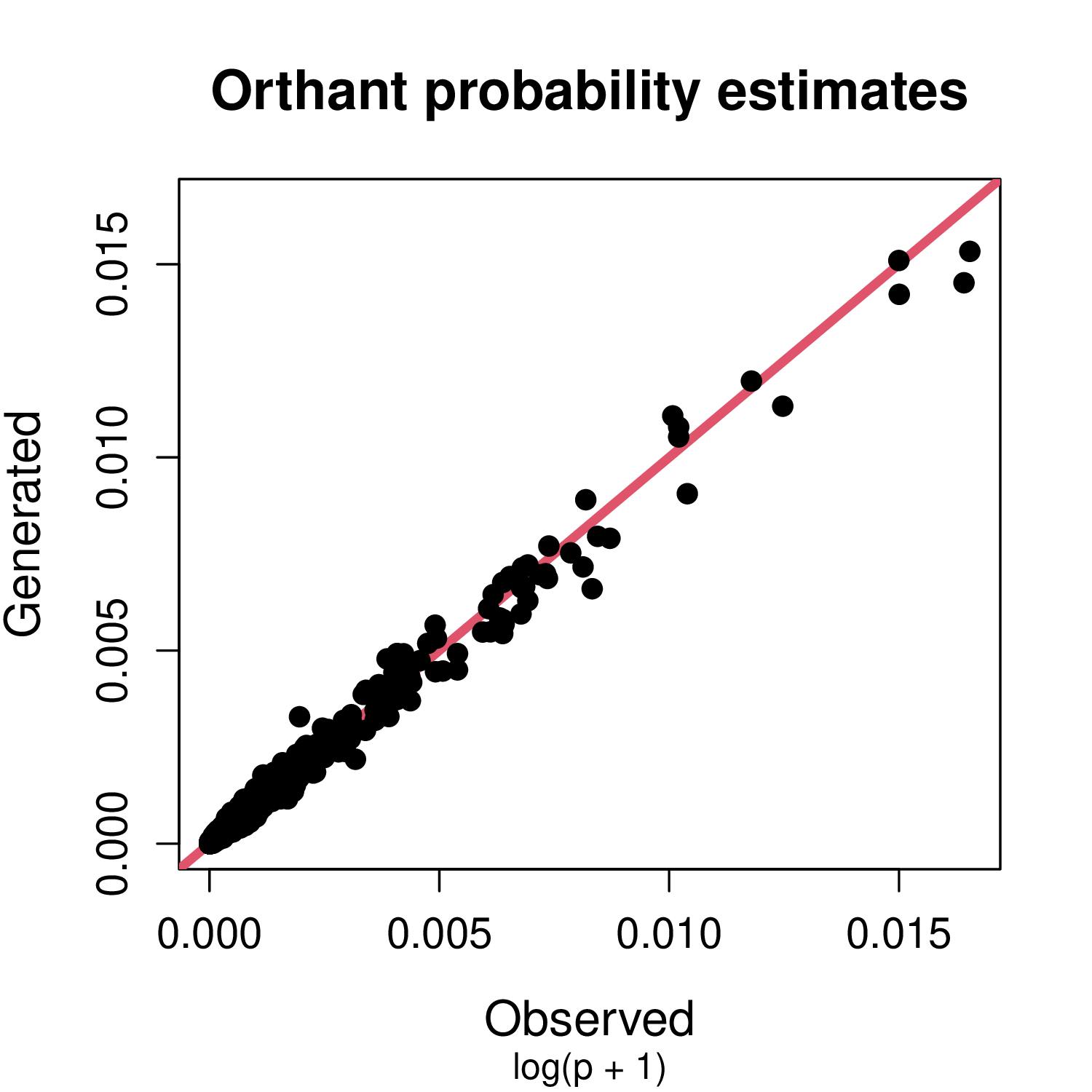}
    \end{subfigure}%
    \caption{Orthant probability plots for copula 2 with $d=10$ and $n=100\ 000$.  }
    \label{fig:orthant_plots_cop2_d10}
\end{figure}

\begin{figure}[h!]
    \centering
    \begin{subfigure}[b]{0.2\textwidth}
        \centering
        \includegraphics[width=\textwidth]{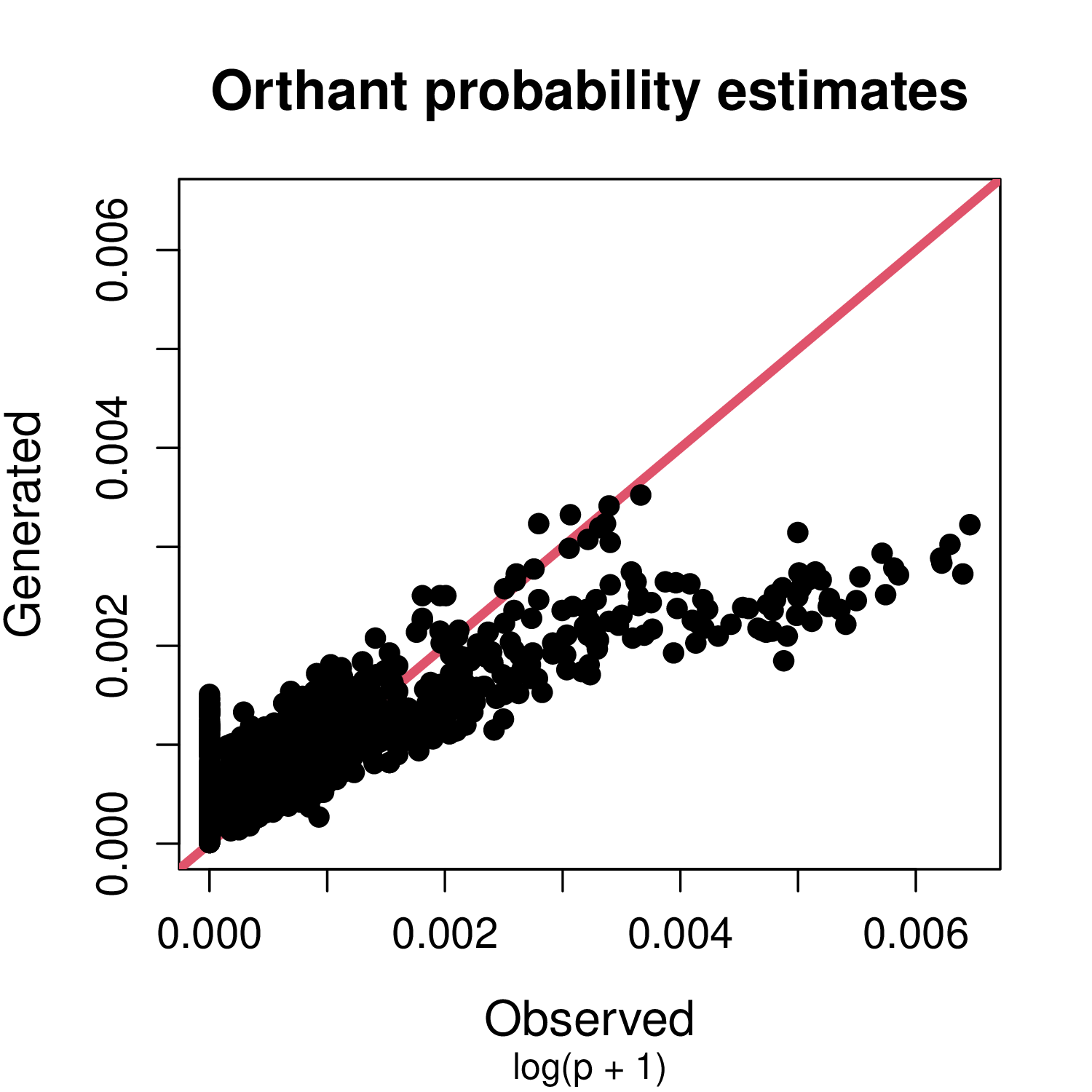}
    \end{subfigure}%
    \begin{subfigure}[b]{0.2\textwidth}
        \centering
        \includegraphics[width=\textwidth]{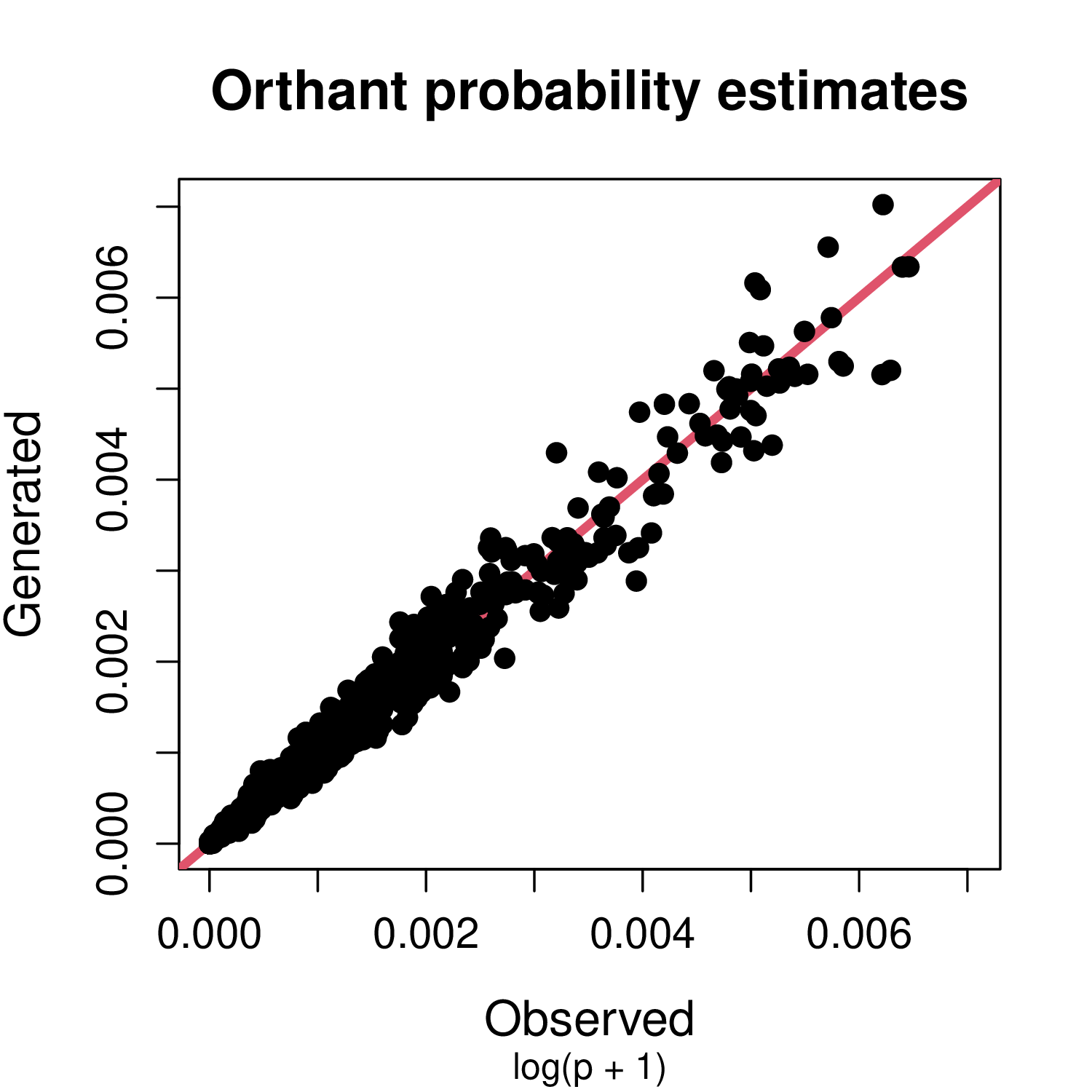}
    \end{subfigure}%
    \begin{subfigure}[b]{0.2\textwidth}
        \centering
        \includegraphics[width=\textwidth]{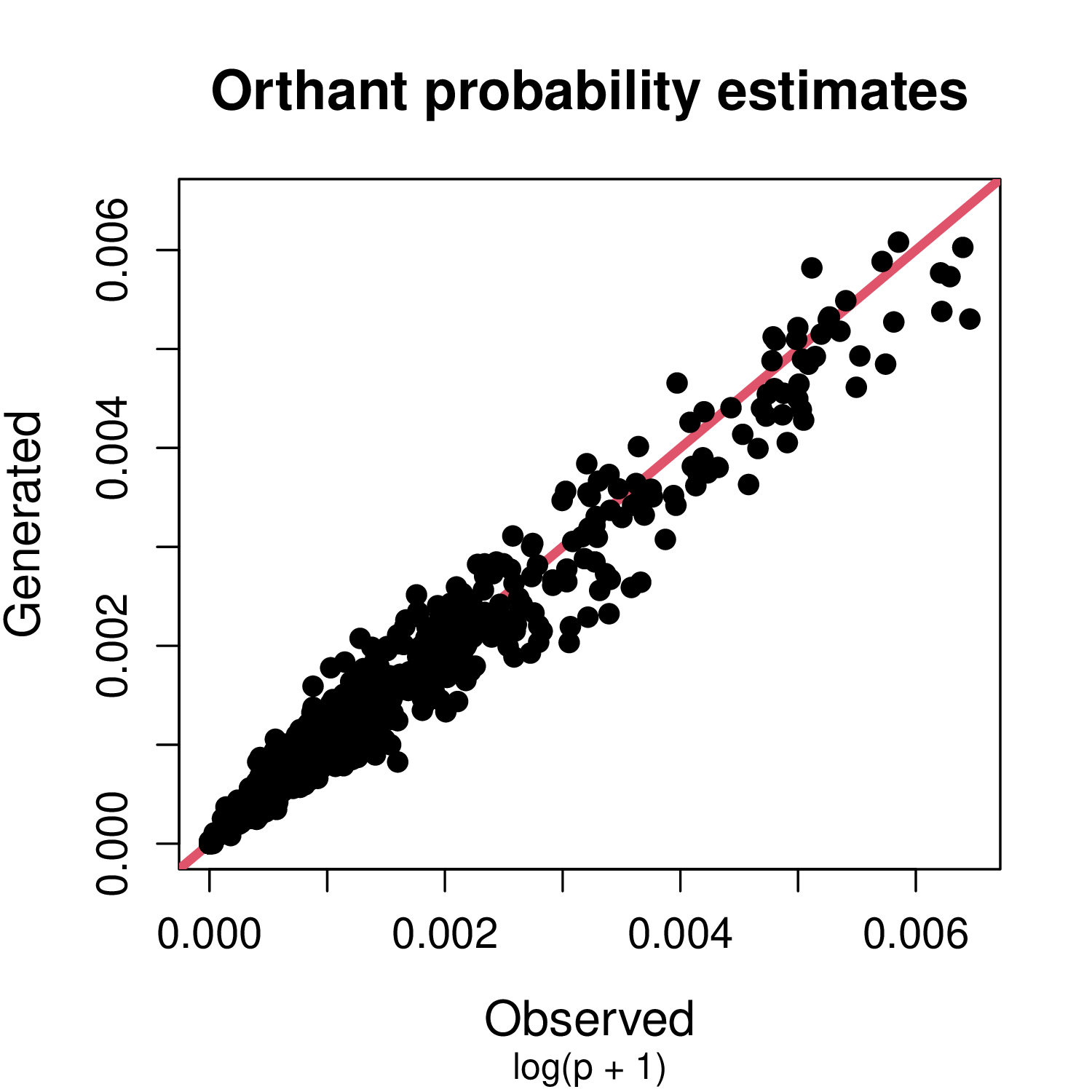}
    \end{subfigure}%
    \begin{subfigure}[b]{0.2\textwidth}
        \centering
        \includegraphics[width=\textwidth]{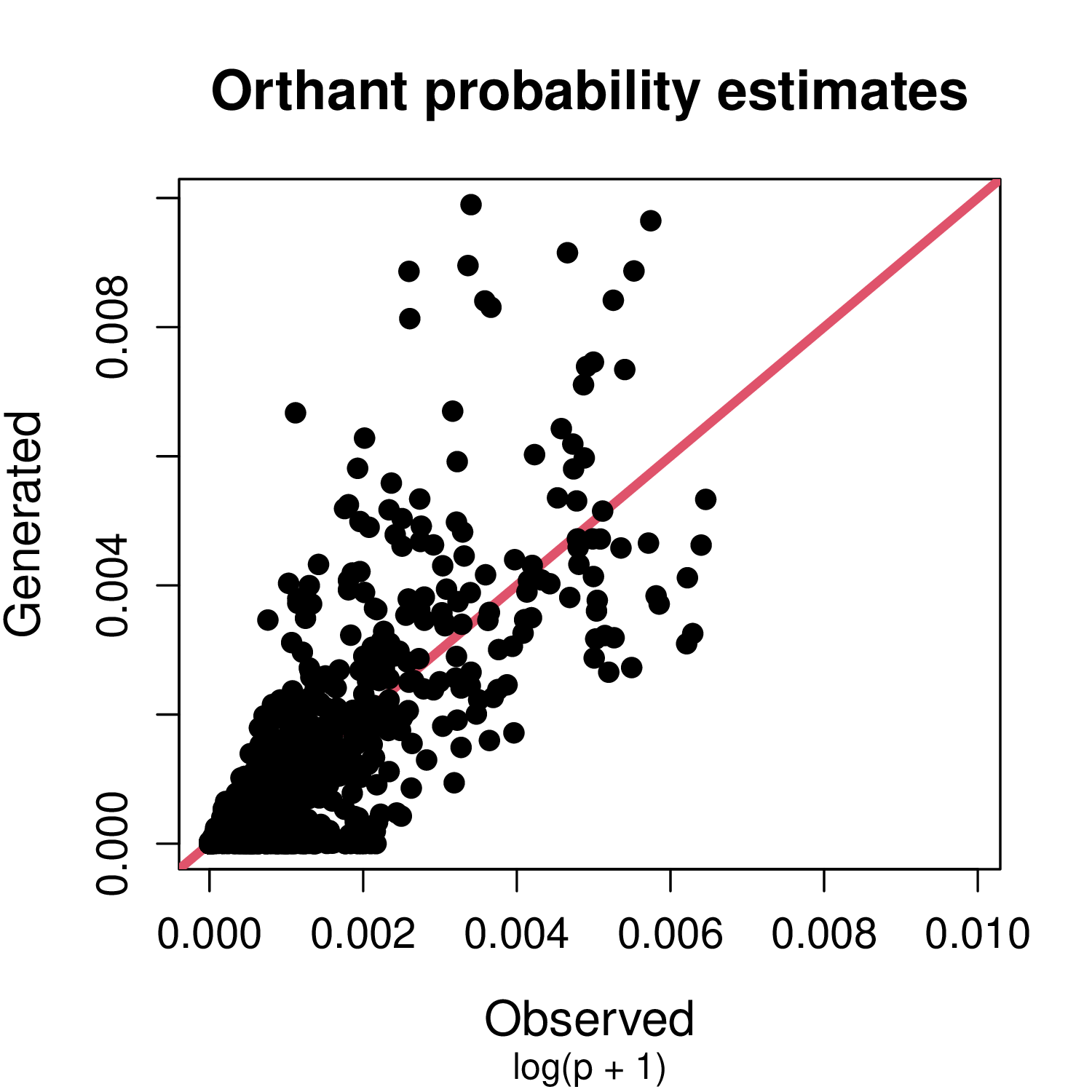}
    \end{subfigure}%
    \begin{subfigure}[b]{0.2\textwidth}
        \centering
        \includegraphics[width=\textwidth]{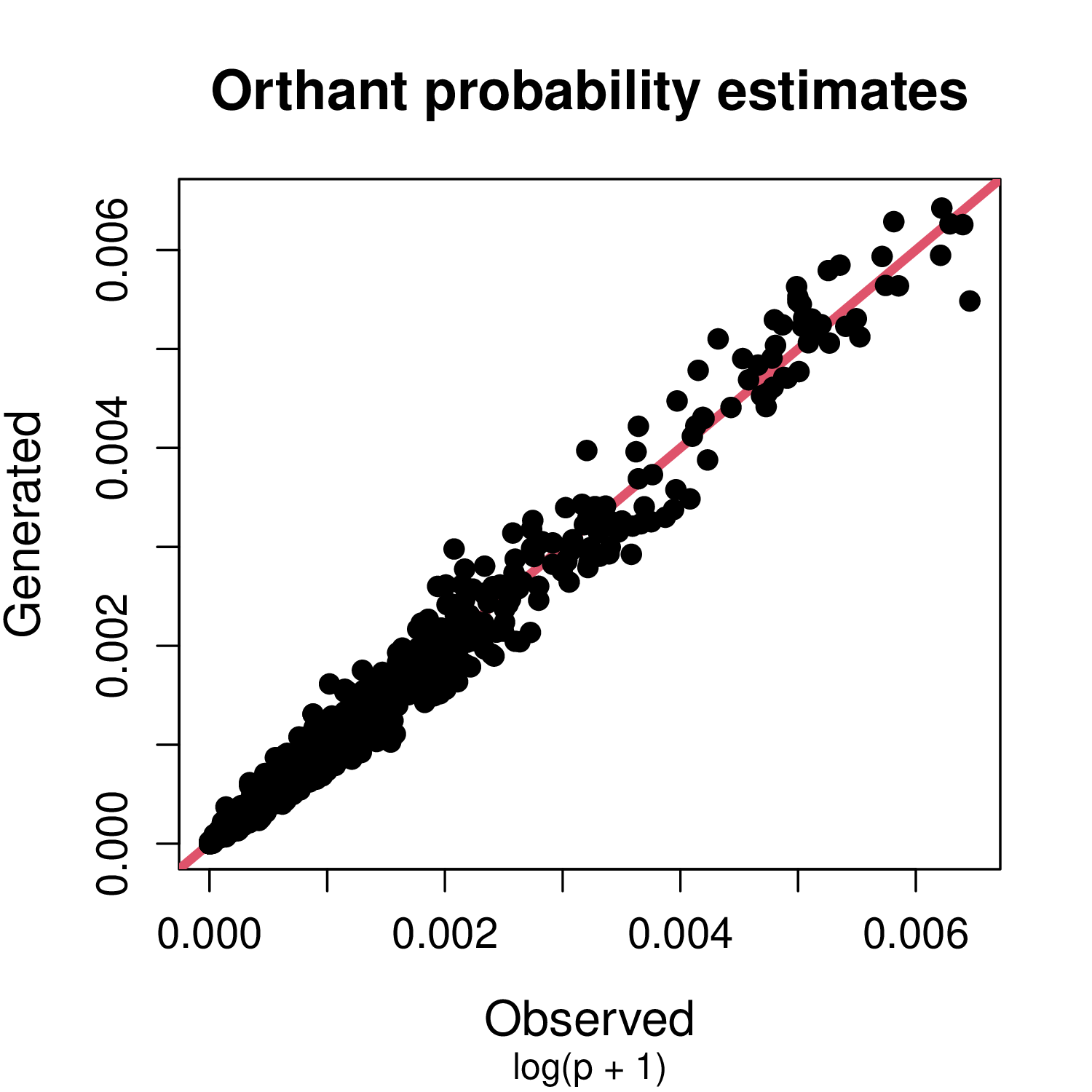}
    \end{subfigure}%

    \begin{subfigure}[b]{0.2\textwidth}
        \centering
        \includegraphics[width=\textwidth]{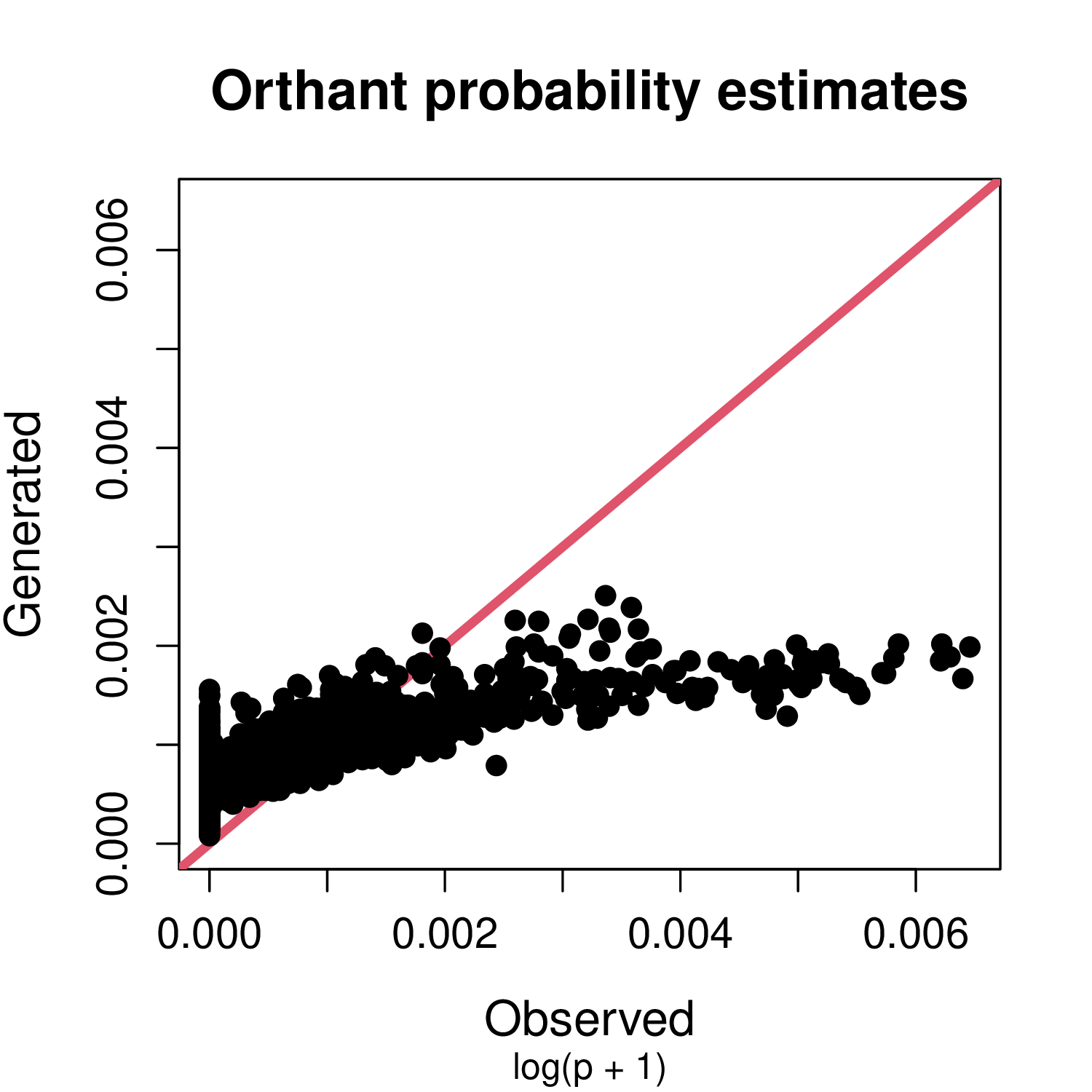}
    \end{subfigure}%
    \begin{subfigure}[b]{0.2\textwidth}
        \centering
        \includegraphics[width=\textwidth]{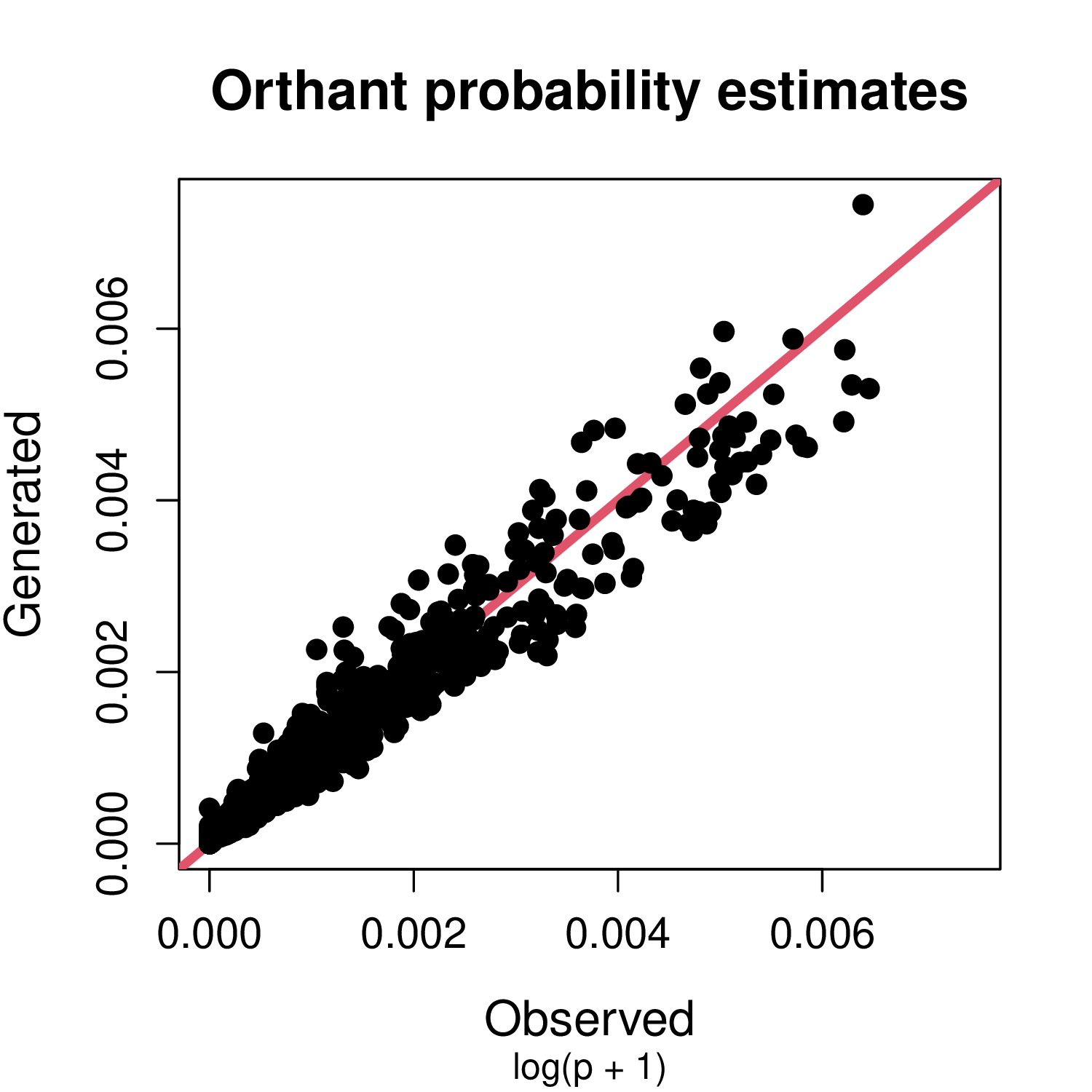}
    \end{subfigure}%
    \begin{subfigure}[b]{0.2\textwidth}
        \centering
        \includegraphics[width=\textwidth]{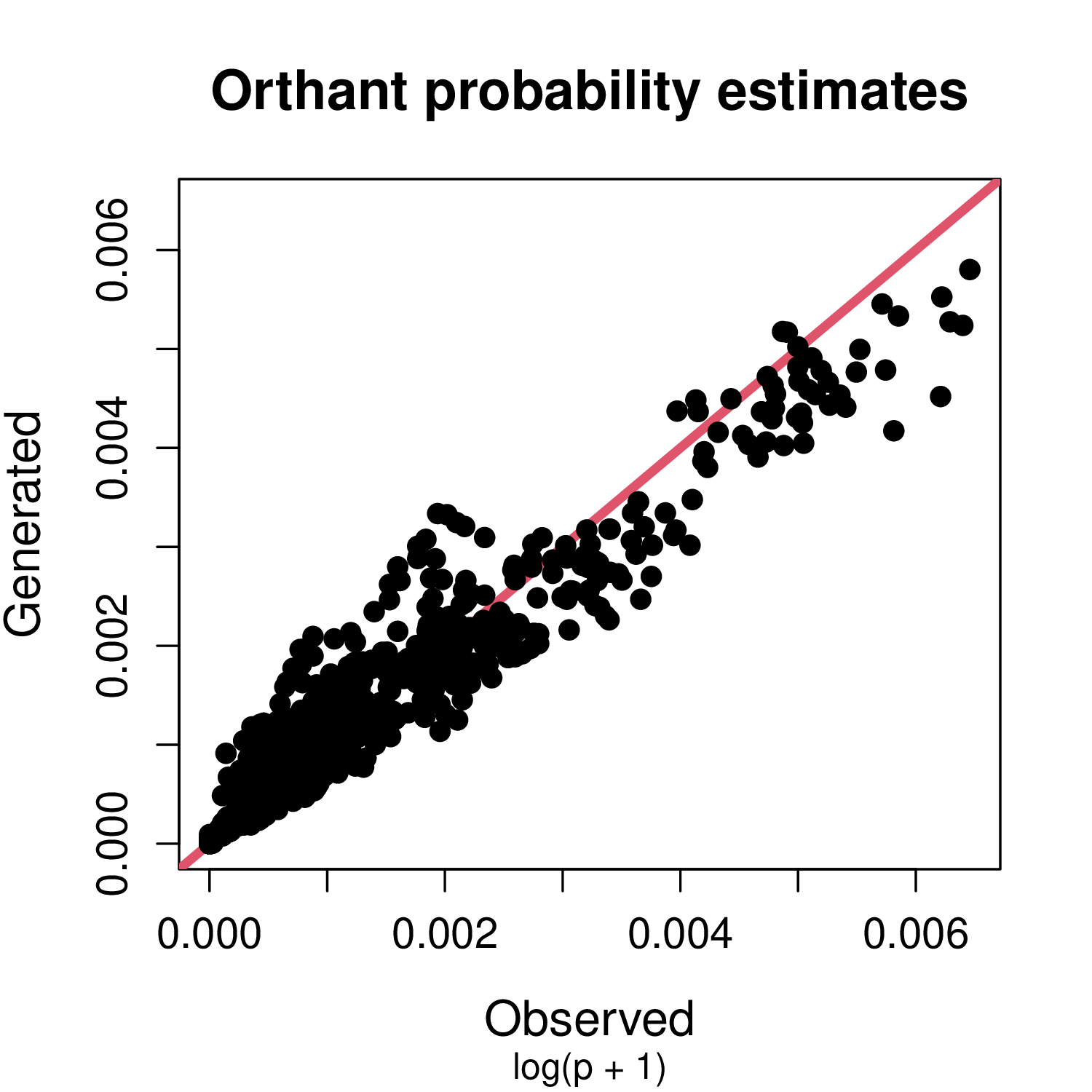}
    \end{subfigure}%
    \begin{subfigure}[b]{0.2\textwidth}
        \centering
        \includegraphics[width=\textwidth]{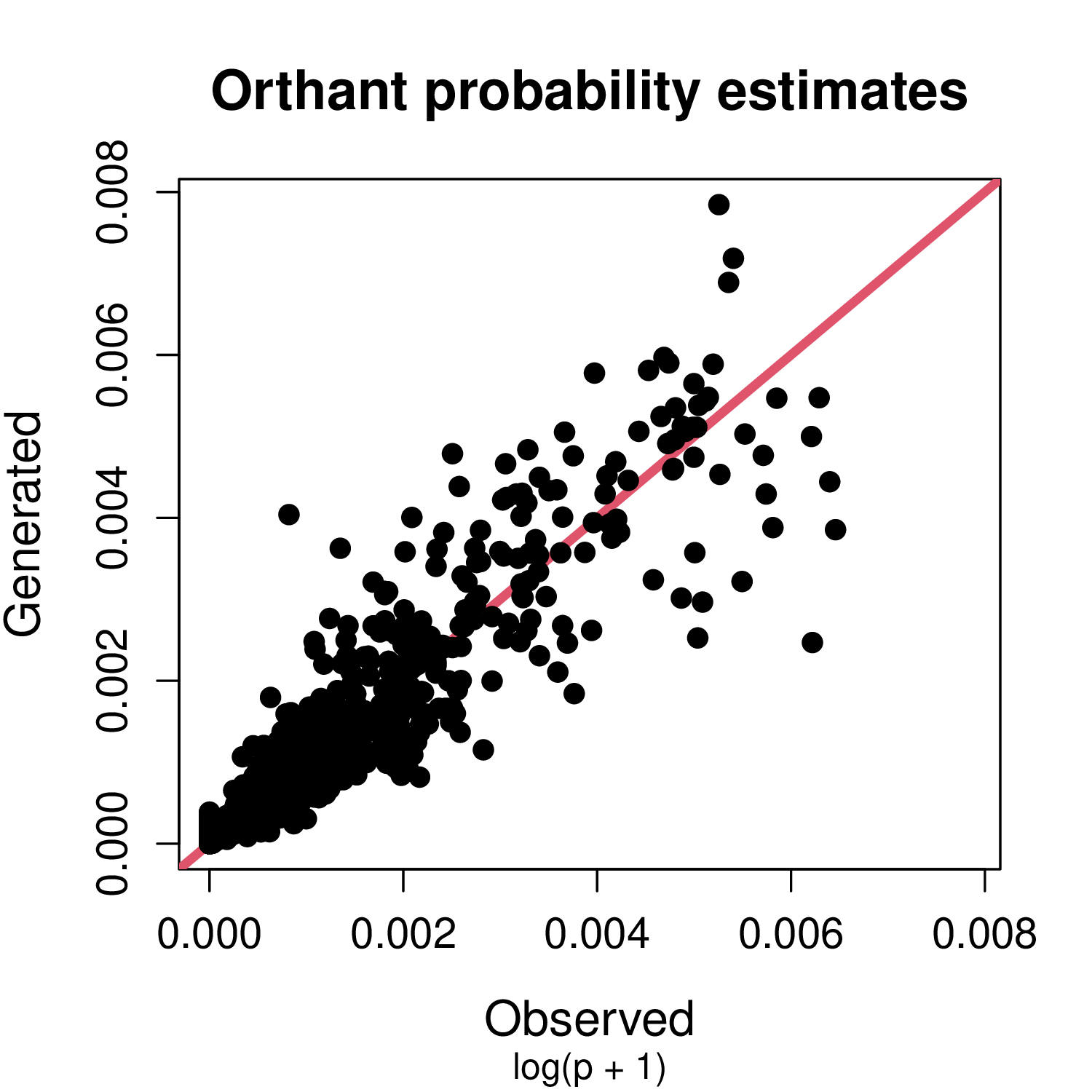}
    \end{subfigure}%
    \begin{subfigure}[b]{0.2\textwidth}
        \centering
        \includegraphics[width=\textwidth]{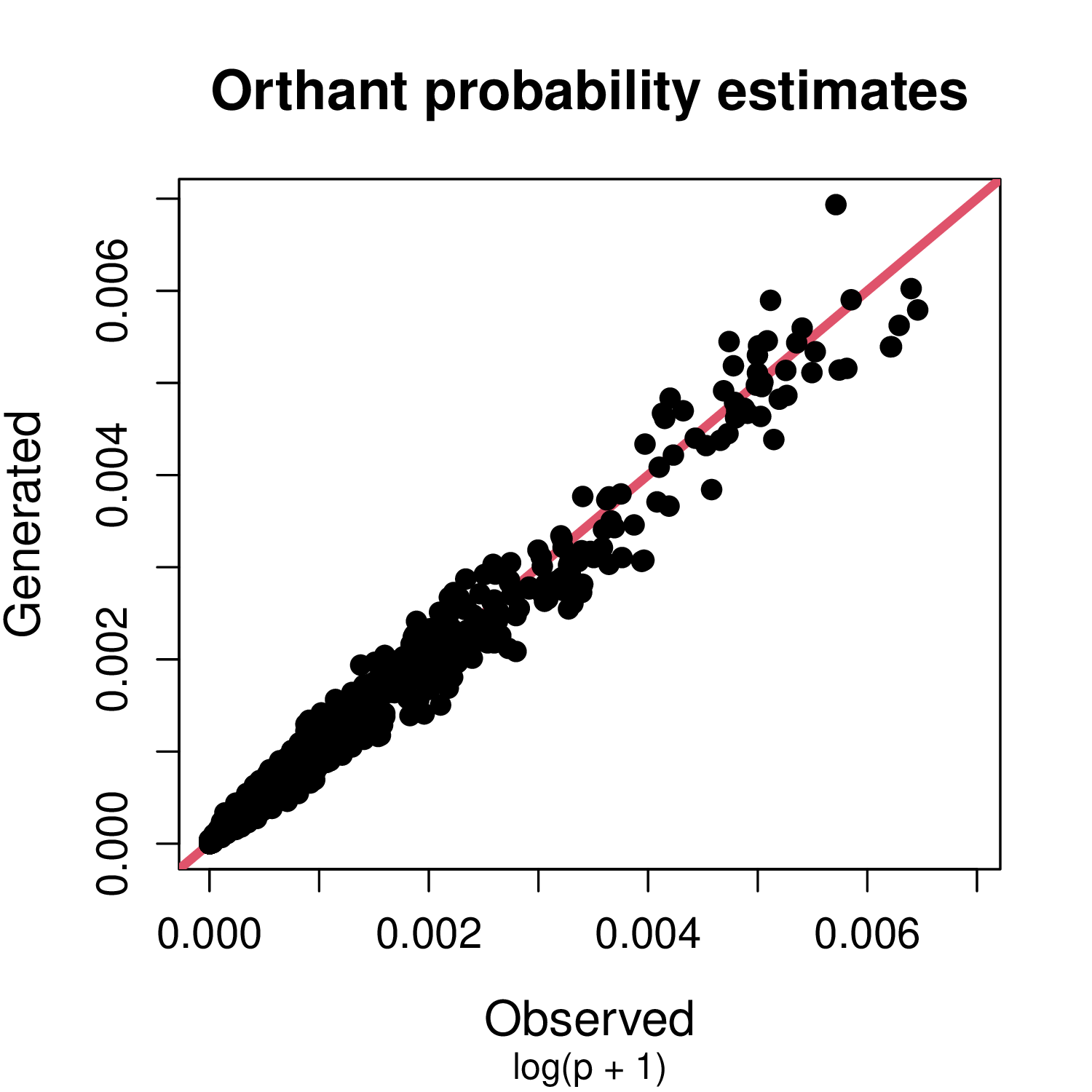}
    \end{subfigure}%
    \caption{Orthant probability plots for copula 5 with $d=10$ and $n=100\ 000$.  }
    \label{fig:orthant_plots_cop5_d10}
\end{figure}

Figures~\ref{fig:qq_plots_cop2_d10}--\ref{fig:orthant_plots_cop5_d10} give the QQ and orthant probability plots obtained from each deep generative approach for copulas 2 and 5 and $d = 10$, with data simulated on both marginal distributions. The ordering of methods and margins is as in Figure~\ref{fig:qq_plots_cop2_d5} of the main text.

\clearpage

\begin{landscape}

\begin{table}[!h]
\centering\centering
\caption{Average $\operatorname{Skill}(F_{*})$ ($* \in \{\text{FM},\text{NFMAF},\text{GAN},\text{NFNSF} \}$) scores (to 6 significant figures) and standard deviations (to 3 significant figures) over 50 simulated samples from copula 5 with $d = 10$ for each deep generative approach.}
\centering
\begin{tabular}[t]{c|c|c|c|c|c|c}
\hline
Copula & Margins & n & FM & NFMAF & GAN & NFNSF\\
\hline
5 & Double Pareto & $10^{3}$ & \textbf{1.00125 (0.000586)} & 1.00189 (0.00085) & 1.02691 (0.0248) & 1.00152 (0.000782)\\
\hline
5 & Double Pareto & $10^{4}$ & 1.00033 (0.00014) & 1.00048 (0.000334) & 1.00085 (0.000342) & \textbf{1.00026 (0.000129)}\\
\hline
5 & Double Pareto & $10^{5}$ & 1.00033 (0.000187) & 1.00031 (0.000259) & 1.0011 (0.000545) & \textbf{1.00013 (7.56e-05)}\\
\hline
5 & Laplace & $10^{3}$ & \textbf{1.00134 (0.00067)} & 1.0028 (0.00131) & 1.04122 (0.0368) & 1.00187 (0.000754)\\
\hline
5 & Laplace & $10^{4}$ & 1.00035 (0.000184) & \textbf{1.00032 (0.000163)} & 1.00355 (0.00144) & 1.00045 (0.000362)\\
\hline
5 & Laplace & $10^{5}$ & 1.00039 (0.000193) & \textbf{1.00027 (0.000162)} & 1.02163 (0.00494) & 1.00046 (0.00041)\\
\hline
\end{tabular}
\end{table}

\end{landscape}

\clearpage

\begin{table}[!h]
\centering\centering
\caption{$\operatorname{Skill}(F_{*})$ scores (to 6 significant figures) of each deep generative approach ($* \in \{\text{FM},\text{NFMAF},\text{GAN},\text{NFNSF} \} $) for the $d = 50$ examples.\\}
\centering
\begin{tabular}[t]{c|c|c|c|c|c|c}
\hline
Copula & n & d & FM & NFMAF & GAN & NFNSF\\
\hline
Sparse Gaussian, Laplace margins & $10^{5}$ & 50 & 1.00212 & \textbf{1.00064} & 1.00880 & 1.00106\\
\hline
Sparse Gaussian, Double Pareto margins & $10^{5}$ & 50 & 1.01489 & 1.00067 & 1.00349 & \textbf{1.00051}\\
\hline
Randomised GAN & $10^{5}$ & 50 & 1.04867 & \textbf{1.00178} & 1.46359 & 1.00249\\
\hline
\end{tabular}
\end{table}

\begin{figure}[h!]
    \centering
    \includegraphics[width=\textwidth]{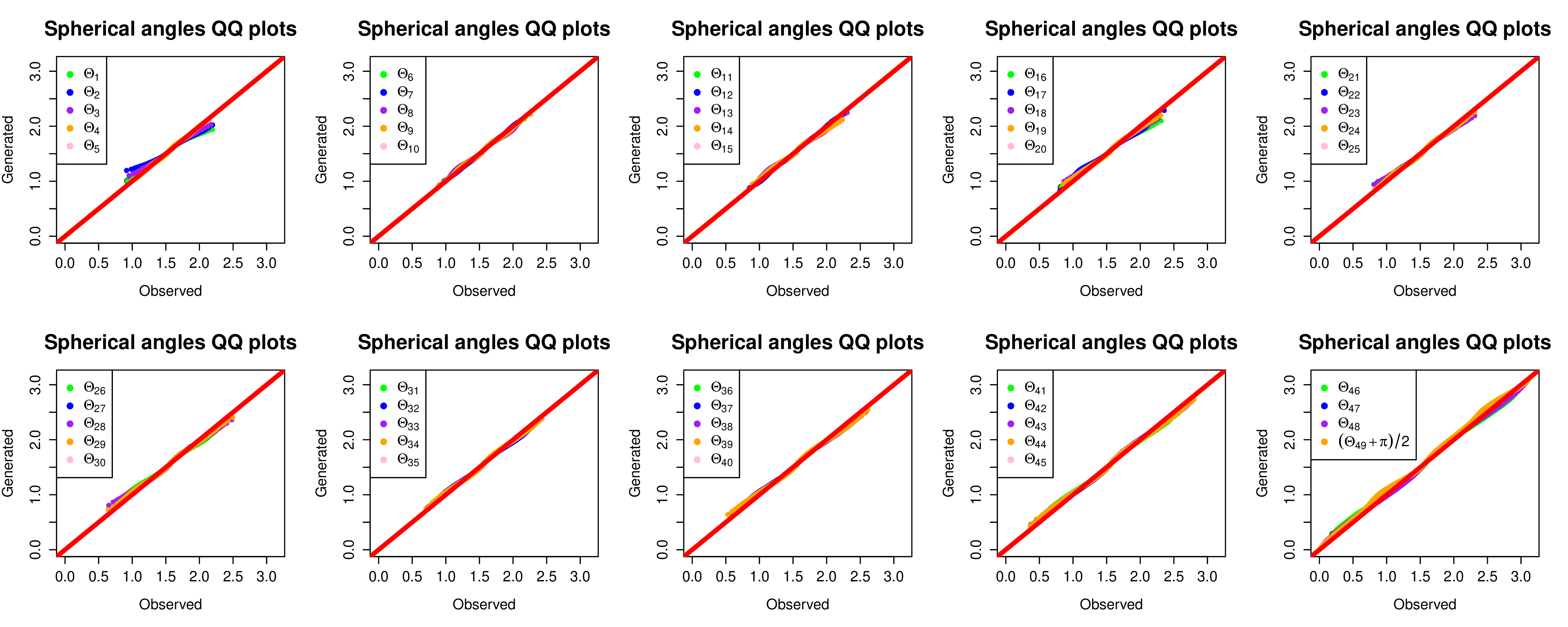}
    \caption{Spherical angle QQ plots for sparse Gaussian copula with $d=50$, $n=100\ 000$, and Laplace margins, with data generated using the von-Mises mixture model.}
    \label{fig:qq_plots_50_1_VMMIX}
\end{figure}

\begin{figure}[h!]
    \centering
    \includegraphics[width=\textwidth]{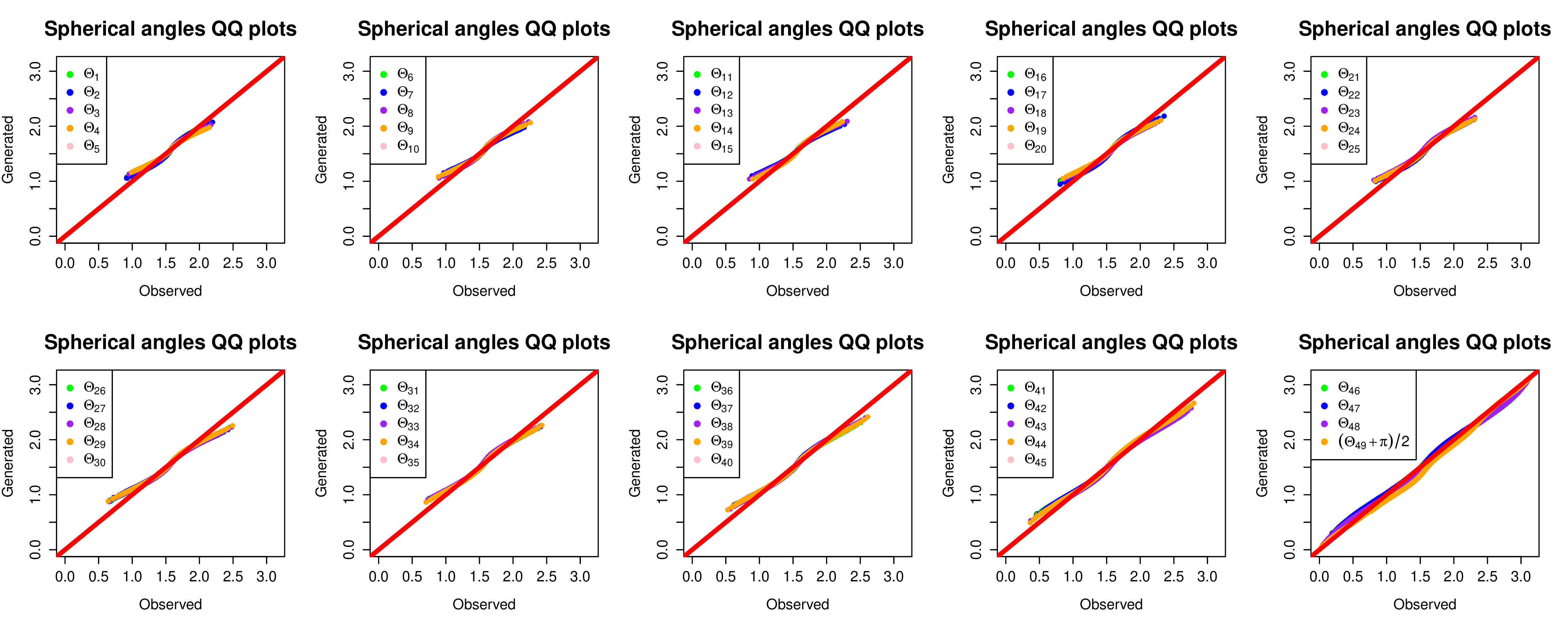}
    \caption{Spherical angle QQ plots for sparse Gaussian copula with $d=50$, $n=100\ 000$, and Laplace margins, with data generated using the FM model.}
    \label{fig:qq_plots_50_1_FM}
\end{figure}

\begin{figure}[h!]
    \centering
    \includegraphics[width=\textwidth]{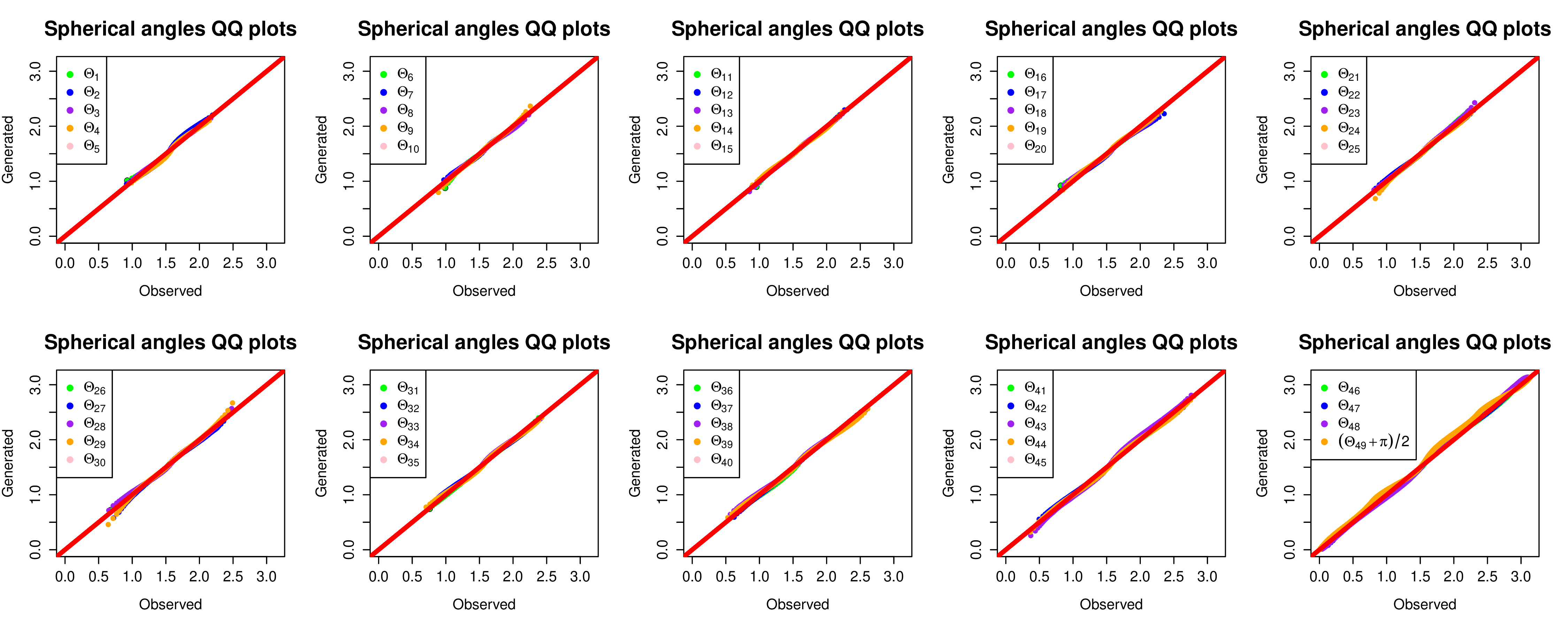}
    \caption{Spherical angle QQ plots for sparse Gaussian copula with $d=50$, $n=100\ 000$, and Laplace margins, with data generated using the NFMAF model.}
    \label{fig:qq_plots_50_1_NFMAF}
\end{figure}

\begin{figure}[h!]
    \centering
    \includegraphics[width=\textwidth]{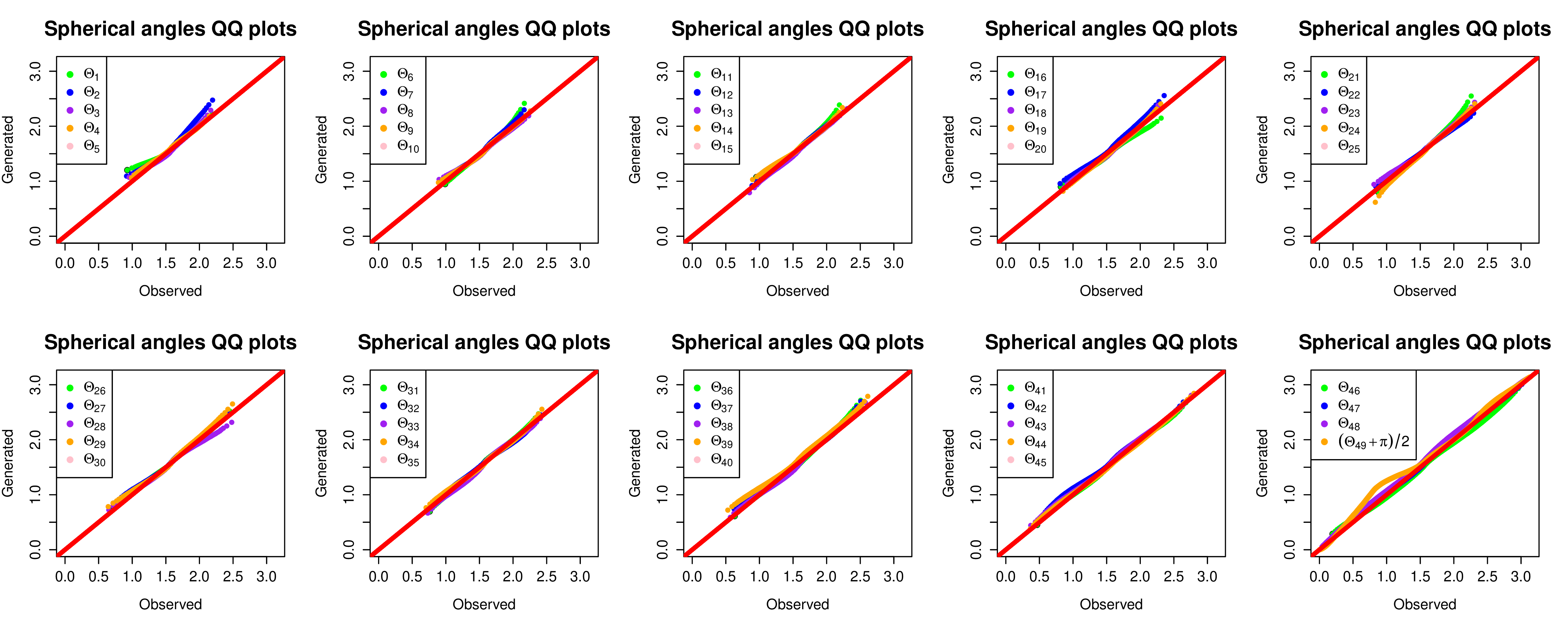}
    \caption{Spherical angle QQ plots for sparse Gaussian copula with $d=50$, $n=100\ 000$, and Laplace margins, with data generated using the GAN model.}
    \label{fig:qq_plots_50_1_GAN}
\end{figure}

\begin{figure}[h!]
    \centering
    \includegraphics[width=\textwidth]{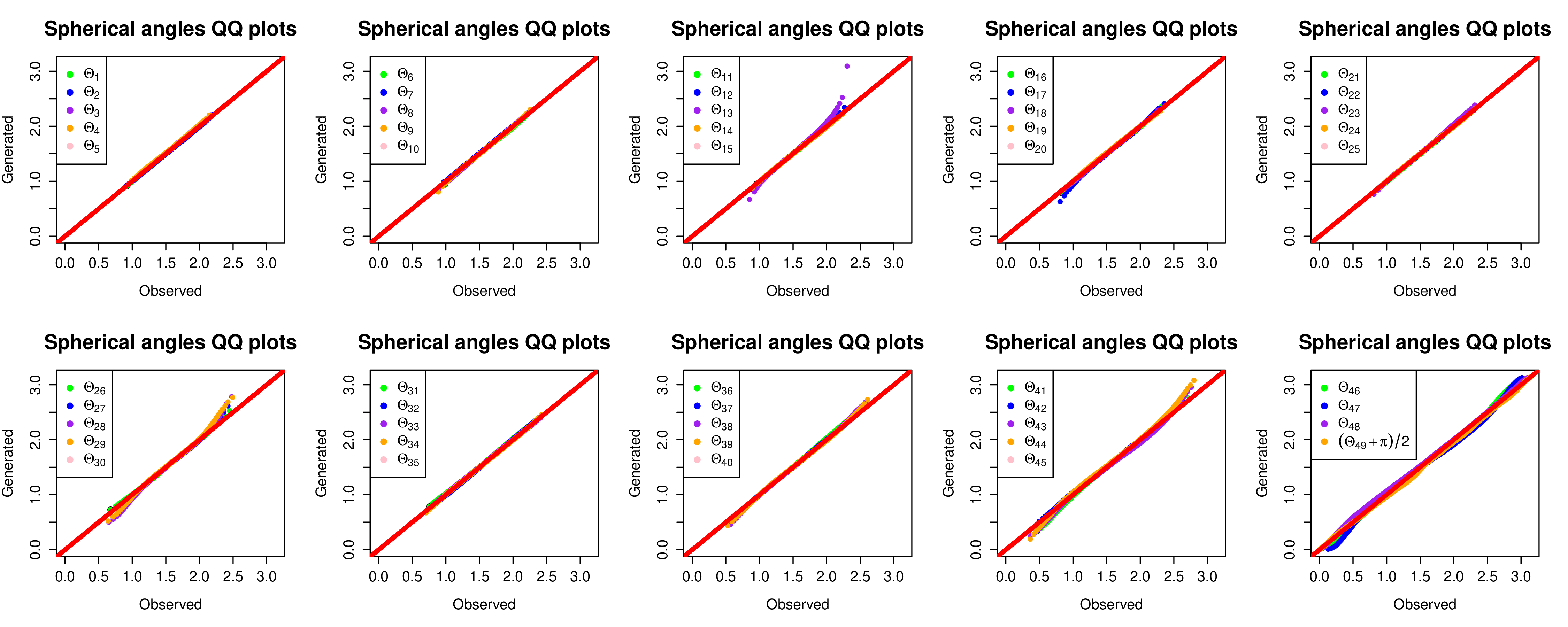}
    \caption{Spherical angle QQ plots for sparse Gaussian copula with $d=50$, $n=100\ 000$, and Laplace margins, with data generated using the NFNSF model.}
    \label{fig:qq_plots_50_1_NFNSF}
\end{figure}

\begin{figure}[h!]
    \centering
    \includegraphics[width=\textwidth]{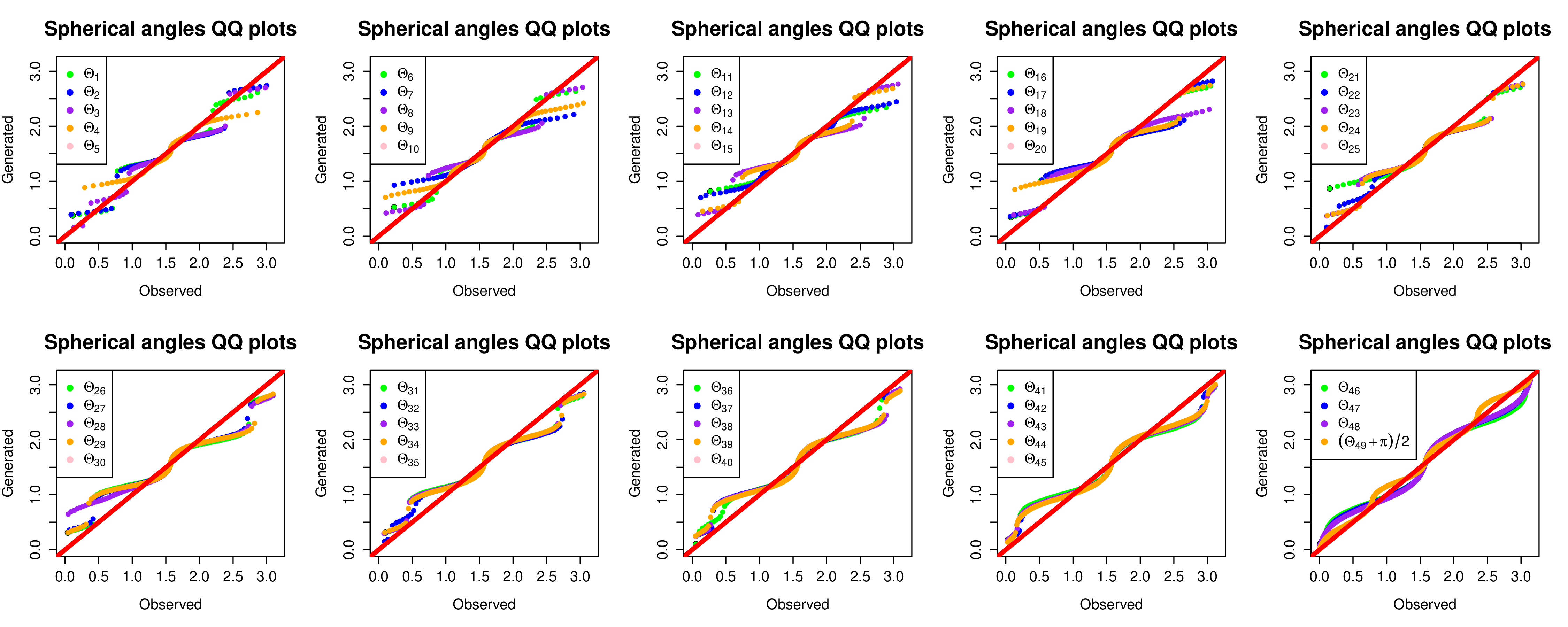}
    \caption{Spherical angle QQ plots for sparse Gaussian copula with $d=50$, $n=100\ 000$, and double Pareto margins, with data generated using the von-Mises mixture model.}
    \label{fig:qq_plots_50_2_VMMIX}
\end{figure}

\begin{figure}[h!]
    \centering
    \includegraphics[width=\textwidth]{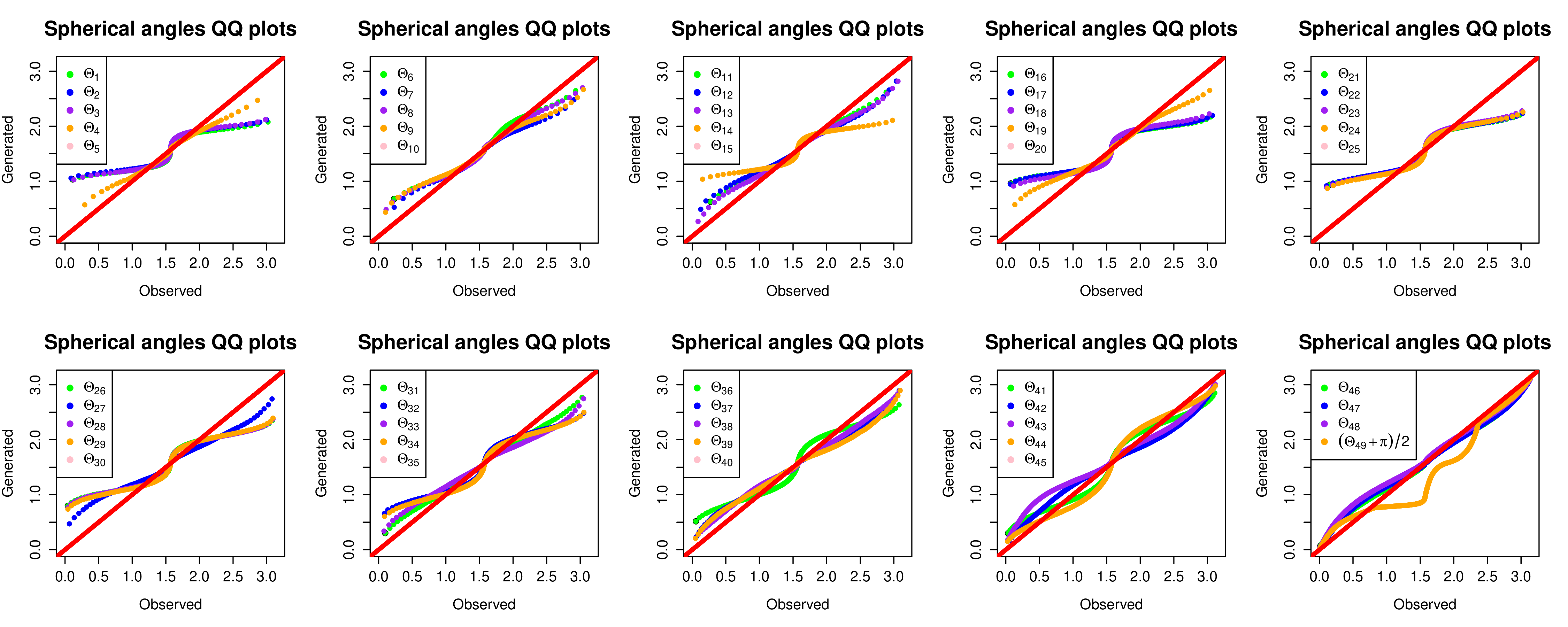}
    \caption{Spherical angle QQ plots for sparse Gaussian copula with $d=50$, $n=100\ 000$, and double Pareto margins, with data generated using the FM model.}
    \label{fig:qq_plots_50_2_FM}
\end{figure}

\begin{figure}[h!]
    \centering
    \includegraphics[width=\textwidth]{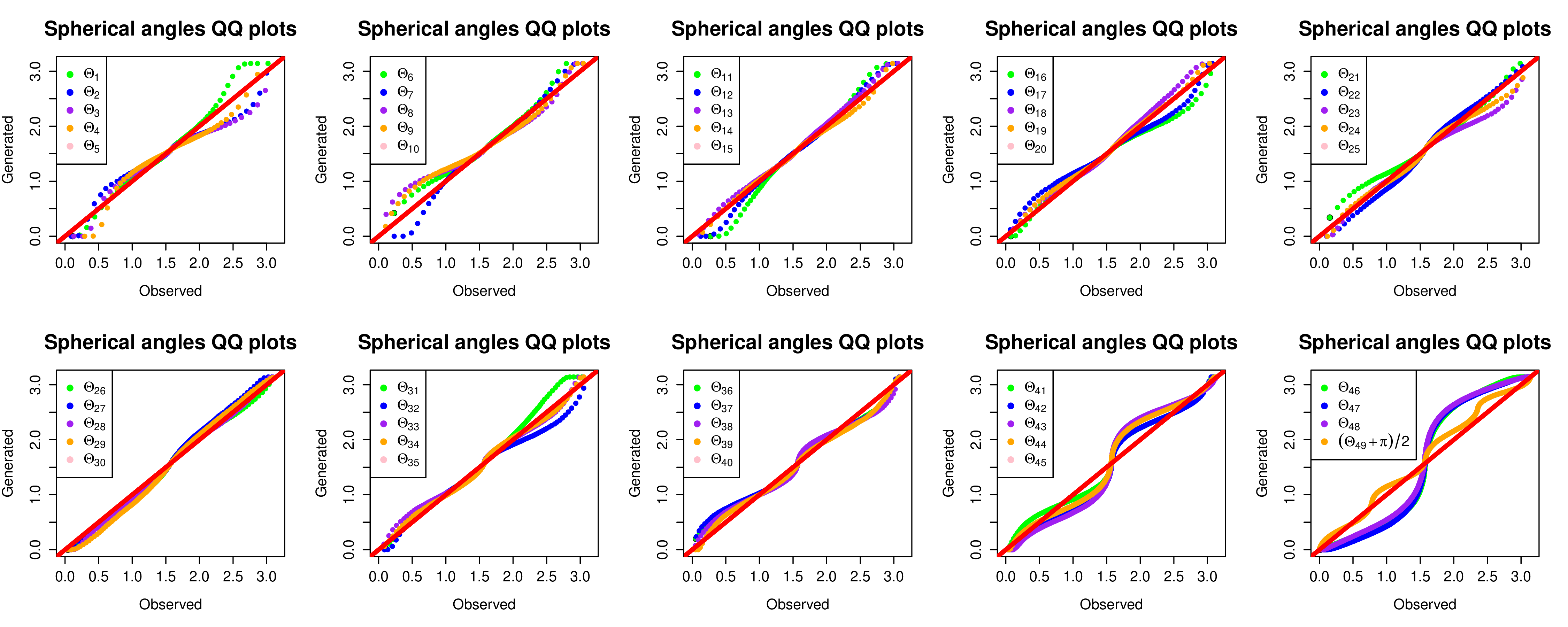}
    \caption{Spherical angle QQ plots for sparse Gaussian copula with $d=50$, $n=100\ 000$, and double Pareto margins, with data generated using the NFMAF model.}
    \label{fig:qq_plots_50_2_NFMAF}
\end{figure}

\begin{figure}[h!]
    \centering
    \includegraphics[width=\textwidth]{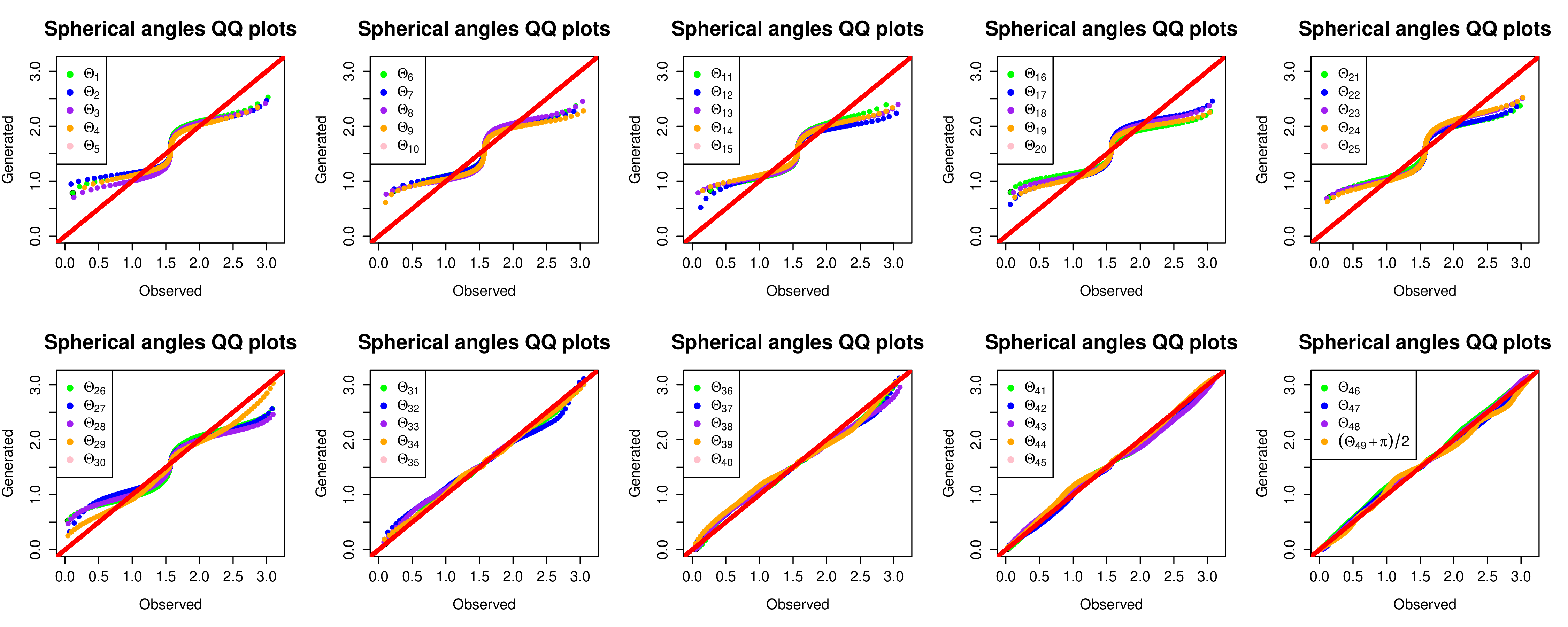}
    \caption{Spherical angle QQ plots for sparse Gaussian copula with $d=50$, $n=100\ 000$, and double Pareto margins, with data generated using the GAN model.}
    \label{fig:qq_plots_50_2_GAN}
\end{figure}

\begin{figure}[h!]
    \centering
    \includegraphics[width=\textwidth]{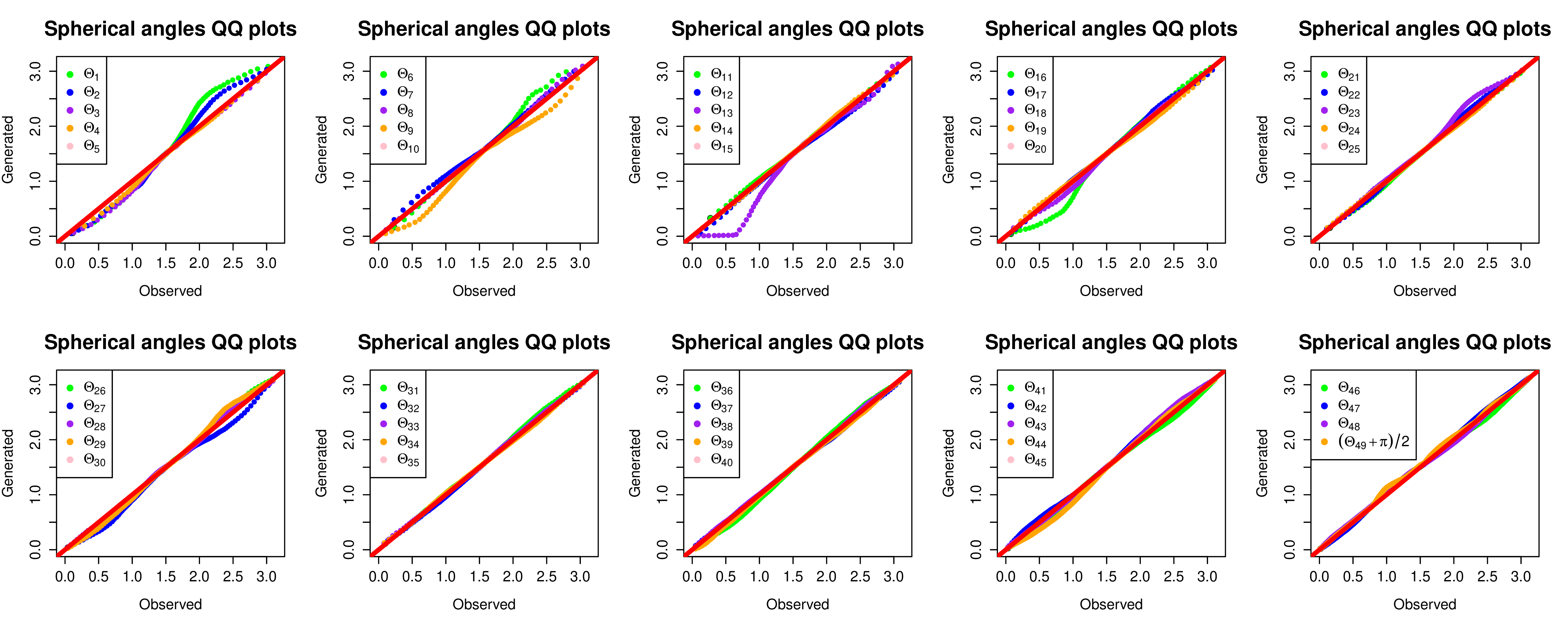}
    \caption{Spherical angle QQ plots for sparse Gaussian copula with $d=50$, $n=100\ 000$, and double Pareto margins, with data generated using the NFNSF model.}
    \label{fig:qq_plots_50_2_NFNSF}
\end{figure}

\begin{figure}[h!]
    \centering
    \includegraphics[width=\textwidth]{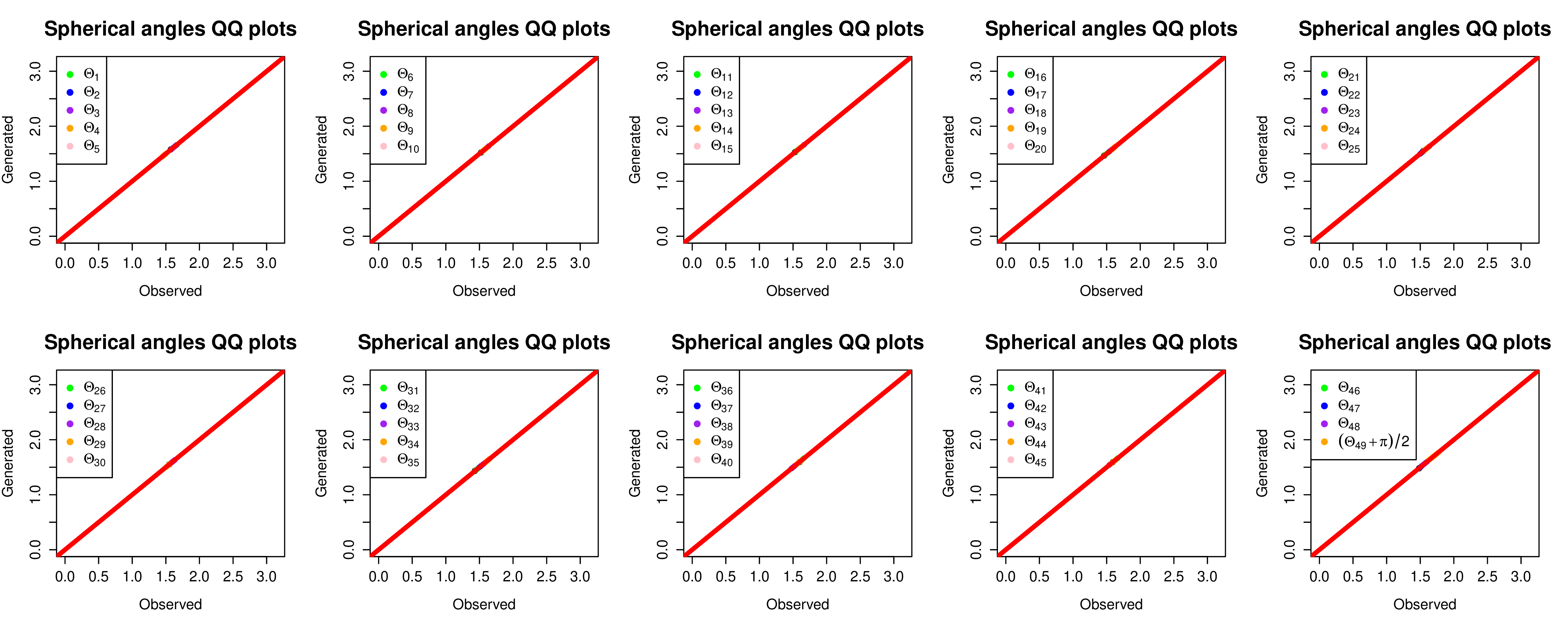}
    \caption{Spherical angle QQ plots for randomised GAN model with $d=50$ and $n=100\ 000$, with data generated using the von-Mises mixture model.}
    \label{fig:qq_plots_50_3_VMMIX}
\end{figure}

\begin{figure}[h!]
    \centering
    \includegraphics[width=\textwidth]{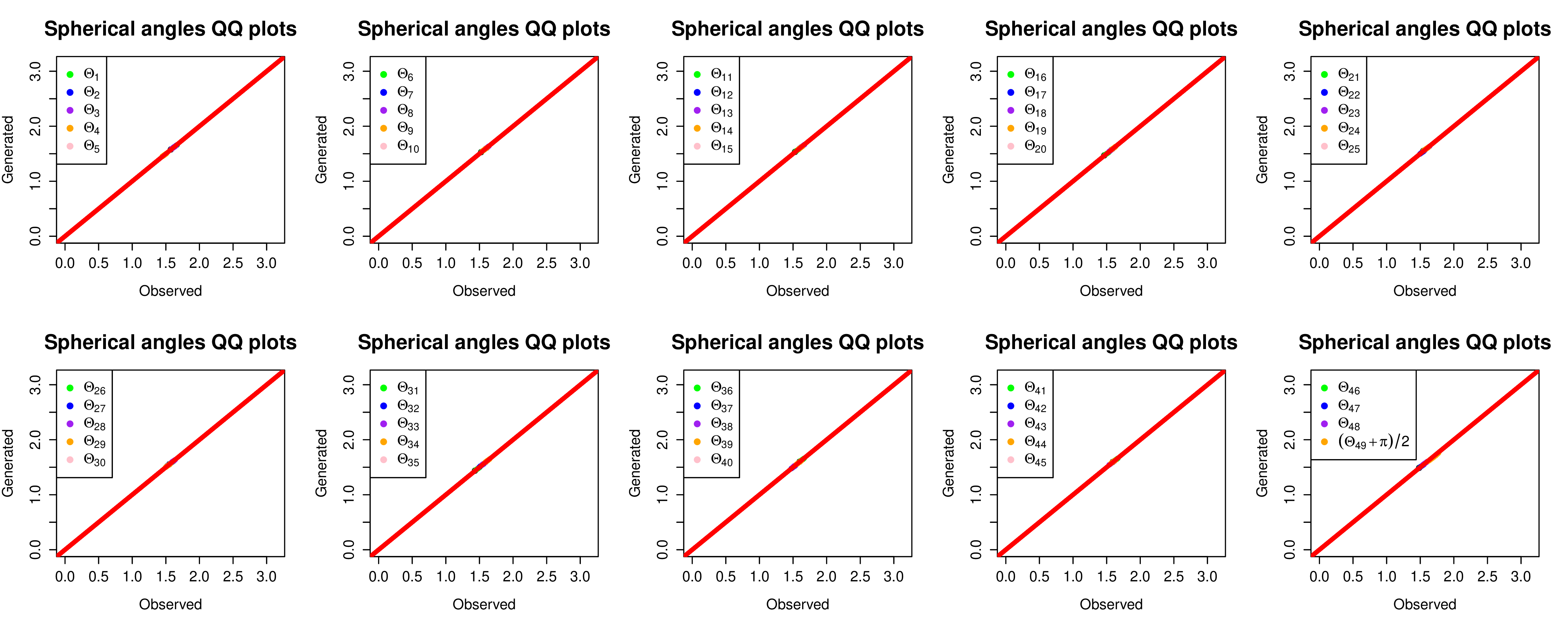}
    \caption{Spherical angle QQ plots for randomised GAN model with $d=50$ and $n=100\ 000$, with data generated using the FM model.}
    \label{fig:qq_plots_50_3_FM}
\end{figure}

\begin{figure}[h!]
    \centering
    \includegraphics[width=\textwidth]{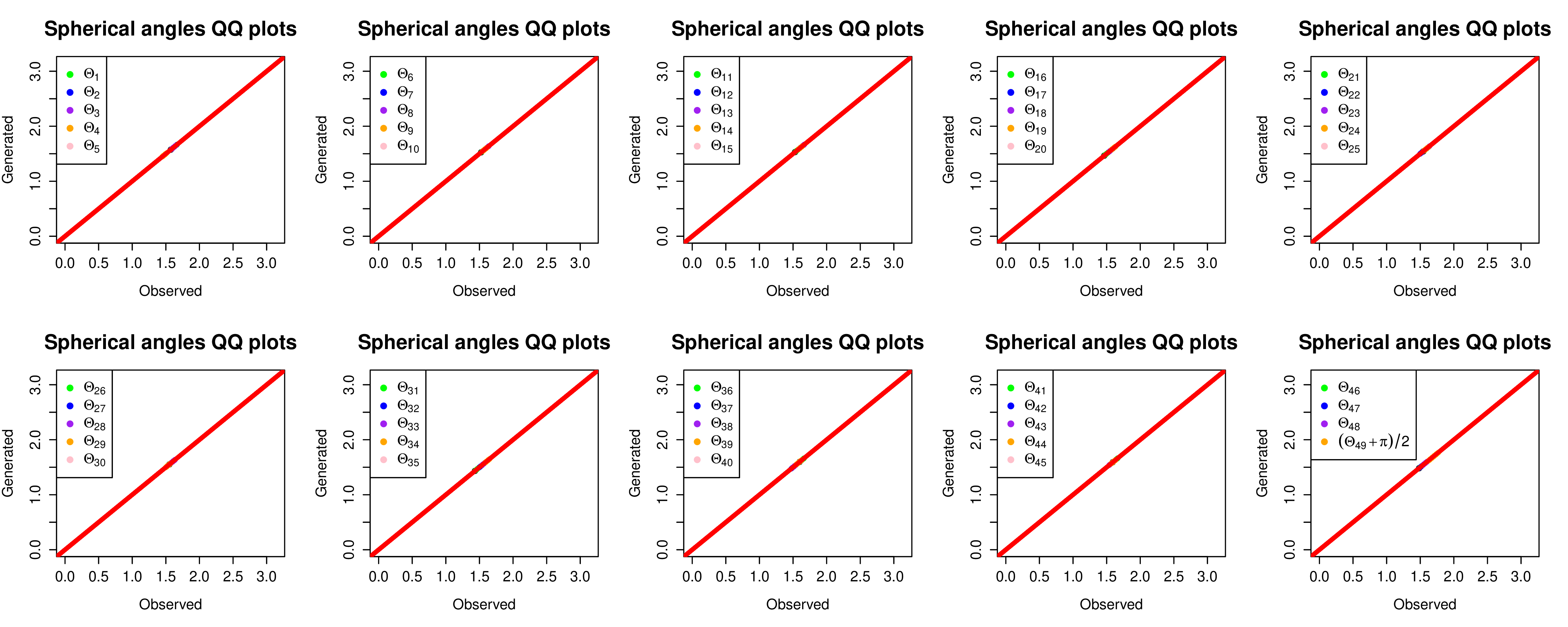}
    \caption{Spherical angle QQ plots for randomised GAN model with $d=50$ and $n=100\ 000$, with data generated using the NFMAF model.}
    \label{fig:qq_plots_50_3_NFMAF}
\end{figure}

\begin{figure}[h!]
    \centering
    \includegraphics[width=\textwidth]{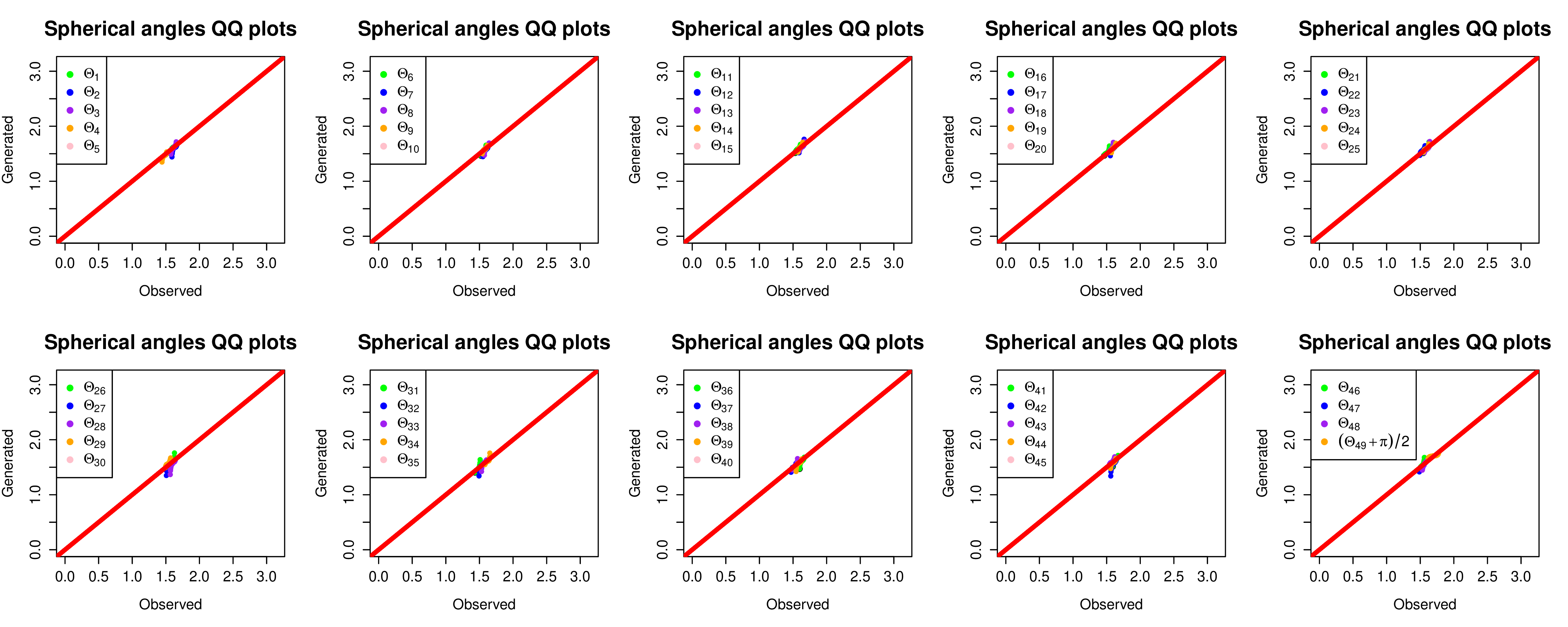}
    \caption{Spherical angle QQ plots for randomised GAN model with $d=50$ and $n=100\ 000$, with data generated using the GAN model.}
    \label{fig:qq_plots_50_3_GAN}
\end{figure}

\begin{figure}[h!]
    \centering
    \includegraphics[width=\textwidth]{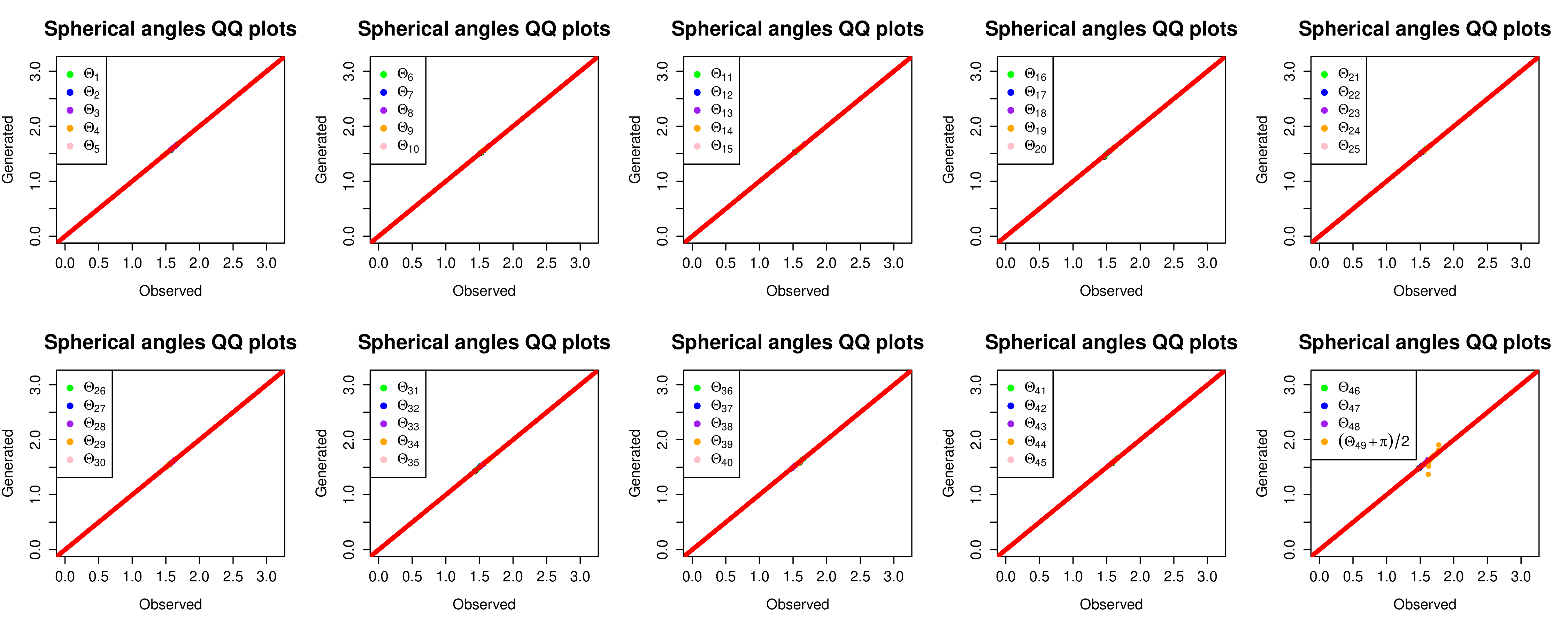}
    \caption{Spherical angle QQ plots for randomised GAN model with $d=50$ and $n=100\ 000$, with data generated using the NFNSF model.}
    \label{fig:qq_plots_50_3_NFNSF}
\end{figure}

\clearpage

\section{Additional case study figures}\label{app:case_figs}

\begin{figure}[!htbp]
    \centering
    \includegraphics[width=\linewidth]{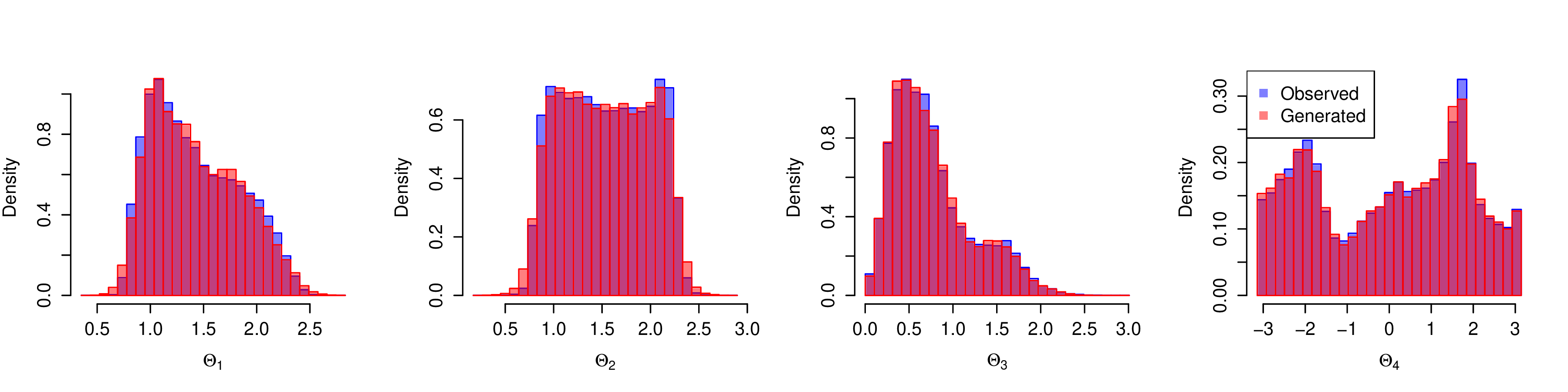}
    \caption{Histogram plots for spherical angles obtained from the baseline approach.}
    \label{fig:hist_wave_baseline}
\vspace{0.5cm}
    \includegraphics[width=.5\linewidth]{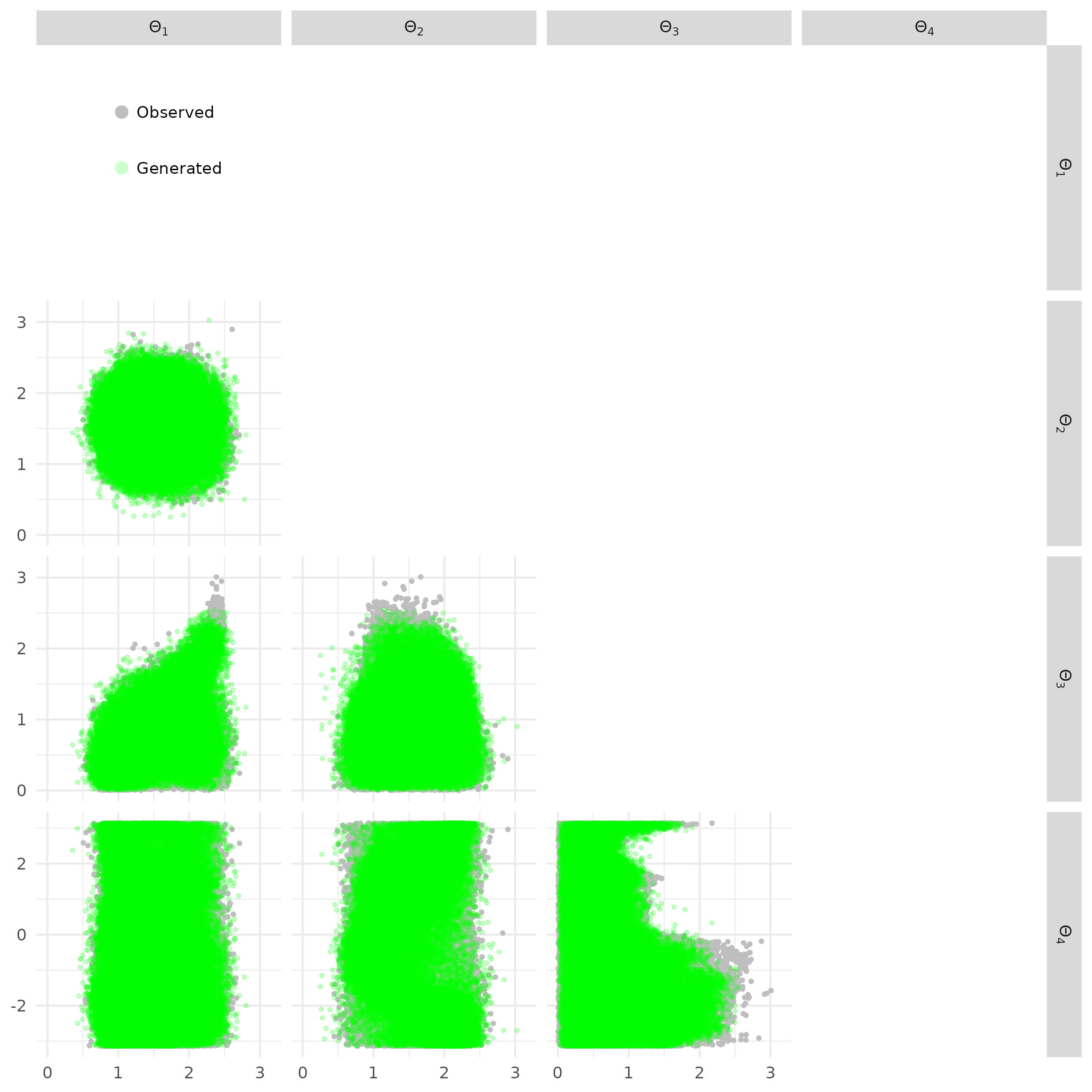}
    \caption{Pairwise scatterplots for spherical angles obtained from the baseline approach.}
    \label{fig:scatter_wave_baseline}
\end{figure}

Figures~\ref{fig:hist_wave_baseline}--\ref{fig:scatter_wave_baseline} give the diagnostics obtained from the baseline approach for the metocean data set. One can observe generally good performance, yet some of the deep learning approaches outperform this technique on the whole, especially in terms of marginal estimation and pairwise dependencies.

\begin{figure}[!htbp]
    \centering
    \includegraphics[width=\linewidth]{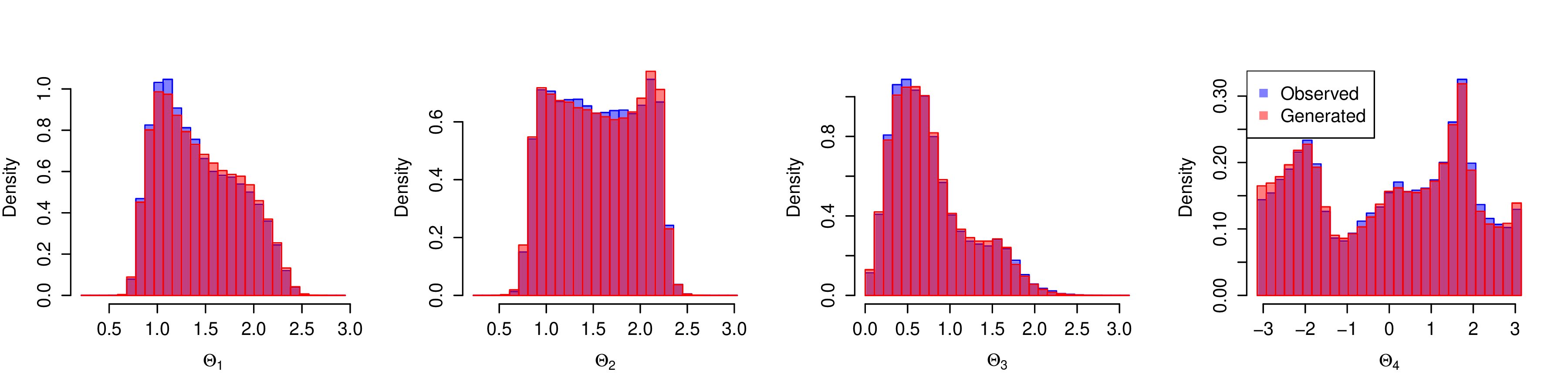}
    \caption{Histogram plots for spherical angles obtained from the FM approach. }
    \label{fig:hist_wave_fm}
\vspace{0.5cm}
    \includegraphics[width=.5\linewidth]{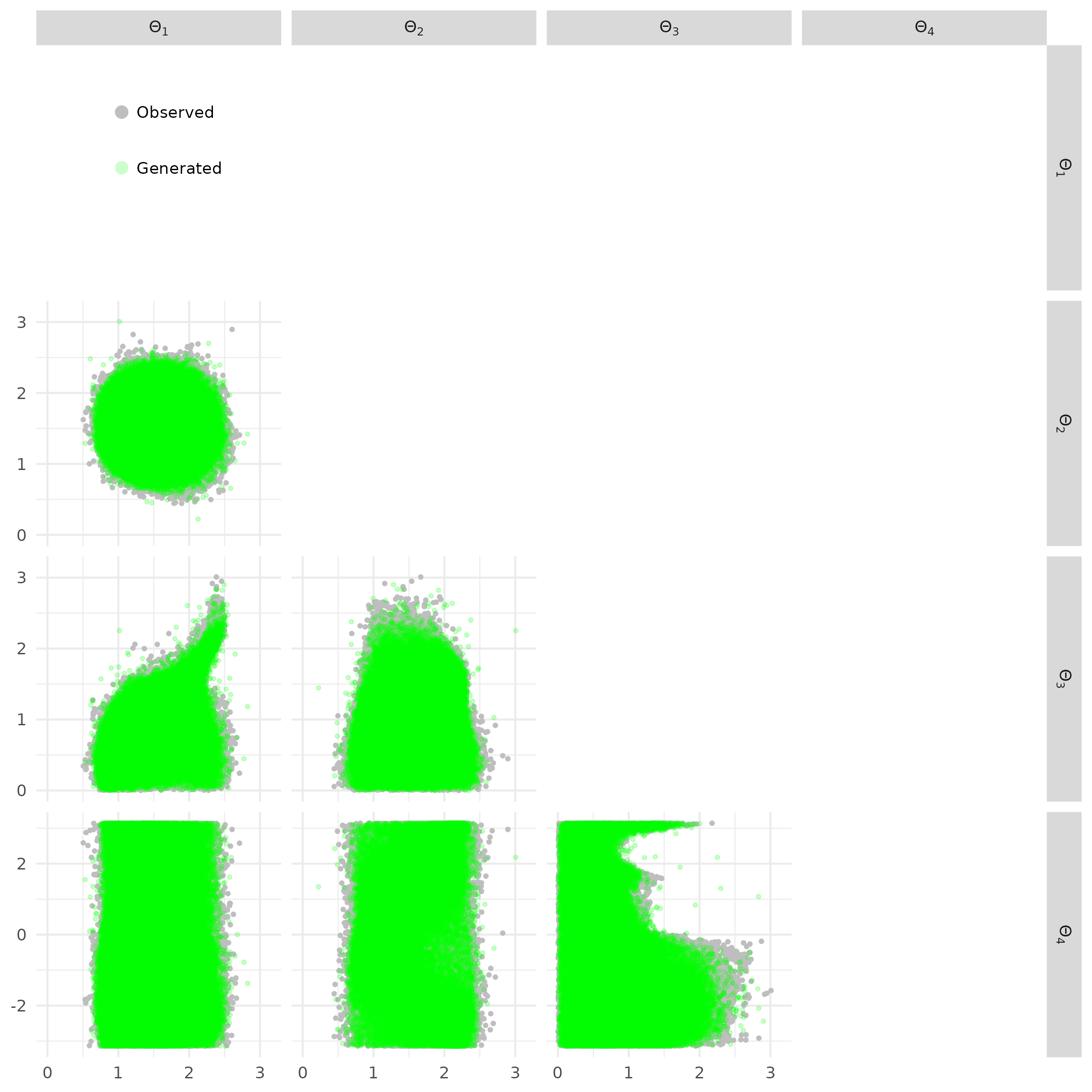}
    \caption{Pairwise scatterplots for spherical angles obtained from the FM approach. }
    \label{fig:scatter_wave_fm}
\end{figure}

\begin{figure}
    \centering
    \includegraphics[width=\linewidth]{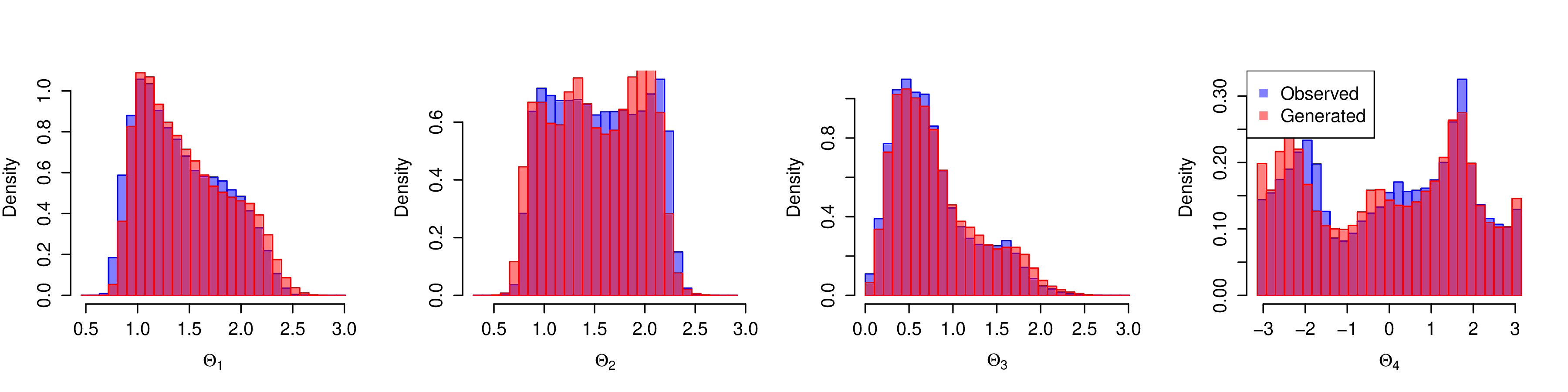}
    \caption{Histogram plots for spherical angles obtained from the GAN approach. }
    \label{fig:hist_wave_gan}
\vspace{0.5cm}
    \centering
    \includegraphics[width=.5\linewidth]{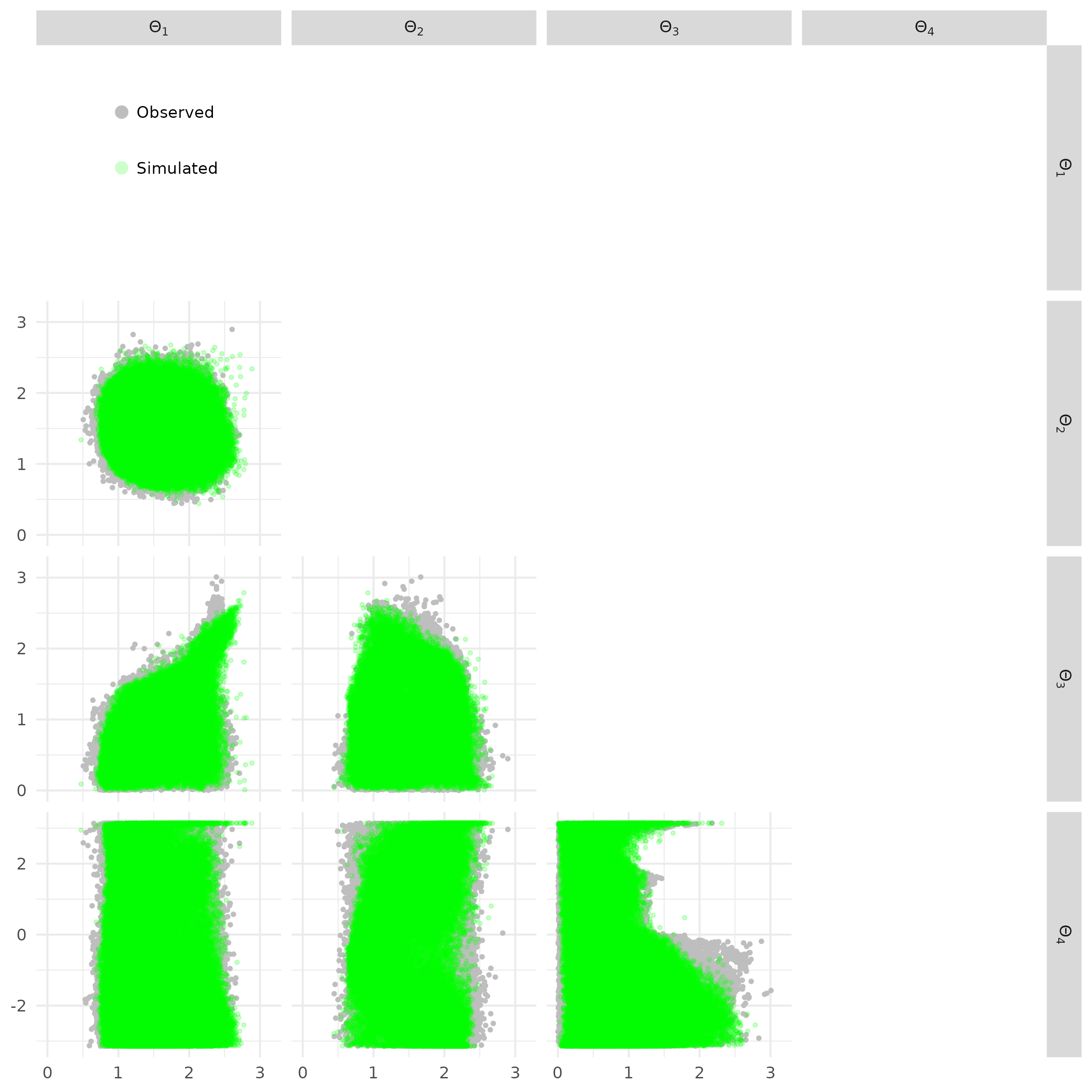}
    \caption{Pairwise scatterplots for spherical angles obtained from the GAN approach. }
    \label{fig:scatter_wave_gan}
\end{figure}

\begin{figure}
    \centering
    \includegraphics[width=\linewidth]{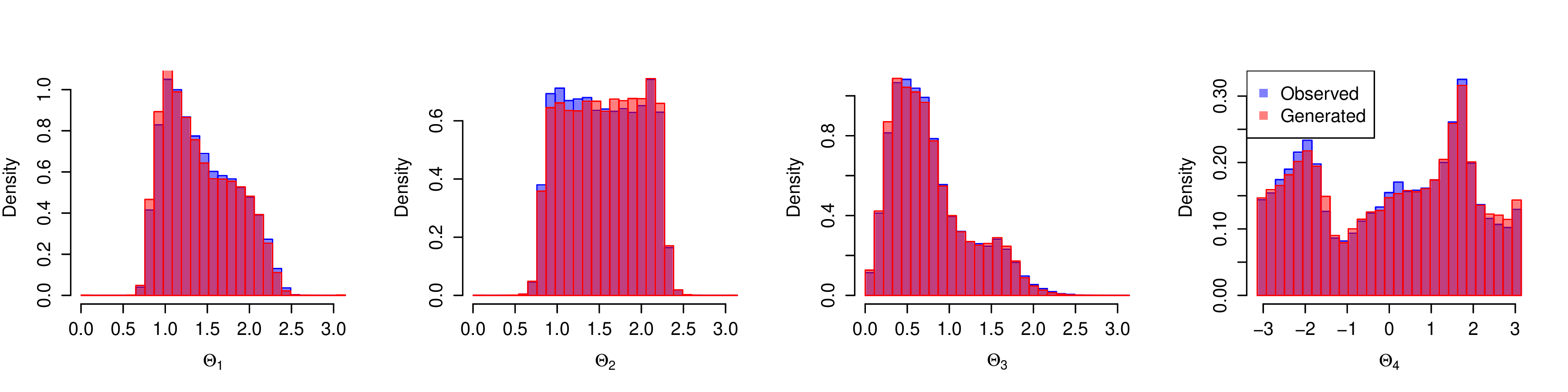}
    \caption{Histogram plots for spherical angles obtained from the NFNSF approach. }
    \label{fig:hist_wave_nf_nsf}
\vspace{0.5cm}
    \includegraphics[width=.5\linewidth]{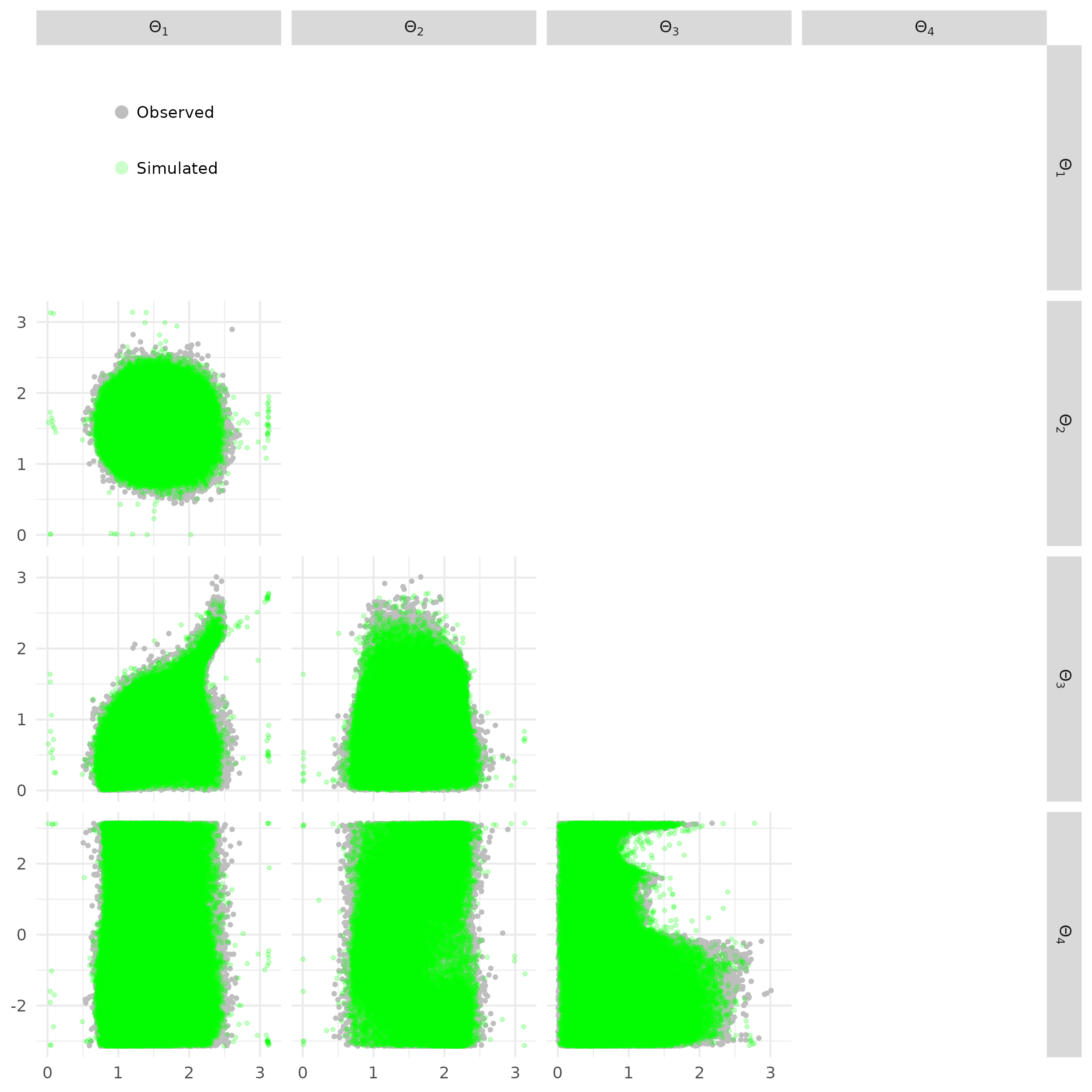}
    \caption{Pairwise scatterplots for spherical angles obtained from the NFNSF approach. }
    \label{fig:scatter_wave_nf_nsf}
\end{figure}

Figures~\ref{fig:hist_wave_fm}--\ref{fig:scatter_wave_nf_nsf} give the spherical histogram and pairwise scatterplots on the metocean data set from the FM, GAN and NFNSF approaches. 

\end{document}